\documentclass[12pt]{ociamthesis}  % default square logo 
\usepackage{etex}

\usepackage{amssymb}
\usepackage{amsmath}
\usepackage{hyperref}
\usepackage{url}

\usepackage{epsfig}
\usepackage[all]{xy}  

\usepackage{setspace}

\usepackage{qtree}

\usepackage{bussproofs}
\usepackage{diagrams}

\usepackage{subfigure}

\usepackage{txfonts}
\usepackage{tikz}
\usetikzlibrary{
    shapes,
    shapes.geometric,
    shapes.symbols,
    shapes.arrows,
    shapes.multipart,
    shapes.callouts,
    shapes.misc}

\newcommand{\semantics}[1]{[\![ #1 ]\!]}

\newenvironment{chabstract}{\begin{quote}\begin{center}\textbf{Chapter Abstract}\end{center}\singlespace\it}{\begin{center}\rule{0.7\textwidth}{1px}\end{center}\end{quote}}

\title{Category-Theoretic Quantitative Compositional Distributional Models of Natural Language Semantics}   %note \\[1ex] is a line break in the title

\author{Edward Grefenstette}             %your name
\college{Balliol College}  %your college

\degree{Doctor of Philosophy}     %the degree
\degreedate{Hilary 2013}         %the degree date

\begin{document}

%this baselineskip gives sufficient line spacing for an examiner to easily
%markup the thesis with comments
% \baselineskip=18pt plus1pt

%set the number of sectioning levels that get number and appear in the contents
\setcounter{secnumdepth}{3}
\setcounter{tocdepth}{3}

\maketitle                  % create a title page from the preamble info

\mbox{}
\thispagestyle{empty}
\newpage
%!TEX root = ../grefenstettethesis.tex

\begin{dedication}
To my parents, Irene and Greg.
\end{dedication}        % include a dedication.tex file
\mbox{}
\thispagestyle{empty}
\newpage
%!TEX root = ../grefenstettethesis.tex

\begin{acknowledgements}

I would like to begin by thanking my supervisors, Bob Coecke, Mehrnoosh Sadrzadeh, and Stephen Pulman, for their guidance and help over the last three years. In particular, I thank Mehrnoosh for her invaluable support in the day-to-day aspects of my work, and for overseeing the production and publication of the papers we co-authored during the course of my DPhil.

On the academic front, I would also like to thank, in no particular order, Stephen Clark, Phil Blunsom, Marco Baroni, Georgiana Dinu, Yao-Zhong Zhang, Tom Melham, Samson Abramsky, Yorick Wilks, and the many other world-class researchers whom I have had the pleasure of interacting with and being challenged by as I learned the ropes of the research world. Thanks also go to the Engineering and Physical Sciences Research Council for funding my doctoral work through a Doctoral Training Account and an Enhanced Stipend.

From home, I thank my loving parents Irene and Greg for their emotional and material support over the last twenty-seven years, and for giving me the opportunity to reach this level of research (as well as proof-reading my dissertation). I thank Grandma and Grandpa Grefenstette for supporting my studies and personal development both at Oxford and throughout life. I also thank my maternal grandparents, Bernard and Olga, my brother Nicholas, my sister Natalie, and the rest of my family for all the love I've received over the years. This work would not have been possible without it.

Finally, thanks to my friends in and out of Oxford, and in particular to Chris, John, Renie, Kelsey, Tom and Emily, Sean, Jenn, Karl, and Jan, for putting up with me throughout the doctoral experience. Listing all those who have made the last few years of my life an adventure is beyond the scope of this thesis, but to all those I've omitted: thank you. It's been a blast.

\end{acknowledgements}   % include an acknowledgements.tex file
\mbox{}
\thispagestyle{empty}
\newpage
%!TEX root = ../grefenstettethesis.tex

\begin{abstract}

{\singlespace This thesis is about the problem of compositionality in distributional semantics. Distributional semantics presupposes that the meanings of words are a function of their occurrences in textual contexts. It models words as distributions over these contexts and represents them as vectors in high dimensional spaces. The problem of compositionality for such models concerns itself with how to produce distributional representations for larger units of text (such as a verb and its arguments) by composing the distributional representations of smaller units of text (such as individual words).}

{\singlespace This thesis focuses on a particular approach to this compositionality problem, namely using the categorical framework developed by Coecke, Sadrzadeh, and Clark, which combines syntactic analysis formalisms with distributional semantic representations of meaning to produce syntactically motivated composition operations. This thesis shows how this approach can be theoretically extended and practically implemented to produce concrete compositional distributional models of natural language semantics. It furthermore demonstrates that such models can perform on par with, or better than, other competing approaches in the field of natural language processing.}

{\singlespace There are three principal contributions to computational linguistics in this thesis. The first is to extend the DisCoCat framework on the syntactic front and semantic front, incorporating a number of syntactic analysis formalisms and providing learning procedures allowing for the generation of concrete compositional distributional models. The second contribution is to evaluate the models developed from the procedures presented here, showing that they outperform other compositional distributional models present in the literature. The third contribution is to show how using category theory to solve linguistic problems forms a sound basis for research, illustrated by examples of work on this topic, that also suggest directions for future research.}

\end{abstract}          % include the abstract
\mbox{}
\thispagestyle{empty}
\newpage

\begin{romanpages}          % start roman page numbering
\tableofcontents            % generate and include a table of contents
\mbox{}
\newpage
\mbox{}
\newpage
\listoffigures              % generate and include a list of figures
\listoftables
\end{romanpages}            % end roman page numbering

\onehalfspacing
% \doublespacing

\mbox{}
\thispagestyle{empty}
\newpage
%now include the files of latex for each of the chapters etc

\clearpage
\setcounter{page}{1}

%!TEX root = ../grefenstettethesis.tex

\chapter{Introduction} % (fold)
\label{cha:introduction}

\begin{quote}
\singlespace
    For a large class of cases---though not for all---in which we employ the word ``meaning'' it can be explained thus: the meaning of a word is its use in the language.

    \hfill\emph{---Ludwig Wittgenstein}
\end{quote}

Language is fundamentally messy. Since the early days of philosophical thought, thinkers have sought to tease structure out of it, from the Aristotlean syllogisms to the development of mathematical logic in the work of Peano and Frege, \emph{inter alia}, at the dawn of the twentieth century. Others, such as the later Wittgensteinian school of thought, held that language was perfectly `in order' as it was, and that it is by measuring the correctness of our speech acts against the linguistic community we live in that we learn how to properly use language, rather than by appealing to some objective standard dictating the correct structure of meaning.

As the world evolved into the present age of information, language became not only the purview of linguists, philosophers and logicians, but also of engineers and scientists. The rapid increase of machine readable and publicly available text through the development of the world wide web prompted a need for new technological tools to systematically classify, search, translate, summarise, and analyse text. As this need grew, new mathematical and computational methods were developed to extract structure from unstructured data.

While some thinkers of the modern age sought to adapt the tools and methods of logicians, leading to the development of mathematical accounts of natural language meaning such as formal semantics, others wished to accept the messiness of language. They instead turned to the large amount of data available in order to model meaning through statistical and distributional means. Thus while formal semanticists treated natural language as a programming language, with grammar as its syntax and logic as its semantics, distributional and statistical semanticists interpreted the meaning of our words as being a function of their contexts, as observed in text from various sources.

Both in the case of philosophers of language and computational linguists, there is an apparent mutual inconsistency between the formal and distributional/empirical approaches. They seem orthogonal, in that the former portrays language meaning as well-organised, tidy, and well structured; while the latter aims to learn the underlying meaning of words without appeal to any underlying structure beyond the superficial relations words hold to one another through grammatical relations, or through simply occurring within the same sentence or document. However, it could be argued that such a distinction between these two views of semantics forms a false dichotomy, and that there is some middle ground on which we could both think of language as being a matter of structured relations and functions on the one hand, and empirically learnable meaning on the other.

This thesis deals with this middle ground. In the past decade, computational linguists have sought to combine the structured approach of formal semantics and the empirical methods of distributional semantics to produce a new class of models of natural language semantics, dubbed \emph{compositional distributional semantic models}, capable of exploiting the structured aspects of language while learning meaning empirically.

I begin, in Chapter~\ref{cha:literature_review}, by giving an overview of these two seemingly different ways of modelling natural language semantics. I discuss formal semantics, which creates an association between grammatical elements and parts of logical expressions, in which syntactic analysis yields the production of logical statements based on how grammatical elements combine to form sentences.
\begin{center}
    {\footnotesize
        \begin{tabular}{ccc}
            \begin{tabular}{l|l}
                Syntactic Analysis & Semantic Interpretation\\
                \hline
                S $\Rightarrow$ NP VP & $\semantics{VP}(\semantics{NP})$\\
                NP $\Rightarrow$ cats, milk, etc. & $\semantics{\textrm{cats}},\,\semantics{\textrm{milk}},\,\ldots$\\
                VP $\Rightarrow$ Vt NP & $\semantics{Vt}(\semantics{NP})$\\
                Vt $\Rightarrow$ like, hug, etc. & $\lambda yx.\semantics{\textrm{like}}(x,y),\,\ldots$ \\
            \end{tabular}
            &
            $\quad \Rightarrow \quad$
            &
            \begin{tabular}{c}
                \Tree [.$\semantics{\textrm{chase}}(\semantics{\textrm{dogs}},\semantics{\textrm{cats}})$ $\semantics{\textrm{dogs}}$ [.$\quad \lambda x.\semantics{\textrm{chase}}(x,\semantics{\textrm{cats}})$ $\quad\lambda yx.\semantics{\textrm{chase}}(x,y)$  $\semantics{\textrm{cats}}$  ] ]
            \end{tabular}
        \end{tabular}}\\[.5cm]
        
    A simple formal semantic model.
\end{center}
I then introduce distributional semantic models, which model the meaning of words as vectors in high dimensional spaces, constructed from the other words they co-occur with.
\begin{center}
            \setlength{\unitlength}{0.5mm}
            \begin{picture}(100, 100)
              \thicklines
              \put(22,82){\text{furry}}
              \linethickness{0.4mm}
              \put(30, 20){\vector(1, 0){50}}
              \put(1,-5) {\text{stroke}}
              \put(30, 20){\vector(0, 1){60}}
              \put(83,18){\text{pet}}
              \put(30, 20){\vector(-1, -1){20}}
                \thinlines
                  \linethickness{0.4mm}
              \put(30, 20){\begin{color}{blue}\vector(1, 2){20}\end{color}}
              \put(30, 20){\begin{color}{blue}\vector(1, 1){25}\end{color}}
              \put(30, 20){\begin{color}{blue}\vector(0.05, -0.05){20}\end{color}}
            %    \put(30, 20){\begin{color}{blue}\vector(0, -2){14}\end{color}}
              \put(45,62){\text{{\bf cat}}}
              \put(53,47){\text{\bf dog}}
              \put(55,0){\text{\bf snake}}
            %    \put(24,0){\text{\bf Cat}}
            \end{picture}\\[.5cm]

            Graphical representation of a simple distributional semantic model.
\end{center}
Furthermore, I discuss past attempts to reconcile both approaches with mathematical operations that compose word vectors to form semantic vectors for larger units of text such as phrases or sentences, surveying the different approaches to this problem in the literature.

In Chapter~\ref{cha:foundations_of_discocat}, I give an overview of an existing framework---DisCoCat---developed by \cite{Coecke2010}, which provides a general account of how grammatical formalisms and distributional semantics can be combined to generate compositional distributional models of semantics. The DisCoCat framework borrows a powerful mathematical tool frequently used in quantum information theory, namely category theory, which allows information to be communicated between different mathematical formalisms, provided that they share some underlying structure. I survey the background knowledge required to understand this framework by introducing pregroup grammars, a syntactic formalism with a convenient algebraic structure, and which is easily represented as a category.
\begin{center}
    \begin{tikzpicture}[thick, every text node part/.style={align=center}]
\node [above] (v1) at (0,0) {John \\ $n$};
\node [above] (v2) at (1,0) {$n^r$};
\node [above] (v5) at (1.5,0) {loves\\ $s$};
\node (v6) at (1.5,-.75) {};
\node [above] (v3) at (2,0) {$n^l$};
\node [above] (v4) at (3,0) {Mary \\$n$};
\draw (0,-.25) .. controls +(0,-.75) and +(0,-.75) .. (1,-.25);
\draw (2,-.25) .. controls +(0,-.75) and +(0,-.75) .. (3,-.25);
\draw [-] (1.5,-.25) -- (1.5,-1);
\end{tikzpicture}\\[.5cm]

Pregroup parse of ``John loves Mary''.
\end{center}
I also introduce the basics of category theory, and the diagrammatic calculus which allows us to reason about compositional operations within a category.
 \begin{center}
     \tikzstyle{bordered} = [draw,outer sep=0,inner sep=1,minimum size=15]
\tikzstyle{vector} = [draw, isosceles triangle, shape border rotate=90, isosceles triangle stretches, outer sep=0,inner sep=1, minimum height=5, minimum width=30]
\tikzstyle{covector} = [draw, isosceles triangle, shape border rotate=-90, isosceles triangle stretches, outer sep=0,inner sep=1, minimum height=5, minimum width=30]
    \begin{tikzpicture}[thick]

\node[vector] (v1) at (-.5,2) {$v$};
\node[vector,minimum width=3cm] (T) at (2,2.1) {$T$};
\node[vector] (v2) at (4.5,2) {$w$};

\draw (v1) -- (-.5,1.5);
\draw (v2) -- (4.5,1.5);
\draw (T) -- (2,1);
\draw (1.25,1.76) -- (1.25,1.5);
\draw (2.75,1.76) -- (2.75,1.5);

\draw (-.5,1.5) .. controls +(0,-.75) and +(0,-.75) .. (1.25,1.5);
\draw (2.75,1.5) .. controls +(0,-.75) and +(0,-.75) .. (4.5,1.5);

\node at (6.25,1.75) {$=$};

\node[vector,minimum width=3cm] (T2) at (9,2.1) {$T$};
\draw (T2) -- (9,1);

\node[covector] (v3) at (8.25,1.3) {$v$};
\node[covector] (v4) at (9.75,1.3) {$w$};

\draw (8.25,1.76) -- (v3);
\draw (9.75,1.76) -- (v4);
\end{tikzpicture}\\[.5cm]

Graphical representation of subject-verb-object composition in DisCoCat.
 \end{center}
Finally, I discuss how distributional semantics fits into the categorical discourse within DisCoCat, and how giving categorical representations to both pregroup grammars and vector spaces allows us to produce syntactically motivated composition operations.

In Chapter~\ref{cha:syntactic_extensions}, I develop new syntactic extensions to the DisCoCat framework. I begin by presenting the notion of a functorial passage, and discuss how functors can be used to extend the framework to incorporate syntactic formalisms other than pregroup grammars. I then illustrate this by showing how Context Free Grammars can be incorporated into the extended framework by translating them into pregroup grammars. I also show how Lambek Grammars can be integrated into the framework by interpreting them as a specific kind of category called a bi-closed monoidal category. I show how an existing graphical calculus for such categories can be applied to develop a functorial passage from these categories to the categories representing our semantic models.
\begin{center}
\tikzset{func/.style={shape=rectangle,rounded corners=8,minimum width=2cm,minimum height=.5cm,draw}}
\tikzset{claspnode/.style={shape=circle,minimum width=0.25cm,fill=white,draw}}
            \begin{minipage}{3cm}
        \begin{tikzpicture}[thick,scale = 0.6]
            
            \begin{scope}
                
                \draw (0, 6.5) node {Men};
                \draw (2,6.5) node {kill};

                \draw[<-] (1.5,6) -- node [left] {$n$} +(0,-1.5);
                \draw[->] (2.5,6) -- node [right] {$s$} +(0,-1.5);
                \draw (2.5,5.5) node (c4) [claspnode] {};
                \draw (1.5,5.5) to (c4);
                
                \draw[->] (0,6) -- node[left] {$n$} (0,4.5);
                
                \draw (1.5,3.5) ellipse (2cm and 1cm);
                \draw (1.5,4.5) -- (1.5,4);
                \draw (0,4.5) -- (0,4);
                \draw (0,4) .. controls +(0,-1) and +(0,-1) .. (1.5,4);
                \draw (2.5,4.5) -- (2.5,3);
                
                \draw[->] (2.5,3) -- node [right] {$s$} +(0,-1);
                
            \end{scope}         
            
        \end{tikzpicture}
\end{minipage} \hspace{3cm}
\begin{minipage}{5cm}
    \begin{tikzpicture}[thick,scale = 0.6]
        
        \begin{scope}
            
            \draw (0, 6.5) node {Men};
            \draw (2.5,6.5) node {kill};
            \draw (5,6.5) node {dogs};

            % \draw[<-] (1.5,6) -- node [left] {$N$} +(0,-1.5);
            \draw[->] (2.5,6) -- node [left] {$n \multimap s\ $} +(0,-1.5);
            \draw (2.5,5.5) node (c4) [claspnode] {};
            \draw (3.5,5.5) to (c4);
            
            \draw[<-] (3.5,6) -- node[right] {$n$} (3.5,4.5);
            \draw[->] (5,6) -- node[left] {$n$} (5,4.5);
            \draw (3.5,4) .. controls +(0,-1) and +(0,-1) .. (5,4);
            \draw (3.5,4.5) -- (3.5,4);
            \draw (5,4.5) -- (5,4);
            \draw (3.75,3.5) ellipse (2cm and 1cm);
            \draw[->] (2.5,3) -- node [left] {$n \multimap s\ $} +(0,-1);
            
            \draw[->] (0,6) -- node[left] {$n$} (0,4.5);

            \draw[dashed] (2.5,2) -- (1.5,1.5);
            \draw[dashed] (2.5,2) -- (2.5,1.5);
            
            \draw[->] (2.5,1.5) -- node [right] {$\ s$} +(0,-1);
            \draw[<-] (1.5,1.5) -- node [left] {$n$} +(0,-1);
            \draw (2.5,1) node [claspnode] (c5) {};
            \draw (1.5,1) to (c5);
            
            \draw (1.5,.5) -- (1.5,0);
            \draw (0,4.5) -- (0,0);
            \draw (0,0) .. controls +(0,-1) and +(0,-1) .. (1.5,0);
            \draw (2.5,4.5) -- (2.5,3);
            \draw (2.5,.5) -- (2.5,-.5);
            \draw (1.25,-.5) ellipse (2cm and .8cm);
            \draw[->] (2.5,-.5) -- node [right] {$s$} +(0,-1.5);

        \end{scope}         
        
    \end{tikzpicture}
    \end{minipage}\\[.5cm]

    Example of the diagrammatic calculus for categorical representation of Lambek Grammar.
\end{center}

In Chapter~\ref{cha:learning_procedures_for_a_discocat}, I present a new learning procedure for generating concrete compositional distributional semantic models from the DisCoCat framework. This procedure learns the semantic representations for relations by summing over the information that the relations' arguments hold in a corpus. I present a reduced representation for such relations which allows us to efficiently compute the semantic representations for relations which would otherwise require large amounts of data to model. I define composition operations for these reduced representations. Furthermore, I demonstrate that, under certain assumptions, the reduced representations are equivalent to the full representations of the relation they model and are embedded within these full representations.
\begin{center}
        \tikzstyle{bordered} = [draw,outer sep=0,inner sep=1,minimum size=15]
\tikzstyle{vector} = [draw, isosceles triangle, shape border rotate=90, isosceles triangle stretches, outer sep=0,inner sep=1, minimum height=5, minimum width=30]
    \begin{tikzpicture}[thick,scale=0.9]

\node[vector,minimum width=7cm,minimum height=3cm] at (0,2) {};
\node[bordered,minimum width=1.5cm] at (0,1.25) {$f^v$};

\draw (-1.25,0) -- (-1.25,1.75);
\draw (1.25,0) -- (1.25,1.75);
\draw (-.25,1.53) -- (-.25,1.75);
\draw (.25,1.53) -- (.25,1.75);
\draw (-.25,.97) -- (-.25,-1);
\draw (.25,.97) -- (.25,-1);

\draw (-1.25,1.75) .. controls +(0,.75) and +(0,.75) .. (-.25,1.75);
\draw (1.25,1.75) .. controls +(0,.75) and +(0,.75) .. (.25,1.75);

\node[vector] (subj) at (-5,.85) {$s$};
\node[vector] (obj) at (5,.85) {$o$};
\draw (subj) -- (-5,0);
\draw (obj) -- (5,0);

\draw (-5,0) .. controls +(0,-1) and +(0,-1) .. (-1.25,0);
\draw (5,0) .. controls +(0,-1) and +(0,-1) .. (1.25,0);

\node at (7,1) {$=$};

\node[bordered,minimum width=2cm] at (9,1.25) {$f^v$};
\node[vector] (subj2) at (8.4,2) {$s$};
\node[vector] (obj2) at (9.6,2) {$o$};
\draw (8.4,1.53) -- (subj2);
\draw (9.6,1.53) -- (obj2);
\draw (8.4,.97) -- (8.4,-1);
\draw (9.6,.97) -- (9.6,-1);
\end{tikzpicture}

Diagrammatic example of embedded reduced representations.
\end{center}
I furthermore provide a simplified alternative learning procedure for these reduced representations, which ignores some aspects of the relations it provides a model for, but is significantly quicker to learn and more efficient to store, while outperforming the model it supplants in certain experiments.
\begin{center}
    \tikzstyle{bordered} = [draw,outer sep=0,inner sep=1,minimum size=15]
\tikzstyle{vector} = [draw, isosceles triangle, shape border rotate=90, isosceles triangle stretches, outer sep=0,inner sep=1, minimum height=5, minimum width=30]
\begin{tikzpicture}[thick]

\begin{scope}
    \node[bordered,minimum width=4cm] at (0,1.25) {$f^R$};
\node[vector] (subj2) at (-1.5,2) {$a_1$};
\node[vector] (obj2) at (1.5,2) {$a_n$};
\draw (-1.5,1.5) -- (subj2);
\draw (1.5,1.5) -- (obj2);
\draw (-1.5,1) -- (-1.5,0);
\draw (1.5,1) -- (1.5,0);

\node at (0,2) {$\ldots$};

\node at (0,.5) {$\ldots$};

\end{scope}

\begin{scope}[xshift=3cm]
\node at (0,1.25) {$=$};
\end{scope}

\begin{scope}[xshift=4cm]
 \node[bordered] (frl) at (0,1.25) {$f^{\bar{R}}$};
 \node[bordered] (frr) at (2,1.25) {$f^{\bar{R}}$};   
\node[vector] (subj2) at (0,2) {$a_1$};
\node[vector] (obj2) at (2,2) {$a_n$};
\draw (frl) -- (subj2);
\draw (frr) -- (obj2);
\draw (frl) -- (0,0);
\draw (frr) -- (2,0);

\node at (1,1.25) {$\ldots$};

\end{scope}

\end{tikzpicture}\\[.5cm]

Diagrammatic form of generalised Kronecker composition model.
\end{center}

In Chapter~\ref{cha:evaluating_a_discocat}, I evaluate the new models described in the previous chapter in a series of sentence similarity detection experiments, comparing these models with other compositional distributional models of semantics. These experiments test how well compositional models can produce sentence representations for sentences containing ambiguous words, and disambiguate the constituent words through the compositional process.
\begin{center}
 \begin{tabular}{|c|c|}
 \hline
 Sentence 1 & Sentence 2\\
 \hline
 \hline
 
 butler bow & butler submit  \\
\hline
 head bow & head stoop\\
 \hline
  company bow & company submit \\
\hline
 government bow & government stoop\\
 \hline
 \end{tabular}\\[.5cm]

 Sample sentence pairs from a phrase similarity detection experiment.
\end{center}
These experiments show that the models generated by the DisCoCat framework and the learning procedures developed in this thesis outperform other approaches to distributional compositionality.

Finally, in Chapter~\ref{cha:applications_and_future_work}, I discuss additional work that I have done on this topic with various collaborators, and outline future work to continue to develop this growing field. Three further developments are described. The first addresses the issue of incorporating logical operations into compositional distributional models. I discuss how a simple predicate calculus might be simulated using tensors, and how this simulation fits into the DisCoCat framework, before considering some of the difficulties that arise when trying to model genuine logical calculi using distributional methods. Second, I discuss how machine learning methods such as linear regression may be adapted to the DisCoCat framework, as an alternative to the learning procedures I initially presented. Third, I provide the foundations for integrating a complex grammatical formalism, Combinatory Categorial Grammars, into the DisCoCat framework, and discuss issues faced when trying to accommodate the many variants this formalism possesses. I conclude by suggesting general directions future research might take based on these three points.

The aim of this thesis is to not only provide an exploration of the ways in which the DisCoCat framework can be extended, but to also give a proof-of-concept, showing that this abstract framework can be instantiated, that concrete models can be developed based on it, and that these models offer an interesting new direction in the search for sophisticated models of natural language semantics.

% chapter introduction (end)

\part{Background}
\mbox{}
\newpage

%!TEX root = ../grefenstettethesis.tex

\chapter{Literature Review} % (fold)
\label{cha:literature_review}

\begin{chabstract}
  This chapter presents the background literature concerning compositionality in distributional semantics. The reader will find an overview of another well-known compositional formalism, formal semantics, as well as a brief reminder of the basic concepts behind distributional semantic models. Then follows a survey of various approaches to the mathematics of vector composition from the last few decades, and the discussing and contrasting of various matrix-vector based approaches to distributional compositionality.
\end{chabstract}

Compositional formal semantic models represent words as parts of logical expressions, composed according to grammatical structure. These models embody classical ideas in logic and philosophy of language, mainly Frege's principle that the meaning of a sentence is a function of the meaning of its parts \cite{Frege1892}. Well studied and robust, logical formalisms offer a scalable theory of meaning which can be used to reason about language using logical tools of proof and inference. In contrast, distributional models are a more recent approach to semantic modelling, representing the meaning of words not as logical formula but as vectors whose values have been empirically learned from corpora. Distributional models have found their way into real world applications such as thesaurus extraction \cite{Grefenstette1994,Curran2004} or automated essay marking \cite{Landauer1997}, and have strong connections to semantically motivated information retrieval \cite{Manning}. This new dichotomy in defining properties of meaning: `logical form' versus `contextual use',  has left the question of the foundational structure of meaning, initially of sole concern to linguists and to philosophers of language, even more of a challenge.

In this chapter, I present an overview of the background to the work developed in this thesis by briefly describing formal and distributional approaches to natural language semantics, and providing a non-exhaustive list of some approaches to compositional distributional semantics. For  a more complete review of the topic, I encourage the reader to consult one of the many excellent surveys of the field \cite{turney2010frequency,Curran2004,Clarkbookchap2013}.

\section{Formal Semantics} % (fold)
\label{sec:montague_semantics}

The approach commonly known as \emph{Formal Semantics}, principally due to the work of Richard Montague in the area of inductive logic, provides methods for translating sentences of natural language into logical formulae, which can then be fed to computer-based reasoning tools \cite{Alshawi1992}.

\begin{figure}[h]
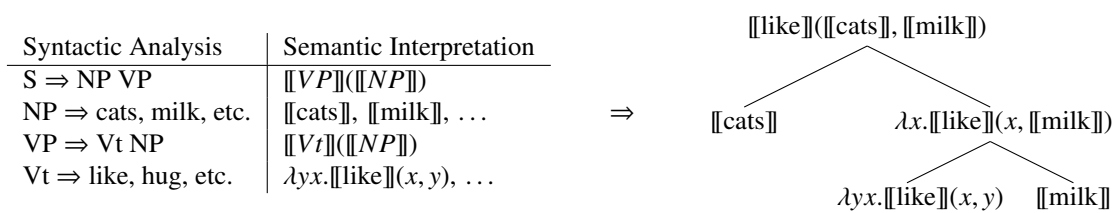

\begin{center}
    \footnotesize
    \begin{tabular}{ccc}
        \begin{tabular}{l|l}
            Syntactic Analysis & Semantic Interpretation\\
            \hline
            S $\Rightarrow$ NP VP & $\semantics{VP}(\semantics{NP})$\\
            NP $\Rightarrow$ cats, milk, etc. & $\semantics{\textrm{cats}},\,\semantics{\textrm{milk}},\,\ldots$\\
            VP $\Rightarrow$ Vt NP & $\semantics{Vt}(\semantics{NP})$\\
            Vt $\Rightarrow$ like, hug, etc. & $\lambda yx.\semantics{\textrm{like}}(x,y),\,\ldots$ \\
        \end{tabular}
        &
        $\quad \Rightarrow \quad$
        &
        \begin{tabular}{c}
            \Tree [.$\semantics{\textrm{like}}(\semantics{\textrm{cats}},\semantics{\textrm{milk}})$ $\semantics{\textrm{cats}}$ [.$\quad \lambda x.\semantics{\textrm{like}}(x,\semantics{\textrm{milk}})$ $\quad\lambda yx.\semantics{\textrm{like}}(x,y)$  $\semantics{\textrm{milk}}$ ] ]
        \end{tabular}
    \end{tabular}
\end{center}
\caption{A simple model of formal semantics.}
\label{fig:formal_semantic_model}
\end{figure}

To compute the meaning of a sentence consisting of $n$ words, the individual meanings of each words must \emph{interact}. In formal semantics, this interaction is represented as a function derived from the grammatical structure of the sentence. Formal models consist of a pairing of syntactic analysis rules (in the form of a grammar) with semantic interpretation rules, as exemplified by the simple model presented on the left hand side of Figure~\ref{fig:formal_semantic_model}.

The semantic representations of words here are expressions of a higher-order logic formed by lambda expressions over parts of first-order logical formulae, which can be combined with one another to form well-formed expressions of first-order logic. The function $\semantics{-}:\mathcal{L} \to \mathcal{M}$ maps elements of the lexicon $\mathcal{L}$ to their interpretation in the logical model $\mathcal{M}$ used. Proper nouns are typically just logical atoms, while nouns, adjectives, verbs, and other relational words are interpreted as predicates and relations. The parse of a sentence such as ``cats like milk'', represented here as a binarised parse tree, is used to produce its semantic interpretation by substituting semantic representations for their grammatical constituents and applying $\beta$-reduction where needed. Such a derivation is shown on the right hand side of Figure~\ref{fig:formal_semantic_model}.

What makes this class of models attractive is that it reduces language meaning to logical expressions, a subject well-known to philosophers of language, logicians, and linguists. The properties of first-order and higher-order logics are well studied, and it is possible to evaluate the meaning of a sentence if given a logical model and domain (the model-theoretic approach to logic), as well as to verify whether or not one sentence entails another according to the rules of logical consequence and deduction, based on syntactic rules (the syntactic approach to logic).

However, such logical analysis says nothing about the closeness in meaning or topic of expressions beyond their truth-conditions and which models satisfy these truth conditions. Therefore, formal semantic approaches to modelling language meaning do not perform well on language tasks where the notion of similarity is not strictly based on truth conditions, such as document retrieval, topic classification, etc. Furthermore, an underlying domain of objects and a valuation function must be provided, as with any logic, leaving open the question of how we might \emph{learn} the meaning of language using such a model, rather than just use it.

% section montague_semantics (end)

\section{Distributional Semantics} % (fold)
\label{sec:distr}

The distributional semantics approach to lexical semantics represents the meaning of words as distributions in a high-dimensional vector space. This approach is based on the \emph{distributional hypothesis} of Harris \cite{Harris1968}, who postulated that the meaning of a word was dictated by the context of its use. The more famous dictum stating this hypothesis is Firth's statement \cite{Firth1957} that ``You shall know a word by the company it keeps''. This view of semantics has furthermore been associated \cite{MScGrefenstette2009,turney2010frequency} with earlier work in philosophy of language by Wittgenstein, presented in \cite{Wittgenstein1953}, who stated that language meaning was equivalent to its real world use.

Practically speaking, in this approach, the meaning of a word can be learned from a corpus by looking at what other words occur with it within a certain \emph{context}. The resulting distribution can be represented as a vector in a semantic vector space. This vectorial representation is convenient because vectors are a familiar structure with a rich set of ways of computing vector distance, allowing us to experiment with different word similarity metrics. The geometric nature of this representation entails that we can not only compare individual words' meaning with various levels of granularity (e.g.~we might, for example, be able to show that cats are closer to kittens than to dogs, but that all three are mutually closer than cats and steam engines), but also apply methods frequently called upon in information retrieval tasks such as those described by \cite{Manning}, to group concepts by topic, sentiment, or other semantic classes.

The distribution underlying word meaning here is a vector in a vector space, the basis vectors of which are dictated by the context. In simple models, the basis vectors will be annotated with words from the lexicon. Traditionally, the vector spaces used in such models are Hilbert spaces, i.e.~vector spaces with orthogonal bases, such that the inner product of any one basis vector with another (other than itself) is zero. The semantic vector for any word can be represented as the weighted superposition (i.e.~the vector sum) of the basis vectors:
\begin{displaymath}
    \overrightarrow{\textrm{some word}} = \sum_i{c_i \overrightarrow{n_i}}
\end{displaymath}
where some set of orthogonal unit vectors $\{\overrightarrow{n_i}\}_i$ is the basis of the vector space which the meaning of the word lives in, and $c_i \in \mathbb{R}$ is the weight associated with basis vector $\overrightarrow{n_i}$.

A common source of confusion when encountering these models for the first time is to think of these annotated basis vectors as representations of the words which annotate them. In fact, they are simply representations of those words as context tokens. For example, the word `furry' may be used to annotate one of the basis elements of a vector space in its role as a context. However, when we wish to reason about the meaning of `furry', we will use the semantic vector associated with it in the vector space, which is distinct from the basis vector annotated with `furry'. This distinction may help explain why the orthogonality of the basis vectors is acceptable despite the fact that some of the words which annotate them may have similar meaning: we assume their context-theoretic properties to be independent, even if the semantic vectors we will eventually construct for these words may be closer in the space.

The construction of the vector for a word is done by counting, for each lexicon word $n_i$ associated with basis vector $\overrightarrow{n_i}$, how many times $n_i$ occurs in the context of each occurrence of the word for which we are constructing the vector. This count is then typically adjusted according to a weighting scheme (e.g.~term frequency inverse document frequency). The ``context'' of a word can be something as simple as the other words occurring in the same sentence as the word or within $k$ words of it, or something more complex, such as using dependency relations or other syntactic features.

\begin{figure}[h]
    \begin{center}
            \setlength{\unitlength}{0.5mm}
            \begin{picture}(100, 100)
              \thicklines
              \put(22,82){\text{furry}}
              \linethickness{0.4mm}
              \put(30, 20){\vector(1, 0){50}}
              \put(1,-5) {\text{stroke}}
              \put(30, 20){\vector(0, 1){60}}
              \put(83,18){\text{pet}}
              \put(30, 20){\vector(-1, -1){20}}
                \thinlines
                  \linethickness{0.4mm}
              \put(30, 20){\begin{color}{blue}\vector(1, 2){20}\end{color}}
              \put(30, 20){\begin{color}{blue}\vector(1, 1){25}\end{color}}
              \put(30, 20){\begin{color}{blue}\vector(0.05, -0.05){20}\end{color}}
            %    \put(30, 20){\begin{color}{blue}\vector(0, -2){14}\end{color}}
              \put(45,62){\text{{\bf cat}}}
              \put(53,47){\text{\bf dog}}
              \put(55,0){\text{\bf snake}}
            %    \put(24,0){\text{\bf Cat}}
            \end{picture}
            \end{center}
            \caption{A simple model of distributional semantics.}
    \label{fig:distributional_semantic_model}
\end{figure}
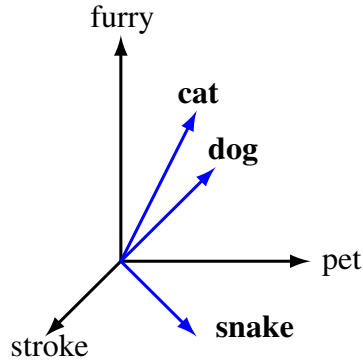

To give an example, suppose that we wish to construct the semantic vectors for the meaning of `dog', `cat' and `snake'. Let us fix the set of possible context words to be $\{$`furry', `pet', `stroke'$\}$. Let us consider `context' to mean ``words occurring in the same sentence as the target word''. We begin by looking at the instances of `dog' in some training corpus of real text usage and see that it occurs twice with `furry', twice with `pet' and once with `stroke'. We therefore learn from the training corpus the vector:
\[
\overrightarrow{\textrm{dog}} = [2\ 2\ 1]^\top
\]
We then observe that `cat' occurs thrice in the same sentence as `furry', once as `pet' and does not occur in the same sentence as stroke. Likewise, we note that `snake' does not occur in the context of `furry', but twice in the context of `pet' and `stroke'. We therefore build the vectors:
\[
\overrightarrow{\textrm{cat}} = [3\ 1\ 0]^\top \qquad \overrightarrow{\textrm{snake}} = [0\ 2\ 2]^\top
\]
Figure~\ref{fig:distributional_semantic_model} shows a graphical representation of the semantic space in which we built these vectors. We can visually observe that `cat' and `dog' seem closer in this space than `cat' and `snake' or `dog' and `snake'. We can also see that `dog' appears closer to `snake' in this space than the vector for `cat' is close to `snake'. The geometric notion of a semantic space therefore yields this non-binary aspect of similarity: concepts are not classified as similar or dissimilar, but are instead part of a continuum of similarity defined by the similarity metric used to compare vectors.

Commonly, the similarity of two semantic vectors is computed by taking their cosine measure, which is the sum of the product of the basis weights of the vectors:
\begin{displaymath}
    cosine( \overrightarrow{a}, \overrightarrow{b}) = \frac{\sum_i{c^a_i c^b_i}}{\sqrt{\sum_i{(c^a_i)^2}\sum_i{(c^b_i)^2}}}
\end{displaymath}
where $c^a_i$ and $c^b_i$ are the basis weights for $\overrightarrow{a}$ and $\overrightarrow{b}$, respectively.
However, other options may be a better fit for certain implementations, typically dependent on the weighting scheme. Some of these are surveyed by \cite{Curran2004}. Throughout this thesis, I will use the cosine measure as the principal similarity metric for vector comparison because it is the most commonly used metric for the models I will consider and compare my work to.

To continue with our example, we wish to compute the relative similarity of `dog' and `cat' versus that of `dog' and `snake', to verify that dogs and cats are more similar in our model than dogs and snakes. We first compute:
\[
\left\langle \overrightarrow{\textrm{dog}} \mid \overrightarrow{\textrm{cat}} \right\rangle =
\left[2\ 2\ 1\right]
\left[
\begin{tabular}{c}
$3$\\
$1$\\
$0$
\end{tabular}
\right]
= 2 \cdot 3 + 2 \cdot 1 + 1 \cdot 0 = 8
\]
To obtain the cosine measure, we must normalise the inner product by the product of vector lengths
\[
\left\Vert \overrightarrow{\textrm{dog}} \right\Vert \cdot \left\Vert \overrightarrow{\textrm{cat}} \right\Vert
= \sqrt{2^2 + 2^2 + 1^2} \cdot \sqrt{3^2 + 1^2 + 0^2} = 3\sqrt{10}
\]
to obtain the cosine measure
\[
cosine(\overrightarrow{\textrm{dog}},\overrightarrow{\textrm{cat}}) = \frac{8}{3\sqrt{10}} \approx 0.84
\]

We apply the same procedure for `dog' and `snake', first calculating the inner product of the semantic vectors
\[
\left\langle \overrightarrow{\textrm{dog}} \mid \overrightarrow{\textrm{snake}} \right\rangle =
\left[2\ 2\ 1\right]
\left[
\begin{tabular}{c}
$0$\\
$2$\\
$2$
\end{tabular}
\right]
= 2 \cdot 0 + 2 \cdot 2 + 1 \cdot 2 = 6
\]
and then the normalisation coefficient
\[
\left\Vert \overrightarrow{\textrm{dog}} \right\Vert \cdot \left\Vert \overrightarrow{\textrm{snake}} \right\Vert
= \sqrt{2^2 + 2^2 + 1^2} \cdot \sqrt{0^2 + 2^2 + 2^2} = 3\sqrt{8}
\]
to obtain the cosine measure
\[
cosine(\overrightarrow{\textrm{dog}},\overrightarrow{\textrm{snake}}) = \frac{6}{3\sqrt{8}} \approx 0.71
\]
Interpreting the cosine measure as a percentage or degree of conceptual similarity, we see that `dog' is more than 10\% closer in meaning to `cat' than it is to `snake'.

Readers interested in learning more about these aspects of distributional lexical semantics are invited to consult \cite{Curran2004}, which contains an extensive overview of implementation options for distributional models of word meaning.

% section distributional_semantics (end)

\section{Compositionality and Vector Space Models} % (fold)
\label{sec:compositionality_and_vector_space_models}

In the above overview of distributional models of lexical semantics, we have seen that distributional semantic models (DSMs) are a rich and innovative way of learning word meaning from a corpus, and obtaining a measure of semantic similarity for words. However, it should be fairly obvious that the same method cannot be applied to sentences, whereby the meaning of a sentence would be given by the distribution of other sentences with which it occurs. 

First and foremost, a sentence typically occurs only once in a corpus, and hence substantial and informative distributions cannot be created in this manner. More importantly, human ability to understand new sentences is a compositional mechanism: we understand sentences we have never seen before because we can generate sentence meaning from the words used, and how they are \emph{put into relation}. To go from word vectors to sentence vectors, we must provide a \emph{composition operation} allowing us to construct a sentence vector from a collection of word vectors. In this section, I will discuss several approaches to solving this problem, their advantages, and their limitations.

\subsection{Additive Models} % (fold)
\label{sub:additive_models}

The simplest composition operation that comes to mind is straightforward vector addition, such that: 
\begin{displaymath}
    \overrightarrow{ab} = \overrightarrow{a} + \overrightarrow{b}
\end{displaymath}
Conceptually speaking, if we view word vectors as semantic information distributed across a set of properties associated with basis vectors, using vector addition as a semantic composition operation states that the semantic information of a set of lemmas in a sentence is simply the sum of the semantic information of the individual lemmas. While crude, this approach is computationally cheap, and appears sufficient for certain NLP tasks: \cite{Landauer1997} shows it to be sufficient for automated essay marking tasks, and \cite{MScGrefenstette2009} shows it to perform better than a collection of other simple similarity metrics for summarisation, sentence paraphrase, and document paraphrase detection tasks.

However there are two principal objections to additive models of composition: first, vector addition is commutative, therefore $\overrightarrow{\textrm{John drank wine}} = \overrightarrow{\textrm{John}} + \overrightarrow{\textrm{drank}} + \overrightarrow{\textrm{wine}} = \overrightarrow{\textrm{Wine drank John}}$, and thus vector addition ignores syntactic structure completely; and second, vector addition sums the information contained in the vectors, effectively jumbling the meaning of words together as sentence length grows.

The first objection is problematic, as the syntactic insensitivity of additive models leads them to equate the representations of sentences with patently different meanings. \cite{mitchell2008vector} propose to add some degree of syntactic sensitivity---namely accounting for word order---by weighting word vectors according to their order of appearance in a sentence as follows:
\begin{displaymath}
    \overrightarrow{ab} = \alpha \overrightarrow{a} + \beta \overrightarrow{b}
\end{displaymath}
where $\alpha,\, \beta \in \mathbb{R}$. Consequently $\overrightarrow{\textrm{John drank wine}} = \alpha \cdot \overrightarrow{\textrm{John}} + \beta \cdot \overrightarrow{\textrm{drank}} + \gamma \cdot \overrightarrow{\textrm{wine}}$ would not have the same representation as $\overrightarrow{\textrm{Wine drank John}} = \alpha \cdot \overrightarrow{\textrm{wine}} + \beta \cdot \overrightarrow{\textrm{drank}} + \gamma \cdot \overrightarrow{\textrm{John}}$. 

The question of how to obtain weights and whether they are only used to reflect word order or can be extended to cover more subtle syntactic information is open, but it is not immediately clear how such weights may be obtained empirically and whether this mode of composition scales well with sentence length and increase in syntactic complexity. \cite{Guevara2010} suggests using machine-learning methods such as partial least squares regression to determine the weights empirically, but states that this approach enjoys little success beyond minor composition such as adjective-noun or noun-verb composition, and that there is a dearth of metrics by which to evaluate such machine learning-based systems, stunting their growth and development at the time of writing.

The second objection states that vector addition leads to increase in ambiguity as we construct sentences, rather than decrease in ambiguity as we would expect from giving words a context. For this reason, \cite{mitchell2008vector} suggest replacing additive models with multiplicative models as discussed in the next section, or combining them with multiplicative models to form mixture models as discussed in $\S$\ref{sub:mixture_models}.
% subsection additive_models (end)

\subsection{Multiplicative Models} % (fold)
\label{sub:multiplicative_models}

The multiplicative model of \cite{mitchell2008vector} is an attempt to solve the ambiguity problem, discussed in the previous section, and provide implicit disambiguation during composition. The composition operation proposed is the component-wise multiplication ($\odot$) of two vectors. Vectors are expressed as the weighted superposition of their basis vectors, and the weights of the basis vectors of the composed vector is the product of the weights of the original vectors; for $\overrightarrow{a} = \sum_i{c_i \overrightarrow{n_i}}$, and $\overrightarrow{b} = \sum_i{c'_i\, \overrightarrow{n_i}}$, we have 
\begin{displaymath}
    \overrightarrow{ab} = \overrightarrow{a} \odot \overrightarrow{b} = \sum_i{c_i c'_i\, \overrightarrow{n_i}}
\end{displaymath}
Such multiplicative models are shown by \cite{mitchell2008vector} to perform better at verb disambiguation tasks than additive models for noun-verb composition, against a baseline set by the original verb vectors. The experiment that they use to demonstrate this improvement will also serve to evaluate our own models, and form the basis for further experiments, as discussed below in Chapter~\ref{cha:evaluating_a_discocat}.

This approach to compositionality still suffers from two conceptual problems: first, component-wise multiplication remains commutative and hence word order is not accounted for; second, rather than `diluting' information during large compositions and creating ambiguity, it may remove too much information through the `filtering' effect of component-wise multiplication.

The first problem is more difficult to deal with for multiplicative models than for additive models, since both scalar multiplication and component-wise multiplication are commutative linear operations and hence $\alpha \overrightarrow{a} \odot \beta \overrightarrow{b} = \alpha \overrightarrow{b} \odot \beta \overrightarrow{a}$ and thus word order cannot be taken into account using scalar weights.

To illustrate how the second problem entails that multiplicative models do not scale well with sentence length, let us look at the structure of component-wise multiplication again: $\overrightarrow{a} \odot \overrightarrow{b} = \sum_i{c_i c'_i\, \overrightarrow{n_i}}$. For any $i$, if $c_i = 0$ or $c'_i = 0$, then $c_i c'_i = 0$, and therefore for any composition, the number of non-zero basis weights of the produced vector is less than or equal to the number of non-zero basis weights of the original vectors: at each composition step information is filtered out (or preserved, but never increased). Hence as the number of vectors to be composed grows, the number of non-zero basis weights of the product vector stays the same or---more realistically---decreases. Therefore for any composition of the form $\overrightarrow{a_1 \ldots a_i \ldots a_n} = \overrightarrow{a_1} \odot \ldots \odot \overrightarrow{a_i} \odot \ldots \odot \overrightarrow{a_n}$, if there exist two subsets of the set of vectors involved such that the component-wise multiplication of the vectors in one subset forms a vector orthogonal to that formed from the component-wise multiplication of the vectors in the other, then $\overrightarrow{a_1 \ldots a_i \ldots a_n} = \overrightarrow{0}$. It follows that purely multiplicative models alone are not apt as a single mode of composition beyond binary composition operations.

One solution to this second problem not discussed by \cite{mitchell2008vector} would be to introduce some smoothing factor $s \in \mathbb{R^{+}}$ for point-wise multiplication such that $\overrightarrow{a} \odot \overrightarrow{b} = \sum_i{(c_i + s) (c'_i + s) \overrightarrow{n_i}}$, ensuring that information is never completely filtered out. Seeing how the problem of syntactic insensitivity still stands in the way of full-blown compositionality for multiplicative models, I leave it to those interested in salvaging purely multiplicative models to determine whether some suitable value of $s$ can be determined.
% subsection multiplicative_models (end)

\subsection{Mixture Models} % (fold)
\label{sub:mixture_models}

The problems faced by multiplicative models presented in $\S$\ref{sub:multiplicative_models} are acknowledged in passing by \cite{mitchell2008vector}, who propose mixing additive and multiplicative models in the hope of leveraging the advantage of each while doing away with their pitfalls. This is simply expressed as the weighted sum of additive and multiplicative models:
\begin{displaymath}
    \overrightarrow{ab} = \alpha \overrightarrow{a} + \beta \overrightarrow{b} + \gamma (\overrightarrow{a} \odot \overrightarrow{b})
\end{displaymath}
where $\alpha$, $\beta$ and $\gamma$ are pre-determined scalar weights.

The problems for these models are threefold. First, the question of how scalar weights are to be obtained still needs to be addressed. Mitchell and Lapata \cite{mitchell2008vector} concede that one advantage of purely multiplicative models over weighted additive or mixture models is that the lack of scalar weights removes the need to optimise the scalar weights for particular tasks (at the cost of not accounting for syntactic structure), and avoids the methodological concerns accompanying this requirement.

Second, the question of how well this process scales from noun-verb composition to more syntactically rich expressions must be addressed. Using scalar weights to account for word order seems ad-hoc and superficial, as there is more to syntactic structure than the mere ordering of words. Therefore an account of how to build sentences vectors for sentences such as ``The dog bit the man'' and ``The man was bitten by the dog'' in order to give both sentences the same (or a similar) representation would need to give a richer role to scalar weights than just token order. Perhaps specific weights could be given to particular syntactic classes (such as nouns) to introduce a more complex syntactic element into vector composition; but it is clear that this alone is not a solution, as the weight for nouns ``dog'' and ``man'' would be the same, allowing for the same commutative degeneracy observed in non-weighted additive models, in which $\overrightarrow{\textrm{the dog bit the man}} = \overrightarrow{\textrm{the man bit the dog}}$. Introducing a mixture of weighting systems accounting for both word order and syntactic roles may be a solution; however, it is not only ad-hoc but also arguably only partially reflects the syntactic structure of the sentence.

The third problem is that \cite{mitchell2008vector} show that in practice, while mixture models perform better at verb disambiguation tasks than additive models and weighted additive models, they perform equivalently to purely multiplicative models with the added burden of requiring parametric optimisation of the scalar weights.

Therefore while mixture models aim to take the best of additive and multiplicative models while avoiding their problems, they are only partly successful in achieving the latter goal, and demonstrably not much better in achieving the former.
% subsection mixture_models (end)

\subsection{Tensor-Based Models} % (fold)
\label{sub:tensor_models}

From $\S\S$\ref{sub:additive_models}--\ref{sub:mixture_models} we observe that the need for incorporating syntactic information into DSMs to achieve true compositionality is pressing, if only to develop a non-commutative composition operation that can take into account word order without the need for ad-hoc weighting schemes, and hopefully to integrate richer syntactic information as well.

An early proposal by Smolensky \cite{smolensky1990tensor,Smolensky2006} to use linear algebraic tensors as a composition operation solves the problem of finding non-commutative vector composition operators. The composition of two vectors is their tensor product, sometimes called the Kronecker product when applied to vectors rather than vector spaces: for $\overrightarrow{a}\in V = \sum_i{c_i \overrightarrow{n_i}}$, and $\overrightarrow{b} \in W = \sum_j{c'_j \overrightarrow{n'_j}}$, we have:
\begin{displaymath}
    \overrightarrow{ab} = \overrightarrow{a} \otimes \overrightarrow{b} = \sum_{ij}{c_ic'_j\, \overrightarrow{n_i} \otimes \overrightarrow{n'_j}}
\end{displaymath}
To illustrate with an example consider the following two vectors:
\[
\overrightarrow{a} = [a_1\ a_2]^{\top} \qquad \overrightarrow{b} = [b_1\ b_2]^{\top}
\]
Their Kronecker product is as follows:
\[
\overrightarrow{a} \otimes \overrightarrow{b}
=
\left[
\begin{tabular}{cc}
$a_1 b_1$ & $a_1 b_2$\\
$a_2 b_1$ & $a_2 b_2$
\end{tabular}
\right]
\]

The composition operation takes the original vectors and maps them to a vector in a larger vector space $V \otimes W$ which is the tensor space of the original vectors' spaces. Here the second instance of $\otimes$ is not a recursive application of the Kronecker product, but rather the pairing of basis elements of $V$ and $W$ to form a basis element of $V \otimes W$. The shared notation and occasional conflation of Kronecker and tensor products may seem confusing, but is fairly standard in multi-linear algebra.

The advantage of this approach is twofold. First, vectors for different words need not live in the same spaces but can be composed nonetheless. This allows us to represent vectors for different word classes (e.g.~topics, syntactic roles, etc.) in different spaces with different bases, which was not possible under additive or multiplicative models. Second, because the product vector lives in a larger space, we obtain the intuitive notion that the information of the whole is richer and more complex than the mere sum or product of the information of the parts.

\subsubsection{Dimensionality Problems} 

However this increase in dimensionality brings two rather large problems for tensor based models. The first is computational: the size of the product vector space is the product of the size of the original vector spaces. If we assume that all words live in the same space $N$ of dimensionality $dim(N)$ then the dimensionality of an $n$-word sentence vector is $dim(N)^n$. If we have as many basis vectors for our word semantic space as there are lexemes in our vocabulary---e.g.~approximately 170k in English\footnote{Source: \texttt{http://www.oxforddictionaries.com/page/howmanywords}}---then the size of our sentence vectors quickly reaches magnitudes for which vector comparison (or even storage) are computationally intractable\footnote{At four bytes per integer, and one integer per basis vector weight, the vector for ``John loves Mary'' would require roughly $(170000 \cdot 4)^3 \approx 280\ \textrm{petabytes}$ of storage, which is over ten times the data Google processes on a daily basis according to \cite{Dean2008a}.}. Even if, as most DSM implementations do, we restrict the basis vectors of word semantic spaces to the $k$ (e.g.~$k=2000$) most frequent words in a corpus, the sentence vector size still grows exponentially with sentence length, and the implementation problems remain.

The second problem is mathematical: sentences of different length live in different vector spaces, and if we assign different vector spaces to different word types (e.g.~syntactic classes), then sentences of different syntactic structure live in different vector spaces. As a result, they cannot be compared directly using inner product or cosine measure, leaving us with no obvious mode of semantic comparison for sentence vectors. If any model wishes to use tensor products in composition operations, it must find some way of reducing the dimensionality of product vectors to some common vector space so that they may be directly compared.

One notable method by which these dimensionality problems can be solved in general are the holographic reduced representations proposed by \cite{plate1991holographic}. The product vector of two vectors is projected into a space of smaller dimensionality by circular convolution to produce a trace vector. The circular correlation of the trace and one of the original vectors produces a noisy version of the other original vector. The noisy vector can be used to recover the clean original vector by comparing it with a pre-defined set of candidates (for example the set of word vectors if our original vectors are word meanings). Traces can be summed to form new traces effectively containing several vector pairs from which original vectors can be recovered. Using this encoding/decoding mechanism, the tensor product of sets of vectors can be encoded in a space of smaller dimensionality, and then recovered for computation without ever having to fully represent or store the full tensor product, as discussed by \cite{widdows2008semantic}.

There are problems with this approach that make it unsuitable for our purposes. First, there is a limit to the information that can be stored in traces, which is independent of the size of the vectors stored, but is a logarithmic function of their number. As we wish to be able to store information for sentences of variable word length without having to directly represent the tensored sentence vector, setting an upper bound to the number of vectors that can be composed in this manner limits the length of the sentences we can represent compositionally using this method.

Second, and perhaps more importantly, there are restrictions on the nature of the vectors that can be encoded in such a way: the vectors must be independently distributed such that the mean euclidean length of each vector is equal to one. Such conditions are unlikely to be met in word semantic vectors obtained from a corpus, and as the failure to do so affects the system's ability to recover clean vectors, holographic reduced representations are not \emph{prima facie} useable for compositional DSMs. However, it is important to note that \cite{widdows2008semantic} considers possible linguistic application areas where they may be of use, although once again these mostly involve noun-verb and adjective-noun compositionality rather than full-blown sentence vector construction. We retain from \cite{plate1991holographic} the importance of finding methods by which to project the tensored sentence vectors into a common space for direct comparison, as will be discussed further in  Chapter~\ref{cha:foundations_of_discocat}.

\subsubsection{Syntactic Expressivity} % (fold)

An additional problem of a more conceptual nature is that using the tensor product as a composition operation simply preserves word order. As I discussed in $\S$\ref{sub:mixture_models}, this is not enough on its own to model sentence meaning. We need to have some means by which to incorporate syntactic analysis into composition operations. 

Early work on including syntactic sensitivity into DSMs by \cite{Grefenstette1992} suggests using syntactic dependency relations to determine the frame in which the distributions for word vectors are collected from the corpus, thereby embedding syntactic information into the word vectors. This idea is built upon by \cite{Clark2006} who suggest incorporating dependency relations into tensor-based composition operations as vectors themselves. For example in the sentence ``Simon loves red wine'', ``Simon'' is the subject of ``loves'', ``wine'' is its object, and ``red'' is an adjective describing ``wine''. Hence from the dependency tree with ``loves'' as root node, its subject and object as children, and their adjectival descriptors (if any) as their children, we read the following structure: $\overrightarrow{\textrm{loves}} \otimes \overrightarrow{subj} \otimes \overrightarrow{\textrm{Simon}} \otimes \overrightarrow{obj} \otimes \overrightarrow{\textrm{wine}} \otimes \overrightarrow{adj} \otimes \overrightarrow{\textrm{red}}$. Using the equality relation for inner products of tensor products:
\begin{displaymath}
    \langle \overrightarrow{a} \otimes \overrightarrow{b} \mid \overrightarrow{c} \otimes \overrightarrow{d} \rangle = \langle \overrightarrow{a} \mid \overrightarrow{c} \rangle \times \langle \overrightarrow{b} \mid \overrightarrow{d} \rangle
\end{displaymath}
we can therefore express inner-products of sentence vectors efficiently without ever having to actually represent the tensored sentence vector: 
{\scriptsize \begin{align*}
    & \langle \overrightarrow{\textrm{Simon loves red wine}} \mid \overrightarrow{\textrm{Mary likes delicious ros\'e}} \rangle \\
    & = \langle \overrightarrow{\textrm{loves}} \otimes \overrightarrow{subj} \otimes \overrightarrow{\textrm{Simon}} \otimes \overrightarrow{obj} \otimes \overrightarrow{\textrm{wine}} \otimes \overrightarrow{adj} \otimes \overrightarrow{\textrm{red}} \mid \overrightarrow{\textrm{likes}} \otimes \overrightarrow{subj} \otimes \overrightarrow{\textrm{Mary}} \otimes \overrightarrow{obj} \otimes \overrightarrow{\textrm{ros\'e}} \otimes \overrightarrow{adj} \otimes \overrightarrow{\textrm{delicious}} \rangle\\
    & = \langle \overrightarrow{\textrm{loves}} \mid \overrightarrow{\textrm{likes}} \rangle \times \langle \overrightarrow{subj} \mid \overrightarrow{subj} \rangle \times \langle \overrightarrow{\textrm{Simon}} \mid \overrightarrow{\textrm{Mary}} \rangle \times \langle \overrightarrow{obj} \mid \overrightarrow{obj} \rangle \times \langle \overrightarrow{\textrm{wine}} \mid \overrightarrow{\textrm{ros\'e}} \rangle \times \langle \overrightarrow{adj} \mid \overrightarrow{adj} \rangle \times \langle \overrightarrow{\textrm{red}} \mid \overrightarrow{\textrm{delicious}} \rangle\\
    & = \langle \overrightarrow{\textrm{loves}} \mid \overrightarrow{\textrm{likes}} \rangle \times \langle \overrightarrow{\textrm{Simon}} \mid \overrightarrow{\textrm{Mary}} \rangle  \times \langle \overrightarrow{\textrm{wine}} \mid \overrightarrow{\textrm{ros\'e}} \rangle \times \langle \overrightarrow{\textrm{red}} \mid \overrightarrow{\textrm{delicious}} \rangle
\end{align*}}
This example shows that this formalism allows for the comparison of sentences with identical dependency trees to be broken down to term-to-term comparison without the need for the tensor products to ever be computed or stored, reducing computation to inner product calculations. 

However while matching up terms with identical syntactic roles in the sentence works well in the above example, this model suffers from the same problems as the original tensor-based compositionality of \cite{smolensky1990tensor} in that, by the authors' own admission, sentences of different syntactic structure live in spaces of different dimensionality and thus cannot be directly compared. Hence we cannot use this to measure the similarity between even small variations in sentence structure, such as the pair ``Simon likes red wine'' and ``Simon likes wine'', which are sentences of different length and grammatical structure, and therefore live in different vector spaces under this approach.
% subsection tensor_models (end)

\subsection{SVS Models} % (fold)
\label{sub:svs_models}

The idea of including syntactic relations to other lemmas in word representations discussed in $\S$\ref{sub:tensor_models}, above, is applied differently in the structured vector space model presented by \cite{Erk2008}. They propose to represent word meanings not as simple vectors, but as triplets:
\begin{displaymath}
    w = (v,R,R^{-1})
\end{displaymath}
where $v$ is the word vector, constructed as in any other DSM, $R$ and $R^{-1}$ are selectional preferences, and take the form of $\mathcal{R} \to \mathcal{D}$ maps where $\mathcal{R}$ is the set of dependency relations and $\mathcal{D}$ is the set of word vectors. Selectional preferences are used to encode the lemmas that $w$ is typically the parent of in the dependency trees of the corpus in the case of $R$, and typically the child of in the case of $R^{-1}$.

Composition takes the form of vector updates according to the following protocol. Let $a = (v_a,R_a,R^{-1}_a)$ and $b = (v_b,R_b,R^{-1}_b)$ be two words being composed, and let $r$ be the dependency relation linking $a$ to $b$. The vector update procedure is as follows:
\begin{align*}
    & a' = (v_a \odot R^{-1}_b(r), R_a - \{r\}, R^{-1}_a)\\
    & b' = (v_b \odot R_a(r), R_b, R^{-1}_b - \{r\})
\end{align*}
where $a',b'$ are the updated word meanings, and $\odot$ is whichever vector composition (addition, component-wise multiplication) we wish to use. The word vectors in the triplets are effectively filtered by combination with the lemma which the word they are being composed with expects to bear relation $r$ to, and this relation between the composed words $a$ and $b$ is considered to be used and hence removed from the domain of the selectional preference functions used in composition. 

This mechanism is therefore a more sophisticated version of the compositional disambiguation mechanism discussed by \cite{mitchell2008vector} in that the combination of words filters the meaning of the original vectors which may be ambiguous (e.g.~if we have one vector for all senses of the word ``bank''); however, contrary to \cite{mitchell2008vector} the information of the original vectors is modified but essentially preserved, allowing for further combination with other terms, rather than directly producing a joint vector for the composed words. The added fact that $R$ and $R^{-1}$ are partial functions associated with specific lemmas forces grammaticality during composition, since if $a$ holds a dependency relation $r$ to $b$ which it never expects to hold (for example a verb having as subject another verb, rather than the reverse) then $R_a$ and $R^{-1}_b$ are undefined for $r$ and the update fails. However, there are some problems with this approach if our goal is true compositionality.

First, this model does not allow some of the `repeated compositionality' we need because of the update of $R$ and $R^{-1}$. For example, we expect that an adjective composed with a noun produces something \emph{like} a noun in order to be further composed with a verb or even another adjective. However here, because the relation \emph{adj} would be removed from $R^{-1}_b$ for some noun $b$ composed with an adjective $a$, this new representation $b'$ would not have the properties of a noun in that it would no longer expect composition with an adjective, rendering representations of simple expressions like ``the new red car'' impossible. Of course, we could remove the update of the selectional preference functions from the compositional mechanism, but then we would lose this attractive feature of grammaticality enforcement through the partial functionality of $R$ and $R^{-1}$.

Second, this model does little more than represent the implicit disambiguation which is expected during composition, rather than actually provide a full blown compositional model. The inability of this system to provide a novel mechanism by which to obtain a joint vector for two composed lemmas---thereby building towards sentence vectors---entails that this system provides no means by which to obtain semantic representations of larger syntactic structures which can be compared by inner product or cosine measure as is done with any other DSM. Of course, this model could be combined with the compositional models presented in $\S\S$\ref{sub:additive_models}--\ref{sub:mixture_models} to produce sentence vectors, but while some syntactic sensitivity would have been obtained, the word ordering and other problems of the aforementioned models would still remain, and little progress would have been made towards true compositionality.

Finally, \cite{Erk2008} state that the selectional preferences for each word must be pre-defined, but provide no procedure by which we would learn these from a corpus. This is an open problem for this model rather than a conceptual objection, but it remains to be seen if such preferences can be empirically determined without affecting the performance or tractability of such a model.

We retain from this attempt to introduce compositionality in DSMs that including information obtained from syntactic dependency relations is important for proper disambiguation. We also note that having some mechanism by which the grammaticality of the expression being composed is a pre-condition for its composition is a desirable feature for any compositional mechanism.
% subsection svs_models (end)

% section compositionality_and_vector_space_models (end)

\section{Matrix-Based Compositionality} % (fold)
\label{sec:matrix_based_composition}

The final class of approaches to vector composition I wish to discuss are three matrix based models.

\subsection{Generic Additive Model}

The first is the Generic Additive Model of \cite{zanzotto2010estimating}. This is a generalisation of the weighted additive model presented in $\S$\ref{sub:additive_models}. In this model, lexical vectors are not just multiplied by fixed parameters $\alpha$ and $\beta$ before adding them to form the representation of their combination. Instead, they are the arguments of matrix multiplication by square matrices $A$ and $B$ as follows:
\begin{displaymath}
    \overrightarrow{ab} = A \overrightarrow{a} + B \overrightarrow{b}
\end{displaymath}
Here $A$ and $B$ represent the added information provided by putting two words into relation. 

The numerical content of $A$ and $B$ is learned by linear regression over triplets $(\overrightarrow{a},\overrightarrow{b},\overrightarrow{c})$ where $\overrightarrow{a}$ and $\overrightarrow{b}$ are lexical semantic vectors, and $\overrightarrow{c}$ is the \emph{expected} output of the combination of $\overrightarrow{a}$ and $\overrightarrow{b}$. This learning system thereby requires the provision of labelled data for linear regression to be performed. Zanzotto et al.~suggest several sources for this labelled data, such as dictionary definitions and word etymologies.

This approach is richer than the weighted additive models, since the matrices act as linear maps on the vectors they take as `arguments', and thus can encode more subtle syntactic or semantic relations. However, this model treats all word combinations as the same operation, e.g.~treating the combination of an adjective with its argument and a verb with its subject as the same sort of composition. This may not be fundamentally wrong, but it would be worthwhile evaluating how well this system works in the context of experiments such as those presented in Chapter~\ref{cha:evaluating_a_discocat}. But because of the diverse ways there are of training such supervised models, we leave it to those who wish to further develop this specific line of research to perform such evaluations.

\subsection{Adjective Matrices}
\label{sub:adjective_matrices}

The second approach is the matrix-composition model of \cite{Baroni2010}, which they develop only for the case of adjective-noun composition, although their approach can seamlessly be used for any other predicate-argument composition. Contrary to most of the approaches above, which aim to combine two lexical vectors to form a lexical vector for their combination, Baroni and Zamparelli suggest giving different semantic representations to different types, or more specifically to adjectives and nouns. 

In this model, nouns are lexical vectors, as with other models. However, embracing a view of adjectives that is more in line with formal semantics than with distributional semantics, they model adjectives as linear maps taking lexical vectors as input and producing lexical vectors as output. Such linear maps can be encoded as square matrices, and applied to their arguments by matrix multiplication. Concretely, let $M^{adjective}$ be the matrix encoding the adjective's linear map, and $\overrightarrow{noun}$ be the lexical semantic vector for a noun; their combination is simply
\[
\overrightarrow{adjective\ noun} = M^{adjective} \times \overrightarrow{noun}
\]

Similarly to the Generic Additive Model described above, the matrix for each adjective is learned by linear regression over a set of pairs $(\overrightarrow{noun},\overrightarrow{c})$ where the vectors $\overrightarrow{noun}$ are the lexical semantic vectors for the arguments of the adjective in a corpus, and $\overrightarrow{c}$ is the semantic vector corresponding to the expected output of the composition of the adjective with that noun.

This may, at first blush, also appear to be a supervised training method for learning adjective matrices from `labelled data', seeing how the expected output vectors are needed. However, Baroni and Zamparelli work around this constraint by automatically producing the labelled data from the corpus by treating the adjective-noun compound as a single token, and learning its vector using the same distributional learning methods they used to learn the vectors for nouns.

This same approach can be extended to other unary relations without change. It bears some similarity to some of the work presented in this document, and a direct comparison of our frameworks would be interesting. However, this requires a method of extending this approach to binary predicates for full comparison to be made. Furthermore, this approach has a larger set of free parameters, such as the method of dimensionality reduction used, the linear regression learning algorithm and its parameters, and so on. While this entails that this class of models can certainly be tweaked to improve results, the difficult task of finding optimal parameters for such a framework is outside of the scope of this thesis, and we will not attempt in our evaluations of Chapter~\ref{cha:evaluating_a_discocat} an unfair direct comparison by using substandard parameters.

\subsection{Recursive Matrix-Vector Model} 
The third approach is the recently-developed Recursive Matrix-Vector Model (MV-RNN) of \cite{socherEMNLP12}\footnote{Other similar related deep-learning approaches are also worth investigating, such as that of \cite{hermann2013role}.}, which claims the two matrix-based models described above as special cases. In MV-RNN, words are represented as a pairing of a lexical semantic vector $\overrightarrow{a}$ with an operation matrix $A$. Within this model, given the parse of a sentence in the form of a binarised tree, the semantic representation $(\overrightarrow{c},C)$ of each branch node in the tree is produced by performing the following two operations on its children $(\overrightarrow{a},A)$ and $(\overrightarrow{b},B)$. 

First, the vector component $\overrightarrow{c}$ is produced by applying the operation matrix of one child to the vector of the other, and vice versa, and then projecting the concatenation of both of the products back into the same vector space as the child vectors using a projection matrix $W$, which must also be learned:
\[
\overrightarrow{c} = g
\left(
W \times
\left[
\begin{tabular}{c}
$B \times \overrightarrow{a}$\\
$A \times \overrightarrow{b}$
\end{tabular}
\right]
\right)
\]
Here, $g$ is a component-wise activation function to be determined by the model implementation (e.g.~a sigmoid function), used to introduce an element of non-linearity to the compositional process.

Second, the matrix $C$ is calculated by projecting the pairing of matrices $A$ and $B$ back into the same space, using a projection matrix $W_M$, which must also be learned:
\[
C = W_M \times
\left[
\begin{tabular}{c}
$A$\\
$B$
\end{tabular}
\right]
\]
The pairing $(\overrightarrow{c},C)$ obtained through these operations forms the semantic representation of the phrase falling under the scope of the segment of the parse tree below that node.

This approach to compositionality yields good results in the experiments described in \cite{socherEMNLP12}. It furthermore has appealing characteristics such as treating relational words differently through their operation matrices, and allowing for recursive composition, as the output of each composition operation is of the same type of object as its inputs. However, the learning procedure for the projection matrices and the operation matrices for each word, described in \cite{socherEMNLP12}, is non-trivial. It relies on a supervised learning algorithm, and thus is open to the same problems of data availability and sparsity as any other supervised learning system. This also makes it difficult to produce a version of the model to run the experiments described in Chapter~\ref{cha:evaluating_a_discocat}, although working towards joint evaluation of the models described in this section with the MV-RNN approach will certainly need to be the object of further work.

The other significant difference with the compositional framework presented in this thesis is that composition in MV-RNN is always a binary operation, e.g.~to compose a transitive verb with its subject and object one would first need to compose it with its object, and then compose the output of that operation with the subject. The framework discussed in this thesis allows for the construction of representations for relations of larger arities, permitting the simultaneous composition of a verb with its subject and object. Whether or not this theoretical difference leads to significant differences in composition quality requires joint evaluation which, once again, I will leave to further work.
% section matrix_based_composition (end)

% chapter literature_review (end)
\mbox{}
\newpage
%!TEX root = ../grefenstettethesis.tex

\chapter{Foundations of DisCoCat} % (fold)
\label{cha:foundations_of_discocat}

\begin{chabstract}
  This chapter describes the DisCoCat framework of \cite{Coecke2010}, which forms the basis for most of the work presented in this thesis. It introduces the concepts of pregroup grammars, categories, diagrammatic calculi for categories, and of how information may be shared between mathematical formalisms through the medium of categories.
\end{chabstract}

In Chapter~\ref{cha:literature_review}, I discussed distributional semantic models (DSMs) and attempts to provide a vector composition operation over word meanings to form distributional sentence representations. In this chapter, I will present an existing formalism aiming to solve this compositionality problem, as well as the mathematical background required to understand it and further extensions, building on the features and failures of previously discussed attempts at syntactically-sensitive compositionality.

\cite{Clark2008,Coecke2010} propose to adapt a category theoretic model initially used to describe information flow in quantum information theory to the task of composing of semantic vectors. Syntactic analysis in the form of pregroup grammars---a type of categorial grammar---is given categorical semantics in order to be represented as a compact closed category $P$ (a concept explained below), the objects of which are syntactic types and the morphisms of which are the reductions forming the basis of syntactic analysis. Vectors for words reside in vector spaces containing semantic vectors for lemmas of a particular syntactic type, and the set of vector spaces is represented as a compact closed  category $\mathbf{FVect}$ with vector spaces as objects and linear maps as morphisms. 

The key feature of category theory exploited here is the ability to express different mathematical formalisms as structures which can be related, even if the original formalisms belong in different branches of mathematics. Hence the product category $P \times \mathbf{FVect}$ allows us to relate syntactic types to vector spaces and syntactic reductions to linear maps so that we obtain a mechanism by which \emph{syntactic analysis guides semantic composition operations}.

This pairing of syntactic analysis and semantic composition ensures both that grammaticality restrictions are in place as in the model of \cite{Erk2008}, and that syntactically-driven semantic composition in the form of inner-products provides the implicit disambiguation features as in the compositional models of \cite{Erk2008} and \cite{mitchell2008vector}. The composition mechanism also involves projection of tensored vectors into a common semantic space without the need for computing the full representation of the tensored vectors (in a manner similar to \cite{plate1991holographic}), but without the added restriction as to the nature of the vector spaces it can be applied to. This avoids the complexity and comparison problems faced by other tensor-based composition mechanisms such as those of \cite{smolensky1990tensor} and \cite{Clark2006}.

The word vectors can be specified model-theoretically and the sentence space can be defined over boolean values to obtain grammatically-driven truth-theoretic semantics in the style of \cite{Montague1974}, as proposed by  \cite{Clark2008}. Some logical operators can be emulated in this setting, such as using swap matrices for negation as shown by \cite{Coecke2010}. Alternatively, corpus-based variations on this formalism have been proposed by \cite{Grefenstette2010} to obtain a non-truth theoretic semantic model of sentence meaning for which logical operations have yet to be defined.

Before explaining how this formalism works, in $\S$\ref{sec:a_categorical_passage_from_grammar_to_semantics}, I will introduce pregroup grammars in $\S$\ref{sec:pregroup_grammars}, and the required basics of category theory in $\S$\ref{sec:categories}.

\section{Pregroup Grammars} % (fold)
\label{sec:pregroup_grammars}

Presented by Lambek in \cite{Lambek1999,Lambek2008} as a successor to his non-commutative type-logical calculus presented in \cite{Lambek1958}, pregroup grammars are a class of categorial grammars with pregroup semantics. They comprise atomic grammatical types which can combine to form compound types. A series of application rules allow for type-reductions, forming the basis of syntactic analysis. The pregroup semantics of this syntactic formalism are what interest us, as will be discussed in  $\S$\ref{sec:a_categorical_passage_from_grammar_to_semantics}. However, our first step will be to show how this syntactic analysis formalism works, which will in turn require an introduction to pregroups.

\subsection{Pregroups} % (fold)
\label{sub:pregroups}

A pregroup is an algebraic structure of the form $(P,\leq,\cdot, 1,(-)^l,(-)^r)$. Let us explain these elements individually:
\begin{itemize}
    \item $P$ is simply a set of objects $\{a,b,c,\ldots\}$. 
    \item $\leq$ is a partial ordering relation on $P$.
    \item $\cdot$ is an associative, non-commutative monoid multiplication operator, and can be conceived of as a function $- \cdot - : P \times P \to P$ such that if $a,b \in P$ then $a \cdot b \in P$. Therefore $P$ is closed under this operation.
    \item $1 \in P$ is the unit, satisfying $a \cdot 1 = a = 1 \cdot a$ for all $a \in P$.
    \item $(-)^l$ and $(-)^r$ are the left and right adjoints, and can be conceived of as functions $(-)^l : P \to P$ and $(-)^r : P \to P$ such that for any $a \in P$, $a^l, a^r \in P$. Adjoints are further described by the following axioms:
    \begin{itemize}
        \item Reversal: if $a \leq b$ then $b^l \leq a^l$ (as for $a^r$, $b^r$).
        \item Ordering: $a \cdot a^r \leq 1 \leq a^r \cdot a$ and $a^l \cdot a \leq 1 \leq a \cdot a^l$.
        \item Cancellation: $a^{lr} = a = a^{rl}$.
        \item Equality of identity: $1^r = 1 = 1^l$.
        \item Self-adjoint multiplication: $(a \cdot b)^r = b^r \cdot a^r$.
    \end{itemize}
\end{itemize}

I say that a pregroup is freely generated by some basic set of types $\{a,b,c,\ldots,z\}$ to mean that all elements of the pregroup, such as the adjoints $\{a^l, b^{rr}, \ldots\}$ and complex types $\{a \cdot b, c \cdot d \cdot e, c^r \cdot a \cdot b^l\}$ are formed by applying the adjoint operations $(-)^r$ and $(-)^l$ and the multiplication operation $\cdot$ to elements of the basic set or those thus-generated from it. Notationally, this means that the only alphabet used in complex types is that used to enumerate objects of the basic set.

As a notational simplification I write $ab$ for $a \cdot b$, and if $abcd \leq cd$ I write $abcd \to cd$ and call this a reduction, omitting the identity wherever it might appear. Monoid multiplication is associative, so parentheses may be added or removed for notational clarity without changing the meaning of the expression as long as they are not directly under the scope of an adjoint operator.

An example reduction in pregroup might be:
\begin{displaymath}
    aa^r b c^lc \to b c^lc \to b
\end{displaymath}
I note here that the reduction order is not always unique, as I could have reduced the expression as follows: $aa^r b c^lc \to aa^rb \to b$. As a further notational simplification, if there exists a chain of reductions $a \to \ldots \to b$ we may simply write $a \to b$ (in virtue of the transitivity of partial ordering relations). Hence in our above example, we can express both reduction paths as $aa^r b c^lc \to b$.

\subsection{Pregroups and Syntactic Analysis} % (fold)
\label{sub:pregroups_and_grammar}

Pregroups can be used for grammatical analysis by freely generating the set $P$ of a pregroup from the combination of basic syntactic types $n,s,\ldots$ and defining one type ($s$) to be the sentence type. As in any categorial grammar, words of the lexicon are assigned one or more possible types (corresponding to different syntactic roles) in a pre-defined \emph{type dictionary}, and the grammaticality of an expression is verified by demonstrating the existence of a reduction from the type of the expression to the sentence type $s$.

For example, let us assign to nouns the type $n$, and to transitive verbs the compound type $n^r s n^l$. We can read from the type of a transitive verb that it is something which `expects' a noun on its left, and one on its right, in order to reduce to a sentence. A sample reduction of ``John loves cake'' with `John' and `cake' being nouns of type $n$ and `loves' being a verb of type $n^r s n^l$ is as follows:
\begin{displaymath}
    n (n^r s n^l) n \to (s n^l) n \to s
\end{displaymath}
And thus we see that the transitive verb has combined with the subject to become something that requires an object, which it obtains and then becomes a sentence. The expression reduces to $s$, and hence the expression is grammatical.

Intransitive verbs can be given the type $n^rs$ such that ``John sleeps'' would be analysed in terms of the reduction $n (n^rs) \to s$. Adjectives can be given the type $nn^l$ such that ``red round rubber ball'' would be analysed by $(nn^l)(nn^l)(nn^l)n \to n$. And so on and so forth for other syntactic classes\ldots

Lambek, in \cite{Lambek2008}, presents the details of a slightly more complex pregroup grammar with a richer hierarchy of types than presented here. It is hand-constructed and iteratively extended by expanding the type hierarchy as previous versions of the grammar encounter unparseable expressions. 

\subsection{A graphical calculus for pregroups} % (fold)
\label{sub:a_graphical_calculus_for_pregroups}

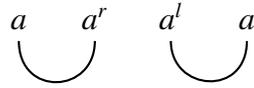
\begin{figure}[!ht]
\begin{center}
    \begin{tikzpicture}[thick]
\node [above] (v1) at (0,0) {$a$};
\node [above] (v2) at (1,0) {$a^r$};
\node [above] (v3) at (2,0) {$a^l$};
\node [above] (v4) at (3,0) {$a$};
\draw (v1) .. controls +(0,-1) and +(0,-1) .. (v2);
\draw (v3) .. controls +(0,-1) and +(0,-1) .. (v4);
\end{tikzpicture}
\end{center}
\caption{Diagrammatic examples of pregroup reductions.}
\label{fig:pregroup_reductions}
\end{figure}

Pregroups can be represented using a simple graphical calculus \cite{sadrzadeh2007high} allowing us to visually exhibit the simultaneous nature of type reductions in an elegant and intuitive manner. Cancellations of the type $aa^r$ or $a^la$ are represented as `cups' as shown in Figure~\ref{fig:pregroup_reductions}. I designate the non-reduction of a type by a single downward line, which can be seen as the `output' of the reduction.

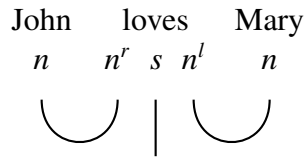
\begin{figure}[!ht]
\begin{center}
    \begin{tikzpicture}[thick, every text node part/.style={align=center}]
\node [above] (v1) at (0,0) {John \\ $n$};
\node [above] (v2) at (1,0) {$n^r$};
\node [above] (v5) at (1.5,0) {loves\\ $s$};
\node (v6) at (1.5,-.75) {};
\node [above] (v3) at (2,0) {$n^l$};
\node [above] (v4) at (3,0) {Mary \\$n$};
\draw (0,-.25) .. controls +(0,-.75) and +(0,-.75) .. (1,-.25);
\draw (2,-.25) .. controls +(0,-.75) and +(0,-.75) .. (3,-.25);
\draw [-] (1.5,-.25) -- (1.5,-1);
\end{tikzpicture}
\end{center}
\caption{Diagrammatic pregroup parse of ``John loves Mary''.}
\label{fig:pregroup_loves}
\end{figure}

Figure~\ref{fig:pregroup_loves} shows the diagrammatic reduction for the pregroup parse of ``John loves Mary'', whereby a noun (``John'') of type $n$ combines with the leftmost adjoint of the compound term for a transitive verb (``loves'') of type $n^rsn^l$, and another noun (``Mary'') combines with the rightmost adjoint of the verb to form a sentence type $s$.

This diagrammatic calculus bears some striking similarities to the diagrammatic calculus for compact closed categories, described in $\S$\ref{sub:a_graphical_calculus_for_categories}. This similarity is no coincidence because of the relation between pregroups and compact closed categories, discussed in $\S$\ref{sub:monoidal_categories}. This diagrammatic similarity will make it easier to visually describe the process of passing from syntactic analysis to semantic interpretation, as discussed in $\S$\ref{sec:a_categorical_passage_from_grammar_to_semantics}.

% subsection a_graphical_calculus_for_pregroups (end)

% section pregroups (end)

\section{Categories} % (fold)
\label{sec:categories}

Category theory is a branch of pure mathematics which allows for the formulation of other mathematical structures and formalisms in terms of objects, arrows, and a few axioms. This simplicity and restricted conceptual language makes category theory both specific and general. It is specific in that the new properties of existing theories can be deduced from categorical axioms. It is general in that properties of these theories can be related to properties of other theories if they bear the same categorical representations. 

It is this ability category theory provides to communicate information both within and across mathematical structures which makes it such a powerful tool. In this function, it has been at the centre of recent work in the foundations of physics and the modelling of quantum information flow, as presented in \cite{abramsky2004categorical}. The connection\footnote{I interpret this connection as one of loose analogy, at best, rather than holding the view that there is some fundamental link between quantum mechanics and language. It just happens that in both quantum mechanics and language, there is a notion of information being communicated, exchanged, or affected by objects with an ``uncertain'' state. In the case of quantum mechanics, this uncertainty takes the form of state superpositions, while in language it takes the form of ambiguity and polysemy. It should therefore come as no surprise that the mathematics developed to deal with one of these domains might be adapted to deal with the other, but I believe that the connection stops there. If anything, the ability to straightforwardly adapt the mathematics of quantum information flow to linguistic information flow exemplifies the advantages provided by the abstractness and generality of category theory.} between the mathematics used for this branch of physics and those potentially useful for linguistic modelling has been noted by several sources, such as \cite{widdowsgeometry,Lambek2010,van2004geometry}.

In this section, in order to demonstrate how these mathematical methods carry over to semantic analysis, I will briefly examine the basics of category theory, monoidal categories, and compact closed categories. The focus will be on defining enough basic concepts to proceed rather than provide a full-blown tutorial on category theory and the modelling of information flow, as several excellent sources already cover both aspects, e.g.~\cite{mac1998categories,walters1991categories,coecke2009categories}. A categories-in-a-nutshell crash course is also provided in \cite{Clark2008,Coecke2010}.

\subsection{The Basics of Category Theory} % (fold)
\label{sub:the_basics}

Let us first consider the simplest definition of a category. A basic category $\mathbf{C}$ is defined in terms of the following elements:
\begin{itemize}
    \item A collection of objects $ob(\mathbf{C})$.
    \item A collection of morphisms $hom(\mathbf{C})$.
    \item A morphism composition operation $\circ$.
\end{itemize}

Each morphism $f$ has a domain $dom(f) \in ob(\mathbf{C})$ and a codomain $codom(f) \in ob(\mathbf{C})$. For $dom(f) = A$ and $codom(f) = B$, I abbreviate these definitions as $f:A \to B$. Despite the notational similarity to function definitions, it is important to state that nothing else is pre-supposed about morphisms, and we should not treat them as functions.

The following axioms hold:
\begin{itemize}
    \item For any $f : A \to B$ and $g : B \to C$ there exists $h : A \to C$ and $h = g \circ f$.
    \item For any $f: A \to B$, $g : B \to C$ and $h : C \to D$, $\circ$ satisfies $(h \circ g) \circ f = h \circ (g \circ f)$.
    \item For every $A \in ob(\mathbf{C})$ there is an identity morphism $id_A : A \to A$ such that for any $f : A \to B$, $f \circ id_A = A = id_B \circ f$.
\end{itemize}

We can express various mathematical formalisms using such basic categories, and verify that these axioms hold. For example there is a category of sets with sets as objects and functions as morphisms, a category of posets with posets as objects and order-preserving maps as morphisms, and a category of groups with groups as objects and group homomorphisms as morphisms, to name a few.

A product category $\mathbf{C} \times \mathbf{D}$ of two categories $\mathbf{C}$ and $\mathbf{D}$ is a category with pairs $(A,B)$ as objects, where $A \in ob(\mathbf{C})$ and $B \in ob(\mathbf{D})$. There exists a morphism $(f,g) : (A,B) \to (C,D)$ in $\mathbf{C} \times \mathbf{D}$ if and only if there exists $f:A \to C \in hom(\mathbf{C})$ and $g: B \to D \in hom(\mathbf{D})$. Product categories are useful in attaining this desired generality of category theory, in that they allow us to relate objects and operations (morphisms) in one mathematical formalism or structure to those in another. However, this method of relating structures is not ideal, as will be discussed in Chapter~\ref{cha:syntactic_extensions}, where a more elegant alternative will be provided (namely functors). For the time being, though, I will use product categories for the sake homogeneity with the work of~\cite{Coecke2010}, which this present work extends.
% subsection the_basics (end)

\subsection{Compact Closed Categories} % (fold)
\label{sub:monoidal_categories}

A slightly more complex class of categories is that of monoidal categories, which allow us to reason not just about objects and the relations between them, but also about combinations of objects in terms of the objects which they comprise. Formally, a (strict) monoidal category $\mathbf{C}$ is a basic category to which we add a bifunctor $\otimes$ (sometimes referred to as a \emph{monoidal tensor}) satisfying the following conditions:
\begin{itemize}
    \item For all $A,B \in ob(\mathbf{C})$ there is an object $A \otimes B \in ob(\mathbf{C})$.
    \item For all $A,B,C \in ob(\mathbf{C})$, we have $(A \otimes B) \otimes C \cong A \otimes (B \otimes C)$.
    \item There exists some $I \in ob(\mathbf{C})$ such that for any $A \in ob(C)$, we have $I \otimes A \cong A \cong A \otimes I$.
    \item For $f : A \to C$ and $g : B \to D$ in $hom(\mathbf{C})$ there is $f \otimes g: A \otimes B \to C \otimes D$ in $hom(\mathbf{C})$.
    \item For $f_1 : A \to C$, $f_2 : B \to D$, $g_1 : C \to E$ and $g_2 : D \to F$ the following equality holds:
    \begin{displaymath}
        (g_1 \otimes g_2) \circ (f_1 \otimes f_2) = (g_1 \circ f_1) \otimes (g_2 \circ f_2)
    \end{displaymath}
\end{itemize}
The strictness of the category entails that the isomorphisms described above are equalities.

A compact bi-closed category $\mathbf{C}$ is a monoidal category with the following additional axioms:
\begin{itemize}
    \item Each object $A \in ob(\mathbf{C})$ has left and right `adjoint' objects $A^l$ and $A^r$ in $ob(\mathbf{C})$. The following isomorphism shows the distribution of adjoints over tensored objects:
    \[
    (A \otimes B)^l \cong (B^l \otimes A^l) \quad \textrm{and} \quad (A \otimes B)^r \cong (B^r \otimes A^r)
    \]
    \item There exist four structural morphisms for each object $A \in ob(\mathbf{C})$:
    \begin{itemize}
        \item $\eta^l_A : I \to A \otimes A^l$.
        \item $\eta^r_A : I \to A^r \otimes A$.
        \item $\epsilon^l_A : A^l \otimes A \to I$.
        \item $\epsilon^r_A : A \otimes A^r \to I$.
    \end{itemize}
    \item All such structural morphisms satisfy the following equalities:
    \begin{itemize}
        \item $(1_A \otimes \epsilon^l_A) \circ (\eta^l_A \otimes 1_A) = 1_A$.
        \item $(\epsilon^r_A \otimes 1_A) \circ (1_A \otimes \eta^r_A) = 1_A$.
        \item $(1_{A^r} \otimes \epsilon^r_A) \circ (\eta^r \otimes 1_{A^r}) = 1_{A^r}$.
        \item $(\epsilon^l_A \otimes 1_{A^l}) \circ (1_{A^l} \otimes \eta^l_A) = 1_{A^l}$.
    \end{itemize}
\end{itemize}
Furthermore, for product categories involving compact closed categories, if there are pairings $(a,A)$ and $(b,B)$, then there is a pairing $(a \otimes b,A \otimes B)$. One might describe compact closed categories as monoidal categories where we not only deal with the combination of objects, but also qualify how such combinations relate to simpler objects through `cancellations' ($\epsilon$ morphisms) and `productions' ($\eta$ morphisms).

It is worth noting the obvious similarity between compact closed categories and the pregroup structures discussed in $\S$\ref{sub:pregroups}. I note that each object in a compact closed category has a left and a right adjoint, as do objects in pregroups. The monoidal tensor behaves identically to monoidal multiplication, and is also associative. There is a unit object $I$ with the same equality properties as $1$ in a pregroup. Furthermore, we note that if morphisms in a compact closed category are considered as ordering relations, the structural morphisms hold the same inequality relations as the object-adjoint pairings do in a pregroup.

We can therefore consider a pregroup as a compact closed category $P$ modelling a poset. The elements of the pregroup's set are the category's objects; the ordering relations are its morphisms, $1$ as $I$, and monoidal multiplication is the bifunctor $\otimes$. Notationally, instead of the single ordering relation symbol $\leq$ we instead can write $\leq_{(a,b)}$ to denote the morphism expressing $a \leq b$. Likewise, the unary operators $(-)^l$ and $(-)^r$ can be turned into a set of morphisms linking types to their adjoints, where each morphism can be individually denoted $(-)^l_a$ for the case $(-)^l :: a \mapsto a^l$, and similarly $(-)^r_a$ for the case $(-)^r :: a \mapsto a^r$, for any such $a$ in $P$.

The procedure described above is called giving categorical semantics to the pregroup. In $\S$\ref{sec:a_categorical_passage_from_grammar_to_semantics}, I will discuss how the little category theory we have seen here and this notion of giving categorical semantics to other formalisms can aid us in achieving our goal of syntax-sensitive compositional DSMs.

% paragraph monoidal_categories (end)

\subsection{A Graphical Calculus for Compact Closed Categories} % (fold)
\label{sub:a_graphical_calculus_for_categories}

Compact closed categories may appear to be very abstract mathematical entities to reason with and about. Fortunately, a graphical calculus, surveyed in \cite{Selinger2010}, has been developed to provide both visual and practical support for these tasks. Proofs in this graphical calculus take the form of applications of diagrammatic rewrite rules which are sound and complete, and correspond to mathematical proofs about compact closed categories. This graphical calculus has basic elements and rewrite rules (and associated categorical meanings), some important ones of which are shown in Tables~\ref{tab:graphical_basic_elements}--\ref{tab:rewrite_rules}.

\begin{table}[ht!]
\begin{center}
\begin{tabular}{c|c|c}
\textbf{Name} & \textbf{Diagram} & \textbf{Meaning}\\
\hline
\hline
    \raisebox{3\height}{Identity} & 
\begin{tikzpicture}[thick]
\node (v1) at (0,1) {};
\node (v2) at (0,-1) {};
\draw [-] (v1) -- node [right] {$A$} (v2);
\end{tikzpicture}
 & \raisebox{3\height}{$1_A: A \to A$}\\
\hline
\raisebox{3\height}{Morphism} & 
\begin{tikzpicture}[thick]
      \tikzstyle{bordered} = [draw,outer sep=0,inner sep=1,minimum size=15]
\node[bordered] (f) at (0,0) {$f$};
\node (v1) at (0,1) {};
\node (v2) at (0,-1) {};
\draw[-]  (v1) -- node [right] {$A$} (f);
\draw[-]  (f) -- node [right] {$B$} (v2);
\end{tikzpicture}
& \raisebox{3\height}{$f: A \to B$}\\
\hline
\raisebox{5\height}{Composition} & 
\begin{tikzpicture}[thick]
      \tikzstyle{bordered} = [draw,outer sep=0,inner sep=1,minimum size=15]
\node[bordered] (f) at (0,0) {$f$};
\node[bordered] (g) at (0,-1) {$g$};
\node (v1) at (0,1) {};
\node (v2) at (0,-2) {};
\draw[-]  (v1) -- node [right] {$A$} (f);
\draw[-]  (f) -- node [right] {$B$} (g);
\draw[-]  (g) -- node [right] {$C$} (v2);
\end{tikzpicture}
& \raisebox{5\height}{$g \circ f: A \to C$}\\
\hline
\raisebox{3\height}{Tensored functions} &
\begin{tikzpicture}[thick]
      \tikzstyle{bordered} = [draw,outer sep=0,inner sep=1,minimum size=15]
\node[bordered] (f) at (0,0) {$f$};
\node (v1) at (0,1) {};
\node (v2) at (0,-1) {};
\draw[-]  (v1) -- node [right] {$A$} (f);
\draw[-]  (f) -- node [right] {$B$} (v2);
\node[bordered] (g) at (1,0) {$g$};
\node (v3) at (1,1) {};
\node (v4) at (1,-1) {};
\draw[-]  (v3) -- node [right] {$C$} (g);
\draw[-]  (g) -- node [right] {$D$} (v4);
\end{tikzpicture}
& \raisebox{3\height}{$f \otimes g : A \otimes B \to C \otimes D$}\\
\hline
\raisebox{3\height}{Morphisms over tensors} &
\begin{tikzpicture}[thick]
      \tikzstyle{bordered} = [draw,outer sep=0,inner sep=1,minimum size=15, minimum width=40]
      \tikzstyle{invisiborder} = [outer sep=0,inner sep=1,minimum height=15]
\node[bordered] (f) at (0,0) {$f$};
\node (v1) at (-.25,1) {};
\node (v2) at (.25,1) {};
\node (v3) at (0,-1) {};
\node[invisiborder] (fl) at (-0.25,0) {};
\draw[-]  (v1) -- node [left] {$A$} (fl);
\node[invisiborder] (fr) at (0.25,0) {};
\draw[-]  (v2) -- node [right] {$B$} (fr);
\draw[-] (f) -- node [right] {$C$} (v3);
\end{tikzpicture}
& \raisebox{3\height}{$h : A \otimes B \to C$}\\
\hline
\raisebox{3\height}{State} &
\begin{tikzpicture}[thick]
\tikzstyle{vector} = [draw, isosceles triangle, shape border rotate=90, isosceles triangle stretches, outer sep=0,inner sep=1, minimum height=5, minimum width=30]
\node at (0,.8) {};
\node[vector] (v) at (0,0) {$\psi$};
\node (v1) at (0,-1) {};
\draw[-] (v) -- node [right] {$A$} (v1);
\end{tikzpicture}
& \raisebox{3\height}{$\psi: I \to A$}\\
\hline
\raisebox{3\height}{Co-state} &
\begin{tikzpicture}[thick]
\tikzstyle{vector} = [draw, isosceles triangle, shape border rotate=-90, isosceles triangle stretches, outer sep=0,inner sep=1, minimum height=5, minimum width=30]
\node[vector] (v) at (0,-1) {$\psi^*$};
\node (v1) at (0,0) {};
\draw[-] (v) -- node [right] {$A$} (v1);
\end{tikzpicture}
& \raisebox{3\height}{$\psi^{*}: A^* \to I$}\\
\end{tabular}
\end{center}
\caption{Basic elements of the graphical calculus for compact closed categories.}
\label{tab:graphical_basic_elements}
\end{table}

In Table~\ref{tab:graphical_basic_elements}, the basic elements are shown. I depict the flow of information as going from the top of the diagram towards the bottom, along the paths of wires typed with objects of the category. In some applications of this calculus, an opposite convention is used, whereby information flows from bottom to top.

Identity is seen as a naked wire. Morphisms are a box which transforms an input wire into an output wire of (possibly) different type. Morphism composition is two boxes on the same wire, while tensored morphisms are two wires side by side (with morphism boxes on them). A function over tensors is a box that take two wires in and outputs one or more wires. States (morphisms from the unit $I$ to other elements of the category) are triangles with one or more output wires. Their co-state (morphisms from objects of the category to $I$)  are represented as upside-down triangles with one or more input wires, where the star ($*$) usually stands for an adjoint, as shown in the `swing' and `float' rewrite rules, discussed below.

\begin{table}[ht!]
\begin{center}
\begin{tabular}{c|c|c}
\textbf{Name} & \textbf{Diagram} & \textbf{Meaning}\\
\hline
\hline
    \raisebox{2\height}{$\epsilon^r$ map} &
\begin{tikzpicture}[thick]
\node (v1) at (0,0) {$A$};
\node (v2) at (1,0) {$A^r$};
\draw  (v1) .. controls +(0,-.75) and +(0,-.75) .. (v2);
\end{tikzpicture}
& \raisebox{2\height}{$\epsilon^r_A : A \otimes A^r \to I$}\\
\hline
\raisebox{2\height}{$\epsilon^l$ map} &
\begin{tikzpicture}[thick]
\node (v1) at (0,0) {$A^l$};
\node (v2) at (1,0) {$A$};
\draw  (v1) .. controls +(0,-.75) and +(0,-.75) .. (v2);
\end{tikzpicture}
& \raisebox{2\height}{$\epsilon^l_A : A^l \otimes A \to I$}\\
\hline
\raisebox{1\height}{$\eta^r$ map} &
\begin{tikzpicture}[thick]
\node (v1) at (0,0) {$A^r$};
\node (v2) at (1,0) {$A$};
\draw  (v1) .. controls +(0,.75) and +(0,.75) .. (v2);
\end{tikzpicture}
& \raisebox{1\height}{$\eta^r_A : I \to A^r \otimes A$}\\
\hline
\raisebox{1\height}{$\eta^l$ map} &
\begin{tikzpicture}[thick]
\node (v1) at (0,0) {$A$};
\node (v2) at (1,0) {$A^l$};
\draw  (v1) .. controls +(0,.75) and +(0,.75) .. (v2);
\end{tikzpicture}
& \raisebox{1\height}{$\eta^l_A : I \to A \otimes A^l$}\\
\end{tabular}
\end{center}
\caption{Structural morphisms in the graphical calculus for compact closed categories.}
\label{tab:graphical_struct_morph}
\end{table}

In Table~\ref{tab:graphical_struct_morph}, the diagrammatic forms for the structural morphisms of a compact closed category are shown. The ``special'' \emph{structural morphisms} of a compact closed category have a specific representation in this diagrammatic calculus, instead of boxes. I represent the $\epsilon$ morphisms as `cups' similar to those found in the diagrammatic pregroup calculus presented in $\S$\ref{sub:a_graphical_calculus_for_pregroups}, and the $\eta$ morphisms as `caps'. Naturally, in similar diagrammatic calculi where the flow of information is from bottom to top, cups and caps stand for $\eta$ and $\epsilon$ morphisms, respectively.

\begin{table}[ht!]
\begin{center}
\begin{tabular}{c|c|c}
\textbf{Name} & \textbf{Diagram} & \textbf{Meaning}\\
\hline
\hline
\raisebox{2\height}{Yank} &
\begin{tikzpicture}[thick]
\node at (0,-.15) {$A$};
\node at (2,.4) {$A$};
\node (v1) at (0,0) {};
\node (v2) at (1,0) {};
\node (v3) at (2,0) {};
\draw[-] (v1.north) .. controls +(0,.75) and +(0,.75) .. (v2.north);
\draw[-] (v2.north) .. controls +(0,-.75) and +(0,-.75) .. (v3.north);
\node at (2.5,0) {$=$};
\draw[-] (3,0.75) -- node [right] {$A$} (3,-.5);
\end{tikzpicture}
& \raisebox{2\height}{$(1_A \otimes \epsilon^l_A) \circ (\eta^l_A \otimes 1_A) = 1_A$}\\
\hline
\raisebox{2\height}{Yank} &
\begin{tikzpicture}[thick]
\node at (0,.4) {$A$};
\node at (2,-.15) {$A$};
\node (v1) at (0,0) {};
\node (v2) at (1,0) {};
\node (v3) at (2,0) {};
\draw[-] (v1.north) .. controls +(0,-.75) and +(0,-.75) .. (v2.north);
\draw[-] (v2.north) .. controls +(0,.75) and +(0,.75) .. (v3.north);
\node at (2.5,0) {$=$};
\draw[-] (3,0.75) -- node [right] {$A$} (3,-.5);
\end{tikzpicture}
& \raisebox{2\height}{$(\epsilon^r_A \otimes 1_A) \circ (1_A \otimes \eta^r_A) = 1_A$}\\
\hline
\raisebox{4\height}{Slide} &
\begin{tikzpicture}[thick]
      \tikzstyle{bordered} = [draw,outer sep=0,inner sep=1,minimum size=15]
    \node (v1) at (0,1) {};
    \node[bordered] (f1) at (0,0) {$f$};
    \node (v2) at (0,-2) {};
    \draw[-] (v1) -- node [right] {$A$} (f1);
    \draw[-] (f1) -- node [right] {$B$} (v2);
    \node at (1,-.5) {$=$};
    \node (v3) at (2,1) {};
    \node[bordered] (f2) at (2,-1) {$f$};
    \node (v4) at (2,-2) {};
    \draw[-] (v3) -- node [right] {$A$} (f2);
    \draw[-] (f2) -- node [right] {$B$} (v4);
\end{tikzpicture}
& \raisebox{4\height}{$1_B \circ f = f \circ 1_A$}\\
\hline
\raisebox{4\height}{Swing} &
\begin{tikzpicture}[thick]
    \tikzstyle{vector} = [draw, isosceles triangle, shape border rotate=90, isosceles triangle stretches, outer sep=0,inner sep=1, minimum height=5, minimum width=30]
    \node at (0,1.25) {};
    \node[vector] (v0) at (0,0.65) {$\psi$};
    \node (v1) at (0,0) {};
    \draw[-] (v0) -- node [left] {$A$} (v1.south);
    \node (v2) at (1,0) {};
    \node (v3) at (1,1) {};
    \draw [-] (v2.south) -- node [right] {$A^r$} (v3);
    \draw [-] (v1) .. controls +(0,-1) and +(0,-1) .. (v2);
    \node at (2,0) {$=$};
    \node[vector, shape border rotate=-90] (v4) at (3,-.25) {$\psi^r$};
    \node (v5) at (3,1) {};
    \draw [-] (v4) -- node [right] {$A^r$} (v5);
    \end{tikzpicture}
&
\raisebox{4\height}{$\epsilon^r_A \circ (\psi \otimes 1_{A^r}) = \psi^l \circ 1_{A^r}$}\\
\hline
\raisebox{4\height}{Swing} &
    \begin{tikzpicture}[thick]
    \tikzstyle{vector} = [draw, isosceles triangle, shape border rotate=90, isosceles triangle stretches, outer sep=0,inner sep=1, minimum height=5, minimum width=30]
    \node at (0,1.25) {};
    \node[vector] (v0) at (1,0.65) {$\psi$};
    \node (v1) at (1,0) {};
    \draw[-] (v0) -- node [right] {$A$} (v1.south);
    \node (v2) at (0,0) {};
    \node (v3) at (0,1) {};
    \draw [-] (v2.south) -- node [left] {$A^l$} (v3);
    \draw [-] (v1) .. controls +(0,-1) and +(0,-1) .. (v2);
    \node at (2,0) {$=$};
    \node[vector, shape border rotate=-90] (v4) at (3,-.25) {$\psi^l$};
    \node (v5) at (3,1) {};
    \draw [-] (v4) -- node [right] {$A^l$} (v5);
    \end{tikzpicture}
&
\raisebox{4\height}{$\epsilon^l_A \circ (1_{A^r} \otimes \psi) = \psi^r \circ 1_{A^l}$}\\
\hline
\raisebox{4\height}{Float} &
    \begin{tikzpicture}[thick]
    \tikzstyle{vector} = [draw, isosceles triangle, shape border rotate=90, isosceles triangle stretches, outer sep=0,inner sep=1, minimum height=5, minimum width=30]
    \node at (0,1) {};
    \node[vector, shape border rotate=-90] (v0) at (0,-.65) {$\psi$};
    \node (v1) at (0,0) {};
    \draw [-] (v0) -- node [left] {$A$} (v1.north);
    \node (v2) at (1,0) {};
    \node (v3) at (1,-1) {};
    \draw [-] (v1) .. controls +(0,1) and +(0,1) .. (v2);
    \draw [-] (v2.north) -- node [right] {$A^l$} (v3);
    \node at (2,0) {$=$};
    \node [vector] (v4) at (3,.25) {$\psi^l$};
    \node (v5) at (3,-1) {};
    \draw [-] (v4) -- node [right] {$A^l$} (v5);
    \end{tikzpicture}
& \raisebox{4\height}{$(\phi \otimes 1_{A^l}) \circ \eta^l_A = 1_{A^l} \circ \psi^l$}\\
\hline
\raisebox{4\height}{Float} &
    \begin{tikzpicture}[thick]
    \tikzstyle{vector} = [draw, isosceles triangle, shape border rotate=90, isosceles triangle stretches, outer sep=0,inner sep=1, minimum height=5, minimum width=30]
    \node at (0,1) {};
    \node[vector, shape border rotate=-90] (v0) at (1,-.65) {$\psi$};
    \node (v1) at (1,0) {};
    \draw [-] (v0) -- node [right] {$A$} (v1.north);
    \node (v2) at (0,0) {};
    \node (v3) at (0,-1) {};
    \draw [-] (v1) .. controls +(0,1) and +(0,1) .. (v2);
    \draw [-] (v2.north) -- node [left] {$A^r$} (v3);
    \node at (2,0) {$=$};
    \node [vector] (v4) at (3,.25) {$\psi^r$};
    \node (v5) at (3,-1) {};
    \draw [-] (v4) -- node [right] {$A^r$} (v5);
    \end{tikzpicture}
& \raisebox{4\height}{$(1_{A^r} \otimes \phi) \circ \eta^r_A = 1_{A^r} \circ \psi^r$}\\
\end{tabular}
\end{center}
\caption{Rewrite rules in the graphical calculus for compact closed categories.}
\label{tab:rewrite_rules}
\end{table}

\begin{figure}[ht!]
\begin{center}
    \begin{tikzpicture}[thick]
      \tikzstyle{bordered} = [draw,outer sep=0,inner sep=1,minimum size=15]
      \tikzstyle{vector} = [draw, isosceles triangle, shape border rotate=90, isosceles triangle stretches, outer sep=0,inner sep=1, minimum height=5, minimum width=30]
    \begin{scope}
    \node (v1) at (0,2) {};
    \node[bordered] (f) at (0,1) {$f$};
    \node (v2) at (0,0) {};
    \node (v3) at (1,0) {};
    \node (v4) at (1,1) {};
    \node (v5) at (2,1) {};
    \node (v6) at (2,-1) {};
    \draw[-] (v1) -- node [left] {$A$} (f);
    \draw[-] (f) -- node [left] {$B$} (v2);
    \draw[-] (v3) -- (v4);
    \draw[-] (v5) -- node [right] {$B$} (v6);
    \draw[-] (v2.north) .. controls +(0,-1) and +(0,-1) .. (v3.north);
    \draw[-] (v4.south) .. controls +(0,1) and +(0,1) .. (v5.south);
    
    \node[vector, shape border rotate=-90, minimum height=20, minimum width= 40] at (0.5,-.5) {};
    
    \node[vector, minimum height=20, minimum width= 40] at (1.5,1.5) {};
    
    \end{scope}
    
    \begin{scope}[xshift=80pt]
    \node at (0,.5) {$=$};
    \end{scope}

    \begin{scope}[xshift=200pt]
    \node at (0,.5) {$=$};
    \end{scope}
    
    \begin{scope}[xshift=110pt]
    \node (v1) at (0,2) {};
    \node [bordered] (f) at (0,1) {$f$};
    \node (v2) at (0,-1) {};
    \draw [-] (v1) -- node [right] {$A$} (f);
    \draw [-] (f) -- node [right] {$B$} (v2);
    \end{scope}
    
    \begin{scope}[xshift=140pt]
    \node at (0,.5) {$=$};
    \end{scope}
    
    \begin{scope}[xshift=170pt]
    \node (v1) at (0,2) {};
    \node [bordered] (f) at (0,0) {$f$};
    \node (v2) at (0,-1) {};
    \draw [-] (v1) -- node [right] {$A$} (f);
    \draw [-] (f) -- node [right] {$B$} (v2);
    \end{scope}
    
    \begin{scope}[xshift=230pt]
    \node (v1) at (0,2) {};
    \node[bordered] (f) at (2,0) {$f$};
    \node (v2) at (0,0) {};
    \node (v3) at (1,0) {};
    \node (v4) at (1,1) {};
    \node (v5) at (2,1) {};
    \node (v6) at (2,-1) {};
    \draw[-] (v5) -- node [right] {$A$} (f);
    \draw[-] (f) -- node [right] {$B$} (v6);
    \draw[-] (v3) -- (v4);
    \draw[-] (v1) -- node [left] {$A$} (v2);
    \draw[-] (v2.north) .. controls +(0,-1) and +(0,-1) .. (v3.north);
    \draw[-] (v4.south) .. controls +(0,1) and +(0,1) .. (v5.south);
    
    \node[vector, shape border rotate=-90, minimum height=20, minimum width= 40] at (0.5,-.5) {};
    \node[vector, minimum height=20, minimum width= 40] at (1.5,1.5) {};
    
    \end{scope}
\end{tikzpicture}
\end{center}
\caption{Examples of yank-slide equalities in the graphical calculus for compact closed categories.}
\label{fig:teleportation-like}
\end{figure}
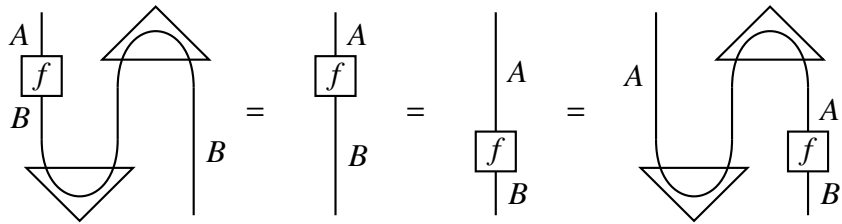

Finally, in Table~\ref{tab:rewrite_rules}, some of the key graphical re-write rules are shown, with names which are (for the most part) not ``official'', but principally there to make it easier to talk about them. First the two ``yank'' rewrites show how the combination of a cup and a cap `cancel' each other to produce an identity morphism, following the definitions of the structural morphisms. The second shows how morphisms can ``slide'' up and down straight wires without changing the categorical meaning of the diagram. Both of these rewrite rules can be combined to show that morphisms can slide along non-straight wires (i.e.~those including cups and caps) without changing the meaning of the diagram. This is exemplified in Figure~\ref{fig:teleportation-like}, which shows how a morphism can slide across such a non-straight wire. From left to right: I first use the yank equality, then the slide equality, then yank again to obtain the rightmost diagram from the leftmost. For those interested, this property is one of the central elements behind the diagrammatic proof of quantum teleportation \cite{abramsky2004categorical,coecke2006kindergarten}, the diagrammatic representation of which closely resembles that shown in Figure~\ref{fig:teleportation-like}, with the inclusion of additional morphisms.

Finally, the swing and float rules show that the yank operations can be separated out into separate steps, allowing us to `move' states along cups and caps.
% subsection a_graphical_calculus_for_categories (end)

% section categories (end)

\section{A Categorical Passage from Grammar to Semantics} % (fold)
\label{sec:a_categorical_passage_from_grammar_to_semantics}

In $\S$\ref{sub:monoidal_categories} I discussed how any pregroup grammar could be represented as a compact closed category $P$. In $\S$\ref{sub:the_basics} I described how product categories allowed us to relate the objects and morphisms of one category to those of another. In this section, I will present how  \cite{Clark2008,Coecke2010} suggest building on this by using categories to relate semantic composition to syntactic analysis in order to achieve syntax-sensitive composition in DSMs.

\subsection{$\mathbf{FVect}$} % (fold)
\label{sub:fvect}
Let $\mathbf{FVect}$ be the symmetric monoidal compact closed category of finite-dimensional Hilbert spaces over $\mathbb{R}$, i.e.~vector spaces over $\mathbb{R}$ with orthogonal bases of finite dimension, and an inner product operation $\langle - \mid - \rangle : A \times A \to \mathbb{R}$ for every vector space $A$. The objects of $\mathbf{FVect}$ are the vector spaces, and the morphisms are linear maps between vector spaces. The unit object is $\mathbb{R}$ and the monoidal tensor is the linear algebraic tensor product of vector spaces. The symmetric aspect of this category means that for any two objects $A \otimes B$ and $B \otimes A$ in the category, there exists an isomorphism $A \otimes B \cong B \otimes A$, corresponding here to the fact that any tensor is isomorphic to its permutations.  

As a result of its symmetric nature, the category is degenerate in its adjoints, in that for any vector space $A$, we have the isomorphisms $A^l \cong A^r \cong A$. This is because the adjoint of a vector space $A$ is its co-vector space $A^r = A^l = A^*$, the elements of which are the conjugate transposes of the vectors from that vector space. Since the conjugate transpose of a real-valued vector is just the transpose of that vector, each vector in some space $A$ can be isomorphically mapped to a covector (its transpose) in $A^*$, hence $A \cong A^*$. As such, we can effectively do away with adjoints in this category, and `collapse' $\epsilon^l$, $\epsilon^r$, $\eta^l$, and $\eta^r$ maps into `adjoint-free' $\epsilon$ and $\eta$ maps. The structural morphisms of the category are the inner product operations $\epsilon$,
\[
\epsilon_A : A \otimes A \to \mathbb{R} ::  \overrightarrow{v} \otimes \overrightarrow{w} \mapsto \langle \overrightarrow{a} \mid \overrightarrow{b} \rangle
\] 
and the $\eta$ maps from real numbers to tensored vector spaces 
\[
\eta_A : \mathbb{R} \to A \otimes A :: 1 \mapsto \overrightarrow{1_{A \otimes A}}
\]
where $\overrightarrow{1_{A \otimes A}}$ is the superposition of all the basis vectors $\{\overrightarrow{a_i} \otimes \overrightarrow{a_j}\}_{ij}$ of $A \otimes A$
\[
\overrightarrow{1_{A \otimes A}} = \sum_{ij}{\overrightarrow{a_i} \otimes \overrightarrow{a_j}}
\]

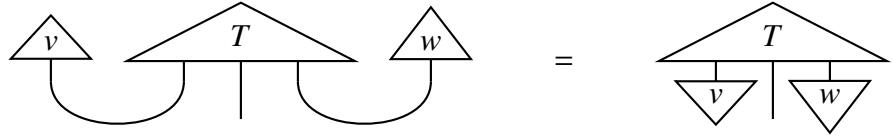
\begin{figure}[ht]
\begin{center}
     \tikzstyle{bordered} = [draw,outer sep=0,inner sep=1,minimum size=15]
\tikzstyle{vector} = [draw, isosceles triangle, shape border rotate=90, isosceles triangle stretches, outer sep=0,inner sep=1, minimum height=5, minimum width=30]
\tikzstyle{covector} = [draw, isosceles triangle, shape border rotate=-90, isosceles triangle stretches, outer sep=0,inner sep=1, minimum height=5, minimum width=30]
    \begin{tikzpicture}[thick]

\node[vector] (v1) at (-.5,2) {$v$};
\node[vector,minimum width=3cm] (T) at (2,2.1) {$T$};
\node[vector] (v2) at (4.5,2) {$w$};

\draw (v1) -- (-.5,1.5);
\draw (v2) -- (4.5,1.5);
\draw (T) -- (2,1);
\draw (1.25,1.76) -- (1.25,1.5);
\draw (2.75,1.76) -- (2.75,1.5);

\draw (-.5,1.5) .. controls +(0,-.75) and +(0,-.75) .. (1.25,1.5);
\draw (2.75,1.5) .. controls +(0,-.75) and +(0,-.75) .. (4.5,1.5);

\node at (6.25,1.75) {$=$};

\node[vector,minimum width=3cm] (T2) at (9,2.1) {$T$};
\draw (T2) -- (9,1);

\node[covector] (v3) at (8.25,1.3) {$v$};
\node[covector] (v4) at (9.75,1.3) {$w$};

\draw (8.25,1.76) -- (v3);
\draw (9.75,1.76) -- (v4);
\end{tikzpicture}
 \end{center}
    \caption{Sample diagrammatic representation of distributional composition.}
    \label{fig:diagcomp}
\end{figure}

On the diagrammatic front, I treat a vector $\overrightarrow{v} \in A$ as a state $\overrightarrow{v}: \mathbb{R} \to A$ and a co-vector as its co-state. This means that the application of $\epsilon$ maps to model the composition of vectors with a tensor corresponds to the application of the swing operation described above, showing how the vectors are brought into relation with the tensor through inner products, as shown in Figure~\ref{fig:diagcomp}.

% subsection fvect (end)

\subsection{Syntax Guides Semantics} % (fold)
\label{sub:syntax_guides_semantics}

If we consider the product category $P \times \mathbf{FVect}$, we observe that is has as objects pairs $(a,A)$ where $a$ is a pregroup type and $A$ is a vector space, and as morphisms pairs $(\leq,f)$ where $\leq$ is a pregroup ordering relation and $f$ is a linear map. By the definition of product categories, for any two vector space-type pairs $(a,A)$ and $(b,B)$, there exists a morphism $(a,A) \to (b,B)$ only if there exists both an ordering $a \leq b$ and a linear map $A \to B$. If we view these pairings as the association of syntactic types with vector spaces containing semantic vectors for words of that type, this restriction effectively states that a linear map from $A$ to $B$ is only ``permitted'' in the product category if $a \leq b$.

Both $P$ and $\mathbf{FVect}$ being compact closed, it is simple to show that $P \times \mathbf{FVect}$ is as well, by considering the pairs of unit objects and structural morphisms from the separate categories: $I$ is now $(1,\mathbb{R})$, and the structural morphisms are $(\epsilon^l_a,\epsilon_A)$, $(\epsilon^r_a,\epsilon_A)$, $(\eta^l_a,\eta_A)$, $(\eta^r_a,\eta_A)$. We are particularly interested in the $\epsilon$ maps, which are defined as follows (from the definition of product categories):
\begin{displaymath}
    (\epsilon^l_A,\epsilon_A) : (a^la,A \otimes A) \to (1,\mathbb{R})  \qquad (\epsilon^r_A,\epsilon_A) : (aa^r,A \otimes A) \to (1,\mathbb{R})
\end{displaymath}
This states that whenever there is a reduction step in the grammatical analysis of a sentence, there is a composition operation in the form of an inner product on the semantic front. Hence if nouns of type $n$ live in some noun space $N$ and transitive verbs of type $n^l s n^r$ live in some space $N \otimes S \otimes N$, then there must be some structural morphism of the form:
\begin{displaymath}
    (\epsilon_n^r 1_s  \epsilon_n^l, \epsilon_N \otimes 1_S \otimes \epsilon_N) : (n (n^r s n^l) n,N \otimes (N \otimes S \otimes N) \otimes N) \to (s,S)
\end{displaymath}
We can read from this morphism the functions required to compose a sentence with a subject noun, a transitive verb, and an object noun, in order to obtain a vector living in some sentence space $S$, namely $(\epsilon_N \otimes 1_S \otimes \epsilon_N)$. Diagrammatically, this composition is represented as in Figure~\ref{fig:diagcomp}, where $\overrightarrow{v} \in N$ is the vector for the subject, $\overrightarrow{w} \in N$ is the vector for the object, and $T \in N \otimes S \otimes N$ is the tensor representing the noun.

The form of a syntactic type is therefore what dictates the structure of the semantic space associated with it. The structural morphisms of the product category guarantee that for every syntactic reduction there is a semantic composition morphism provided by the product category: \emph{syntactic analysis guides semantic composition}.

% subsection syntax_guides_semantics (end)

\subsection{Example} % (fold)
\label{sub:fvect_examples}

To give an example, we can give syntactic type $n$ to nouns, and $n^rs$ to intransitive verbs. The parse for ``kittens sleep'', namely $n n^rs \to s$, corresponds to the morphism $\epsilon^r_n \otimes 1_s$ in $P$. The syntactic types dictate that the noun $\overrightarrow{\textrm{kittens}}$ lives in some vector space $N$, and the intransitive verb $\overrightarrow{\textrm{sleep}}$ in $N \otimes S$. The reduction morphism $(\epsilon_n^r1_s)$ gives us the composition morphism $(\epsilon_N \otimes 1_S)$, which we can apply to $\overrightarrow{\textrm{kittens}} \otimes \overrightarrow{\textrm{sleep}}$. 

Since we can express any vector as the weighted superposition of its basis vectors, let us expand $\overrightarrow{\textrm{kittens}} = \sum_i{c_i^{\textrm{kittens}} \overrightarrow{n_i}}$ and $\overrightarrow{\textrm{sleep}} = \sum_{ij}{c_{ij}^{\textrm{sleep}} \overrightarrow{n_i} \otimes \overrightarrow{s_j}}$. We can then express the composition as follows:{ \small
\begin{align*}
    \overrightarrow{\textrm{kittens sleep}} & = (\epsilon_N \otimes 1_S) (\overrightarrow{\textrm{kittens}} \otimes \overrightarrow{\textrm{sleep}})\\
    & = (\epsilon_N \otimes 1_S) \left(\sum_i{c_i^{\textrm{kittens}} \overrightarrow{n_i}} \otimes \sum_{jk}{c_{jk}^{\textrm{sleep}} \overrightarrow{n_j} \otimes \overrightarrow{s_k}}\right)\\
    & = (\epsilon_N \otimes 1_S) \left(\sum_{ijk}{c_{i}^{\textrm{kittens}} c_{jk}^{\textrm{sleep}} \overrightarrow{n_i} \otimes \overrightarrow{n_j} \otimes \overrightarrow{s_k}}\right)\\
    & = \sum_{ijk}{ c_i^{\textrm{kittens}} c_{jk}^{\textrm{sleep}} \langle \overrightarrow{n_i}  \mid \overrightarrow{n_j} \rangle \overrightarrow{s_k} }\\
    & = \sum_{ik}{ c_i^{\textrm{kittens}} c_{ik}^{\textrm{sleep}} \overrightarrow{s_k} }
\end{align*}}

The above equations are hopefully fairly clear at this stage: I express the vectors in their explicit form. I consolidate the sums by virtue of distributivity of the linear algebraic tensor product over addition; I then apply the tensored linear maps to the vector components (as the weights are scalars); and finally, I simplify the indices since $\langle \overrightarrow{n_i} \mid \overrightarrow{n_j} \rangle = 1$ if $\overrightarrow{n_i} = \overrightarrow{n_j}$ and $0$ otherwise. I obtain a vector that lives in sentence space $S$.

Transitive sentences can be dealt with in a similar fashion:
\begin{align*}
    & \overrightarrow{\textrm{kittens chase mice}}  \\
    &  \quad= (\epsilon_N \otimes 1_S \otimes \epsilon_N) (\overrightarrow{\textrm{kittens}} \otimes \overrightarrow{\textrm{chase}} \otimes \overrightarrow{\textrm{mice}}) \\
    &  \quad = (\epsilon_N \otimes 1_S \otimes \epsilon_N) \left(\sum_i{c_i^{\textrm{kittens}} \overrightarrow{n_i}} \otimes \left(\sum_{jkl}{c_{jkl}^{\textrm{chase}} \overrightarrow{n_j} \otimes \overrightarrow{s_k} \otimes \overrightarrow{n_l}}\right) \otimes \sum_m {c_m^{\textrm{mice}} \overrightarrow{n_m}} \right) \\
    &  \quad = (\epsilon_N \otimes 1_S \otimes \epsilon_N) \left(\sum_{ijklm}{c_{i}^{\textrm{kittens}} c_{jkl}^{\textrm{chase}} c_{m}^{\textrm{mice}} \overrightarrow{n_i} \otimes \overrightarrow{n_j} \otimes \overrightarrow{s_k} \otimes \overrightarrow{n_l} \otimes \overrightarrow{n_m} }\right) \\
    &  \quad = \sum_{ijklm}{ c_i^{\textrm{kittens}} c_{jkl}^{\textrm{chase}} c_{m}^{\textrm{mice}} \langle \overrightarrow{n_i}  \mid \overrightarrow{n_j} \rangle \overrightarrow{s_k} \langle \overrightarrow{n_l}  \mid \overrightarrow{n_m} \rangle }\\
    & \quad = \sum_{ikm}{ c_i^{\textrm{kittens}} c_{ikm}^{\textrm{chase}} c_{m}^{\textrm{mice}} \overrightarrow{s_k} } 
\end{align*}

In both cases, it is important to note that the tensor product passed as argument to the composition morphism, namely $\overrightarrow{\textrm{kittens}} \otimes \overrightarrow{\textrm{sleep}}$ in the intransitive case and $\overrightarrow{\textrm{kittens}} \otimes \overrightarrow{\textrm{chase}} \otimes \overrightarrow{\textrm{mice}}$ in the transitive case, never needs to be computed.

% subsection fvect_examples (end)

% section a_categorical_passage_from_grammar_to_semantics (end)

% chapter foundations_of_discocat (end)

\part{Theory}
\mbox{}
\newpage
%!TEX root = ../grefenstettethesis.tex

\chapter{Syntactic Extensions} % (fold)
\label{cha:syntactic_extensions}

\begin{chabstract}
  This chapter discusses a general methodology for the integration of syntactic formalisms into the DisCoCat framework by means of a functorial passage between the categorical representations of such formalisms and the category of vector spaces $\mathbf{FVect}$. It presents such categorical semantics and functorial passages for two grammatical formalisms: Context Free Grammars, and Lambek Grammars.
\end{chabstract}

The DisCoCat framework presented in Chapter~\ref{cha:foundations_of_discocat} makes use of pregroup grammars for syntactic analysis, principally because of their convenient algebraic structure which allows us to interpret any pregroup as a compact closed category. However, on a general level, the core strength of the framework is its ability to define the passage from syntax to semantics without commitment to the particular syntactic formalism used, or indeed the particular semantic mode of representation. 

In this chapter, I will discuss how some other syntactic analysis formalisms can be substituted for the pregroup grammars of Chapter~\ref{cha:foundations_of_discocat}. In $\S$\ref{sec:functorial_passages}, I discuss the notion of a functor and of a functorial passage between categories, and demonstrate that the pair category $P \times \mathbf{FVect}$ used in $\S$\ref{sub:syntax_guides_semantics} gives rise to a functor mapping syntactic analyses to semantic representations. In $\S$\ref{sec:supporting_context_free_grammars}, I present a way in which generative grammars such as context free grammars may be supported in the framework. In $\S$\ref{sec:supporting_lambek_grammar}, I will present work describing how such a functor can be defined for Lambek Grammar, how this requires the use of non-compact categories, and how an existing diagrammatic calculus can be used to reason about such categories. 

\section{Functorial Passages} % (fold)
\label{sec:functorial_passages}

In Chapter~\ref{cha:foundations_of_discocat}, a product category was used to model the passage from syntactic aspects of natural language composition to semantic representations of compositional operations and their arguments. There is, however, no requirement that we use a product category to model this passage. In fact---considering the compositional properties of functors described above---a functorial passage from syntax to semantics might fit our needs better. The idea of using a functorial passage from pregroup categories to $\mathbf{FVect}$ had been previously suggested by \cite{Preller2010}, and such a functorial passage has been described in a paper I co-authored with Coecke and Sardzadeh \cite{Sadrzadeh2013}. In this section, I will show that we can also define such a functorial passage from a product category to the category of vector spaces given certain restrictions, which are provided by the association made between pregroup types and vector spaces introduced in $\S$\ref{sub:syntax_guides_semantics}.

In $\S$\ref{sub:syntax_guides_semantics} I discussed the construct developed by \cite{Clark2008} of the pair category $P \times \mathbf{FVect}$ formed from a pregroup category $P$ and the category of vector spaces $\mathbf{FVect}$. We saw that this category was compact closed because $P$ and $\mathbf{FVect}$ were, and that for any linear map $f: A \to B$ in $\mathbf{FVect}$ there are morphisms $(\leq,f):x \otimes A \to y \otimes B$ for every $x$ and $y$ such that $x \leq y$. This definition actually covers a larger number of objects than we actually need, since it gives no explicit association between pregroup types and vector spaces. Yet this association is implicitly relied upon so that for every pregroup parse in $P$ there is one and only one linear map in $\mathbf{FVect}$ corresponding to the composition operation associated with the parse. In other words, we rely on a pre-defined mapping $M: ob(P) \to ob(\mathbf{FVect})$ of pregroup types to vector spaces in order to be able to talk about \emph{the} association between syntactic analysis and semantic composition, rather than \emph{an} association between syntactic analysis and semantic composition. To make the implicit explicit: we say that a morphism $(\leq,f) : x \otimes A \to y \otimes B$ describes how $f$ composes semantic vectors in vector space $A$ standing for words of type $x$ to produce a phrase vector in $B$ for words of type $y$ \emph{if and only if} $M(x) = A$ and $M(y) = y$. This restriction effectively allows us to ignore all the other products in $ob(P \times \mathbf{FVect})$, and focus only on those pairs of products for which the morphisms between elements of the pair correspond to unique semantic interpretations of a pregroup parse.

To give a concrete example, consider the sentence ``John loves Mary'', and suppose we give nouns the pregroup type $n$ and transitive verbs the type $n^rsn^l$. We wish to model the semantic vectors for nouns in some vector space $N$ and sentences of type $s$ in some vector space $S$. Given the isomorphic relation between adjoints in $\mathbf{FVect}$, we'd expect the vectors for transitive verbs to live in $N \otimes S \otimes N$, for the composition of nouns with the `argument places' of the verb to even be possible. The parse of ``John loves Mary'' corresponds to the ordering $n(n^rsn^l)n \leq s$. This ordering corresponds to an arrow standing for the function $\epsilon^r_n \otimes 1_s \otimes \epsilon^l_n$ in $P$ mapping the object $n(n^rsn^l)n$ to $s$. To find the morphism in $\mathbf{FVect}$ allowing for the composition of the word vectors for `John' in $N$ with `loves' in $N \otimes S \otimes N$ and `Mary' in $N$ to form a vector in the sentence space $S$, we look at the product category to find what maps $(n(n^rsn^l)n , N \otimes (N \otimes S \otimes N) \otimes N)$ to $(s,S)$. This morphism is, of course, one of the structural morphisms in $P \times \mathbf{FVect}$, namely $(\epsilon^r_n \otimes 1_s \otimes \epsilon^l_n, \epsilon_N \otimes 1_S \otimes \epsilon_N)$, as described in $\S$\ref{sub:syntax_guides_semantics}. However it should be noted that there are an infinite number of other pairings of vector spaces and linear maps with this parse which could have been considered, such as $(\epsilon^r_n \otimes 1_s \otimes \epsilon^l_n, \epsilon_N) : (n(n^rsn^l)n, N \otimes N) \to (s,1)$ or $(\epsilon^r_n \otimes 1_s \otimes \epsilon^l_n, id_S) : (n(n^rsn^l)n, S) \to (s,S)$, amongst others, simply by virtue of the definition of product categories. These other objects and morphisms of $P \times \mathbf{FVect}$ are completely irrelevant to our task of associating syntax with semantics, yet they are there for us to consider. What allows us to discount them is the assumption, made earlier in this paragraph, that $n$ should be associated with $N$, $n^r s n^l$ with $N \otimes S \otimes N$, and $s$ with $S$, which is essentially an instance of the sort of mapping $M$ discussed above. This shows us how the product-category approach works if such a mapping is defined, but that product categories come with a lot of extra material that is irrelevant, whereas the idea of defining a functor between $P$ and $\mathbf{FVect}$ has no such superfluous objects or morphisms and focuses on just the association of syntactic parses with semantic morphisms.

In this section, I begin by briefly describing functors in $\S$\ref{sub:functors}, before demonstrating in $\S$\ref{sub:from_product_categories_to_functors} that the mapping $M$ between pregroup types and vector spaces and the product category $P \times \mathbf{FVect}$ can be used to canonically define a functor between $P$ and $\mathbf{FVect}$.

\subsection{Functors} % (fold)
\label{sub:functors}

The simplest way to think of a functor is as a map between two categories $\mathbf{A}$ and $\mathbf{B}$ which associates each object $A$ in $ob(\mathbf{A})$ with some object $F(A)$ in $ob(\mathbf{B})$, and each morphism $f:A_1 \to A_2$ in $hom(\mathbf{A})$ with some morphism $F(f): F(A_1) \to F(A_2)$ in $hom(\mathbf{B})$. Functors must furthermore preserve the structure of $\mathbf{A}$ in $\mathbf{B}$ by ensuring that the following two equations hold for all morphisms $f$ and $g$ of $hom(\mathbf{A})$ and all objects $a$ of $ob(\mathbf{A})$:
\begin{enumerate}
    \item $F(id_A) = id_{F(A)}$, i.e.~if $A$ maps to some $F(A)$ then its identity morphism maps to that same object's identity morphism.
    \item $F(f \circ g) = F(f) \circ F(g)$, i.e.~if two morphisms $f$ and $g$ compose to some $h = f \circ g$ in $hom(\mathbf{A})$, then there must be some $F(h)$ in $hom(\mathbf{B})$ which is the composition of $F(f)$ and $F(g)$.
\end{enumerate}

Furthermore, a \emph{strict} monoidal functor between monoidal categories is such that the following equality holds for two such categories $\mathbf{A}$ and $\mathbf{B}$, equipped with monoidal tensors $\otimes_\mathbf{A}$ and $\otimes_\mathbf{B}$. For any objects $A$ and $B$ of $ob(\mathbf{A})$:
\[
F(A \otimes_\mathbf{A} B) \cong F(A) \otimes_\mathbf{B} F(B)
\]

To give a trivial example of a functor, consider the following two pregroups:
\begin{itemize}
    \item $P_1$, freely generated from the set $\{a,b\}$ with:
    \begin{itemize}
         \item the ordering relation $\leq^1$,
         \item the multiplicative operation $\cdot_1$,
         \item the adjoint maps $(-^1)^l$ and $(-^1)^r$,
         \item and the unit $1_1$.
     \end{itemize} 
    \item $P_2$, freely generated from $\{c,d,e\}$, with:
    \begin{itemize}
         \item the ordering relation $\leq^2$,
         \item the multiplicative operation $\cdot_2$,
         \item the adjoint maps $(-^2)^l$ and $(-^2)^r$,
         \item the unit $1_2$,
         \item and the stipulation that $c \leq^2 d$.
     \end{itemize}  
\end{itemize}

We can define a functor $F$ as follows:
\begin{itemize}
    \item $F(1_1) = 1_2$, $F(a) = c$, $F(b) = d$ (by the definition of functors between monoidal categories, we therefore also have $F(ab)=cd$, $F(aab) = ccd$, and so on).
    \item $F((-^1)^l_i) = (-^2)^l_{F(i)}$ and $F((-^1)^r_i) = (-^2)^r_{F(i)}$ for any $i$ in $P_1$. E.g.~$F((-^1)^l_a) = (-^2)^l_{c}$, so we map the adjoints of $a$ in $P_1$, such as $a^l$, to those of $a$'s ``image'' $c$ in $P_2$, such as $c^l$.
    \item $F(\leq^1_{(i,j)}) = \leq^2_{(F(i),F(j))}$ for any $i,j$ in $P_1$. E.g.~$F(\leq^1_{(ba^la,b)}) = \leq^2_{(dc^lc,d)}$, so effectively we map the reduction $ba^la \to b$ in $P_1$ to $cd^ld \to d$ in $P_2$.
\end{itemize}

The first condition for functors is satisfied by mapping identity morphisms of $P_1$ to those of $P_2$. The second is satisfied primarily through the natural transitivity of ordering relations in both categories.

The important point to note here is that no type in $P_1$ is mapped by $F$ to the type $c$ of $P_2$, or indeed to any compound types containing $c$ (e.g.~$ac$, $bca$, $cc$, etc.). Likewise, even though $F$ maps $a$ to $c$ and $b$ to $d$, the absence of a morphism $\leq^1_{(a,b)}$ in $P_1$ means that there exist ordering morphisms in $P_2$, such as $\leq^2_{(a,b)}$, which are \emph{not} images of morphisms in $P_1$ even though the objects they connect are images of objects in $P_1$. Functors can, in this sense, act like surjective mappings. They may, in this context, be seen as a way of explicitly embedding one mathematical structure into another (possibly larger) structure.

A functor is said to be `forgetful' if it maps complex objects in one category to simpler objects in another category, where the simpler objects are ``parts'' of the more complex objects of the original category. A trivial example is a functor $F$ between a category of pairs of atoms (of the form $(a,b), (b,d), \ldots$) and a category of atoms (e.g.~$a, b, c, \ldots$) which maps each pair in the original category to the atom corresponding to the left element of the pair in the simpler category, and morphisms between pairs to morphisms between the left elements of the pair. For example $F$ would map $(a,b)$ and $(a,c)$ to $a$, and some morphism $f: (a,b) \to (a,c)$ to the morphism $Ff : a \to a$. We can see here that $F$ `forgets' some of the information from the objects and morphisms of the original category by ignoring aspects of morphisms and objects relating to the right element of each pair.

Finally, functors can be interpreted as morphisms in the category of categories $\mathbf{Cat}$, and hence the usual morphism composition rules apply for functors. Explicitly, for categories $\mathbf{A}$, $\mathbf{B}$ and $\mathbf{C}$, the functors $F: \mathbf{A} \to \mathbf{B}$ and $G: \mathbf{B} \to \mathbf{C}$ define a functor $H = G \circ F: \mathbf{A} \to \mathbf{C}$.

% subsection functors (end)

\subsection{From Product Categories to Functors} % (fold)
\label{sub:from_product_categories_to_functors}

We have just seen that functors can be composed like morphisms to define a functor from a category $\mathbf{A}$ to a category $\mathbf{C}$ through composition, if we are provided with `intermediate' functors from $\mathbf{A}$ to $\mathbf{B}$ and $\mathbf{B}$ to $\mathbf{C}$. This is an important property of functors within the context of our desire to link syntax to semantics. If a functor is defined between a pregroup category $P$ and $\mathbf{FVect}$ (which, as will be discussed below, can be done for any pregroup category), then to use another grammatical formalism in lieu of pregroup grammars, it suffices to show that this other grammatical structure can be given categorical semantics as well, and that a functor can be defined from the relevant categorification of the grammar to some pregroup category. A functor from the new grammatical category to $\mathbf{FVect}$ then arises directly from composition. Putting this more formally: if it can be shown that there is a functor $F$ mapping pregroup types in some pregroup category $P$ to vector spaces in $\mathbf{FVect}$, and reduction morphisms in $P$ to linear maps in $\mathbf{FVect}$, then if some grammatical formalism (e.g.~a context free grammar, or combinatorial grammar, etc.) can be represented as a category $P'$ for which there exists a functor $G: P' \to P$, it follows that the types and analysis morphisms of $P'$ can be mapped directly to $\mathbf{FVect}$ by some functor $H = F \circ G : P' \to \mathbf{FVect}$.

Let us then consider how we might define a functor using a mapping $M$ between pregroup types and vector spaces, and the structure of the product category $P \times \mathbf{FVect}$. We begin by formally defining $M$ as follows. Let $\mathcal{T}$ be the set of basic pregroup types generating a pregroup $P$. This can be formalised by stating that any type $t$ in $P$ is written as
\[
t : w_1^{m_1} \ldots w_n^{m_n}
\] 
where $1 \leq n$, $w_i \in \mathcal{T}$ for all $i$, and $m_i$ is an optional adjoint marker for $w_i$ (e.g.~$l$, $ll$, $r$, etc.~, or no adjoint).

We associate, via $M$, a vector space $A$ to each element $a$ of the generating set $\mathcal{T}$, such that $M(a)=A$, without any commitments to the uniqueness of this association. This allows us to associate one vector space to several elements of the generating set, which might be desirable if, for example, we wish for nouns and noun phrases to have separate pregroup types (say $n$ and $\bar{n}$), but for words/phrases of both types to reside in the same vector space so that we can compare the noun ``dog'' with the phrase ``Mary's favourite pet''. 

We then recursively define the association of compound pregroup types with vector spaces as follows:
\begin{align*}
    & M(1) = \mathbb{R} \\
    & M(ab) = M(a) \otimes M(b)\\
    & M(a^r) = M(a^l) = M(a)^{\ast} = A^{\ast}
\end{align*}
for any types $a$ and $b$ in $ob(P)$, and where $(V_1 \otimes \ldots \otimes V_n)^\ast \cong (V_n \otimes \ldots \otimes V_1)$. Because the set of pregroup types is freely generated, we can view the objects of a category $M(P)$ as being freely generated by a set $\{M(x) \mid x \in \mathcal{T} \}$, such that any object of the category $M(t)$ can be written as follows:
\[
M(t) : M(w_1) \ldots M(w_n) \quad \forall i.[w_i \in \mathcal{T}]
\] 

To give an example, if $M(n)=N$ and $M(s)=S$, we can systematically generate the vector space for transitive verbs of type $n^rsn^l$ as being $M(n^rsn^l) = M(n^r) \otimes M(s) \otimes M(n^l) = N \otimes S \otimes N$.

Next, we consider the full subcategory $\mathbf{M}_{P\times \mathbf{FVect}}$ of $P\times \mathbf{FVect}$. A subcategory of a category $\mathbf{C}$ is a category $\mathbf{C'}$ where the objects of $\mathbf{C'}$ are objects of $\mathbf{C}$ (hence $ob(\mathbf{C'}) \subseteq ob(\mathbf{C})$), and likewise the morphisms of $\mathbf{C'}$ are also morphisms of $\mathbf{C}$ (hence $hom(\mathbf{C'}) \subseteq hom(\mathbf{C})$). A subcategory is full if for all $X, Y$ in $ob(\mathbf{C'})$, the sets $\{f | f:X \to Y \land f \in hom(\mathbf{C'})\}$ and $\{f | f:X \to Y \land f \in hom(\mathbf{C})\}$ are equal, i.e.~if all morphisms between $X$ and $Y$ in $\mathbf{C}$ are also morphisms in $\mathbf{C'}$. The subcategory $\mathbf{M}_{P\times \mathbf{FVect}}$ of $P\times \mathbf{FVect}$ is defined as follows:
\begin{itemize}
    \item For all $(a,A) \in ob(P\times \mathbf{FVect})$, $(a,A) \in \mathbf{M}_{P\times \mathbf{FVect}}$ if and only if $M(a) = A$.
    \item For all morphisms $(f, g): (x,X) \to (y,Y)$ in $hom(P\times \mathbf{FVect})$, $(f,g) \in hom(\mathbf{M}_{P\times \mathbf{FVect}})$ if and only if $(x,X)$ and $(y,Y)$ are in $ob(\mathbf{M}_{P\times \mathbf{FVect}})$, and for $f = f_1 \otimes \ldots \otimes f_n$ and $g=g_1 \otimes \ldots \otimes g_m$ the following conditions hold: $m=n$, and $g_i$ is a structural morphism of $\mathbf{FVect}$ if $f_i$ is a structural morphism of $P$.
\end{itemize}
The first condition ensures that we only consider the object pairings relevant to our task of associating syntax with semantics, and the second condition states that the subcategory is full, and places restrictions on which morphisms exist in the subcategory. We can verify that this category is still compact closed by the same reasoning used in $\S$\ref{sub:syntax_guides_semantics}, namely that if $f: x \to y$ is a structural morphism of $P$ and $g: X \to Y$ is a structural morphism of $\mathbf{FVect}$, then if $(x,X)$ and $(y,Y)$ are in $ob(\mathbf{M}_{P\times \mathbf{FVect}})$, $(f,g): (x,X) \to (y,Y)$ is also a structural morphism of $\mathbf{M}_{P\times \mathbf{FVect}}$, by virtue of the subcategory being full.

The map $M$ ensures that for every $a\in ob(P)$ there is a unique $(a,A) \in ob(\mathbf{M}_{P\times \mathbf{FVect}})$, and that for every morphism $f \in hom(P)$ there is a unique morphism $(f,g)\in hom(\mathbf{M}_{P\times \mathbf{FVect}})$ by virtue of the fact that all morphisms in $P$ are composed of structural morphisms, and that there is an injective map from the set of structural morphisms in $P$ to those of $\mathbf{FVect}$. It follows that there is a functor $F: P \to \mathbf{M}_{P\times \mathbf{FVect}}$ defined as above, using $M$. 

Furthermore, there is a trivial forgetful functor $G: \mathbf{M}_{P\times \mathbf{FVect}} \to \mathbf{FVect}$ which associates any object $(a,A) \in ob(\mathbf{M}_{P\times \mathbf{FVect}})$ to $A \in \mathbf{FVect}$, and any morphism $(f,g)\in hom(\mathbf{M}_{P\times \mathbf{FVect}})$ to a morphism $g \in \mathbf{FVect}$. It follows that the functor $G \circ F$ maps the objects and morphisms of $P$ to those of $\mathbf{FVect}$. I have therefore shown that from the product category $P \times \mathbf{FVect}$ and a mapping $M$ between pregroup types and vector spaces, a functor between the pregroup category $P$ and the category of vector spaces $\mathbf{FVect}$ can naturally be defined.
% subsection from_product_categories_to_functors (end)

% section functorial_passages (end)

\section{Supporting Context Free Grammars} % (fold)
\label{sec:supporting_context_free_grammars}

While the term ``context free grammar'' technically refers to any grammar which is context free in the language theoretic sense, computational linguists usually refer to a specific sort of phrase-structure-grammar when talking about ``Context Free Grammars'' (CFGs). In this sense, the expression refers to a well-known and well-studied syntactic formalism for analysing the phrase structure of sentences, formalised by Chomsky in \cite{chomsky1956three}. As a topic, it is covered in virtually any introduction to language theory, computational linguistics, or mathematical linguistics. Efficient parsing algorithms have been developed to recover the CFG structure from digital text (e.g.~CYK \cite{cocke1969programming,kasami1965efficient,younger1967recognition}, Earley \cite{earley1970efficient}), and a large number of corpora and other resources have been created to assist the learning of such generative grammars (e.g.~the Penn Treebank \cite{marcus1993building}). It is a popular tool for computational linguistics, and it would make little sense to develop a formalism for syntactically motivated compositional distributional semantics that does not offer the opportunity to leverage the vast amounts of work done on and with CFGs.

In this section, I introduce a way of integrating context free grammars into the DisCoCat formalism described in Chapter~\ref{cha:foundations_of_discocat}. I begin, in $\S$\ref{sub:context_free_grammar} by briefly introducing CFGs and specifically the sort of CFGs I will be discussing later in this section. In $\S$\ref{sub:defining_a_functor_cfg}, I then show how a CFG can be given a simple categorical structure, and how a functor from this category to $\mathbf{FVect}$ can be defined by showing that there exists at least one functor between every CFG-as-a-category and some pregroup category $P$, guided by the fact that pregroup grammars and CFGs are weakly equivalent \cite{Buszkowski2008}.

\subsection{Context Free Grammar} % (fold)
\label{sub:context_free_grammar}

In this section, I begin by giving a general definition of context free grammar, followed by the specification of some restrictions on CFGs which will be useful when giving CFGs categorical structure.

\subsubsection{General Definition} % (fold)
\label{ssub:general_definition}

Formally, a context free grammar is defined by four elements: 
\begin{itemize}
    \item a finite set $V$ of non-terminal symbols, which can be seen as being the syntactic types of words and phrases. In simple grammars, these are usually given intuitive names such as $NP$ for noun-phrases, $VP$ for verb phrases, $S$ for sentences, and so on.
    \item a set $\Sigma$ of terminals, which we can interpret, within the context of linguistics, as being the words of our language.
    \item a finite set $R$ of production rules, usually written in the form $A \Rightarrow B\ C$, $A \Rightarrow B$, $A \Rightarrow B\ C\ D$ and so on, where $B$, $C$ and $D$ may be elements of $V$ or $\Sigma$, but $A$ must be an element of $V$ since it `generates' the other symbols (and therefore is a non-terminal).
    \item the starting symbol of the grammar (from the set $V$). We will use $S$ as starting symbol in this section, but there is no obligation to do so.
\end{itemize}

Rules of the form $A \Rightarrow B\ C$ can be read as ``A generates the types B C'', or, going the other way, as saying that ``a B and a C combine to form an A''. On the one hand, the generative direction from ``top'' to ``bottom'' is usually followed when parsing a string to assign to each word a syntactic type. The ``bottom-to-top'' approach, on the other hand, is more akin to what we do with pregroups when the types of words are known, and the CFG can be used to describe the phrasal structure of a sentence. Going this way, the generative rules can be seen as substitution rules where the element on the left of the arrow is substituted for the elements on the right to determine the type of the phrase comprising them.

\begin{table}[ht]
\begin{center}
    \begin{tabular}{lcl}
    $S$ & $\Rightarrow$ & $NP\ VP$\\
    $VP$ & $\Rightarrow$ & $V_t\ NP$\\
    $NP$ & $\Rightarrow$ & $Det\ N$\\
    $Det$ & $\Rightarrow$ & $\textrm{the}$\\
    $N$ & $\Rightarrow$ & $\textrm{dog}|\textrm{cat}$\\
    $V_t$ & $\Rightarrow$ & $\textrm{chases}$
    \end{tabular}
\end{center}
\caption{A simple CFG.}
\label{tab:simpleCFG}
\end{table} 

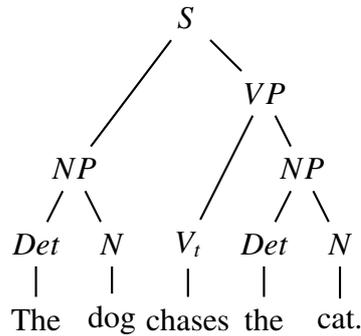
\begin{figure}[ht]
    \begin{center}
        \begin{tikzpicture}[thick]
\node (the1) at (0,0) {The};
\node (dog) at (1,0) {dog};
\node (chases) at (2,0) {chases};
\node (the2) at (3,0) {the};
\node (cat) at (4,0) {cat.};

\node (det1) at (0,1) {$Det$};
\node (n1) at (1,1) {$N$};
\node (vt) at (2,1) {$V_t$};
\node (det2) at (3,1) {$Det$};
\node (n2) at (4,1) {$N$};

\node (np1) at (.5,2) {$NP$};
\node (np2) at (3.5,2) {$NP$};

\node (vp) at (3,3) {$VP$};

\node (s) at (2,4) {$S$};

\draw (s) -- (np1);
\draw (s) -- (vp);

\draw (vp) -- (np2);
\draw (vp) -- (vt);

\draw (np1) -- (det1);
\draw (np1) -- (n1);
\draw (np2) -- (det2);
\draw (np2) -- (n2);

\draw (det1) -- (the1);
\draw (det2) -- (the2);
\draw (n1) -- (dog);
\draw (n2) -- (cat);
\draw (vt) -- (chases);

\end{tikzpicture}
    \end{center}
    \caption{Sample parse tree for the CFG in Table~\ref{tab:simpleCFG}.}
    \label{fig:simpleCFGparse}
\end{figure}

To give an example, a simple CFG is shown in Table~\ref{tab:simpleCFG}. The set of non-terminal symbols is $V = \{S,\,NP,\,VP,\,V_t,\,Det\}$, the set of terminal symbols is $\{\textrm{the},\, \textrm{dog},\, \textrm{cat},\, \textrm{chases}\}$, and the production rules are as shown. The $N \Rightarrow \textrm{dog}|\textrm{cat}$ is used as shorthand for the definition of several rules (namely $N \Rightarrow \textrm{dog}$ and $N \Rightarrow \textrm{cat}$) on one line. A parse tree for the sentence ``The dog chases the cat.'' is shown in Figure~\ref{fig:simpleCFGparse}. Such parse trees are generated from the grammar's production rules by interpreting each rule as the formation of a tree with, as root, the symbol on the left of the arrow; and with, as children of the root, the symbols on the right of the arrow, themselves being the roots of subtrees of one or more nodes. In this case, we see that `the' and `dog', as well as `the' and `cat', both being determiner-noun pairs $DT\ N$, each combine to form a small tree with a noun phrase $NP$ as a root. The second of these noun phrases (``the cat'') combines with the transitive verb $V_t$, namely ``chases'', to form a tree with a verb phrase $VP$ as a root. This $VP$ combines with the first $NP$ to produce a tree with a sentence $S$ as root, which is the whole parse tree for this sentence.

As a notational convenience, we state that if for some symbols $A,\,B_1,\, \ldots,\, B_n$ there exists a set of rules such that $A$ generates $B_1\ \ldots\ B_n$, we write $A \Rightarrow^+ B_1\ \ldots\ B_n$ and call this a substitution sequence. We treat such substitution sequences as being uniquely characterised by a set of production rules used to generate $B_1\ \ldots\ B_n$ from $A$. In this sense, each distinct $A \Rightarrow^+ B_1\ \ldots\ B_n$ stands for a different subtree with $A$ as root and $B_1,\, \ldots,\, B_n$ as leaves. Naturally, any two substitution sequences $A \Rightarrow^+ B$ and $B \Rightarrow^+ C$ can be combined to form a unique substitution sequence $A \Rightarrow^+ C$.

To give a short example, in the sample CFG in Table~\ref{tab:simpleCFG}, we can see that there exists a substitution sequence $S \Rightarrow^+ NP\ V_t\ NP$ because $S \Rightarrow NP\ VP$ and $VP \Rightarrow V_t\ NP$.

% subsubsection general_definition (end)

\subsubsection{Restrictions} % (fold)
\label{ssub:restrictions}

While the above description of CFGs suffices to fully define the basic mechanics and components of phrase-structure CFGs, I add a further four restrictions on the form of the CFGs we will consider in this section. The reason for these restrictions will be explained at the end of $\S$\ref{ssub:from_translation_dictionaries_to_functors}.

First, we will only consider pseudo-proper CFGs. A CFG is pseudo-proper if:
\begin{itemize}
    \item Every non-terminal except $I$ is on the left-hand side of some production rule;
    \item For every terminal or non-terminal, there is some sequence of production rules such that the root symbol $S$ generates a set of symbols including that terminal or non-terminal;
    \item There are no cycles, i.e.~for any symbol $A$ there is no sequence of production rules $A \Rightarrow^+ A$ beginning with that symbol on the left and ending with that symbol as the sole symbol on the right.
    \item Any sequence of symbols containing the empty string $\varepsilon$ is equivalent to the same sequence of symbols with all instances of $\varepsilon$ removed. Any sequence of symbols is trivially equivalent to itself.
\end{itemize}
These rules entail that for any sequence of symbols, there is at most a finite set of finite trees with $S$ as root that form valid parses of that sequence. I call these CFGs pseudo-proper because of the last rule about $\varepsilon$. The definition of proper CFGs replaces this rule with one stating that the empty string will not be on the right-hand side of any production rule, as a way of ensuring that there are a finite number of trees with a finite sequence of symbols as leaves and $S$ as a root. The modification of this rule which I introduced seeks to preserve this property without removing $\varepsilon$ from our list of symbols, as it will be useful in the next section. The last rule instead simply equates all sequences of symbols which are identical once $\varepsilon$ is removed, thereby treating it as an `invisible' symbol potentially present in any sequence, and effectively eliminating it from the parsing process.

Second, we only consider CFGs that produce binarised parsed trees, meaning that every production rule has at least one symbol on the right of the arrow, and at most two. We do not count the `invisible' symbol $\varepsilon$ in this restriction. This restriction can be applied without loss of generality, as any CFG can be modified to fit this constraint by introducing a finite number of additional symbols while still recognising the same language. For any rules of the form
\[
A \Rightarrow B_1\ B_2\ \ldots\ B_n
\]
I introduce a new symbol $C_n$ and a rule
\[
C_n \Rightarrow B_1\ B_2\ \ldots\ B_{n-1}
\]
and produce an amended version of the first rule
\[
A \Rightarrow B_1\ C_n
\]
We apply this rewrite rule recursively to the newly produced rule in order to produce symbols $C_{n-1}$, $C_{n-2}$, etc.~and associated production rules, until we obtain some symbol $C_i$ which has as production rule
\[
C_i \Rightarrow B_1\ B_2
\]
at which point the original rule has been fully binarised. To give a short example, assume we have the rule
\[
A \Rightarrow B\ C\ D\ E
\]
in our CFG. We begin by rewriting it as
\[
A \Rightarrow F\ E
\]
where $F$ is a newly introduced symbol. We then introduce the rule
\[
F \Rightarrow B\ C\ D
\]
which is not binarised, so we must rewrite it as
\[
F \Rightarrow G\ D
\]
where $G$ is newly introduced symbol. We then introduce the rule
\[
G \Rightarrow B\ C
\]
which is binarised, so the rewrite process is complete. We have therefore binarised the rule
\[
A \Rightarrow B\ C\ D\ E
\]
by replacing it with the set of rules
\begin{align*}
     & A \Rightarrow F\ E\\
     & F \Rightarrow G\ D\\
     & G \Rightarrow B\ C
 \end{align*}
This rewrite system is part of the process of translating any grammar to Chomsky Normal Form (CNF), although the thus-produced binarised CFG is not technically in CNF, as we allow rules for the form
\[
A \Rightarrow B
\]
where $B$ is a non-terminal (whereas a CFG in CNF has only binary production rules with non-terminals on the right, and unary production rules with terminals on the right). The point of binarising a grammar will be made clear in $\S$\ref{sub:defining_a_functor_cfg} when we discuss the categorical structure of CFGs, but full-blown translation into CNF is not required.

Third, we will disallow production rules which have a mix of terminals and non-terminals, excluding $\varepsilon$, on the right-hand side of the arrow. This assumption can be made without loss of generality as any rule with such a mixture of non-terminals and terminals can be rewritten by replacing each terminal symbol with a new non-terminal, and introducing a new production rule in which the newly created non-terminal produces the terminal symbol.

Finally, the fourth restriction is that for any non-terminal which appears on the left-hand side of a production rule with non-terminals on the right-hand side, there is no rule with that symbol on the left-hand side and non-terminals, excluding $\varepsilon$, on the right-hand side. Again, this restriction can be applied without loss of generality: we modify any rule violating the restriction, and of the form
\[
A \Rightarrow n
\]
where $n$ is one or more non-terminals (in the form $n_1|n_2|\ldots|n_k$), by rewriting it as
\[
A \Rightarrow N
\]
where $N$ is a new non-terminal symbol, and adding the rule
\[
N \Rightarrow n
\]
We call such non-terminals, which are on the left-hand side of production rules with terminals on the right-hand side, \emph{basic types}. All other non-terminals will be referred to as \emph{complex types}. Rules with basic types on the left-hand side will be referred to as \emph{terminal rules}, while rules with complex types on the left-hand side will be referred to as \emph{non-terminal rules}. We call $R_{NT}$ the subset of $R$ consisting of all the non-terminal rules.

% subsubsection restrictions (end)

% subsection context_free_grammar (end)

\subsection{CFGs as Categories} % (fold)
\label{sub:cfgs_as_categories}

To use CFGs in the DisCoCat formalism instead of pregroup grammars, two things must be done. First, we need to show that CFGs can be given a categorical interpretation. Second, we need to show that there is a functor between any CFG-as-a-category and the category $\mathbf{FVect}$.

Here, I show that any CFG satisfying the restrictions described above can be interpreted as a category $\mathbf{CFG}$:
\begin{enumerate}
    \item Let $V$ be the set of non-terminals of our CFG.
    \item Let $ob(\mathbf{CFG})$ be a set such that $V \subset ob(\mathbf{CFG})$ and $\varepsilon \in ob(\mathbf{CFG})$.
    \item Let the sequence of symbols of form $A\ B$ found on the right-hand side of production rules in a CFG be the concatenation of the objects $A$ and $B$. 
    \item Let the concatenation operation for types be the associative monoidal tensor $\otimes$ in the category, such that for any $A$ and $B$ in $ob(\mathbf{CFG})$, the object $A \otimes B$ is in $ob(\mathbf{CFG})$.
    \item Let $hom(\mathbf{CFG})$ be a set of arrows, where for any substitution sequence $A \Rightarrow^+ B$ there is one unique arrow $f: B \to A$ in $hom(\mathbf{CFG})$.
    \item Let $A \otimes \varepsilon \cong A \cong \varepsilon \otimes A$ for any $A \in ob(\mathbf{CFG})$, by the rule for equality of sequences containing $\varepsilon$, defined above.
    \item Let self equivalence of sequences of symbols define an arrow from each object to itself.
\end{enumerate}

I hypothesise that these stipulations define a monoidal category $\mathbf{CFG}$. Stipulations 1--4 define the set of objects in the category. Stipulation 5 defines the set of morphisms in the category. Stipulations 4 and 6 state that if $\mathbf{CFG}$ is a category, then it is a monoidal category with $\varepsilon$ as unit. To confirm that this is a valid definition of a category, we verify that it satisfies the axioms outlined in $\S$\ref{sub:the_basics}:
\begin{itemize}
    \item By the definition of substitution sequences, any two sequences $A \Rightarrow^+ B$ and $B \Rightarrow^+ C$ can be combined to form a substitution sequence $A \Rightarrow^+ C$. This combination corresponds to arrow composition: for $A \Rightarrow^+ B$ there is some arrow $f: B \to A$; for $B \Rightarrow^+ C$ there is some arrow $g: C \to B$; for $A \Rightarrow^+ C$ there is some arrow $h: C \to A$, and the combination of substitution sequences states $f \circ g = h$.
    \item Substitution sequences are uniquely characterised by the production rules involved, hence if $A \Rightarrow^+ B$ and $B \Rightarrow^+ C$ combine to form a sequence $A \Rightarrow^+ C$, and $B \Rightarrow^+ C$ and $C \Rightarrow^+ D$ combine to form a sequence $B \Rightarrow^+ D$, then the combinations of $A \Rightarrow^+ B$ with $B \Rightarrow^+ D$ and $A \Rightarrow^+ C$ with $C \Rightarrow^+ D$ form the same sequence $A \Rightarrow^+ D$. With morphisms $f: B \to A$ as $A \Rightarrow^+ B$, $g: C \to B$ as $B \Rightarrow^+ C$ and $h: D \to C$ as $C \Rightarrow^+ D$, then $(f \circ g) \circ h = f \circ (g \circ h)$, and therefore arrow composition is associative.
    \item The self equivalence arrows $f_A: A \to A$ from any object $A$ to itself correspond to replacing a sequence of symbols with itself. Doing this before applying a substitution sequence $A \Rightarrow^+ B$ associated with some morphism $g: B \to A$ does not change the outcome of the substitution, hence $g \circ f_A = g$. Furthermore, because the replacing of the sequence of symbols $B$ with itself after the application of a substitution sequence $A \Rightarrow^+ B$ does not change the outcome of the substitution sequence, it follows that $f_B \circ g = g$. It follows that self equivalence arrows satisfy the identity arrow axiom.
\end{itemize}
So we have a collection of objects, a collection of morphisms, and a composition operation, all of which satisfy the three axioms defining a category. Furthermore, we have an identity object and a monoidal tensor, so it therefore follows that any $\mathbf{CFG}$ defined from a context free grammar satisfying the restrictions outlined in $\S$\ref{sub:context_free_grammar} is a monoidal category.
% subsection cfgs_as_categories (end)

\subsection{Defining a Functor} % (fold)
\label{sub:defining_a_functor_cfg}

We now turn to the task of defining a functor between some CFG-as-a-category $\mathbf{CFG}$ and $\mathbf{FVect}$. Contrary to what we did with pregroup categories $P$ where we relied on the compact closed nature of both $P$ and $\mathbf{FVect}$ to define a functor, things are not as simple with $\mathbf{CFG}$. While with pregroup categories, the structure of pregroup types dictates that a transitive verb of form $n^rsn^l$ be mapped to a vector space $N \otimes S \otimes N$, nothing about the transitive verb `type' in CFGs tells us that we should map $V_t$ to something of the structure $N \otimes S \otimes N$ rather than any other object in $\mathbf{FVect}$. Likewise for arrows: the structure of pregroup reduction morphisms in $P$ dictates the structure of the morphisms they are mapped to in $\mathbf{FVect}$, such that for example $\epsilon^r_n \otimes 1_s \otimes \epsilon^l_n: n^rsn^l \to s$ in $hom(P)$ should map to $\epsilon_N \otimes 1_S \otimes \epsilon_N: N \otimes S \otimes N \to S$ in $hom(\mathbf{FVect})$ rather than some other linear map between $N \otimes S \otimes N$ and $S$. In short, contrary to pregroups which directly reveal their `functional structure' for both objects and reductions, CFGs conceal it, and an additional step must be taken before defining a functor between some category $\mathbf{CFG}$ and $\mathbf{FVect}$. Here, I will show how for every $\mathbf{CFG}$ there exists one or more pregroup categories $P_i$ such that a functor $F_i$ can be systematically defined between $\mathbf{CFG}$ and each $P_i$, thereby allowing functorial passage from $\mathbf{CFG}$ to $\mathbf{FVect}$ via some $P_i$. I do this by first presenting an algorithm translating any CFG with $|R_{NT}|$ non-terminal production rules into at most $2^{|R_{NT}|}$ pregroup grammars. Second, I will show that each such CFG-to-pregroup grammar translation constitutes a functorial passage between a CFG-as-a-category $\mathbf{CFG}$ and some pregroup category $P$.

\subsubsection{CFG to Pregroup Translation} % (fold)
\label{ssub:cfg_to_pregroup_translation}

\paragraph{Inner Structure} To translate a CFG to a pregroup, we must ``reveal'', or rather make assumptions about, the inner structure of types and generative rules. In other words, we wish to treat a production rule as a type reduction, and spell out the mechanisms of such a reduction, in order to add compact closure to the categorical representation of the grammar. Recalling that for each production rule in a CFG of the form
\[
A \Rightarrow B\ C
\]
there is some arrow $f$ in the categorification of that CFG such that $f: B \otimes C \to A$, the simplest way to reinterpret this in a compact closed setting is to assume that one of the types $B$ or $C$ contains the adjoint of the other, and the type of $A$. Stating this explicitly, let's assume that CFG type $A$ corresponds to some pregroup type $a$. Then either $C$ corresponds to some pregroup type $c$ and $B$ to some compound type $ac^l$, in which case $f: B \otimes C \to A$ in $hom(\mathbf{CFG})$ corresponds to $1_a \otimes \epsilon^l_c : (ac^l)c \to a$ in $hom(P)$; or $B$ corresponds to some pregroup type $b$ and $C$ to some compound type $b^ra$, in which case $f: B \otimes C \to A$ in $hom(\mathbf{CFG})$ corresponds to $\epsilon^r_b \otimes 1_a : b(b^ra) \to a$ in $hom(P)$. 

It is important to note that these two ways of translating a production rule make no assumptions about whether the types which are \emph{not} inferred to be compound types---namely $a$ and $c$ in the first case, and $a$ and $b$ in the second---are compound or simple types themselves. For example, if we have already inferred that $A$ must be of some pregroup type $a = d^re$, then the type of $B$ in the first case will be $ac^l = d^rec^l$, and the type of $C$ in the second case will be $b^ra = b^rd^re$. Similarly, if we had previously inferred the pregroup type of $C$ to be $c=po^l$, then the type of $B$ in the first case becomes $ac^l = a(po^l)^l$, and similarly for the type of $C$ in the second case if that of $B$ had previously been inferred.

 \paragraph{Simple Type Inference} Since each production rule in a binarised parse tree has at most two elements on the right-hand side, and each such rule generates two options in terms of pregroup interpretation of the CFG types, we immediately get the result that each CFG with $|R_{NT}|$ production rules generates at most $2^{|R_{NT}|}$ different pregroup translations using this process. At first blush, it seems that all we need to do is set some convention such as always interpreting the left element of the pair of types on the right-hand side of each production rule as being a compound type in order to obtain a canonical pregroup translation of a CFG. For example, we could always treat a generative rule of the form
\[
A \Rightarrow B\ C
\]
as a pregroup reduction of the form
\[
(ac^l)c \to a
\]
hence interpreting $A$ as some type $a$, $B$ as some type $ac^l$ and $C$ as some type $C$. However, translating a CFG into a pregroup is not quite so straightforward, as there are two cases where a rule of the form
\[
A \Rightarrow B\ C
\]
may only be translated in one way.

The first case, where $A=B$ or $A=C$, is fairly trivial. If $A=B$, then we cannot interpret $B$ as $ac^l$, since this would entail $a = ac^l$, which generates an infinite type if $c \neq 1$, as $a = ac^l = ac^lc^l = ac^lc^lc^l$ and so on. Hence if $A=B$, then the rule must be interpreted as $a (a^ra) \to a$, and so $B$ must be translated as $a$, and $C$ must therefore be translated as $a^ra$. By a similar argument, if $A=C$, then the rule must be interpreted as $(aa^l)a \to a$, $C$ must be translated as $a$, and $B$ must therefore be translated as $aa^r$.

The second case is a little more subtle, and regards the fairly frequent situation where a non-terminal symbol appears on the right-hand side of more than one rule. Let us assume that the following two rules are in our CFG:
\begin{align*}
    &A \Rightarrow B\ C\\
    &D \Rightarrow B\ E
\end{align*}
If we infer from the first rule that the pregroup types of $A$ and $C$ are $a$ and $c$ respectively, and that $B$ has the compound type $ac^l$, then we cannot make a new type inference for $B$ in the second case whereby $B$ is a compound type. If we could, then we would, assuming $D$ and $E$ have pregroup types $d$ and $e$ respectively, obtain the following equality:
\[
ac^l = de^l
\]
which holds only for the case where $a=d$ and $c=e$. A similar argument holds for $B$ in the pairings:
\begin{align*}
    & A \Rightarrow B\ C \quad \text{and} \quad D \Rightarrow E\ B\\
    & A \Rightarrow C\ B \quad \text{and} \quad D \Rightarrow B\ E\\
    & A \Rightarrow C\ B \quad \text{and} \quad D \Rightarrow E\ B
\end{align*}
thereby covering all cases without loss of generality.

\paragraph{Type Inference Algorithm} Considering both of these cases and formalising them, we can produce an algorithm, shown in Figure~\ref{fig:gen-pregroup}, which takes a CFG and outputs a set of dictionaries each mapping the symbols of the CFG to pregroup types in a pregroup grammar $P$. These dictionaries then allow for the production rules of the CFG to be interpreted as reduction rules in $P$.

\begin{figure}[h!]
\fbox{\begin{minipage}{\textwidth}
\begin{enumerate} 
\item Let $D_0$ be a dictionary mapping each non-terminal $X$ in a CFG to a unique symbol $D_0[X] = x$. Let $D_0[\varepsilon] = 1$.
\item Let $B$ be a list initially containing only $D_0$.
\item Let $R_{NT}$ be a list of CFG production rules named $R_1,\,R_2,\,\ldots$
\item If $R_{NT}$ is not empty, remove the first $R_i$ element for $R$, otherwise terminate and return $B$.
\item Let $B'$ be an empty list. For each $D_j \in B$ follow these steps:
\begin{enumerate}
    \item Let $D_k$, $D_m$ be new dictionaries.
    \item If $R_i$ is of the form $A \Rightarrow B$, then let $D_k[X] = D_j[X] \langle D_j[B]:=D_j[A] \rangle$ for all $X$ in $D_j$, where $v\langle w:=y \rangle$ means that all instances of $w$ in $v$ are to be replaced with $y$, and add $D_k$ to $B'$.
    \item If $R_i$ is of the form $A \Rightarrow B\ C$ and $D_k[B] = D_0[B]$ and $B \neq A$, then let $D_k[X] = D_j[X] \langle D_j[B]:=D_j[A] \cdot D_j[C]^l \rangle$ for all $X$ in $D_j$ and add $D_k$ to $B'$.
    \item If $R_i$ is of the form $A \Rightarrow B\ C$ and $D_k[C] = D_0[C]$ and $C \neq A$, then let $D_m[X] = D_j[X] \langle D_j[C]:=D_j[B]^r \cdot D_j[A] \rangle$ for all $X$ in $D_j$ and add $D_m$ to $B'$.
\end{enumerate}
\item Let B = B'. Go to step 4.
\end{enumerate}
\end{minipage}}
\caption{Procedure for translating a CFG with $|R_{NT}|$ production rules into $k \leq 2^{|R_{NT}|}$ pregroup grammars.}
\label{fig:gen-pregroup}
\end{figure}

Let's unpack the algorithm shown in Figure~\ref{fig:gen-pregroup} a little. Step 1 defines an initial injective mapping from CFG symbols to hypothetical types in a pregroup, assuming that each CFG non-terminal stands for a unique pregroup type. This mapping, represented as a type dictionary, is used to intialise the algorithm, but also during the running of the algorithm to verify whether or not a type has been already been set by a rule (i.e.~whether or not the pregroup type assigned to a CFG non-terminal has been changed since the beginning of the algorithm). We also assign the empty string $\varepsilon$ to the pregroup unit $1$, an assignment which will not be changed during the algorithm.

Step 2 defines a boundary list. As the algorithm is effectively constructing a tree, this list can be seen as containing all the leaves in the tree at any particular point in the algorithm's running. Step 3 defines a list of production rules $R_{NT}$ which we will use as a stack, and which initially contains all the non-terminal rules in the CFG. Step 4 forms the beginning of a loop which runs while $R_{NT}$ is not empty, and which at each iteration pops a new production rule off the $R_{NT}$ stack and runs the steps which follow. If the stack is empty, it returns the boundary list $B$ which contains the leaves of the fully constructed tree.

Steps 5 is the body of a loop run for each new rule $R_i$ popped off the $R_{NT}$ stack, iterating over the CFG-to-pregroup type mapping dictionaries in the boundary $B$. It begins by creating an empty list $B'$ which will become the new boundary (as we are turning leaves in the old boundary into branch nodes). Step 5a creates two new empty dictionaries $D_k$ and $D_m$ for each dictionary $D_j$ in the old boundary. If the $R_i$ is a unary rule (one non-terminal), step 5b is run. Step 5b interprets rules of the form $A \Rightarrow B$ as mappings $b \to a$ where $b$ and $a$ are obtained from $D_j[B]$ and $D_j[A]$. It copies all the mappings in $D_j$ to the new dictionary $D_k$ while replacing every instance of $b$ in the value of $D_k$ with $a$. This new dictionary is added to the new boundary $B'$, discarding $D_m$. If the rule is binary, steps 5c and 5d are run. Step 5c interprets rules of the form $A \Rightarrow B\ C$ as mappings $(ac^l) a \to a$ where $c$ and $a$ are obtained from $D_j[C]$ and $D_j[A]$. If non-terminals $B$ and $A$ do not use the same symbol (i.e.~$B \neq A$), and if the pregroup type of $B$ has not already been defined by some other rules (i.e.~$D_k[B] = D_0[B]$, namely if the pregroup interpretation of $B$ is still the same as the initial assignment in $D_0$), then the mappings of $D_j$ are copied to the new dictionary $D_k$, replacing all instances of $b = D_0[B]$ with $ac^l$, and then adding $D_k$ to the new boundary list $B'$. Similarly, step 5c interprets rules of the form $A \Rightarrow B\ C$ as mappings $b b^ra \to a$ where $b$ and $a$ are obtained from $D_j[B]$ and $D_j[A]$. If non-terminals $C$ and $A$ do not use the same symbol (i.e.~$C \neq A$), and if the pregroup type of $C$ has not already been defined by some other rules (i.e.~$D_k[C] = D_0[C]$, namely if the pregroup interpretation of $C$ is still the same as the initial assignment in $D_0$), then the mappings of $D_j$ are copied to the new dictionary $D_m$, replacing all instances of $c = D_0[C]$ with $b^ra$, and then adding $D_m$ to the new boundary list $B'$.

Finally, step 6 sets the boundary $B$ to the newly produced boundary list $B'$ by redefining $B$ as being equal to $B'$. The algorithm then returns to step 4.

Revisiting this algorithm in even more general terms, steps 1--3 initialise a tree construction algorithm. Step 4 creates a loop over non-terminal rules in the CFG, creating a new set of nodes to be added to leaves of the tree. And step 5 does the actual work of creating new nodes for each leaf of the tree, the number of which depends on the structure of the CFG rule being considered, and whether or not the types being inferred have already been inferred by a previous rule, or whether there are restrictions on what inferences can take place. Step 6 completes the loop started in step 4.

\paragraph{Algorithm Output} The thus constructed tree starts with a single type dictionary assuming every CFG terminal corresponds to a unique atomic pregroup type. As we progress down the tree, nodes correspond modified copies of their parents where the pregroup structure of CFG types is progressively inferred by considering the pregroup interpretations of CFG rules. Each branch node has one or more children generated by different interpretations of the CFG rule being considered at that depth. Left children correspond to the inference made from the interpretation of unary rules and of binary rules where the left element of the right-hand side is interpreted as a compound type; while right children correspond to the inference made from the interpretation of binary rules where the right element of the right-hand side is interpreted as a compound type. The leaves of the tree are the type dictionaries inferred once all the CFG non-terminal rules have been considered.

\paragraph{Algorithm Complexity}

Let $m = |R_{NT}|$ be the number of production rules in the grammar being processed by this algorithm, and $n = |V|$ be the number of non-terminals. The algorithm creates a binary tree of depth $m$ where each node costs $n$ operations to construct. The time complexity of the algorithm is therefore $O(m^2n)$.

The algorithm can be modified to run in $O(mn)$ time by skipping to step 6 once one of steps 5b, 5c, or 5d has been executed (i.e.~produced a new leaf). This corresponds to a dynamic algorithm descending the tree produced by the full-blown algorithm by following the left-most path.

\paragraph{Algorithm Termination}

As a byproduct of the complexity analysis, it is easy to see that the algorithm is guaranteed to terminate: each loop in Step 5 changes the definition of exactly one of the remaining unchanged types, and hence the loop will run at most $n$ times before terminating the algorithm. The algorithm is furthermore guaranteed to succeed if the CFG satisfies the restrictions presented above, as is discussed at the end of this section.

% subsubsection cfg_to_pregroup_translation (end)

\subsubsection{From Translation Dictionaries to Functors} % (fold)
\label{ssub:from_translation_dictionaries_to_functors}

We now complete this section by showing that each of these dictionaries $D_i \in B$ is sufficient to define a functorial passage $F_i : \mathbf{CFG} \to P_i$ from the categorification $\mathbf{CFG}$ of a CFG to some pregroup $P_i$ associated with $D_i$. Each dictionary $D_i$ assigns to each CFG non-terminal symbol a pregroup type in some pregroup $P_i$, freely generated by the set of atomic symbols in the dictionary $\{D_i[X] | X \in D_i \land D_i[X] = D_0[X]\}$. Therefore we obtain our functorial map for objects:
\[
F_i : ob(\mathbf{CFG}) \to ob(P_i) :: X \mapsto D_i[X]
\]
which covers symbol concatenation as follows:
\[
F_i : ob(\mathbf{CFG}) \to ob(P_i) :: X \otimes_{\mathbf{CFG}} Y \mapsto D_i[X] \otimes_{P_i} D_i[Y]
\]
Therefore if some $D_i$ defines
\begin{align*}
    & D_i[V_t] = \bar{n}^rs\bar{n}^l\\
    & D_i[N] = n\\
    & D_i[Det] = \bar{n}n^l
\end{align*}
then the CFG parse in of a sentence such as ``The dog chased the cat'', namely $S \Rightarrow^{+} Det\ N\ V_t\ Det\ N$, would translate via $F_i$ to $(\bar{n}n^l)n(\bar{n}^rs\bar{n}^l)(\bar{n}n^l)n$ in $P_i$, which reduces to $s$.

This brings us to the second aspect of the functor, which is the map between morphisms which would allow us to formalise how a parse in a CFG maps to a reduction operation in $P_i$. We essentially get this ``for free'' by the uniqueness up to isomorphism of morphisms between any two objects in any pregroup category $P_i$, since the morphisms stand for ordering relations. The idea is then that for any morphism $f: A \to B$ in $hom(\mathbf{CFG})$ we map $f$ to the unique morphism $g = F(f): F(A) \to F(B)$ in $P_i$. This presupposes that there is a morphism $g: F(A) \to F(B)$ in $hom(P_i)$ when there is some $f: A \to B$ in $hom(\mathbf{CFG})$, but this is precisely what the algorithm I presented here ensures: we assign pregroup types to CFG non-terminals by exploiting the fact that whatever the concatenation of symbols on the right of the production rule is translated as reduces to whatever the translation of the symbol on the left is. This is to say that for every concatenation of CFG symbols in the grammar there is some morphism in $P_i$ which stands for the reduction of their corresponding pregroup types.

Throughout this section we have been talking about non-terminals. However, an essential part of a pregroup grammar is the type dictionary for terminal symbols which assigns to each word in our language one or more pregroup types. We get this type dictionary for free from the pregroup translation dictionaries defined above and the terminal rules in our CFG. We simply state that for every map $D_i$ translating the non-terminals and productions of some context free grammar into the types and reductions of some pregroup grammar $P_i$ there exists a term-to-type dictionary $T_i$. This term-to-type dictionary is defined as follows: for any terminal rule in the CFG of the form
\[
A \Rightarrow w_1|w_2|\ldots|w_n
\]
producing one or more terminal symbols from a non-terminal $A$, the pregroup type of those non-terminals is simply the type of the non-terminal producing them; in other words:
\[
T_i[w_1] = D_i[A] \quad T_i[w_2] = D_i[A] \quad \ldots \quad T_i[w_n] = D_i[A]
\]

To conclude this section, let's examine why the restrictions on CFGs presented in $\S$\ref{ssub:restrictions} were helpful in defining this algorithm and functorial passage:
\begin{enumerate}
    \item We only considered pseudo-proper CFGs so that:
    \begin{itemize}
        \item All non-terminals (except $I$) are on the left-hand side of some rule, allowing us to infer the type of all non-terminals with complex types, and produce the type dictionaries for the basic types.
        \item All symbols are connected to the root $S$ by some sequence of rules, ensuring once again that each type can be inferred.
        \item The prohibition of cycles ensures that the algorithm does not reach a deadlock.
        \item Equivalence under concatenation with $\varepsilon$ allows us to use $\varepsilon$ as special `empty' type, purely so it can be mapped to the unit object in a pregroup category.
    \end{itemize}
    \item Only considering binarised rules allowed for a fairly simple algorithm. This isn't necessary \emph{per se}, but was set as a restriction principally out of convenience.
    \item Disallowing production rules with a mix of terminals and non-terminals on the right-hand side served to separate type inference from the creation of a type dictionary mapping words to sets of basic types.
    \item The restriction prohibiting a symbol from being on the left-hand side of both rules with non-terminals on the right and rules with terminals on the right served to create a distinction between basic and complex types, which was, once again, done to separate type inference from the production of the type dictionary.
\end{enumerate}

% subsubsection from_translation_dictionaries_to_functors (end)

% subsection defining_a_functor_cfg (end)

% section supporting_context_free_grammars (end)

\section{Supporting Lambek Grammar} % (fold)
\label{sec:supporting_lambek_grammar}

Another grammatical formalism which might be useful to support in the DisCoCat framework is the categorial grammar developed by Lambek in the 50s, generally referred to as Lambek Grammar. Aside from showcasing the ability of DisCoCat to adapt to a wide-range of grammatical systems, supporting Lambek Grammar has the additional advantage of laying groundwork for supporting a more expressive categorial grammar, namely Combinatorial Categorial Grammar (CCG), which I will discuss in $\S$\ref{sec:supporting_combinatory_categorial_grammar}. In this section, I will briefly present Lambek Grammar, and the sort of category it will be represented as, before discussing how a functorial passage can be defined from this category to the category of vector spaces $\mathbf{FVect}$.

\subsection{Lambek Grammar} % (fold)
\label{sub:lambek_grammar}

Lambek Grammar (LG), first described in \cite{Lambek1958}, is a context free categorial grammar consisting of an infinite set of types, and a set of type operations, which allow for the analysis of syntax to be completed in a manner akin to logical proofs.

Types in LG are recursively defined as follows:
\begin{itemize}
    \item Atomic types are represented as single letters $A$, $B$, \ldots
    \item $A\backslash_L B$ is a type if $A$ and $B$ are types\footnote{Here, I write backslashes with the subscript $L$ to indicate that I am following Lambek's slash notation, which differs from the slash notation commonly used in CCG, discussed in $\S$\ref{sec:supporting_combinatory_categorial_grammar}.}.
    \item $A/B$ is a type if $A$ and $B$ are types.
    \item $A \cdot B$ is a type if $A$ and $B$ are types, with $\cdot$ being associative.
\end{itemize}

\begin{table}[ht!]
\begin{center}
    \begin{tabular}{ll}
    \AxiomC{}
    \LeftLabel{(Axiom)}
    \UnaryInfC{$A \vdash A$}
    \DisplayProof
     & 
     \AxiomC{$\Gamma, A, \Pi \vdash B$}
     \AxiomC{$\Delta \vdash A$}
     \LeftLabel{(Cut)}
     \BinaryInfC{$\Gamma, \Delta, \Pi \vdash B$}
     \DisplayProof\\
     & \\
     \AxiomC{$\Gamma, A, B, \Pi \vdash C$}
     \LeftLabel{(L$\cdot$)}
     \UnaryInfC{$\Gamma, A \cdot B, \Pi \vdash C$}
     \DisplayProof
     &
     \AxiomC{$\Gamma \vdash A$}
     \AxiomC{$\Delta \vdash B$}
     \LeftLabel{(R$\cdot$)}
     \BinaryInfC{$\Gamma, \Delta \vdash A \cdot B$}
     \DisplayProof\\
     & \\
     \AxiomC{$\Gamma,B,\Pi \vdash C$}
     \AxiomC{$\Delta \vdash A$}
     \LeftLabel{(L$\backslash_L$)}
     \BinaryInfC{$\Gamma, \Delta, A \backslash_L B, \Pi \vdash C$}
     \DisplayProof
     &
     \AxiomC{$A, \Gamma \vdash B$}
     \LeftLabel{(R$\backslash_L$)}
     \UnaryInfC{$\Gamma \vdash A \backslash_L B$}
     \DisplayProof\\
     & \\
     \AxiomC{$\Gamma, A, \Pi \vdash C$}
     \AxiomC{$\Delta \vdash B$}
     \LeftLabel{(L$/$)}
     \BinaryInfC{$\Gamma, A/B, \Delta, \Pi \vdash C$}
     \DisplayProof
     &
     \AxiomC{$\Gamma, B \vdash A$}
     \LeftLabel{(R$/$)}
     \UnaryInfC{$\Gamma \vdash A/B$}
     \DisplayProof
    \end{tabular}
\end{center}
\caption{Axioms of Lambek Grammar.}
\label{tab:LGrules} 
\end{table}

The axioms of LG are shown in Table~\ref{tab:LGrules}, in the form presented in \cite{buszkowski2001lambek}. These axioms allow `empty' values for $\Gamma$, $\Delta$ and $\Pi$. We therefore assume the existence of an empty type $I$ acting as a multiplicative unit for the operation $\cdot$ such that $A \cdot I = A = I \cdot A$ for any $A$.

From these axioms, the following type operations are inferred \cite{moortgat2010, fowler2009parsing}:
\begin{itemize}
    \item \textbf{Slash introduction:} If $A \cdot B \vdash C$ then $A \vdash C / B$ and $B \vdash A \backslash_L C$.
    \item \textbf{Slash elimination}: $(A / B) \cdot B \vdash A$ and $B \cdot (B \backslash_L A) \vdash A$.
    \item \textbf{Composition:} $A/B \cdot B/C \vdash A/C$ and $A \backslash_L B \cdot B\backslash_L C \vdash A \backslash_L C$.
    \item \textbf{Type-raising:} $A \vdash B/ (A \backslash_L B)$ and $A \vdash (B / A) \backslash_L B$.
\end{itemize}

To complete the definition of a Lambek Grammar, a type dictionary is defined mapping each word in our lexicon to the set of types it can hold. A sentence $w_1 w_2 \ldots w_n$ is grammatically correct if there is some $T_i$ for each word $w_i$ in the type dictionary such that $T_1 \cdot T_2 \cdot \ldots \cdot T_n \vdash S$, where $S$ the type associated with sentences, can be logically derived from the axioms of $LG$.

% subsection lambek_grammar (end)

\subsection{Lambek Grammars as Monoidal Bi-Closed Categories} % (fold)
\label{sub:closed_monoidal_categories}

Here, we show that a Lambek Grammar LG can be modelled as a monoidal bi-closed category $\mathbf{LG}$, as was discussed and formalised by Bob Coecke, Mehrnoosh Sadrzadeh and myself in \cite{Sadrzadeh2013}, and previously observed in the original work on Lambek Grammars by Lambek himself.

A monoidal bi-closed category is a monoidal category with unit $I$ and associative monoidal tensor $\otimes$. What differentiates it from a normal monoidal category is that for every pair of objects $A$ and $B$ in the category, there are objects $A \multimapinv B$ and $A \multimap B$, and the morphisms
\begin{itemize}
    \item $ev^l_{A,B} : A \otimes (A \multimap B) \to B$
    \item $ev^r_{A,B} : (A \multimapinv B) \otimes B \to A$
    \item $\Lambda^l(f) : C \to A \multimap B$ for any $f: A \otimes C \to B$
    \item $\Lambda^r(g) : C \to A \multimapinv B$ for any $g: C \otimes B \to A$
\end{itemize}
such that the following diagrams commute:
\begin{diagram}
A \otimes C &\rTo^{1_A \otimes \Lambda^l(f)} &A \otimes (A \multimap B)
\hspace{2cm} & C\otimes B &\rTo^{\Lambda^r(g)\otimes 1_B}&(A\multimapinv B) \otimes B\\
&\rdTo_{f}&\dTo_{ev^l_{A,B}} 
\hspace{2cm} &&\rdTo_{g} &\dTo_{ev^r_{A,B}}\\
&&\qquad \qquad \quad B\hspace{2cm}& &&A
\end{diagram}
The morphisms $ev^l$ and $ev^r$ are referred to as evaluation morphisms, while the $\Lambda$ morphisms $\Lambda^l$ and $\Lambda^r$ are called currying morphisms.

It is fairly straightforward to show that any LG can be represented as a closed monoidal category $\mathbf{LG}$ which is free in its objects. The empty type $I$ is the unit object $I$. For any atomic type $A$ there is an object $A$ in the category. For each type of the form $A \cdot B$ there is some object $A \otimes B$. For each type $A \backslash_L B$ there is an object $A \multimap B$ and for each type $A / B$ there is an object $A \multimapinv B$. In short: the objects of the category are freely generated from the set of of atomic types. This feature will be essential to the definition of a functor between the categorical representation of Lambek Grammars and the category of vector spaces $\mathbf{FVect}$.

As for our type operations:
\begin{itemize}
    \item \textbf{Slash introduction:} let $A \cdot B \vdash C$ be the morphism $f: A \otimes B \to C$. Slash introduction $B \vdash A \backslash_L C$ corresponds to the morphism $\Lambda^l(f) : B \to A \multimap C$, and $A \vdash C / B$ corresponds to the morphism $\Lambda^r(f): A \to C \multimapinv B$.
    \item \textbf{Slash elimination} simply corresponds to the evaluation morphisms.
    \item \textbf{Composition:} for any pair of objects $A \multimap B$ and $B \multimap C$ in the category, there is a morphism 
    \[comp^r_{A \multimap B, B \multimap C}: (A \multimap B) \otimes (B \multimap C) \to A \multimap C\]
     Likewise, for any pair of objects $A \multimapinv B$ and $B \multimapinv C$ in the category, there is a morphism 
     \[comp^l_{A \multimapinv B, B \multimapinv C}: (A \multimapinv B) \otimes (B \multimapinv C) \to A \multimapinv C\]
     \item \textbf{Type raising} morphisms $raise^l_{A,B}$ and $raise^r_{A,B}$ are just the right and left currying of the left and right evaluation morphisms:
     \begin{itemize}
         \item $raise^l_{A,B} = \Lambda^r(ev^l_{A,B}) : A \to B \multimapinv (A \multimap B)$
         \item $raise^r_{A,B}  = \Lambda^l(ev^r_{B,A}) : A \to (B \multimapinv A) \multimap B$
     \end{itemize}
\end{itemize}

\usetikzlibrary{arrows,decorations,backgrounds,positioning,fit}

\tikzset{func/.style={shape=rectangle,rounded corners=8,minimum width=2cm,minimum height=.5cm,draw}}
\tikzset{claspnode/.style={shape=circle,minimum width=0.25cm,fill=white,draw}}

We saw, in Chapter~\ref{cha:foundations_of_discocat}, that compact closed categories have a sound and complete graphical calculus allowing us to reason diagrammatically about structures within such categories. Do we need to give up on this powerful tool when dealing with closed monoidal categories as we do here? Fortunately not.

As I discussed in \cite{Sadrzadeh2013}, with Mehrnoosh Sadrzadeh and Bob Coecke, a diagrammatic calculus for such categories has been developed by Baez and Stay~\cite{Baez2011}. Apart from depicting the flow of meaning, this calculus can also be applied to depict grammatical reduction of Lambek monoids, resembling the constituency parse trees obtained from context free grammars.
    
    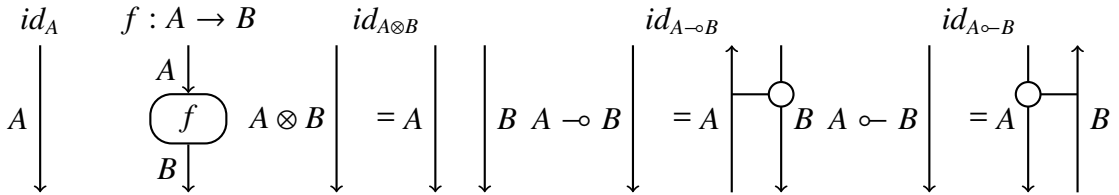
\begin{figure}[htbp]
        \centering
            \begin{tikzpicture}[thick,xscale=.65, yscale=0.65]

            \tikzset{func/.style={shape=rectangle,rounded corners=8,minimum width=1cm,draw}}
            \tikzset{claspnode/.style={shape=circle,minimum width=0.25cm,fill=white,draw}}

            \draw (-14,2.5) node {$id_A$};
            \draw[->] (-14,2) -- node [left] {$A$} (-14,-1);

            \draw (-11,2.5) node {$f : A \to B$};
            \node[func] (f) at (-11,.5) {$f$};
            \draw[->] (-11,2) -- node [left] {$A$} (f);
            \draw[->] (f) -- node [left] {$B$} (-11,-1);

            \draw[->] (-8,2) -- node [left] {$A \otimes B$} (-8,-1);

            \draw (-7,2.5) node {$id_{A \otimes B}$};
            \draw (-7,0.5) node {$=$};

            \draw[->] (-6,2) -- node [left] {$A$} (-6,-1);
            \draw[->] (-5,2) -- node [right] {$B$} (-5,-1);

            % A => B

            \draw[->] (-2,2) -- node [left] {$A \multimap B$} (-2,-1);

            \draw (-1, 2.5) node {$id_{A \multimap B}$};

            \draw (-1, 0.5) node {$=$};

            \draw[->] (0,-1) -- node [left] {$A$} (0,2);
            \draw[->] (1,2) -- node [right] {$B$} (1,-1);
            \filldraw[fill=white] (1,1) circle (0.25);
            \draw (0,1) -- (0.75,1);

            % A <= B

            \draw[->] (4,2) -- node [left] {$A \multimapinv B$} (4,-1);

            \draw (5, 2.5) node {$id_{A \multimapinv B}$};

            \draw (5, 0.5) node {$=$};

            \draw[<-] (6,-1) -- node [left] {$A$} (6,2);
            \draw[<-] (7,2) -- node [right] {$B$} (7,-1);
            \filldraw[fill=white] (6,1) circle (0.25);
            \draw (6.25,1) -- (7,1);

            \end{tikzpicture}
        \caption{Basic diagrammatic language constructs.}
        \label{fig:dlangspecific}
    \end{figure}

    The basic constructs of the diagrammatic language for  monoidal bi-closed categories is shown in Figure~\ref{fig:dlangspecific}. These are read from top to bottom, such that sequential composition of morphisms in the category corresponds to the downwards extension of the diagram. Arrows are annotated with objects of the category, and morphisms are represented as `blobs' with one or more `input' arrows and one or more `output' arrows standing for the domain and codomain. The tensoring of two objects corresponds to the side by side placement of their arrows. Topological equivalence between diagrams indicates an isomorphism between the corresponding categorical objects and vice versa, hence the diagram for $(A \otimes B) \otimes C$ is identical to that of $A \otimes (B \otimes C)$. Finally, there exists a rewrite rule for the objects of the form $A \multimap B$ and $A \multimapinv B$ as shown in the last two diagrams of Figure~\ref{fig:dlangspecific}. The clasp is a restriction in the diagrammatic calculus which prevents us from treating both arrows as separate entities, e.g.~such that a function cannot be applied to one and not to the other.

    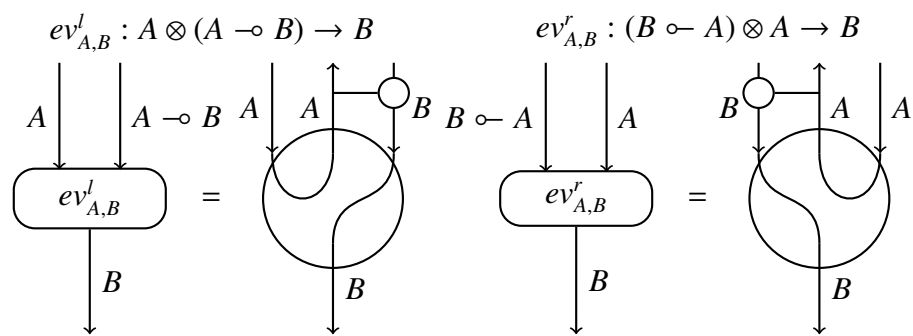
\begin{figure}[htbp]
        \centering
            \begin{tikzpicture}[thick,scale = 0.8]

            % evall

            \node [func,minimum width=2cm] (evall) at (0,.25) {$ev^l_{A,B}$};

            \draw[->] (-.5,2.5) -- node[left] {$A$} (-.5,.73);
            \draw[->] (.5,2.5) -- node[right] {$A \multimap B$} (.5,.73);
            \draw[->] (evall) -- node[right] {$B$} (0,-2);

            \draw (2,3) node {$ev^l_{A,B} : A \otimes (A \multimap B) \to B$};
            \draw (2,0.25) node {$=$};

            \draw[->] (3,2.5) -- node[left] {$A$} (3,1);
            \draw[->] (4,1) -- node[left] {$A$} (4,2.5);
            \draw[->] (5,2.5) -- node[right] {$\, B$} (5,1);
            \node[claspnode] (c1) at (5,2) {};
            \draw (4,2) -- (c1);

            \draw (3,1) .. controls +(0,-1) and +(0,-1) .. (4,1);
            \draw (5,1) .. controls +(0,-1) and +(0,1) .. (4,-.5);
            \draw[->] (4,-.5) -- node[right] {$B$} (4,-2);

            \draw (4,0.25) circle (1.15cm);

            % evalr

            \node [func,minimum width=2cm] (evalr) at (8,.25) {$ev^r_{A,B}$};

            \draw[->] (7.5,2.5) -- node[left] {$B \multimapinv A$} (7.5,.69);
            \draw[->] (8.5,2.5) -- node[right] {$A$} (8.5,.69);
            \draw[->] (evalr) -- node[right] {$B$} (8,-2);

            \draw (10,3) node {$ev^r_{A,B} : (B \multimapinv A) \otimes A \to B$};
            \draw (10,0.25) node {$=$};

            \draw[->] (13,2.5) -- node[right] {$A$} (13,1);
            \draw[->] (12,1) -- node[right] {$A$} (12,2.5);
            \draw[->] (11,2.5) -- node[left] {$B\ $} (11,1);
            \node[claspnode] (c2) at (11,2) {};
            \draw (12,2) -- (c2);

            \draw (13,1) .. controls +(0,-1) and +(0,-1) .. (12,1);
            \draw (11,1) .. controls +(0,-1) and +(0,1) .. (12,-.5);
            \draw[->] (12,-.5) -- node[right]{$B$} (12,-2);

            \draw (12,0.25) circle (1.15cm);

            \end{tikzpicture}
        \caption{Rewrite rules for $ev^l$ and $ev^r$.}
        \label{fig:evrewrite}
    \end{figure}

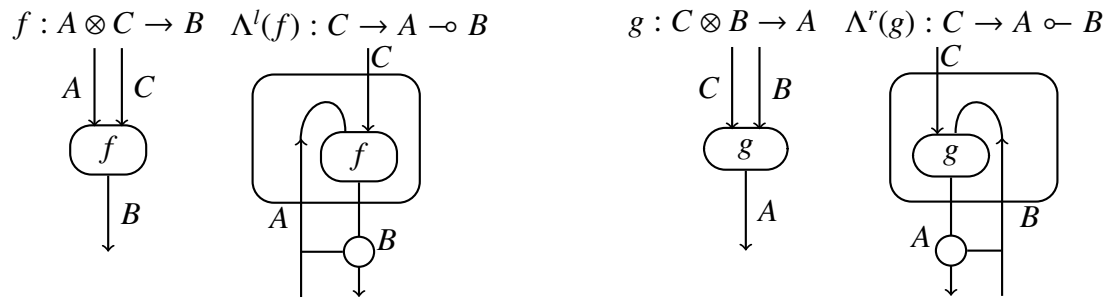
\begin{figure}[ht!]
        \centering
        
        \begin{minipage}{6cm}
                   \begin{tikzpicture}[thick,scale = 0.6]

                   % evall

                   \node [func,minimum width=1cm] (evall) at (0,.25) {$f$};

                   \draw[->] (-.3,2.5) -- node[left] {$A$} (-.3,.77);
                   \draw[->] (.3,2.5) -- node[right] {$C$} (.3,.77);
                   \draw[->] (evall) -- node[right] {$B$} (0,-2);

                   \draw (0,3) node {$f : A \otimes C \to B$};
                   %\draw (2,0.25) node {curried to};
                   \draw (5.5,3) node {$\Lambda^l(f) :  C \to A \multimap B$};

                   \node [func,minimum height=1.7cm,minimum width=2.2cm]  (f) at (5,.5) {\ };
                   \draw[->] (5.7,2.5) -- node {} (5.7,.65);
                               \node at (6,2.3) {$C$};
                   \node [func,minimum width=1cm] (evall) at (5.5,.1) {$f$};
                   \draw[<-] (4.23,.5) -- node[left] {$A$} (4.23,-3);
                   \draw[->] (evall) -- node[right] {$\, B$} (5.5,-3);
                   \node[claspnode] (c1) at (5.5,-2) {};
                   \draw (4.23,-2) -- (c1);
                   \draw (4.22,.45) .. controls +(0,1) and +(0,1) .. (5.2,.65);

                   \end{tikzpicture}
\end{minipage}\hspace{2cm}
\begin{minipage}{6cm} 
           \begin{tikzpicture}[thick,scale = 0.6]

                   \node [func,minimum width=1.1cm] (evall) at (0,.25) {$g$};

                   \draw[->] (-.3,2.5) -- node[left] {$C$} (-.3,.68);
                   \draw[->] (.3,2.5) -- node[right] {$B$} (.3,.68);
                   \draw[->] (evall) -- node[right] {$A$} (0,-2);

                   \draw (-0.5,3) node {$g : C \otimes B \to A$};
                  % \draw (2,0.25) node {curried to};
                   \draw (5,3) node {$\Lambda^r(g) :  C \to A \multimapinv B$};

                   \node [func,minimum height=1.7cm,minimum width=2.2cm]  (g) at (5,.5) {\ };
                   \draw[->] (4.2,2.5) -- node[left] {} (4.2,.57);
                   \node at (4.5,2.3) {$C$};
                   %\draw[->] (3.9,1.3) -- node [right] {$C$} (3.6,.5);
                   \node [func,minimum width=1cm] (evall) at (4.5,.1) {$g$};

                   \draw[->] (evall) -- node[left] {$A\ $} (4.5,-3);
                   \draw[<-] (5.62,.5) -- node[right] {$\, B$} (5.62,-3);
                   \node[claspnode] (c1) at (4.5,-2) {};
                   \draw (5.6,-2) -- (c1);
                   \draw (4.6,.6) .. controls +(0,1) and +(0,1) .. (5.62,.45);
    
                \end{tikzpicture}
\end{minipage}

        \caption{Diagrams for currying rules.}
        \label{fig:curryingdiag}
    \end{figure}

    Another more complex rewrite rule is that provided for the $ev^l$ and $ev^r$ morphisms, shown in Figure~\ref{fig:evrewrite}. These each  take two arrows in and output one: the in arrow of the form $A \multimap B$ or $B \multimapinv A$ is rewritten in clasp form using the rewrite rules from Figure~\ref{fig:dlangspecific}; and the `internal structure' of the morphism is exposed by drawing a bubble around cups and twists that redirect the non-clasped input arrow into the upward arrow from the clasped pair with a cup. Figure~\ref{fig:curryingdiag} shows the diagrams for currying, taken from \cite{Baez2011}, which represent the slash introduction rules. Figure~\ref{fig:composediag} shows the diagrams for the composition rules. 

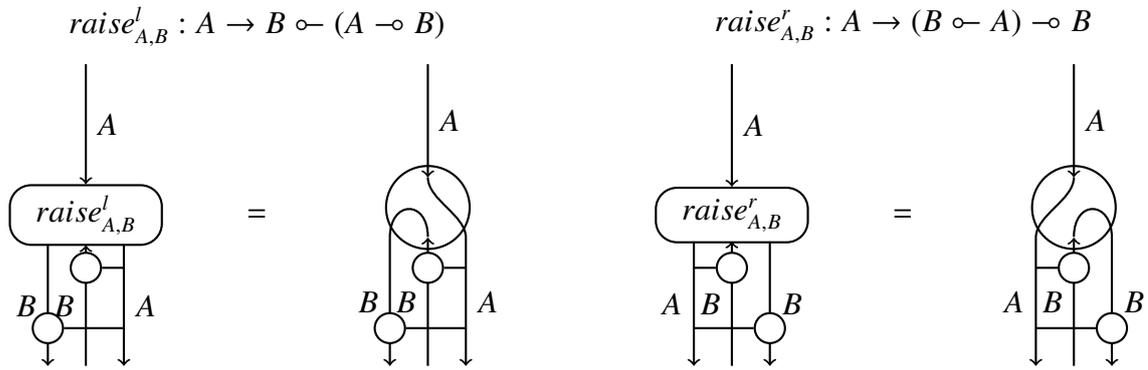
\begin{figure}[ht!]
        \centering
        \usetikzlibrary{arrows,decorations,backgrounds,positioning,fit}

\tikzset{func/.style={shape=rectangle,rounded corners=8,minimum width=2cm,minimum height=.5cm,draw}}
\tikzset{claspnode/.style={shape=circle,minimum width=0.25cm,fill=white,draw}}

\begin{tikzpicture}[thick]

\begin{scope}
    \node[func,minimum width=2cm] (typer) at (0,0) {$raise^l_{A,B}$};

\draw[->] (0,2) --  node [right] {$A$} (typer);
\draw[->] (-.5,-.4) --  node [left] {$B$} (-.5,-2);
\draw[<-] (0,-.4) -- node[left] {$B$} (0,-2);
\draw[->] (.5,-.4) -- node[right] {$A$} (.5,-2);

\node[claspnode] (a1) at (0,-.7) {};
\draw (.5,-.7) -- (a1);
\node[claspnode] (a2) at (-.5,-1.5) {};
\draw (.5,-1.5) -- (a2);

\node at ( 2.25,2.5) {$raise^l_{A,B} : A \to B \multimapinv (A \multimap B)$};
\node at (2.25,0) {$=$};

\draw[->] (4.5,2) --  node [right] {$A$} (4.5,.5);
\draw[->] (4,-.3) --  node [left] {$B$} (4,-2);
\draw[<-] (4.5,-.3) -- node[left] {$B$} (4.5,-2);
\draw[->] (5,-.3) -- node[right] {$A$} (5,-2);

\node[claspnode] (a1) at (4.5,-.7) {};
\draw (5,-.7) -- (a1);
\node[claspnode] (a2) at (4,-1.5) {};
\draw (5,-1.5) -- (a2);

\draw (4.5,.5) .. controls +(0,-.3) and +(0,.3) .. (5,-.3);
\draw (4,-.3) .. controls +(0,.5) and +(0,.5) .. (4.5,-.29);

\draw (4.5,.1) circle (.55cm);
\end{scope}

\begin{scope}[xshift=8.5cm]
    \node[func,minimum width=2cm] (typer) at (0,0) {$raise^r_{A,B}$};

\draw[->] (0,2) --  node [right] {$A$} (typer);
\draw[->] (-.5,-.35) --  node [left] {$A$} (-.5,-2);
\draw[<-] (0,-.35) -- node[left] {$B$} (0,-2);
\draw[->] (.5,-.35) -- node[right] {$B$} (.5,-2);

\node[claspnode] (a1) at (0,-.7) {};
\draw (-.5,-.7) -- (a1);
\node[claspnode] (a2) at (.5,-1.5) {};
\draw (-.5,-1.5) -- (a2);

\node at ( 2.25,2.5) {$raise^r_{A,B}  : A \to (B \multimapinv A) \multimap B$};
\node at (2.25,0) {$=$};

\draw[->] (4.5,2) --  node [right] {$A$} (4.5,.5);
\draw[->] (4,-.3) --  node [left] {$A$} (4,-2);
\draw[<-] (4.5,-.3) -- node[left] {$B$} (4.5,-2);
\draw[->] (5,-.3) -- node[right] {$B$} (5,-2);

\node[claspnode] (a1) at (4.5,-.7) {};
\draw (4,-.7) -- (a1);
\node[claspnode] (a2) at (5,-1.5) {};
\draw (4,-1.5) -- (a2);

\draw (4.5,.5) .. controls +(0,-.3) and +(0,.3) .. (4,-.3);
\draw (4.5,-.29) .. controls +(0,.5) and +(0,.5) .. (5,-.3);

\draw (4.5,.1) circle (.55cm);
\end{scope}

\end{tikzpicture}

        \caption{Diagrams for type raising rules.}

        \label{fig:typeraising}
    \end{figure}

    Finally, Figure~\ref{fig:typeraising} shows the diagrams for the type raising rules, obtained by currying the evaluation rules. This last figure may require a bit of explanation: note that the clasps on the bottom don't seem to fully match the direction of the $\multimap$ and $\multimapinv$ operators in the mathematical definition shown above them. This is no error: since expressions of the form $A \multimap B$ and $B \multimapinv A$ are represented diagrammatically as an upwards pointing arrow labelled $A$ clasped with a downward pointing arrow labelled $B$; when $A$ is of the form $C \multimap D$ or $D \multimapinv C$ we must in turn expand the upward pointing arrow into two clasped arrows. However, as the arrow we are `expanding' is pointing upwards, we must turn the clasped arrow representations upside-down, thereby reversing the up/down direction of clasped arrows and swapping their order (effectively rotating the clasped arrows by $180^{\circ}$). This means that if $A$ is pointing upwards, then its clasped expansion will have a clasp pointing from left to right (from $C$ to $D$) if $A = D \multimapinv C$ and from right to left (again, from $C$ to $D$) if $A = C \multimap D$, thereby showing the clasp in the opposite direction as it is written in the formula.

    \begin{figure}[ht!]
        \centering

        \usetikzlibrary{arrows,decorations,backgrounds,positioning,fit}

\tikzset{func/.style={shape=rectangle,rounded corners=8,minimum width=2cm,minimum height=.5cm,draw}}
\tikzset{claspnode/.style={shape=circle,minimum width=0.25cm,fill=white,draw}}

\begin{tikzpicture}[thick]

\node[func,minimum width=2cm] (cutr) at (0,0) {$comp^r_{A,B,C}$};

\draw[->] (-.5,2.5) --  node [left] {$A \multimap B$} (-.5,.37);
\draw[->] (.5,2.5) --  node [right] {$B \multimap C$} (.5,.37);
\draw[->] (cutr) -- node[right] {$A \multimap C$} (0,-2);

\draw[->] (3,1) -- node [left] {$A$} (3,2.5);
\draw[->] (4,2.5) -- node [right] {$B$} (4,1);
\draw[->] (5,1) -- node [left] {$B$} (5,2.5);
\draw[->] (6,2.5) -- node [right] {$C$} (6,1);

\node at ( 2.25,3) {$comp^r_{A,B,C} : (A \multimap B) \otimes (B \multimap C) \to (A \multimap C)$};
\node at (2.25,0) {$=$};

\node[claspnode] (c1) at (4,2) {};
\draw (3,2) --(c1);
\node[claspnode] (c2) at (6,2) {};
\draw (5,2) --(c2);

\draw (4,1) .. controls +(0,-1) and +(0,-1) .. (5,1);
\draw (3,1) .. controls +(0,-1) and +(0,1) .. (4,-.5);
\draw (6,1) .. controls +(0,-1) and +(0,1) .. (5,-.5);

\draw[<-] (4,-.5) -- node[left] {$A$} (4,-2);
\draw[->] (5,-.5) -- node[right] {$\ C$} (5,-2);

\node[claspnode] (c3) at (5,-1.2) {};
\draw (4,-1.2) -- (c3);

\draw (4.5,0) circle (1cm);

\end{tikzpicture}

\vspace{1cm}

\begin{tikzpicture}[thick]
\node[func,minimum width=2cm] (cutl) at (9.5,0) {$comp^l_{A,B,C}$};

\draw[->] (9,2.5) --  node [left] {$A \multimapinv B$} (9,.4);
\draw[->] (10,2.5) --  node [right] {$B \multimapinv C$} (10,.4);
\draw[->] (cutl) -- node[right] {$A \multimapinv C$} (9.5,-2);

\draw[->] (12.5,1) -- node [left] {$A$} (12.5,2.5);
\draw[->] (13.5,2.5) -- node [right] {$B$} (13.5,1);
\draw[->] (14.5,1) -- node [left] {$B$} (14.5,2.5);
\draw[->] (15.5,2.5) -- node [right] {$C$} (15.5,1);

\node at (11.75,3) {$comp^l_{A,B,C} : (A \multimapinv B) \otimes (B \multimapinv C) \to (A \multimapinv C)$};
\node at (11.75,0) {$=$};

\node[claspnode] (c4) at (12.5,2) {};
\draw (13.5,2) --(c4);
\node[claspnode] (c5) at (14.5,2) {};
\draw (15.5,2) --(c5);

\draw (13.5,1) .. controls +(0,-1) and +(0,-1) .. (14.5,1);
\draw (12.5,1) .. controls +(0,-1) and +(0,1) .. (13.5,-.5);
\draw (15.5,1) .. controls +(0,-1) and +(0,1) .. (14.5,-.5);

\draw[<-] (13.5,-.5) -- node[left] {$A\ $} (13.5,-2);
\draw[->] (14.5,-.5) -- node[right] {$C$} (14.5,-2);

\node[claspnode] (c6) at (13.5,-1.2) {};
\draw (14.5,-1.2) -- (c6);

\draw (14,0) circle (1cm);
\end{tikzpicture}

        \caption{Diagrams for composition rules.}
        \label{fig:composediag}
\end{figure}

\begin{figure}[ht!]
        \centering
        \begin{minipage}{3cm}
        \begin{tikzpicture}[thick,scale = 0.6]
            
            \begin{scope}
                
                \draw (0, 6.5) node {Men};
                \draw (2,6.5) node {kill};

                \draw[<-] (1.5,6) -- node [left] {$n$} +(0,-1.5);
                \draw[->] (2.5,6) -- node [right] {$s$} +(0,-1.5);
                \draw (2.5,5.5) node (c4) [claspnode] {};
                \draw (1.5,5.5) to (c4);
                
                \draw[->] (0,6) -- node[left] {$n$} (0,4.5);
                
                \draw (1.5,3.5) ellipse (2cm and 1cm);
                \draw (1.5,4.5) -- (1.5,4);
                \draw (0,4.5) -- (0,4);
                \draw (0,4) .. controls +(0,-1) and +(0,-1) .. (1.5,4);
                \draw (2.5,4.5) -- (2.5,3);
                
                \draw[->] (2.5,3) -- node [right] {$s$} +(0,-1);
                
            \end{scope}         
            
        \end{tikzpicture}
\end{minipage} \hspace{3cm}
\begin{minipage}{5cm}
    \begin{tikzpicture}[thick,scale = 0.6]
        
        \begin{scope}
            
            \draw (0, 6.5) node {Men};
            \draw (2.5,6.5) node {kill};
            \draw (5,6.5) node {dogs};

            % \draw[<-] (1.5,6) -- node [left] {$N$} +(0,-1.5);
            \draw[->] (2.5,6) -- node [left] {$n \multimap s\ $} +(0,-1.5);
            \draw (2.5,5.5) node (c4) [claspnode] {};
            \draw (3.5,5.5) to (c4);
            
            \draw[<-] (3.5,6) -- node[right] {$n$} (3.5,4.5);
            \draw[->] (5,6) -- node[left] {$n$} (5,4.5);
            \draw (3.5,4) .. controls +(0,-1) and +(0,-1) .. (5,4);
            \draw (3.5,4.5) -- (3.5,4);
            \draw (5,4.5) -- (5,4);
            \draw (3.75,3.5) ellipse (2cm and 1cm);
            \draw[->] (2.5,3) -- node [left] {$n \multimap s\ $} +(0,-1);
            
            \draw[->] (0,6) -- node[left] {$n$} (0,4.5);

            \draw[dashed] (2.5,2) -- (1.5,1.5);
            \draw[dashed] (2.5,2) -- (2.5,1.5);
            
            \draw[->] (2.5,1.5) -- node [right] {$\ s$} +(0,-1);
            \draw[<-] (1.5,1.5) -- node [left] {$n$} +(0,-1);
            \draw (2.5,1) node [claspnode] (c5) {};
            \draw (1.5,1) to (c5);
            
            \draw (1.5,.5) -- (1.5,0);
            \draw (0,4.5) -- (0,0);
            \draw (0,0) .. controls +(0,-1) and +(0,-1) .. (1.5,0);
            \draw (2.5,4.5) -- (2.5,3);
            \draw (2.5,.5) -- (2.5,-.5);
            \draw (1.25,-.5) ellipse (2cm and .8cm);
            \draw[->] (2.5,-.5) -- node [right] {$s$} +(0,-1.5);

        \end{scope}         
        
    \end{tikzpicture}
    \end{minipage}

        \caption{Diagrams for sample Lambek Grammar parses.}
        \label{fig:lambekparse}
    \end{figure}
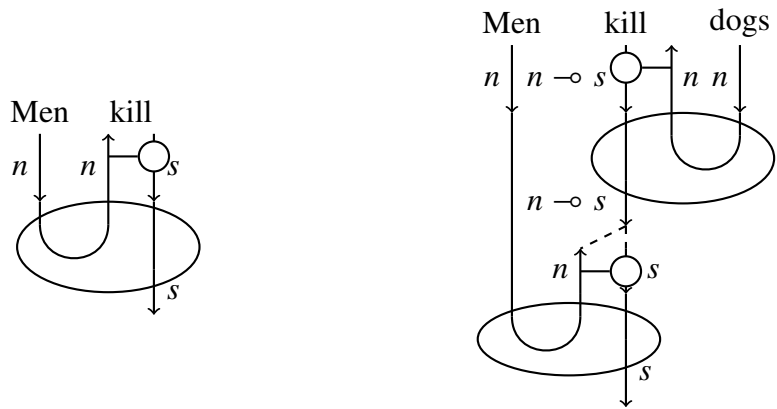

    To give a short example of how parses are represented diagrammatically, the diagrams for ``men kill'' and ``men kill dogs'' are shown in Figure~\ref{fig:lambekparse}. We assume nouns have the type $N$, intransitive verbs have type $N \multimap S$ and transitive verbs have the type $(N \multimap S) \multimapinv N$. The parse for ``men kill'' is just an $ev^l_{N,S}$.  For the parse of ``men kill dogs'', we start with one line with the type $N \multimap S$, do an $ev^r_{N\multimap S, N}$ with the object, then rewrite (marked with the dotted lines) this line  to a clasp form with two lines of type $N$ and $S$ respectively, then do an $ev^l_{N,S}$ with the subject.

% subsection closed_monoidal_categories (end)

\subsection{Defining a Functor} % (fold)
\label{sub:defining_a_functor_lg}

Equipped with this categorical representation $\mathbf{LG}$ of a Lambek Grammar, we now turn to the task of defining a functorial passage from a category $\mathbf{LG}$ to the category of vector spaces $\mathbf{FVect}$. Such a categorical passage was described in \cite{Sadrzadeh2013} which I co-authored with Mehrnoosh Sadrzadeh and Bob Coecke. I am indebted in particular to Mehrnoosh, who worked on the definition of this functor, and has permitted me to reproduce her work here in my own words.

The bi-closed monoidal categories we used to represent $\mathbf{LG}$ are structurally similar to the compact closed monoidal categories used to model pregroup categories $P$ and $\mathbf{FVect}$, in that the evaluation morphisms `act' like epsilon maps; for example:
\[
ev^l_{A,B}: A \otimes (A \multimap B) \to B \quad \cong \quad \epsilon^r_A \otimes 1_B: A \otimes (A^r \otimes B) \to B
\]

To generalise this similarity to a strict functor $F$, let each atomic object $A_{\mathbf{LG}}$ in $ob(\mathbf{LG})$ be assigned to a vector space $F(A_{\mathbf{LG}}) = A_{\mathbf{FVect}}$ in $ob(\mathbf{FVect})$. To simplify things notationally, we will drop the subscripts and just use the same letter. As a special case, the unit type $I$ in $\mathbf{LG}$ maps to the unit type in $\mathbf{FVect}$, namely $\mathbb{R}$, such that $F(I) = \mathbb{R}$.

The functorial passage for the monoidal tensor is simply the map of the monoidal tensor from $\mathbf{LG}$ to $\mathbf{FVect}$, such that for any $A$ and $B$ in $ob(\mathbf{LG})$ we have:
\[
F(A \otimes_{\mathbf{LG}} B) = F(A) \otimes_{\mathbf{FVect}} F(B)
\]
Here too, we will dispense with indicating the subscript on the tensor operator.

Next comes the functorial definition for the $\multimap$ and $\multimapinv$ operations. Generally speaking, for the passage from a bi-closed monoidal category to a compact closed monoidal category, we would have the following two functorial passages:
\begin{align*}
    & F(A \multimap B) = F(A)^r \otimes F(B)\\
    & F(A \multimapinv B) = F(A) \otimes F(B)^l
\end{align*}
However let us remember that in $\mathbf{FVect}$ the adjoints are degenerate such that $A^l = A^r = A^{*}$ and $A^{*} \cong A$ for all vector spaces $A$, so we can do away with the adjoint notation. We therefore obtain a simpler functorial definition:
\begin{align*}
    & F(A \multimap B) = F(A) \otimes F(B)\\
    & F(A \multimapinv B) = F(A) \otimes F(B)
\end{align*}

We now turn to the functorial passage for morphisms. The most complex case to consider here is slash introduction. Slash introduction states that for any morphism $f: A \otimes B \to C$ in $hom(\mathbf{LG})$ there are morphisms $\Lambda^l(f): B \to A \multimap C$ and $\Lambda^r(f): A \to C \multimapinv B$ in $hom(\mathbf{LG})$. Translating this to $\mathbf{FVect}$, this would entail that for any morphism $F(f) : F(A) \otimes F(B) \to F(C)$ in $hom(\mathbf{FVect})$ there must be $F(\Lambda^l(f)): F(B) \to F(A) \otimes F(C)$ and $F(\Lambda^r(f)): F(A) \to F(C) \otimes F(B)$ in $hom(\mathbf{FVect})$. For this passage to hold, we must show that there are such curried morphisms for every such $F(f)$ in $hom(\mathbf{FVect})$.

\begin{figure}[ht!]
        \centering
        
        \begin{minipage}{6cm}
                   \begin{tikzpicture}[thick,scale = 0.6]

                   % evall

                   \node [func,minimum width=1cm] (evall) at (0,.25) {$f$};

                   \draw[->] (-.3,2.5) -- node[left] {$A$} (-.3,.78);
                   \draw[->] (.3,2.5) -- node[right] {$C$} (.3,.78);
                   \draw[->] (evall) -- node[right] {$B$} (0,-2);

                   \draw (0,3) node {$f : A \otimes C \to B$};
                   %\draw (2,0.25) node {curried to};
                   \draw (5.5,3) node {$\Lambda^l(f) :  C \to A \otimes B$};

                   \draw[->] (5.7,2.5) -- node {} (5.7,.65);
                               \node at (6,2.3) {$C$};
                   \node [func,minimum width=1cm] (evall) at (5.5,.1) {$f$};
                   \draw[<-] (4.23,.5) -- node[left] {$A$} (4.23,-2);
                   \draw[->] (evall) -- node[right] {$B$} (5.5,-2);
                   \draw (4.22,.45) .. controls +(0,1) and +(0,1) .. (5.2,.65);

                   \end{tikzpicture}
\end{minipage}\hspace{2cm}
\begin{minipage}{6cm} 
           \begin{tikzpicture}[thick,scale = 0.6]

                   \node [func,minimum width=1.1cm] (evall) at (0,.25) {$g$};

                   \draw[->] (-.3,2.5) -- node[left] {$C$} (-.3,.71);
                   \draw[->] (.3,2.5) -- node[right] {$B$} (.3,.71);
                   \draw[->] (evall) -- node[right] {$A$} (0,-2);

                   \draw (-0.5,3) node {$g : C \otimes B \to A$};
                  % \draw (2,0.25) node {curried to};
                   \draw (5,3) node {$\Lambda^r(g) :  C \to A \otimes B$};

                   \draw[->] (4.2,2.5) -- node[left] {} (4.2,.57);
                   \node at (4.5,2.2) {$C$};
                   %\draw[->] (3.9,1.3) -- node [right] {$C$} (3.6,.5);
                   \node [func,minimum width=1cm] (evall) at (4.5,.1) {$g$};

                   \draw[->] (evall) -- node[left] {$A$} (4.5,-2);
                   \draw[<-] (5.62,.5) -- node[right] {$B$} (5.62,-2);
                   \draw (4.6,.6) .. controls +(0,1) and +(0,1) .. (5.62,.45);
    
                \end{tikzpicture}
\end{minipage}

        \caption{Diagrams for compact closed currying.}
        \label{fig:compactcurryingdiag}
    \end{figure}
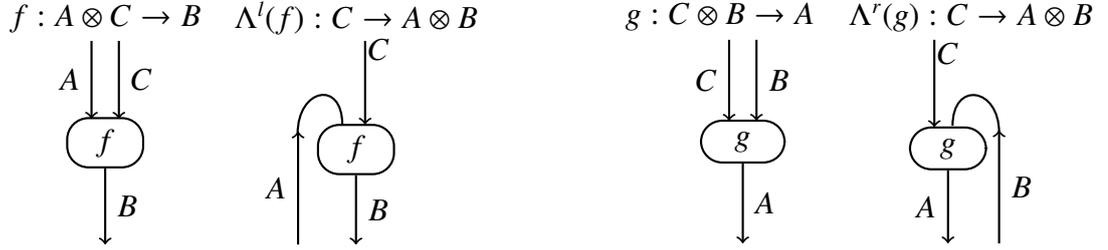

This is where a powerful feature of the graphical calculus for bi-closed categories of \cite{Baez2011} comes into play: we can go from the diagrams of morphisms in bi-closed monoidal categories to those in compact closed categories simply by removing the bubbles and the clasps. This allows us to produce the diagrams for currying in $\mathbf{FVect}$ simply by `cleaning up' those for currying in $\mathbf{LG}$ (or any other bi-closed monoidal category). I therefore show, in Figure~\ref{fig:compactcurryingdiag}, the diagrams for currying in a compact closed category, based on those shown earlier in Figure~\ref{fig:curryingdiag}. From these diagrams, we can simply read the morphisms which are needed for currying rules to hold in $\mathbf{FVect}$, and these turn out to require nothing more than the $\eta_A: \mathbb{R} \to A \otimes A$ maps, which we represent as caps in the diagrammatic calculus, and which we know must exist in $\mathbf{FVect}$ for any object $A$ by virtue of the category being compact closed. This `magical' use of the diagrams implicitly corresponds to the functorial passage from the bi-closed monoidal category representing the Lambek Grammar to a strict compact closed monoidal category which is free in its objects, and then passing from this category to $\mathbf{FVect}$ using a similar functor as was defined for the passage from $P$ to $\mathbf{FVect}$ in $\S$\ref{sub:from_product_categories_to_functors}. Using the diagrammatic calculus to avoid dealing with the underlying mathematics greatly simplifies the operation of defining a function in such a way, but it is important to be aware of the reliance on the freely generated nature of the source category $\mathbf{LG}$.

We can therefore read the following from these diagrams: for any morphism $F(f) : F(A) \otimes F(B) \to F(C)$ in $hom(\mathbf{FVect})$ there must be $F(\Lambda^l(f)): B \to F(A) \otimes F(C)$ and $F(\Lambda^r(f)): F(A) \to F(C) \otimes F(B)$ in $hom(\mathbf{FVect})$, where:
\begin{align*}
    & F(\Lambda^l(f)) = (1_A \otimes F(f)) \circ (\eta_{F(A)} \otimes 1_{F(B)}) : F(B) \to F(A) \otimes F(C)\\
    & F(\Lambda^r(f)) = (F(f) \otimes 1_{F(B)}) \circ (1_{F(A)} \otimes \eta_{F(B)}) : F(A) \to F(C) \otimes F(B)
\end{align*}
So slash introduction is well defined under the functorial passage $F$.

The rest of the operations are far simpler to define, and can be read from Figures~\ref{fig:evrewrite},~\ref{fig:composediag} and~\ref{fig:typeraising}:
\begin{align*}
    & F(ev^l_{A,B}) = \epsilon_{F(A)} \otimes 1_{F(B)} : F(A) \otimes F(A) \otimes F(B) \to F(B)\\
    & F(ev^r_{A,B}) = 1_{F(B)} \otimes \epsilon_{F(A)} : F(B) \otimes F(A) \otimes F(A) \to F(B)\\
    & F(comp^r_{A,B,C}) =  1_{F(A)} \otimes \epsilon_{F(B)} \otimes 1_{F(C)} : F(A) \otimes F(B) \otimes F(B) \otimes F(C) \to F(A) \otimes F(C)\\
    & F(comp^l_{A,B,C}) = 1_{F(A)} \otimes \epsilon_{F(B)} \otimes 1_{F(C)} : F(A) \otimes F(B) \otimes F(B) \otimes F(C) \to F(A) \otimes F(C)\\
    & F(raise^l_{A,B}) = (\eta_{F(B)} \otimes 1_{F(A)}): F(A) \to F(B) \otimes F(B) \otimes F(A)\\
    & F(raise^r_{A,B}) = (1_{F(A)} \otimes \eta_{F(B)}): F(A) \to F(A) \otimes F(B) \otimes F(B)\\
\end{align*}
This completes the definition of a functorial passage between $\mathbf{LG}$ and $\mathbf{FVect}$.
% subsection defining_a_functor_lg (end)

% section supporting_lambek_grammar (end)

% chapter syntactic_extensions_and_non_linear_maps (end)
\mbox{}
\newpage
%!TEX root = ../grefenstettethesis.tex

\chapter{Learning Procedures for a DisCoCat} % (fold)
\label{cha:learning_procedures_for_a_discocat}

\begin{chabstract}
  This chapter presents a procedure for generating concrete compositional distributional models of semantics from the DisCoCat framework presented in Chapter~\ref{cha:foundations_of_discocat}. It discusses several considerations as to the shape and structure of the semantic vector space for sentences, before presenting a learning procedure for the production of the semantic representation for words and relations based on corpus data.
\end{chabstract}

Previously, in $\S$\ref{sec:a_categorical_passage_from_grammar_to_semantics} I presented a categorical formalism which relates syntactic analysis steps to semantic composition operations. The structure of the syntactic representation used dictates the structure of the semantic spaces, but in exchange for this constraint, we are provided with natural composition functions by the syntactic analysis, rather than having to stipulate them \emph{ad hoc}. Whereas the approaches to compositional DSMs presented in Chapter~\ref{cha:literature_review} either failed to take syntax into account during composition, or did so at the cost of not being able to compare sentences of different structure in a common space, this categorical approach projects all sentences into a common sentence space where they can be directly compared. However, this alone does not give us a compositional DSM.

As we have seen in the above examples, the structure of semantic spaces varies with syntactic types. We therefore cannot construct vectors for different syntactic types in the same way, as they live in spaces of different structure and dimensionality. Furthermore, nothing has yet been said about the structure of the sentence space $S$ into which expressions reducing to type $s$ are projected. If we wish to have a compositional DSM which leverages all the benefits of lexical DSMs and extends them to sentence-level distributional representations, we must specify a new sort of vector construction procedure.

In the original formulation of this formalism by \cite{Clark2008,Coecke2010}, examples of how such a compositional DSM could be used for logical evaluation are presented, where $S$ is defined as a boolean space with True and False as basis vectors. However, the word vectors used are hand-written and specified model-theoretically, as the authors leave it for future research to determine how such vectors might be obtained from a corpus. In this chapter, I will discuss a new way of constructing vectors for Compositional DSMs, and of defining the sentence space $S$, in order to reconcile this powerful categorical formalism with the applicability and flexibility of standard distributional models.

\section{Defining Sentence Space} % (fold)
\label{sec:defining_sentence_space}

\subsection{Intuition} % (fold)
\label{sub:intuition}

Assume the following sentences are all true:
\begin{enumerate}
    \item The dogs chased the cats.
    \item The dogs annoyed the cats.
    \item The puppies followed the kittens.
    \item The train left the station.
    \item The president followed his agenda.
\end{enumerate}
If asked which sentences have similar meaning, we would most likely point to the pair (1) and (3), and perhaps to a lesser degree (1) and (2), and (2) and (3). Sentences (4)  and (5) obviously speak of a state of the world unrelated to that which is described by the other sentences.

If we compare these by truth value, we obviously have no means of making such distinctions. If we compare these by lexical similarity, (1) and (2) seem to be a closer match than (1) and (3). If we are classifying these sentences by some higher order relation such as `topic', (5) might end up closer to (3) than (1). What, then, might cause us to pair (1) and (3)?

Intuitively, this similarity seems to be because the subjects and objects brought into relation by similar verbs are themselves similar. Abstracting away from tokens to some notion of property, we might say that both (1) and (3) express the fact that something furry and feline and furtive is being pursued by something aggressive (or playful) and canine. Playing along with the idea that lexical distributional semantics presents concepts (word meanings) as `messy bundles of properties', it seems only natural to have the way these properties are acted upon, qualified, and related as the basis for sentence-level distributional representations. In this respect, I here suggest that the sentence space $S$, instead of qualifying the truth value of a sentence, should express how the properties of the semantic objects within are qualified or brought into relation by verbs, adjectives, and other predicates and relations.

The above intuitions serve primarily to guide the development of a learning procedure as described below. The question of what are the ``true semantics'' of phrase and sentence vectors, of what they represent, is fundamental and difficult to answer. Such vectors could be interpreted as being the distributions of contexts in which sentences occur in an infinite idealised corpus; alternatively, they can be viewed simply as vectors of feature weights to be used in supervised machine learning algorithms such as classifiers, or as pre-training for unsupervised machine learning algorithms such as clustering algorithms. Either way, the construction procedure described below is general, and makes no commitments in terms of the interpretation of phrase and sentence vectors.
% subsection intuition (end)

\subsection{A Concrete Proposal} % (fold)
\label{sub:concrete_proposal}

More specifically, I examine two suggestions for defining the sentence space, namely $S_I \cong N$ for sentences with intransitive verbs and $S_T \cong N \otimes N$ for sentences with transitive verbs.  These definitions mean that the sentence space's dimensions are commensurate with either those of $N$, or those of $N \otimes N$. These are by no means the only options, but as I will discuss here, they offer practical benefits.

In the case of $S_I$, the basis elements are labelled with unique basis elements of $N$, hence $\overrightarrow{s_1} = \overrightarrow{n_1}$, $\overrightarrow{s_2}= \overrightarrow{n_2}$, and so on. In the case of $S_T$,  the basis elements are labelled with unique ordered pairs of elements from $N$, for example $\overrightarrow{s_1} = \overrightarrow{(n_1, n_1)}$, $\overrightarrow{s_2} = \overrightarrow{(n_2, n_1)}$, $\overrightarrow{s_3} = \overrightarrow{(n_1, n_2)}$, and so on. Because of the isomorphism between $S_T$ and $N \otimes N$, I will use the notations $\overrightarrow{(n_i, n_j)}$ and $\overrightarrow{n_i} \otimes \overrightarrow{n_j}$ interchangeably, as both constitute appropriate ways of representing the basis elements of such a space.

To propagate this distinction to the syntactic level, I define types $s_I$ and $s_T$ for intransitive and transitive sentences, respectively.

% subsection concrete_proposal (end)

% section defining_sentence_space (end)

\section{Noun-Oriented Types} % (fold)
\label{sec:noun_oriented_types}

\subsection{Dealing with Nouns} % (fold)
\label{sub:dealing_with_nouns}

Lambek's pregroup types presented in \cite{Lambek2008} include a rich array of basic types and hand-designed compound types in order to capture specific grammatical properties. Here, for the sake of simplicity I will use a simpler set of grammatical types for experimental purposes, similar to some common types found in Combinatory Categorial Grammar (CCG) \cite{steedman2000syntactic}.

I assign a basic pregroup type $n$ for all nouns, with an associated vector space $N$ for their semantic representations. Furthermore, I will treat noun-phrases as nouns, assigning to them the same pregroup type and semantic space.

% subsection dealing_with_nouns (end)

\subsection{Dealing with Relational Words} % (fold)
\label{sub:dealing_with_relational_words}

CCG treats intransitive verbs as functions $NP\backslash S$ that consume a noun phrase and return a sentence, and transitive verbs as functions $(NP\backslash S)/NP$ that consume a noun phrase and return an intransitive verb function, which in turn consumes a noun phrase and returns a sentence. Using my distinction between intransitive and transitive sentences, I give intransitive verbs the type $n^rs_I$ associated with the semantic space $N \otimes S_I$, and transitive verbs the type $n^rs_Tn^l$ associated with the semantic space $N \otimes S_T \otimes N$.

Adjectives, in CCG, are treated as functions $NP/NP$ consuming a noun phrase and returning a noun phrase, and hence I give them the type $nn^l$ and associated semantic space $N \otimes N$.

With the provision of a learning procedure for vectors in these semantic spaces, we can use these types to construct sentence vector representations for simple intransitive verb-based and transitive verb based sentences, with and without adjectives applied to subjects and objects.

% subsection dealing_with_relational_words (end)

% section noun_oriented_types (end)

\section{Learning Procedures} % (fold)
\label{sec:learning_procedures}

\subsection{Groundwork} % (fold)
\label{sub:groundwork}

To begin, I construct the semantic space $N$ for all nouns in my lexicon (typically limited by the words available in the corpus used). Any distributional semantic model can be used for this stage, such as those presented in \cite{Curran2004}, or the lexical semantic models used by \cite{mitchell2008vector}. It seems reasonable to assume that higher quality lexical semantic vectors---as measured by metrics such as the WordSim353 test of \cite{finkelstein2001placing}---will produce better relational vectors from the procedure designed below. I will not test this hypothesis here, but note that it is an underlying assumption in most of the current literature on the subject \cite{Erk2008,mitchell2008vector,Baroni2010}.

Building upon the foundation of the thus-constructed noun vectors, I construct semantic representations for relational words. In pregroup grammars (or other combinatorial grammars such as CCG), we can view such words as functions taking as arguments those types present as adjoints in the compound pregroup type, and returning a syntactic object whose type is that of the corresponding reduction. For example, an adjective $nn^l$ takes a noun or noun phrase $n$ and returns a noun phrase $n$ from the reduction $(nn^l)n \to n$. It can also compose with another adjective to return an adjective $(nn^l) (nn^l) \to nn^l$. I wish for my semantic representations to be viewed in the same way, such that the composition of an adjective with a noun $(1_N \otimes \epsilon_N)((N \otimes N) \otimes N)$ can be viewed as the application of a function $f: N \to N$ to its argument of type $N$.

To learn the representations of such functions, I assume that their meaning can be characterised by the properties that their arguments hold in the corpus, rather than just by their context as is the case in lexical distributional semantic models. To give an example, rather than learning what the adjective ``angry'' means by observing that it co-occurs with words such as ``fighting'', ``aggressive'' or ``mean'', we can learn its meaning by observing that it typically takes, as argument, words that \emph{co-occur with} words such as ``fighting'', ``aggressive'' and ``mean''. While in the lexical semantic case, such associations might only rarely occur in the corpus, in this indirect method we learn what properties the adjective relates to even if they do not co-occur with it directly.

In turn, through composition with its argument, I expect the function for such an adjective to \emph{strengthen} the properties that characterise it in the representation of the object it takes as argument. Let us assume ``angry'' is characterised by arguments that have high basis weights for basis elements corresponding to the concepts (or context words) ``fighting'', ``aggressive'' and ``mean'', and relatively low counts for semantically different concepts such as ``passive'', ``peaceful'' and ``loves''. When I apply ``angry'' to ``dog'' the vector for the compound ``angry dog'' should contain some of the information found in the vector for ``dog''. But this vector should also have \emph{higher} values for the basis weights of ``fighting'', ``aggressive'' and ``mean'', and correspondingly lower values for the basis weights of ``passive'', ``peaceful'', ``loves''.

% subsection groundwork (end)

\subsection{A Learning Algorithm} % (fold)
\label{sub:a_learning_algorithm}
To turn this idea into a concrete algorithm for constructing the semantic representation for relations of any arity, as first presented in \cite{grefenstette2011concrete}, let's examine how we would deal with this for binary relations such as transitive verbs. If a transitive verb of semantic type $N \otimes S_T \otimes N$ is viewed as a function $f: N \times N \to S_T$ which expresses the extent to which the properties of subject and object are brought into relation by the verb, we learn the meaning of the verb by looking at what properties are brought into relation by the verb in terms of what arguments it takes in a corpus. Recall that the vector for a verb $v$, $\overrightarrow{v} \in N \otimes S_T  \otimes N$, can be expressed as the weighted superposition of its basis elements:
\[
\overrightarrow{v} = \sum_{ijk}{c^v_{ijk} \overrightarrow{n_i} \otimes \overrightarrow{s_j} \otimes \overrightarrow{n_k}}
\]
I take the set of vectors for the subject and object of $v$ in the corpus to be the set of pairs $arg_v = \{(\overrightarrow{SUBJ_l},\overrightarrow{OBJ_l})\}_l$. I wish to calculate the basis weightings $\{c^v_{ijk}\}_{ijk}$ for $v$. Exploiting my earlier definition of the basis $\{s_j\}_j$ of $S_T$ which states that for any value of $i$ and $k$ there is some value of $j$ such that $s_j = (n_i, n_k)$, I define $\Delta_{ijk} = 1$ if $s_j = (n_i, n_k)$ and $0$ otherwise. Using all of the above, I define the calculation of each basis weight $c^v_{ijk}$ as:

\[
c^v_{ijk} = \sum_l{\Delta_{ijk} c^{SUBJ_l}_i c^{OBJ_l}_k}
\]

This allows for a full formulation of $\overrightarrow{v}$ as follows:

\[
\overrightarrow{v} = \sum_{l} { \sum_{ijk}{ \Delta_{ijk} c^{SUBJ_l}_i c^{OBJ_l}_k \overrightarrow{n_i} \otimes \overrightarrow{s_j} \otimes \overrightarrow{n_k} }}
\]

The nested sums here may seem computationally inefficient, seeing how this would involve computing $size(arg_v) \times dim(N)^2 \times dim(S) = size(arg_v) \times dim(N)^4$ products. However, using the decomposition of basis elements of $S$ into pairs of basis elements of $N$ (effectively basis elements of $N \otimes N$), we can remove the $\Delta_{ijk}$ term and ignore all values of $j$ where $s_j \neq (n_i, n_k)$, since the basis weight for this combination of indices would be $0$. Therefore I simplify the formulation of $\overrightarrow{v}$:

\begin{align*}
\overrightarrow{v} & = \sum_{l} { \sum_{ik}{ c^{SUBJ_l}_i c^{OBJ_l}_k \overrightarrow{n_i} \otimes \overrightarrow{(n_i,n_k)} \otimes \overrightarrow{n_k} }}\\
\end{align*}

This representation is still bloated: we perform less calculations, but still obtain a vector in which all the basis weights where $s_j \neq (n_i, n_k)$ are $0$, hence where only $dim(N)^2$ of the $dim(N)^4$ values are non-zero. In short, the vector weights for $\overrightarrow{v}$ are, under this learning algorithm, entirely characterised by the values of a $dim(N)$ by $dim(N)$ matrix, the entries of which are products $c^{SUBJ_l}_i c^{OBJ_l}_k$ where $i$ and $k$ have become row and column indices.

Using this and my definition of $S_T$ as a space isomorphic to $N \otimes N$, I can formulate a compact, `reduced' expression of $\overrightarrow{v}$ as follows. Let the Kronecker product of two vectors $\overrightarrow{u}, \overrightarrow{w} \in N$, written $\overrightarrow{u} \otimes \overrightarrow{w} \in N \otimes N$, be as follows:
\[
\overrightarrow{u} \otimes \overrightarrow{w} = \sum_{ij}{c^u_i c^w_j \overrightarrow{n_i} \otimes \overrightarrow{n_j}}
\]
Equipped with this definition, I can formulate the compact form of $\overrightarrow{v}$:
\begin{align*}
\overrightarrow{v} = & \sum_l {\sum_{ik} {c^{SUBJ_l}_i c^{OBJ_l}_k\overrightarrow{n_i} \otimes \overrightarrow{n_k} } }\\ 
= & \sum_l {\overrightarrow{SUBJ_l} \otimes \overrightarrow{OBJ_l}}
\end{align*}

In short, we are only required to iterate through the corpus once, taking for each instance of a transitive verb $v$ the Kronecker product of its subject and object, and summing these across all instances of $v$. It is simple to see that no information was discarded relative to the previous definition of $\overrightarrow{v}$: the dimensionality reduction by a factor of $dim(N)^2$ simply discards all basis elements for which the basis weight was $0$ by default.

% subsection a_learning_algorithm (end)

\subsection{Problems with Reduced Representations} % (fold)
\label{sub:problems_with_reduced_representations}

This raises a small problem though: this compact representation can no longer be used in the compositional mechanism presented in $\S$\ref{sec:a_categorical_passage_from_grammar_to_semantics}, as the dimensions of $\overrightarrow{v}$ no longer match those which it is required to have according to its syntactic type. However, a solution can be devised if we return to the sample calculation, shown in $\S$\ref{sub:fvect_examples}, of the composition of a transitive verb with its arguments. The composition is as follows:
\begin{align*}
    (\epsilon_N \otimes 1_S \otimes \epsilon_N) (\overrightarrow{SUBJ} \otimes \overrightarrow{v} \otimes \overrightarrow{OBJ}) = \sum_{ikm}{ c_i^{SUBJ} c_{ikm}^{v} c_{m}^{OBJ} \overrightarrow{s_k}} 
\end{align*}
where the verb $v$ is represented in its non-compact form. By introducing the compact representation permitted by the isomorphism $S_T \cong N \otimes N$ I can express this as
\[
\overrightarrow{SUBJ\ v\ OBJ} = \sum_{im}{ c_i^{SUBJ} c_{im}^{v} c_{m}^{OBJ} \overrightarrow{n_i} \otimes \overrightarrow{n_m} } 
\]
where  $v$ is represented in its compact form. Furthermore, by introducing the component wise multiplication operation $\odot$:
\[
\overrightarrow{u} \odot \overrightarrow{v} = \sum_i{c^u_i c^v_i \overrightarrow{n_i}}
\]
I can show the general form of transitive verb composition using the reduced verb representation to be as follows:
\begin{align*}
    \overrightarrow{SUBJ\ v\ OBJ}  & = \sum_{im}{ c_i^{SUBJ} c_{im}^{v} c_{m}^{OBJ} \overrightarrow{n_i} \otimes \overrightarrow{n_m} } \\
    & = \left( \sum_{im}{  c_{im}^{v}  \overrightarrow{n_i} \otimes \overrightarrow{n_m} }  \right) \odot \left( \sum_{im}{ c_i^{SUBJ} c_{m}^{OBJ} \overrightarrow{n_i} \otimes \overrightarrow{n_m} } \right)\\
    & = \overrightarrow{v} \odot \left( \overrightarrow{SUBJ} \otimes \overrightarrow{OBJ} \right)
\end{align*}

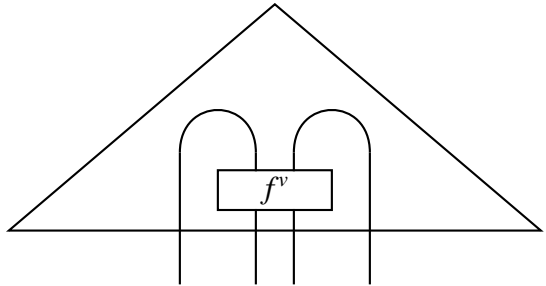
\begin{figure}[ht]
\centering
\tikzstyle{bordered} = [draw,outer sep=0,inner sep=1,minimum size=15]
\tikzstyle{vector} = [draw, isosceles triangle, shape border rotate=90, isosceles triangle stretches, outer sep=0,inner sep=1, minimum height=5, minimum width=30]
\begin{tikzpicture}[thick]

\node[vector,minimum width=7cm,minimum height=3cm] at (0,2) {};
\node[bordered,minimum width=1.5cm] at (0,1.25) {$f^v$};

\draw (-1.25,0) -- (-1.25,1.75);
\draw (1.25,0) -- (1.25,1.75);
\draw (-.25,1.5) -- (-.25,1.75);
\draw (.25,1.5) -- (.25,1.75);
\draw (-.25,1) -- (-.25,0);
\draw (.25,1) -- (.25,0);

\draw (-1.25,1.75) .. controls +(0,.75) and +(0,.75) .. (-.25,1.75);
\draw (1.25,1.75) .. controls +(0,.75) and +(0,.75) .. (.25,1.75);

\end{tikzpicture}
\caption{Diagrammatic form of reduced representations.}
\label{fig:reducedreprdiag}
\end{figure}

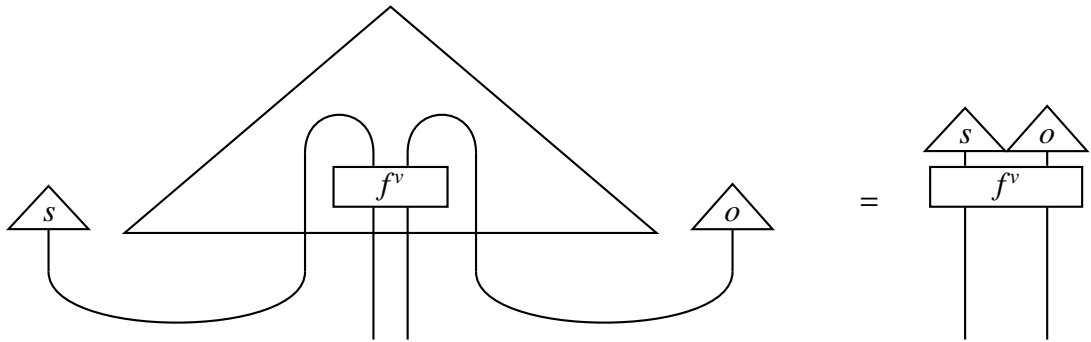
\begin{figure}[ht]
    \centering
    \tikzstyle{bordered} = [draw,outer sep=0,inner sep=1,minimum size=15]
\tikzstyle{vector} = [draw, isosceles triangle, shape border rotate=90, isosceles triangle stretches, outer sep=0,inner sep=1, minimum height=5, minimum width=30]
    \begin{tikzpicture}[thick,scale=0.9]

\node[vector,minimum width=7cm,minimum height=3cm] at (0,2) {};
\node[bordered,minimum width=1.5cm] at (0,1.25) {$f^v$};

\draw (-1.25,0) -- (-1.25,1.75);
\draw (1.25,0) -- (1.25,1.75);
\draw (-.25,1.56) -- (-.25,1.75);
\draw (.25,1.56) -- (.25,1.75);
\draw (-.25,0.95) -- (-.25,-1);
\draw (.25,0.95) -- (.25,-1);

\draw (-1.25,1.75) .. controls +(0,.75) and +(0,.75) .. (-.25,1.75);
\draw (1.25,1.75) .. controls +(0,.75) and +(0,.75) .. (.25,1.75);

\node[vector] (subj) at (-5,.85) {$s$};
\node[vector] (obj) at (5,.85) {$o$};
\draw (subj) -- (-5,0);
\draw (obj) -- (5,0);

\draw (-5,0) .. controls +(0,-1) and +(0,-1) .. (-1.25,0);
\draw (5,0) .. controls +(0,-1) and +(0,-1) .. (1.25,0);

\node at (7,1) {$=$};

\node[bordered,minimum width=2cm] at (9,1.25) {$f^v$};
\node[vector] (subj2) at (8.4,2) {$s$};
\node[vector] (obj2) at (9.6,2) {$o$};
\draw (8.4,1.56) -- (subj2);
\draw (9.6,1.56) -- (obj2);
\draw (8.4,.95) -- (8.4,-1);
\draw (9.6,.95) -- (9.6,-1);
\end{tikzpicture}
\caption{Composition under reduced representations.}
\label{fig:redcompdiag}
\end{figure}

Considering the diagrammatic form of reduced representations may help understand their construction and how composition works. Let $\overrightarrow{v}$ be the reduced representation of a verb, and the function $f^v$ be defined as follows:
\[
f^v : N \otimes N \to N \otimes N :: \overrightarrow{a} \otimes \overrightarrow{b} \mapsto \overrightarrow{v} \odot \left( \overrightarrow{a} \otimes \overrightarrow{b}\right)
\]
The diagrammatic representation of a verb is shown in Figure~\ref{fig:reducedreprdiag} as being embedded within the full representation of the verb. Composition of a subject and object with the full representation results in a yank which feeds the vectors to the function $f^v$ which applies the reduced representation to the Kronecker product of the subject and object vectors, as shown in Figure~\ref{fig:redcompdiag}. In sum, I am doing exactly the same sort of computation when I use reduced representations as I do when I used full representations. The `simplification' of reduced representations is only a reduction in the cost of the computation; in all other respects, it is identical to computing using $\epsilon$ maps.

To summarise what I have done with transitive verbs:
\begin{enumerate}
    \item I have treated them as functions taking two nouns and returning a sentence in a space $S_T \cong N \otimes N$.
    \item I have built the semantic representations for these functions by counting which properties of subject and object noun vectors are related by the verb in transitive sentences in a corpus.
    \item I have assumed that output properties are a function of input properties by the use of the $\Delta_{ijk}$ function, i.e.~the weights associated to $n_i$ from the subject argument and with $n_k$ from the object argument only affect the `output' weight for the sentence basis element $\overrightarrow{s_j}=\overrightarrow{n_i}\otimes\overrightarrow{n_k}$.
    \item I have shown that this leads to a compact representation of the verb's semantic form, and an optimised learning procedure.
    \item I have shown that the composition operations of this formalism can be adapted to this compact representation.
\end{enumerate}

The compact representation and amended composition operation hinge on the choice of $S_T \cong N \otimes N$ as output type for $N \otimes N$ as input type (the pair of arguments the verb takes), justifying my choice of a transitive sentence space. In the intransitive case, the same phenomenon can be observed, since such a verb takes as argument a vector in $N$ and produces a vector in $S_I \cong N$. Furthermore, my choice to make all other types dependent on one base type---namely $n$ (with associated semantic space $N$)---yields the same property for every relational word we wish to learn: the output type is the same as the concatenated (on the syntactic level) or tensored (on the semantic level) input types. It is this symmetry between input and output types that guarantees that any $m$-ary relation, expressed in the original formulation as an element of tensor space $\underbrace{N \otimes \ldots \otimes N}_{2m}$ has a compact representation in $\underbrace{N \otimes \ldots \otimes N}_{m}$ where the $i$th basis weight of the reduced representation stands for the degree to which the $i$th element of the input vector affects the $i$th element of the output vector.
% subsection problems_with_reduced_representations (end)

\subsection{A Generalised Algorithm for Reduced Repesentations} % (fold)
\label{sub:a_generalised_algorithm_for_reduced_repesentations}

The learning procedure discussed above allows the specification of a generalised learning algorithm for reduced representations, first presented in \cite{Grefenstette2011a}, which is as follows. Each relational word $P$ (i.e.~those words with compound pregroup types) with grammatical type $\pi$ and $m$ adjoint types $\alpha_1, \alpha_2, \cdots, \alpha_m$  is encoded as an  $({r\times \ldots \times r})$ multi-dimensional array with $m$ degrees of freedom (i.e.~an order $m$ tensor). Since the vector space $N$ has a fixed basis, each such array is represented in vector form as follows:
\[
\overrightarrow{\text{P}} = \sum_{\underbrace{ij \cdots \zeta}_{m}} c_{ij \cdots  \zeta} \ \underbrace{(\overrightarrow{n}_i\otimes \overrightarrow{n}_j \otimes \cdots \otimes \overrightarrow{n}_\zeta)}_{m}
\]
This vector lives in the tensor space $\underbrace{N \otimes N \otimes \cdots \otimes N}_m$. Each $c_{ij \cdots  \zeta}$ is computed according to the procedure described in Figure~\ref{fig:gen-matrix}. 

\begin{figure}[h!]
\fbox{\begin{minipage}{\textwidth}
{\bf 1)}\ Consider a sequence of words containing a relational word `P' and its arguments w$_1$, w$_2$, $\cdots$, w$_m$, occurring in the same order as described in P's grammatical type $\pi$. Refer to these sequences as `P'-relations. Suppose there are $K$ of them.\\
{\bf 2)}  Retrieve the vector $\overrightarrow{\text{w}}_l$ of each argument w$_l$.\\
{\bf 3)}\ Suppose  w$_1$ has weight $c^1_i$ on basis vector $\overrightarrow{n}_i$, w$_2$ has weight $c^2_j$ on basis vector $\overrightarrow{n}_j$,  $\cdots$, and  w$_m$ has weight $c^m_\zeta$ on basis vector $\overrightarrow{n}_\zeta$.  Multiply   these weights

\[
c^1_i \times c^2_j \times \cdots \times c^m_\zeta
\]

{\bf 4)} Repeat the above steps for all the $K$ `P'-relations, and sum the corresponding weights

\[
c_{ij \cdots  \zeta} = \sum_{k=1}^K \left(c^1_i \times c^2_j \times \cdots \times c^m_\zeta \right)_k
\]
\end{minipage}}
\caption{Procedure for learning  weights for matrices of words `P' with relational types $\pi$ of $m$ arguments.}
\label{fig:gen-matrix}
\end{figure}

\medskip
\noindent
Linear algebraically, this procedure corresponds to computing the following sum:
\[
\overrightarrow{\text{P}} = \sum_k \left(\overrightarrow{\text{w}}_1 \otimes \overrightarrow{\text{w}}_2 \otimes \cdots \otimes \overrightarrow{\text{w}}_m
\right)_k
\]

The general formulation of composing a relational word $P$ with its arguments $arg_1,\ldots,arg_m$ is now expressed as
\[
\overrightarrow{P} \odot \left(\overrightarrow{arg_1} \otimes \ldots \otimes \overrightarrow{arg_m} \right)
\]
For example the computation of ``furry cats nag angry dogs'' would correspond to the following operation:
\[
\overrightarrow{\textrm{furry cats nag angry dogs}} = \overrightarrow{\textrm{nag}} \odot \left( \left(\overrightarrow{\textrm{furry}} \odot \overrightarrow{\textrm{cat}}\right) \otimes \left(\overrightarrow{\textrm{angry}} \odot \overrightarrow{\textrm{dog}}\right) \right)
\]

% subsection a_generalised_algorithm_for_reduced_repesentations (end)

\subsection{Example} % (fold)
\label{sub:example}

To give an example, taken from \cite{Grefenstette2011a}, I demonstrate how the meaning of the world `show' might be learned from a corpus and then composed. Suppose there are two instances of the verb `show' in the corpus:

\vspace{0mm}
\begin{tabular}{ccll}
$s_1$ &\quad = \quad & table show result\\
$s_2$ & \quad = \quad & map show location
\end{tabular}

\smallskip
\noindent
The vector of `show' is

\[\overrightarrow{\text{show}} \ = \  
\overrightarrow{\text{table}} \otimes \overrightarrow{\text{result}} \ + \
\overrightarrow{\text{map}} \otimes \overrightarrow{\text{location}}
\]

\noindent
Consider a vector space $N$ with four basis vectors `far', `room', `scientific', and `elect'. The TF/IDF-weighted values for vectors of selected nouns (built from the British National Corpus) are as shown in Table~\ref{tab:sample_weights}.

\begin{table}[h!]
    \begin{center}
        \begin{tabular}{c|c|c|c|c|c|c|c}
        \hline
        $\mathbf{i}$ & $\overrightarrow{\mathbf{n_i}}$ &  table & map & result & location & master & dog\\
        \hline
        \hline

        1&far &6.6 & 5.6 &7& 5.9 & 4.3 & 5.5\\

        2&room &  27 & 7.4 &1.0& 7.3 & 9.1 & 7.6\\

        3&scientific& 0& 5.4& 13& 6.1 & 4.1 & 0\\

        4&elect & 0  & 0&4.2&0 & 6.2 & 0
        \end{tabular}
    \end{center}
    \caption{Sample  weights for selected noun vectors.}
    \label{tab:sample_weights}
\end{table}

\noindent
Part of the matrix compact representation of `show' is presented in Table~\ref{tab:show_matrix}.

\begin{table}[h]
    \begin{center}
        \begin{tabular}{c|c|c|c|c}
        \hline

        &far & room & scientific & elect\\
        \hline

        far &\fbox{79.24} &47.41&119.96&27.72\\
        room& 232.66 & 80.75&396.14& 113.2\\
        scientific &32.94&31.86 &32.94&0\\
        elect&0&0&0&0\\

        \end{tabular}
        \end{center}
    \caption{Sample semantic matrix for  `show'.}
    \label{tab:show_matrix}
\end{table}
 
 \medskip
 \noindent
As a sample computation, the weight  $c_{11}$ for basis element $(\overrightarrow{n_1},\overrightarrow{n_1})$, i.e.~$(\overrightarrow{\text{far}}, \overrightarrow{\text{far}})$, is computed by multiplying weights of `table' and `result' on $\overrightarrow{\text{far}}$,  i.e.~$6.6 \times 7$, multiplying weights of `map' and `location' on $\overrightarrow{\text{far}}$, i.e.~$5.6 \times 5.9$ then adding these $46.2 + 33.04$ and obtaining the total weight $79.24$. 

I now wish to compute the vector for the sentence ``[the] master shows [his] dog'' (omitting the determiner and possessive for simplicity). The calculation will be:

\begin{align*}
    & \overrightarrow{\textrm{master show dog}} \\ 
    & \qquad = \overrightarrow{\textrm{show}} \odot \left( \overrightarrow{\textrm{master}} \otimes \overrightarrow{\textrm{dog}} \right)\\
    & \qquad = 
    \left[
    \begin{tabular}{cccc}
    79.24 &47.41&119.96&27.72\\
    232.66 & 80.75&396.14& 113.2\\
    32.94&31.86 &32.94&0\\
    0&0&0&0\\
    \end{tabular}
    \right]
    \odot
    \left[
    \begin{tabular}{cccc}
    23.65 & 32.68 & 0 & 0\\
    50.05 & 69.16 & 0 & 0\\
    22.55 & 31.16 & 0 & 0\\
    34.1  & 47.12 & 0 & 0\\
    \end{tabular}
    \right]\\
    & \qquad =
    \left[
    \begin{tabular}{cccc}
     1874.03 &   1549.36&      0    &      0\\
    11644.63 &   5584.67  &      0    &      0\\
    742.80 &    992.76&      0    &      0\\
    0    &      0    &      0    &      0\\
    \end{tabular}
    \right]
\end{align*}

The row-wise flattening of the final matrix representation gives us the result we seek, namely the sentence vector in $S_T$ for ``[the] master shows [his] dog'':
\[
\overrightarrow{\textrm{master show dog}} = [1874 ,   1549,      0    ,      0    ,  11645,
         5585  ,      0,      0,   743 , 993, 0, 0, 0, 0,  0, 0]
\]

% subsection example (end)

\subsection{An Efficient Alternative} % (fold)
\label{sub:an_efficient_alternative}

To complete this chapter, let us consider an alternative and efficient way of learning the representations for transitive verbs, which I will call the \textbf{Kronecker} method. I first presented this efficient learning method with Mehrnoosh Sadrzadeh in \cite{Grefenstette2011b}, where we observed that the compact representation of a verb in the DisCoCat framework, under the assumptions presented earlier in this chapter, can be viewed as $dim(N) \times dim(N)$ matrices in $N \otimes N$. We considered alternatives to the algorithm presented earlier for the construction of such matrices, and were surprised by the results of the \textbf{Kronecker} method, wherein we replaced the matrix learned by our algorithm with the Kronecker product of the lexical semantic vectors for the verb. Concretely this means that given the lexical vector $\overrightarrow{v_{lex}}$ for a transitive verb, built as we would build noun vectors, we construct the matrix representation $\mathbf{verb}$ of the verb as follows:
\[
\mathbf{verb} = \overrightarrow{v_{lex}} \otimes \overrightarrow{v_{lex}}
\] 

Further analysis performed since the publication of that paper can help to understand why this method might work. Using the following property that for any vectors $\overrightarrow{a},\overrightarrow{b},\overrightarrow{c},\overrightarrow{d}$, we have
\[
(\overrightarrow{a} \otimes \overrightarrow{b}) \odot (\overrightarrow{c} \otimes \overrightarrow{d}) =
(\overrightarrow{a} \odot \overrightarrow{c}) \otimes (\overrightarrow{b} \odot \overrightarrow{d})
\]
we can see that the \textbf{Kronecker} model's composition operation can be expressed as
\[
\textbf{Kronecker}:\ \overrightarrow{subject\ verb\ object} = \left(\overrightarrow{verb} \odot \overrightarrow{subject}\right) \otimes \left(\overrightarrow{verb} \odot  \overrightarrow{object}\right)
\]
Bearing in mind that the cosine measure we are using as a similarity metric is equivalent to the inner product of two vectors normalised by the product of their length
\[
cosine(\overrightarrow{a}, \overrightarrow{b}) = \frac{\langle \overrightarrow{a} \vert \overrightarrow{b} \rangle}{\Vert \overrightarrow{a} \Vert \times \Vert \overrightarrow{b} \Vert}
\]
 and the following property of the inner product of Kronecker products
\[
\langle \overrightarrow{a} \otimes \overrightarrow{b} \vert \overrightarrow{c} \otimes \overrightarrow{d} \rangle = \langle \overrightarrow{a} \vert \overrightarrow{c} \rangle \times \langle \overrightarrow{b} \vert \overrightarrow{d} \rangle
\]
we finally observe that comparing two sentences under \textbf{Kronecker} corresponds to the following computation:
\begin{align*}
    & cosine(\overrightarrow{subject\ verb_1\ object}, \overrightarrow{subject\ verb_2\ object})\\
    & \qquad = \alpha \left\langle \left(\overrightarrow{verb_1} \otimes \overrightarrow{verb_1}\right) \odot \left(\overrightarrow{subject} \otimes  \overrightarrow{object}\right) \vert \left(\overrightarrow{verb_2} \otimes \overrightarrow{verb_2}\right) \odot \left(\overrightarrow{subject} \otimes  \overrightarrow{object}\right) \right\rangle\\
    & \qquad = \alpha \left\langle \left(\overrightarrow{verb_1} \odot \overrightarrow{subject}\right) \otimes \left(\overrightarrow{verb_1} \odot  \overrightarrow{object}\right)\vert \left(\overrightarrow{verb_2} \odot \overrightarrow{subject}\right) \otimes \left(\overrightarrow{verb_2} \odot  \overrightarrow{object}\right) \right\rangle\\
    & \qquad = \alpha \left\langle \left( \overrightarrow{verb_1} \odot \overrightarrow{subject} \right) \vert \left( \overrightarrow{verb_2} \odot \overrightarrow{subject} \right) \right\rangle \left\langle \left( \overrightarrow{verb_1} \odot \overrightarrow{object} \right)\vert \left( \overrightarrow{verb_2} \odot \overrightarrow{object} \right) \right\rangle
\end{align*}
where $\alpha$ is the normalisation factor
\[
\alpha = \frac{1}{\Vert\overrightarrow{subject\ verb_1\ object}\Vert \times \Vert \overrightarrow{subject\ verb_2\ object}\Vert}
\]
I note here that the \textbf{Kronecker} model is effectively a parallel application of the multiplicative model presented in $\S$\ref{sub:multiplicative_models}, combining the subject and verb, and object and verb separately. In short, it more or less constitutes the introduction of some mild syntactic sensitivity into the multiplicative model of \cite{mitchell2008vector}.

I can generalise this Kronecker model to any relation of any arity, provided that it takes $n$ inputs and produces $n$ outputs of matching types (since all I am doing is component-wise multiplication of reduced representations). The general learning procedure for learning an $n$-ary relation using the Kronecker method is as follows:
\begin{enumerate}
    \item We learn the lexical vector $R$ for the relation.
    \item We produce its tensor representation $T^R$ by computing
    \[
    T^R = \bigotimes_{i=1}^n{R}
    \]
    \item To compose $T^R$ with its $N$ arguments $\{\overrightarrow{a_1},\ldots,\overrightarrow{a_n}\}$ we compute:
    \[
        T^R \odot \left(\bigotimes_{i=1}^n{\overrightarrow{a_i}}\right)
    \]
\end{enumerate}
This is, in some sense, a Kronecker product-based generalisation of the multiplicative model of \cite{mitchell2008vector}, as the case for $n=1$ is simply component-wise multiplication.

\begin{figure}[ht]
    \centering

    \tikzstyle{bordered} = [draw,outer sep=0,inner sep=1,minimum size=15]
\tikzstyle{vector} = [draw, isosceles triangle, shape border rotate=90, isosceles triangle stretches, outer sep=0,inner sep=1, minimum height=5, minimum width=30]
    \begin{tikzpicture}[thick]

\begin{scope}
    \node[bordered,minimum width=2cm] at (0,1.25) {$f^v$};
\node[vector] (subj2) at (-.6,2) {$s$};
\node[vector] (obj2) at (0.6,2) {$o$};
\draw (-.6,1.5) -- (subj2);
\draw (0.6,1.5) -- (obj2);
\draw (-.6,1) -- (-.6,0);
\draw (0.6,1) -- (0.6,0);
\end{scope}

\begin{scope}[xshift=2cm]
\node at (0,1.25) {$=$};
\end{scope}

\begin{scope}[xshift=4cm]
 \node[bordered] at (-.6,1.25) {$f^{\bar{v}}$};
 \node[bordered] at (.6,1.25) {$f^{\bar{v}}$};   
\node[vector] (subj2) at (-.6,2) {$s$};
\node[vector] (obj2) at (0.6,2) {$o$};
\draw (-.6,1.5) -- (subj2);
\draw (0.6,1.5) -- (obj2);
\draw (-.6,1) -- (-.6,0);
\draw (0.6,1) -- (0.6,0);

\end{scope}

\end{tikzpicture}

\caption{Composition under the Kronecker model.}
\label{fig:kroncompdiag}
\end{figure}
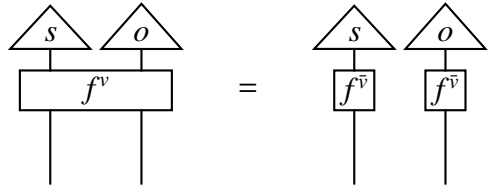

\begin{figure}[ht]
    \centering
    \tikzstyle{bordered} = [draw,outer sep=0,inner sep=1,minimum size=15]
\tikzstyle{vector} = [draw, isosceles triangle, shape border rotate=90, isosceles triangle stretches, outer sep=0,inner sep=1, minimum height=5, minimum width=30]
\begin{tikzpicture}[thick]

\begin{scope}
    \node[bordered,minimum width=4cm] at (0,1.25) {$f^R$};
\node[vector] (subj2) at (-1.5,2) {$a_1$};
\node[vector] (obj2) at (1.5,2) {$a_n$};
\draw (-1.5,1.5) -- (subj2);
\draw (1.5,1.5) -- (obj2);
\draw (-1.5,1) -- (-1.5,0);
\draw (1.5,1) -- (1.5,0);

\node at (0,2) {$\ldots$};

\node at (0,.5) {$\ldots$};

\end{scope}

\begin{scope}[xshift=3cm]
\node at (0,1.25) {$=$};
\end{scope}

\begin{scope}[xshift=4cm]
 \node[bordered] at (0,1.25) {$f^{\bar{R}}$};
 \node[bordered] at (2,1.25) {$f^{\bar{R}}$};   
\node[vector] (subj2) at (0,2) {$a_1$};
\node[vector] (obj2) at (2,2) {$a_n$};
\draw (0,1.5) -- (subj2);
\draw (2,1.5) -- (obj2);
\draw (0,1) -- (0,0);
\draw (2,1) -- (2,0);

\node at (1,1.25) {$\ldots$};

\end{scope}

\end{tikzpicture}
    \caption{Composition under the generalised Kronecker model.}
    \label{fig:genkroncompdiag}
\end{figure}
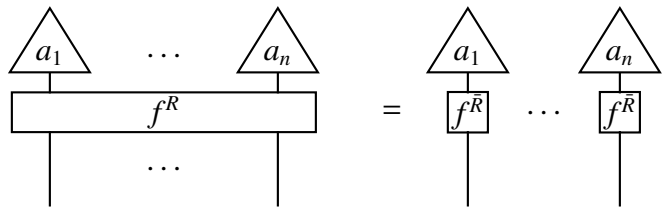

To conclude this section, let us examine the diagrammatic representation of composition under the Kronecker model. We saw, in Figure~\ref{fig:redcompdiag}, that reduced representations could be seen as being diagrammatically embedded inside full representations. Furthermore, these diagrams can be simplified through yank operations to show the application of some function $f^v$ to the Kronecker product of subject and object, where $f^v$ corresponds to the component-wise multiplication of Kronecker product two arguments with the reduced representation of the verb. In the case of the Kronecker model, this reduced representation is itself the Kronecker product of the lexical vectors for the verb, which I will call $\bar{v}$. Let a function $f^{\bar{v}}$ be defined as follows:
\[
f^{\bar{v}} : N \to N :: \overrightarrow{a} \mapsto \overrightarrow{a} \odot \bar{v}
\]
I can therefore express $f^v$ in terms of $f^{\bar{v}}$ as follows:
\[
f^v = f^{\bar{v}} \otimes f^{\bar{v}}
\]
which allows us to formulate the diagrammatic representation of composition under the Kronecker model as shown in Figure~\ref{fig:kroncompdiag}. The general form of the Kronecker model can be shown in a similar way. If we construct the representation of some $n$-ary relation $R$ according to the Kronecker model, by taking the Kronecker product of its lexical vector $\bar{R}$ $n$ times, then we have:
\[
f^R = \bigotimes_{i=1}^n{f^{\bar{R}}}
\]
and the composition of $R$ with its $n$ arguments $\{\overrightarrow{a_1},\ldots,\overrightarrow{a_n}\}$ can be represented diagrammatically as shown in Figure~\ref{fig:genkroncompdiag}.

% subsection an_efficient_alternative (end)

% section learning_procedures (end)

% chapter learning_procedures_for_a_discocat (end)

\mbox{}
\newpage

\part{Practice}
\mbox{}
\newpage
%!TEX root = ../grefenstettethesis.tex

\chapter{Evaluating a DisCoCat} % (fold)
\label{cha:evaluating_a_discocat}

\begin{chabstract}
  This chapter presents three phrase-similarity detection experiments designed to evaluate the models discussed in Chapter~\ref{cha:learning_procedures_for_a_discocat}, and compare them to selected other models discussed in Chapter~\ref{cha:literature_review} within the context of these experiments. It shows the concrete models generated by the DisCoCat framework to outperform other competing models.
\end{chabstract}

Evaluating compositional models of semantics is no easy task: first, we are trying to evaluate how well the compositional process works; second, we also are trying to determine how useful the final representation---the \emph{output} of the composition---is, relative to our needs. 

The scope of this second problem covers most phrase and sentence-level semantic models, from bag-of-words approaches in information retrieval to logic-based formal semantic models, via language models used for machine translation. It is heavily task dependent, in that a representation that is suitable for machine translation may not be appropriate for textual inference tasks, and one that is appropriate for information retrieval may not be ideal for paraphrase detection. Therefore this aspect of semantic model evaluation ideally should take the form of application-oriented testing, e.g.~to test semantic representations designed for machine translation purposes, we should use a machine translation evaluation task.

The DisCoCat framework of \cite{Clark2008,Coecke2010}, described above in Chapter~\ref{cha:foundations_of_discocat}, allows for the composition of any words of any syntactic type. The general learning algorithm presented in $\S$\ref{sec:learning_procedures} technically can be applied to learn and model relations of any semantic type. However, many open questions remain, such as how to deal with logical words, determiners and quantification, and how to reconcile the different semantic types used for sentences with transitive and intransitive sentences. I will leave these questions for future work, briefly discussing some of them in Chapter~\ref{cha:applications_and_future_work}. In the meantime, we concretely are left with a way of satisfactorily modelling only simple sentences without having to answer these bigger questions.

With this in mind, in this chapter I present a series of experiments centred around evaluating how well various models of semantic vector composition (along with the one described in Chapter~\ref{cha:learning_procedures_for_a_discocat}) perform in a phrase similarity comparison task. This task aims to test the quality of a compositional process by determining how well it forms a clear joint meaning for potentially ambiguous words. The intuition here is that tokens, on their own, can have several meanings, and that it is through the compositional process---through giving them \emph{context}---that we understand their specific meaning. For example, ``bank'' itself could (amongst other meanings) mean a river bank or a financial bank; yet in the context of a sentence such as ``The bank refunded the deposit'' it is likely we are talking about the financial institution.

I will present three datasets designed to evaluate how well word-sense disambiguation occurs as a by-product of composition. I begin by describing the first dataset, based around noun-intransitive verb phrases, in $\S$\ref{sec:first_experiment}. In $\S$\ref{sec:second_experiment}, I present a dataset based around short transitive-verb phrases (a transitive verb with subject and object). In $\S$\ref{sec:third_experiment}, I discuss a new dataset, based around short transitive-verb phrases where the subject and object are qualified by adjectives. I leave discussion of these results for $\S$\ref{sec:discussion}.

\section{First Experiment} % (fold)
\label{sec:first_experiment}

This first experiment, which is based on that of Mitchell and Lapata \cite{mitchell2008vector}, and which I presented in \cite{Grefenstette2011a} with Mehrnoosh Sadrzadeh, evaluates the degree to which an ambiguous intransitive verb (e.g.~``draws'') is disambiguated by combination with its subject. 

\subsection{Dataset Description}
 The dataset\footnote{Available at \texttt{http://homepages.inf.ed.ac.uk/s0453356/results}.} comprises 120 pairs of intransitive sentences, each of the form ``NOUN VERB''. These sentence pairs are generated according the following procedure, which will be the basis for the construction of the other datasets discussed below:

\begin{enumerate}
    \item A number of ambiguous intransitive verbs (15, in this case) are selected from frequently occurring verbs in the corpus.
    \item For each verb $V$, two mutually exclusive synonyms $V_1$ and $V_2$ of the verb are produced, and each is paired with the original verb separately (for a total of 30 verb pairs). These are generated by taking maximally distant pairs of synonyms of the verb on Wordnet (e.g.~), but any method could be used here.
    \item For each \emph{pair of verb pairs} $(V,V_1)$ and $(V, V_2)$, two frequently occurring nouns $N_1$ and $N_2$ are picked from the corpus, one for each synonym of $V$. For example if $V$ is ``glow'' and the synonyms $V_1$ and $V_2$ are ``beam'' and ``burn'', we might choose ``face'' as $N_1$ because a face glowing and a face beaming mean roughly the same thing, and ``fire'' as $N_2$ because a fire glowing and a fire burning mean roughly the same thing.
    \item By combining the nouns with the verb pairs, we form two high similarity triplets $(V,V_1,N_1)$ and $(V,V_2,N_2)$, and two low similarity triplets $(V,V_1,N_2)$ and $(V,V_2,N_1)$.
\end{enumerate}
The last two steps can be repeated to form more than four triplets per pair of verb pairs. In \cite{mitchell2008vector}, eight triplets are generated for each pair of verb pairs, obtaining a total of 120 triplets from the 15 original verbs. Each triplet, along with its HIGH or LOW classification (based on the choice of noun for the verb pair) is an entry in the dataset, and can be read as a pair of sentences: $(V,V_i,N)$ translates into the intransitive sentences ``$N$ $V$'' and ``$N$ $V_i$''.

Finally, the dataset is presented, without the HIGH/LOW ratings, to human annotators. These annotators are asked to rate the similarity of meaning of the pairs of sentences in each entry on a scale of 1 (low similarity) to 7 (high similarity). The final form of the dataset is a set of lines each containing:
\begin{itemize}
    \item A $(V, V_i, N)$ triplet
    \item A HIGH or LOW label for that triplet
    \item An annotator identifier and the annotator's score for that triplet
\end{itemize}

Sample sentences from this dataset are shown in Table~\ref{tab:intrans-sent}.

\begin{table}[h]
\begin{center}
 \begin{tabular}{|c|c|}
 \hline
 Sentence 1 & Sentence 2\\
 \hline
 \hline
 
 butler bow & butler submit  \\
\hline
 head bow & head stoop\\
 \hline
  company bow & company submit \\
\hline
 government bow & government stoop\\
 \hline
 \end{tabular}
\end{center}
\caption{Example entries from the intransitive dataset without annotator score, first experiment.}
\label{tab:intrans-sent}
\end{table}

\subsection{Evaluation Methodology} This dataset is used to compare various compositional distributional semantic models according to the following procedure:
\begin{enumerate}
    \item For each entry in the dataset, the representation of the two sentences ``$N$ $V$'' and ``$N$ $V_i$'', which we will name $S_1$ and $S_2$, formed from the entry triple $(V, V_i, N)$ is constructed by the model.
    \item The similarity of these sentences according to the model's semantic distance measure constitutes the \emph{model score} for the entry.
    \item The rank correlation of entry model scores against entry annotator scores is calculated using Spearman's rank correlation coefficient $\rho$.
\end{enumerate}

The Spearman $\rho$ scores are values between $-1$ (inverse rank correlation) and $1$ (perfect correlation). The higher the $\rho$ score, the higher the compositional model can be said to produce sentence representations that match human understanding of sentence meaning, when it comes to comparing the meaning of sentences. As such, I will rank the models evaluated using the task by decreasing order of $\rho$ score.

One of the principal appealing features of Spearman's $\rho$ is that the coefficient is rank based: it does not require models' semantic similarity metrics to be normalised for a comparison to be made. One consequence is that a model providing excellent rank correlation with human scores, but producing model scores on a small scale (e.g.values between $0.5$ and $0.6$) will obtain a higher $\rho$ score than a model producing models scores on a larger scale (e.g.~between $0$ and $1$) but with less perfect rank correlation. If we wished to then use the former model in a task requiring some greater degree of numerical separation (let's say $0$ for non-similar sentences and $1$ for completely similar sentences), we could simply renormalise the model scores to fit the scale. By eschewing score normalisation as an evaluation factor, we minimise the risk of erroneously ranking one model over another.

Finally, in addition to computing the rank alignment coefficient between model scores and annotator scores, \cite{mitchell2008vector} calculate the mean models scores for entries labelled HIGH, and for entries labelled LOW. This information is reported in their paper as additional means for model comparison. However, for the same reason I considered Spearman's $\rho$ to be a fair means of model comparison, namely in that it required no model score normalisation procedure and thus was less likely to introduce error by adding such a degree of freedom, we consider the HIGH/LOW means to be inadequate grounds for comparison precisely because it \emph{requires} normalised model scores for comparison to be meaningful. As such, I will not include these mean scores in the presentation of this experiment's results, or any further experiments in this Chapter.

\subsection{Models Compared}

In this experiment, I compare the best and worst performing models of \cite{mitchell2008vector} to my own. I begin by building a vector space $W$ for all words in the corpus, using standard distributional semantic model construction procedures ($n$-word window) with the parameters of \cite{mitchell2008vector}. Specifically, the basis elements of this vector space are the 2000 most frequently occurring words in the British National Corpus (BNC), excluding common stop words. For my evaluation, the corpus was lemmatised using Carroll's Morpha \cite{minnen2001applied}, which was applied as a by-product of my parsing of the BNC with the C\&Ctools parser of \cite{curran-clark-bos:2007:PosterDemo}. 

The context of each occurrence of a word in the corpus was defined to be five words on either side. After the vector for each word $w$ was produced, the basis weights $c^w_i$ associated with context word $b_i$ were normalised by the following ratio of probabilities weighting scheme:
\[
c^w_i = \frac{P(b_i|w)}{P(b_i)} = \frac{f_{w \land b_i}f_{b_*}}{f_{b_i}f_w}
\]
where $f_{b_*}$ is the total number of context word in the corpus.

Let $\overrightarrow{verb}$ and $\overrightarrow{noun}$ be the lexical semantic vectors for the verb and the noun, from $W$. It should be evident that there is no significant difference in using $W$ for nouns in lieu of building a separate vector space $N$ strictly for noun vectors.

As a first baseline, \textbf{Verb Baseline}, I ignored the information provided by the noun in constructing the sentence representation, effectively comparing the semantic content of the verbs:
\[\textbf{Verb Baseline}:\  \overrightarrow{noun\ verb} = \overrightarrow{verb}\]

The models from \cite{mitchell2008vector} which I evaluate here are those which were strictly unsupervised (i.e.~no free parameters for composition). These are the additive model \textbf{Add}, wherein
\[
\textbf{Add}:\ \overrightarrow{noun\ verb} = \overrightarrow{noun} + \overrightarrow{verb}
\]
and the multiplicative model \textbf{Multiply}, wherein
\[
\textbf{Multiply}:\ \overrightarrow{noun\ verb} = \overrightarrow{noun} \odot \overrightarrow{verb}
\]
Other models with parameters which must be optimised against a held-out section the dataset are presented in \cite{mitchell2008vector}. I omitted them here principally because they do not perform as well as \textbf{Multiply}, but also because the need to optimise the free parameters for this experiment and other datasets makes comparison with completely unsupervised models more difficult, and less fair.

I also evaluate the \textbf{Categorical} model from Chapter~\ref{cha:learning_procedures_for_a_discocat} here, wherein
\[
\textbf{Categorical}:\ \overrightarrow{noun\ verb} = \overrightarrow{verb_{cat}} \odot \overrightarrow{noun}
\]
where $\overrightarrow{verb_{cat}}$ is the compact representation of the relation the verb stands for, computed according to the procedure described in Chapter~\ref{cha:learning_procedures_for_a_discocat}. It should be noted that for the case of intransitive verbs, this composition operation is mathematically equivalent to that of the \textbf{Multiply} model (as component-wise multiplication $\odot$ is a commutative operation), the difference being the learning procedure for the verb vector.

For all such vector-based models, the similarity of two sentences $s_1$ and $s_2$ is taken to be the cosine similarity of their vectors, defined in $\S$\ref{sec:distr}:
\[
similarity(s_1, s_2) = cosine(\overrightarrow{s_1}, \overrightarrow{s_2})
\]

In addition to these vector-based methods, I define an additional baseline and an upper-bound. The additional baseline, \textbf{Bigram Baseline}, is a bigram-based language model trained on the BNC with SRILM \cite{stolcke2002srilm}, using the standard language model settings for computing log-probabilities of bigrams. To determine the semantic similarity of sentences $s_1 = noun\ verb_1$ and $s_2 = noun\ verb_2$ I assumed sentences have mutually conditionally independent properties, and computed the joint probability:
\[
similarity(s_1, s_2) = logP(s_1 \land s_2) = log(P(s_1)P(s_2)) = logP(s_1) + logP(s_2)
\]

The upper bound of the dataset, \textbf{UpperBound}, was taken to be the inter-annotator agreement: the average of how each annotator's score aligns with other annotator scores, using Spearman's $\rho$. The procedure for this was to take all possible pairs of annotators, measure their agreement using $\rho$, and take the average $\rho$ across all such pairs to be the inter-annotator agreement.

\subsection{Results} The results of the first experiment are shown in Table~\ref{tab:results1}. As expected from the fact that \textbf{Multiply} and \textbf{Categorical} differ only in how the verb vector is learned, the results of these two models are virtually identical, outperforming both baselines and the \textbf{Additive} model by a significant margin. However, the distance from these models to the upper bound is even greater, demonstrating that there is still a lot of progress to be made.

\begin{table}[h]
\begin{center}
\begin{tabular}{|ll|}
\hline
Model &  $\rho$\\
\hline
\hline
Verb Baseline &  0.08\\
Bigram Baseline & 0.02\\
\hline
\hline
Add & 0.04\\
Multiply &  0.17 \\
\textbf{Categorical}  & \textbf{0.17}\\
\hline
\hline
UpperBound & 0.40 \\
\hline
\end{tabular}
\end{center}
\caption{Model correlation coefficients with human judgements, first experiment. $p < 0.05$ for each $\rho$.}
\label{tab:results1}
\end{table}

% section first_experiment (end)

\section{Second Experiment} % (fold)
\label{sec:second_experiment}

This second experiment, which I initially presented in \cite{Grefenstette2011a} with Mehrnoosh Sadrzadeh, is an extension of the first experiment to the case of sentences centred around transitive verbs, composed with a subject and an object. The results of the first experiment did not demonstrate any difference between the multiplicative model, which takes into account no syntactic information or word ordering, and the syntactically motivated categorical compositional model. By running the same experiment over a new dataset, where the relations expressed by the verb have a higher arity than in the first, I hope to demonstrate that added structure leads to better results for our syntax-sensitive model.

\subsection{Dataset Description}

The construction procedure for this dataset\footnote{Available at \url{http://www.cs.ox.ac.uk/activities/compdistmeaning/GS2011data.txt}.} is almost exactly as for the first dataset, with the following differences:
\begin{itemize}
    \item Verbs are transitive instead of intransitive.
    \item For each verb pair, I select a set of subject and object nouns to use as context, as opposed to just a subject noun.
\end{itemize}
Lemmatised sentences from sample entries of this dataset are shown in Table~\ref{tab:trans-sent}.

\begin{table}[h]
\begin{center}
 \begin{tabular}{|c|c|}
 \hline
 Sentence 1 & Sentence 2\\
 \hline
 \hline
 
 table show result & table express  result  \\
\hline
 map show location & map picture location\\
 \hline
  table show result & table picture result\\
\hline
 map show location & map express location\\
 \hline
 \end{tabular}
\end{center}
\caption{Example entries from the transitive dataset without annotator score, second experiment.}
\label{tab:trans-sent}
\end{table}

The dataset was passed to a group of annotators, as was done for the previous dataset, who scored each pair of sentences on the same scale of $1$ (not similar in meaning) to $7$ (similar in meaning).

\subsection{Evaluation Methodology}

The methodology for this experiment is exactly that of the previous experiment. Models compositionally construct sentence representations, and compare them using a distance metric (all vector-based models once again used cosine similarity). The rank correlation of model scores with annotator scores is calculated using Spearman's $\rho$, which is used in turn to rank models.

\subsection{Models Compared}

The models compared in this experiment are those of the first experiment, with the addition of an extra trigram-based baseline (trained with SRILM, using addition of log-probability of a sentence as a similarity metric), and the \textbf{Kronecker} variation on our categorical model, previously presented in $\S$\ref{sub:an_efficient_alternative}. With $W$ as the distributional semantic space for all words in the corpus, trained using the same parameters as in the first experiment, $\overrightarrow{subj},\overrightarrow{verb},\overrightarrow{object} \in W$ as the vectors for subject, verb and object of a sentence, respectively, and with $verb_{cat}$ as the compact representation of a transitive verb learned using the algorithm presented in the previous chapter, we have the following compositional methods:
\begin{align*}
    &\textbf{Add}:\ \overrightarrow{subject\ verb\ object} = \overrightarrow{subject} + \overrightarrow{verb} + \overrightarrow{object}\\
    &\textbf{Multiply}:\ \overrightarrow{subject\ verb\ object} = \overrightarrow{subject} \odot \overrightarrow{verb} \odot \overrightarrow{object}\\
    &\textbf{Categorical}:\ \overrightarrow{subject\ verb\ object} = \overrightarrow{verb_{cat}} \odot \left(\overrightarrow{subject} \otimes \overrightarrow{object}\right)\\
    &\textbf{Kronecker}:\ \overrightarrow{subject\ verb\ object} = \left(\overrightarrow{verb} \otimes \overrightarrow{verb}\right) \odot \left(\overrightarrow{subject} \otimes  \overrightarrow{object}\right)
\end{align*}
The upper bound \textbf{UpperBound} here is, again, the inter-annotator agreement.

\subsection{Results}

The results for the second experiment are shown in Table~\ref{tab:results2}. The baseline scores fall in the $0.14$--$0.18$ range, with best results being obtained for the \textbf{Trigram Baseline}. The additive model \textbf{Add} performs comparably to the additive model in the first experiment. This is most likely due to the fact that, through addition, the semantic noise caused by the overlapping word senses present in semantic vectors is increased rather than cancelled out. In other words, information from the verb vectors (namely relatively low or high values) which might be used to implicitly discriminate senses through composition is instead ``drowned out'' by the information from the subject and object vectors through summation, rather than being used to provide a more finely-grained representation. The multiplicative model \textbf{Multiply} performs on par with the version used for the intransitive experiment, but obtains a score comparable to the baselines. 

The best performing models here are those developed and described in this thesis. The \textbf{Categorical} model now outperforms both the baselines and \textbf{Multiply}, obtaining a score of $0.21$. Finally, the newly introduced \textbf{Kronecker} model leads the pack by a steady margin, with a score of $0.28$.

The inter-annotator agreement \textbf{UpperBound} is much higher in this experiment than in the previous experiment, indicating even more room for improvement.

\begin{table}[h]
\begin{center}
\begin{tabular}{|ll|}
\hline
Model & $\rho$\\
\hline
\hline
Verb Baseline &  0.16\\
Bigram Baseline & 0.14\\
Trigram Baseline & 0.18\\
\hline
\hline
Add &  0.05\\
Multiply &  0.17 \\
Categorical & 0.21\\
\textbf{Kronecker} & \textbf{0.28}\\
\hline
\hline
UpperBound &  0.62 \\
\hline
\end{tabular}
\end{center}
\caption{Model correlation coefficients with human judgements, second experiment. $p < 0.05$ for each $\rho$.}
\label{tab:results2}
\end{table}

% section second_experiment (end)

\section{Third Experiment} % (fold)
\label{sec:third_experiment}

The third and final experiment I present is a modified version of the second dataset presented above, where the nouns in each entry are under the scope of adjectives applied to them. The intuition behind the datasets presented in $\S$\ref{sec:first_experiment} and $\S$\ref{sec:second_experiment} was that ambiguous verbs are disambiguated through composition with nouns. These nouns themselves may also be ambiguous, and a good compositional model will be capable of separating the noise produced by other meanings through its compositional mechanism to produce less ambiguous phrase representations. The intuition behind this dataset is similar, in that adjectives provide both additional information for disambiguation of the nouns they apply to, but also additional semantic noise. Therefore a good model will also be able to separate  the useful information of the adjective from its semantic noise when composing it with its argument, in addition to doing this when composing the noun phrases with the verb.

\subsection{Dataset Description}

The construction procedure for this dataset was to take the dataset from $\S$\ref{sec:second_experiment}, and, for each entry, add a pair of adjectives from those most frequently occurring in the corpus. The first adjective from the pair is applied to the first noun (subject) of the entry when forming the sentences, and the second adjective is applied to the second noun (object).

This new dataset\footnote{Available at \url{http://www.cs.ox.ac.uk/activities/compdistmeaning/GS2012data.txt}.} was then annotated again by a group of 50 annotators using Amazon's Mechanical Turk service, asked to give each sentence pair a meaning similarity score between $1$ and $7$, as for the previous datasets. As a form of quality control, I inserted ``gold standard'' sentences in the form of identical sentence pairs and rejected annotators that did not score these gold standard sentences with a high score of $6$ or $7$. Some 94 users returned annotations, of which I kept 50 according to gold standard tests. I did not  apply this ``gold standard'' quality test to the second dataset as it was annotated by friends, and am unaware of whether or not it was applied in the production of the first dataset, but believe that this can only lead to the production of higher quality annotations.

Sample sentences from this dataset are shown in Table~\ref{tab:adjtrans-sent}.

\begin{table}[h]
\begin{center}
 \begin{tabular}{|c|c|}
 \hline
 Sentence 1 & Sentence 2\\
 \hline
 \hline
 
 statistical table show good result & statistical table express good  result  \\
\hline
statistical table show good result & statistical table depict good result\\
 \hline
 \end{tabular}
\end{center}
\caption{Example entries from the adjective-transitive dataset without annotator score, third experiment.}
\label{tab:adjtrans-sent}
\end{table}

\subsection{Evaluation Methodology}

The evaluation methodology in this experiment is identical to that of the previous experiments.

\subsection{Models Compared}

In this experiment, in lieu of simply comparing compositional models ``across the board'', e.g.~using the multiplicative model for both adjective-noun composition and verb-argument composition, I experimented with different \emph{combinations of models}. This evaluation procedure was chosen because I believe that adjective-noun composition need not necessarily be the same kind of compositional process as subject-verb-object composition, but also because different models may latch onto different semantic features during the compositional process, and it would be interesting to see what model mixtures work well together.

Each mixed model has two components: a verb-argument composition model and an adjective-noun composition model. For verb-argument composition, I used the three best models from the previous experiment, namely \textbf{Multiply}, \textbf{Categorical} and \textbf{Kronecker}. For adjective-noun composition I used three different methods. With $\overrightarrow{adjective}$ and $\overrightarrow{noun}$ being the vectors for an adjective and a noun in the distributional lexical semantic space $W$ (built using the same procedure as the previous experiments) and $\overrightarrow{adj_{cat}}$ being the compact representation in the \textbf{Categorical} model, built according to the algorithm from $\S$\ref{sec:learning_procedures}, we have the following models:
\begin{align*}
    & \textbf{AdjMult}:\ \overrightarrow{adjective\ noun} = \overrightarrow{adjective} \odot \overrightarrow{noun}\\
    & \textbf{Categorical}:\ \overrightarrow{adjective\ noun} =  \overrightarrow{adjective_{cat}} \odot \overrightarrow{noun}
\end{align*}  
The third model, \textbf{AdjNoun}, is a holistic (non-compositional) model, wherein the adjective-noun compound was treated as a single token, as its semantic vector $\overrightarrow{(adjective\ noun)_{lex}} \in W$ was learned from the corpus using the same learning procedure applied to construct other vectors in $W$. Hence the model defines adjective-noun ``composition'' as:
\[
\textbf{AdjNoun}:\ \overrightarrow{adjective\ noun} =  \overrightarrow{(adjective\ noun)_{lex}}
\]

In addition to these models, I also evaluated three baselines: \textbf{Verb Baseline}, \textbf{Bigram Baseline}, and \textbf{Trigram Baseline}. As in previous experiments, the verb baseline uses the verb vector as a sentence vector, ignoring the information provided by other words. The bigram and trigram baselines are calculated from the same language model as used in the second experiment. In both cases, the log-probability of each sentence is calculated using SRLIM, and the sum of log-probabilities of two sentences is used as a similarity measure. 

Finally, three addition-based models of sentence formation were evaluated as well, using basic linear algebraic operations. With 
\[\overrightarrow{sentence} = \overrightarrow{adjective_1\ noun_1\ verb\ adjective_2\ noun_2}\] 
these are:
\begin{align*}
    & \textbf{Additive}:\ \overrightarrow{sentence} = \overrightarrow{adjective_1} + \overrightarrow{noun_1} + \overrightarrow{verb} + \overrightarrow{adjective_2} + \overrightarrow{noun_2}\\
    & \textbf{AddMult}:\ \overrightarrow{sentence} = (\overrightarrow{adjective_1} + \overrightarrow{noun_1}) \odot \overrightarrow{verb} \odot (\overrightarrow{adjective_2} + \overrightarrow{noun_2})\\
    & \textbf{MultAdd}:\ \overrightarrow{sentence} =  (\overrightarrow{adjective_1} \odot \overrightarrow{noun_1}) + \overrightarrow{verb} + (\overrightarrow{adjective_2} \odot \overrightarrow{noun_2})
\end{align*}  

\subsection{Results}

The results for the third experiment are shown in Table~\ref{tab:results3}. Going through the combined models, we can notice that in most cases the results stay the same whether the adjective-noun combination method is \textbf{AdjMult} or \textbf{CategoricalAdj}. This is because, as was shown in the first experiment, composition of a unary-relation such as an adjective or intransitive verb with its sole argument under the categorical model with reduced representations is mathematically equivalent to the multiplicative model. The sole difference is the way the adjective or intransitive verb vector is constructed. I note, however, that with \textbf{Categorical} as a verb-argument composition method, the \textbf{CategoricalAdj} outperforms \textbf{AdjMult} by a non-negligible margin ($0.19$ vs.~$0.14$), indicating that the difference in learning procedure can lead to different results depending on what other models it is combined with.

Overall, the best results are obtained for \textbf{AdjMult+Kronecker} ($\rho =0.26$) and \textbf{CategoricalAdj+Kronecker} ($\rho = 0.27$). Combinations of the adjective composition methods with other composition methods than the two listed above at best matches the best-performing baseline, \textbf{Verb Baseline}. In all cases, the holistic model \textbf{AdjNoun} provides the worst results.

Finally, I note that the fairly simple model \textbf{AddMult} trails the best performing models by only a few points.

\begin{table}[h]
\begin{center}
\begin{tabular}{|ll|}
    \hline
    Model & $\rho$\\
    \hline
    \hline
    Verb Baseline & 0.20\\
    Bigram Baseline & 0.14\\
    Trigram Baseline & 0.16\\
    \hline
    \hline
    Additive & 0.10\\
    \textbf{AddMult} & \textbf{0.24}\\
    MultAdd & 0.05\\
    \hline
    \hline
    Multiplicative & \\
    \hline
    \textbf{AdjMult} & \textbf{0.20}\\
    AdjNoun & 0.05\\
    \textbf{CategoricalAdj} & \textbf{0.20}\\
    \hline
    \hline
    Categorical & \\
    \hline
    AdjMult & 0.14\\
    AdjNoun & 0.16\\
    \textbf{CategoricalAdj} & \textbf{0.19}\\
    \hline
    \hline
    Kronecker & \\
    \hline
    \textbf{AdjMult}  & \textbf{0.26}\\
    AdjNoun & 0.17\\
    \textbf{CategoricalAdj} & \textbf{0.27}\\
    \hline
    \hline
    Upperbound & 0.48\\
    \hline
\end{tabular}
\end{center}
\caption{Model correlation coefficients with human judgements, third experiment. $p < 0.05$ for each $\rho$.}
\label{tab:results3}
\end{table}

% section third_experiment (end)

\section{Discussion} % (fold)
\label{sec:discussion}

In this Chapter, I evaluated the framework described in Chapter~\ref{cha:foundations_of_discocat}, implemented using the learning procedures described in Chapter~\ref{cha:learning_procedures_for_a_discocat}, and variants described earlier in this Chapter against other unsupervised compositional distributional models. I used non-compositional models and $n$-gram language models as baselines. These experiments show that the concrete categorical model developed here, and the Kronecker product-based variant presented alongside it, outperform all other models in each experiment save the first, where they perform on par with the leading model. As the experiments involved progressively more syntactically complicated sentences, the increased reliability of the categorical approaches relative to competing models as sentence complexity rises seems to indicate that both the Categorical and Kronecker models successfully leverage the added information provided by additional terms and syntactic structures. While these experiments demonstrate that the added structure provided by the categorical passage from syntax to semantics yields improvements in the quality of semantic representations obtained through composition, they also reveal that there is still significant ground to cover in order to approach levels of performance comparable to that of human annotators.

The third experiment also served to show that using different combinations of composition operations depending on the syntactic type of the terms being combined can yield better results, and that some models combine better than others. Notably, the adjective-noun combination models \textbf{AdjMult} and \textbf{CategoricalAdj}, despite their mathematical similarity, produce noticeably different results when combined with the categorical verb-argument composition operation, while they perform equally with most other verb-argument composition operations. Likewise, the combination of additive and multiplicative models in \textbf{AddMult} performs surprisingly well, despite its simplicity (although the choice of operations was chosen by hand, and there is no canonical way of extending this to every sentence structure). We can conclude that different models combine different semantic aspects more prominently than others, and that through combination we can obtain better representations by assuming that different kinds of composition play on different semantic properties. For example, predicates such as intersective adjectives often add information to their argument (a red ball is a ball  that is also red), hence it may be no surprise that using addition in \textbf{AddMult} performs better than across-the-board component-wise multiplication. This opens the question of how to design models of composition that systematically select which operations will match the semantic aspects of the words being combined based on their syntactic type. This is an open question, which I believe warrants further investigation.

% section discussion (end)

% chapter evaluating_a_discocat (end)
%!TEX root = ../grefenstettethesis.tex

\chapter{Further Work} % (fold)
\label{cha:applications_and_future_work}

\begin{chabstract}
  This chapter presents three areas in which further work has been undertaken, based on the foundations established by the rest of this thesis. It describes the issues surrounding the integration of logic into compositional distributional models, presents a new machine learning algorithm for learning semantic representations, and outlines how Combinatory Categorial Grammars may be integrated into the DisCoCat framework. It concludes by suggesting directions future work might take in this field.
\end{chabstract}

Before concluding this thesis, let us examine some topics which have constituted further research on this topic, both from my own work and from work done in collaboration with colleagues. In this chapter, I will examine three areas in which further developments of the DisCoCat framework and models stemming from it have occurred, before discussing how these three areas exemplify and provide a foundation for future work. In $\S$\ref{sec:distributional_logic}, I will discuss the issues surrounding the modelling of logical operators within the DisCoCat framework, and what difficulties our reliance on multilinear maps modelled as tensors gives rise to. In $\S$\ref{sec:learning_tensors_by_multi_step_regression}, I will report new work done on the topic of integrating sophisticated machine learning methods into the learning procedures for our semantic representations. In $\S$\ref{sec:supporting_combinatory_categorial_grammar}, I will present the foundations for further syntactic extensions allowing us to integrate a more expressive grammatical formalism, Combinatory Categorial Grammar, into the DisCoCat framework. Finally, in $\S$\ref{sec:further_work}, I conclude by showing how these three areas of research set the tone for future work to be done on the topic of categorical compositional distributional semantics.

\section{Distributional Logic} % (fold)
\label{sec:distributional_logic}

The DisCoCat formalism of \cite{Coecke2010} presented in Chapter~\ref{cha:foundations_of_discocat} and the models derived from it presented in Chapter~\ref{cha:learning_procedures_for_a_discocat} allow us to represent and learn syntactically-motivated semantic objects which can be composed to form distributional sentence representations. The learning algorithms provided allow us to model a large class of word-types, technically including logical words present in every day language use, such as ``and'', ``or'', and ``not''. However, this closed lexical class has a very specific use in language, which historically has been modelled as operations in propositional or predicate logical calculus. As such, it may seem counter-intuitive to learn these from data; in fact, it would be more reasonable to take the finite class of words used to express quantification, conjunction, implication, and other logical concepts, as semantic objects which are \emph{designed}. 

A good question following this observation might therefore be: how can logic be modelled in a compositional distributional semantic model? Furthermore, how would logical operators work when applied to non-truth theoretic representations such as those developed in Chapter~\ref{cha:learning_procedures_for_a_discocat}? This second question is an open problem I leave for future work, but to even begin to answer it, it would be helpful to see how classical predicate logic can be simulated within the DisCoCat framework (or something similar). The hope is that we could use such a simulation as a basis or inspiration for the development of non-truth theoretic distributional logic operations in future research. In this section, I discuss how such a logic might be simulated, based on work presented in \cite{GrefenstetteTFDS}, as well as the issues faced by our models in doing so.

\subsection{Distributional Formal Semantics} % (fold)
\label{sub:distributional_formal_semantics}

A popular approach to compositionality in formal semantics is to
derive a formal representation of a phrase from its grammatical
structure and a set of associated rules. I represent the semantics of words as functions and
arguments, and use the grammatical structure to dictate the order and
scope of function application. For example, formal semantic models in
the style of~\cite{Montague1974} will associate a semantic rule
with each syntactic rule in a context-free grammar. A simple formal
semantic model is shown in Figure~\ref{fig:simpleformalsemantics}.

\begin{figure}[htb]
\begin{displaymath}
{
\footnotesize
\begin{tabular}{l|l}
\textbf{Syntax} & \textbf{Semantics}\\
\hline
S $\Rightarrow$ NP VP & $\semantics{S} \Rightarrow \semantics{VP} \left(\semantics{NP}\right)$\\
NP $\Rightarrow$ N & $\semantics{NP} \Rightarrow \semantics{N}$\\
N $\Rightarrow$ ADJ N & $\semantics{N} \Rightarrow \semantics{ADJ}\left(\semantics{N}\right)$\\
VP $\Rightarrow$ Vt NP & $\semantics{VP} \Rightarrow \semantics{Vt}\left(\semantics{NP}\right)$\\
VP $\Rightarrow$ Vi & $\semantics{VP} \Rightarrow \semantics{Vi}$\\
\end{tabular}
\qquad
\begin{tabular}{l|l}
\textbf{Syntax (cont'd)} & \textbf{Semantics (cont'd)}\\
\hline
Vt $\Rightarrow$ $\{\textit{verbs$_t$}\}$ & $\semantics{Vt} \Rightarrow \semantics{\textit{verb$_t$}} $\\
Vi $\Rightarrow$ $\{\textit{verbs$_i$}\}$ & $\semantics{Vi} \Rightarrow \semantics{\textit{verb$_i$}} $\\
ADJ $\Rightarrow$ $\{\textit{adjs}\}$ & $\semantics{ADJ} \Rightarrow \semantics{\textit{adj}}$\\
N $\Rightarrow$ $\{\textit{nouns}\}$ & $\semantics{N} \Rightarrow \semantics{\textit{noun}}$
\end{tabular}
}
\end{displaymath}
\caption{A simple formal semantic model.}
\label{fig:simpleformalsemantics}
	
\end{figure}

Following these rules shown, the parse of a simple sentence like `angry dogs
chase furry cats' yields the following interpretation:
$\semantics{\textit{chase}}\left( \semantics{\textit{furry}}
  \left(\semantics{\textit{cats}}\right),
  \semantics{\textit{angry}}\left(\semantics{\textit{dogs}}\right)
\right)$. This model is very simplistic, and typically, a higher order logic such as a lambda
calculus will be used to provide additional structure, but the key aspect retained by this simple model is that the grammar
dictates the translation from natural language to the functional
form. Generally, in formal semantic models, such functions will be
logical relations and predicates applied to arguments, namely objects from the logical domain. In this section, I consider a way of defining such functions as multilinear maps over geometric objects
instead. This framework is generally applicable to other formal
semantic models than that which is presented here.

% subsection distributional_formal_semantics (end)

\subsection{Tensors as Functions} % (fold)
\label{sub:tensors_as_functions}

The bijective correspondence between linear maps and matrices is a well known property in linear algebra: every linear map $f: A \to B$ can be encoded as a $dim(B)$ by $dim(A)$ matrix $M$, and conversely every such matrix encodes a class of linear maps determined by the dimensionality of the domain and co-domain. The application of a linear map $f$ to a vector $\overrightarrow{v} \in A$ producing a vector $\overrightarrow{w} \in B$ is equivalent to the matrix multiplication:
\[
f(\overrightarrow{v}) = M \times \overrightarrow{v} = \overrightarrow{w}
\]
This correspondence generalises for multilinear maps to a correlation
between $n$-ary maps and order $n+1$ tensors \cite{Bourbaki:1989,Lee:1997}.

Tensors are best described as a generalisation of the notion of vectors and matrices to \emph{larger degrees of freedom} referred to as tensor orders (one for vectors, two for matrices). To illustrate this generalisation, consider how vectors, which are order 1 tensors, may be written as the weighted superposition (summation) of their basis elements: for some vector space $V$ with basis $\{\overrightarrow{b_i}\}_i$, any vector $\overrightarrow{v} \in V$ can be written
\[
\overrightarrow{v} = \sum_i{c^{{v}}_i \overrightarrow{b_i}} = \left[c^{{v}}_1,\, \ldots,\, c^{{v}}_i,\, \ldots,\, c^{{v}}_{dim(V)}\right]
\]
where the weights $c^{{v}}_i$ are elements of the underlying
field (e.g.~$\mathbb{R}$), and thus vectors can be fully described by
such a one-index summation. Likewise, matrices, which are order 2
tensors, can be seen as a collection of row vectors from some
space $V_r$ with basis $\{\overrightarrow{a_i}\}_i$, or of column vectors from some space
$V_c$ with basis $\{\overrightarrow{d_j}\}_j$. Such a matrix $M$ is an
element of the space $V_r \otimes V_c$, and can be fully described by
the two index summation:
\[
M = \sum_{ij}{c^{M}_{ij} \overrightarrow{a_i} \otimes \overrightarrow{d_j}}
\]
where, once again, $c^{M}_{ij}$ is an element of the underlying field
which in this case is simply the element from the $i$th row and
$j$th column of the matrix $M$, and the basis element $\overrightarrow{a_i}
\otimes \overrightarrow{d_j}$ of $V_r \otimes V_c$ is formed by a pair of
basis elements from $V_r$ and $V_c$. The number of indices (or degrees
of freedom) used to fully describe a tensor in this superposition
notation is its order, e.g., an order 3 tensor $T \in A \otimes B \otimes
C$ would be described by the superposition of weights $c^T_{ijk}$
associated with basis elements $\overrightarrow{e_i} \otimes \overrightarrow{f_j}
\otimes \overrightarrow{g_k}$.

Matrix multiplications, inner products, and traces all generalise to tensors as the non-commutative tensor contraction operation
($\times$). For tensors $T \in A \otimes \ldots \otimes B \otimes C$
and $U \in C \otimes D \otimes \ldots \otimes E$, with bases
$\{\overrightarrow{a}_i \otimes \ldots \otimes \overrightarrow{b_j} \otimes
\overrightarrow{c_k}\}_{i \ldots jk}$ and $\{\overrightarrow{c_k} \otimes \overrightarrow{d_l}
\otimes \ldots \otimes \overrightarrow{e_m}\}_{kl \ldots m}$, the tensor
contraction of $T \times U$ is calculated:
\[
\sum_{i \ldots jkl \ldots m}{c^T_{i \ldots jk} c^U_{kl \ldots m} \overrightarrow{a_i} \otimes \ldots \otimes \overrightarrow{b_j} \otimes \overrightarrow{d_l} \otimes \ldots \otimes \overrightarrow{e_m}}
\]
where the resulting tensor is of order equal to two less than the sum of the orders of the input tensors;  the subtraction reflects the elimination of matching basis elements through summation during contraction.

For every curried multilinear map $g: A \to \ldots \to Y \to Z$, there is a tensor $T^g \in Z \otimes Y \otimes \ldots \otimes A$ encoding it \cite{Bourbaki:1989,Lee:1997}. The application of a curried $n$-ary map $h : V_1 \to \ldots \to V_n \to W$ to input vectors $\overrightarrow{v_1} \in V_1,\,\ldots,\, \overrightarrow{v_n} \in V_n$ to produce output vector $\overrightarrow{w} \in W$ corresponds to the tensor contraction of the tensor $T^h \in W \otimes V_n \otimes \ldots \otimes V_1$ with the argument vectors:
\[
h(\overrightarrow{v_1})\ldots(\overrightarrow{v_n}) = T^h \times \overrightarrow{v_1} \times \ldots \times \overrightarrow{v_n}
\]
This isomorphism between tensors (objects) and multilinear maps (maps) is also known in the quantum information theory literature (e.g.~\cite{zyczkowski2004duality,choi1975completely}) as \emph{map-state duality}, whereby for each map there is a bi-partite entangled state, with the map and state each forming part of the dual representation of the same piece of information.

% subsection tensors_as_functions (end)

\subsection{Formal Semantics with Tensors} % (fold)
\label{sub:formal_semantics_with_tensors}

Using the correspondence between $n$-ary maps and tensors of order
$n+1$ discussed in $\S$\ref{sub:tensors_as_functions}, and the correspondence between function arguments and semantic
vectors in standard distributional semantic models, we can turn any
formal semantic model into a compositional distributional semantic
model. This is done by first running a type inference algorithm on the semantic interpretations of generative rules and obtaining both basic (i.e.~uninferred) and function (i.e.~inferred) types, then assigning to each basic type a vector space and to each function type a tensor space. Following this, we can represent arguments by vectors and functions by tensors, and finally, we can model function application by tensor contraction.

To give an example, in the simple formal semantic model presented in Figure~\ref{fig:simpleformalsemantics}, a
type inference algorithm in the vein of \cite{hindley1969principal,milner1978theory} would give the following type assignments: 
\begin{itemize}
    \item It would leave the types of $\semantics{N}$ and
$\semantics{S}$ uninferred, as they are not treated as functions in
any rule; therefore, we treat nouns and sentences as vectors in
some spaces $N$ and $S$.
\item Noun phrases are simply assigned the same type as
$\semantics{N}$, hence elements of $\semantics{NP}$ would also be
vectors in $N$.
\item Verb phrases map noun phrase interpretations to
sentence interpretations, hence $\semantics{VP} : type(\semantics{NP})
\to type(\semantics{S})$ or, as we have inferred the relevant spaces,
$\semantics{VP} : N \to S$.
\item Intransitive verb interpretations are assigned the same types as verb phrase interpretations, and hence an intransitive verb ``vi'' can be represented by
some tensor $T^{\textrm{vi}} \in S \otimes N$.
\item Transitive verbs are interpreted as maps $\semantics{Vt} : type(\semantics{NP}) \to
type(\semantics{VP})$, which can be expanded to $\semantics{Vt} : N \to N
\to S$, giving us, for some transitive verb ``vt'', the tensor
form $T^{\textrm{vt}} \in S \otimes N \otimes N$.
\item Finally, adjectives are maps of form $\semantics{Adj}: type(\semantics{N}) \to type(\semantics{N})$, and hence some
adjective ``adj'' has the tensor form $T^{\textrm{adj}} \in N
\otimes N$.
\end{itemize}

Putting all this together with tensor contraction
($\times$) as function application, let us examine the computation of a sample
sentence ``angry dogs chase furry cats'' with interpretation
$\semantics{\textit{chase}}\left( \semantics{\textit{furry}}
  \left(\semantics{\textit{cats}}\right),
  \semantics{\textit{angry}}\left(\semantics{\textit{dogs}}\right)
\right)$. This would simply be the calculation of $\left(T^{\textrm{chase}}
  \times \left(T^{\textrm{furry}} \times T^{\textrm{cats}} \right)
\right) \times \left( T^{\textrm{angry}} \times T^{\textrm{dogs}}
\right)$, where $T^{\textrm{cats}}$ and $T^{\textrm{dogs}}$ would be
lexical semantic vectors, $T^{\textrm{furry}}$ and
$T^{\textrm{angry}}$ would be square matrices, and
$T^{\textrm{chase}}$ would be an order 3 tensor.

It should be noted that the semantic tensors generated in this framework are in fact identical to those found in the DisCoCat framework, as tensor contraction is just iterated application of $\epsilon$ maps. To give an example, let $T^f$ be tensor in $C \otimes B \otimes A$ modelling a multilinear map $f: A \times B \to C$. Let $\overrightarrow{v} \in A$ and $\overrightarrow{w} \in B$ be its arguments. Then the tensor contraction corresponding to the application of $T^f$ to its arguments can be expressed as $\epsilon$ maps as follows:
\[
\left(T^f \times \overrightarrow{v}\right) \times \overrightarrow{w}
=
((1_C \otimes \epsilon_B) \circ (1_C \otimes \epsilon_A \otimes 1_B))(T^f \otimes \overrightarrow{v} \otimes \overrightarrow{w})
\]
The principal (superficial) difference between this framework and DisCoCat is that tensors always `absorb' arguments on the right through contraction, forming a linear $\lambda$-calculus, whereas in DisCoCat, contraction can happen on the left of a tensor. This difference is indeed superficial, in that any tensor can be transposed to `take arguments' on the left or the right, thereby allowing us to produce an isomorphism between the tensors in a DisCoCat model and those in this distributional formal semantics framework, provided that the vector spaces match.
% subsection formal_semantics_with_tensors (end)

\subsection{Simulating Simple Predicate Calculi} % (fold)
\label{sub:simulating_simple_predicate_calculi}

The approach described above allows us to adapt very simple formal semantic models to work with tensors in a way similar to the DisCoCat framework discussed throughout this thesis. Before turning to the question of how logical connectives and quantifiers might be modelled using tensors in either the distributional formal semantics described above, or in the DisCoCat framework, I begin by showing how tensors can represent model-theoretic predicates from a finite-domain predicate calculus.

In \cite{Coecke2010}, a short example shows how simple predicate logic can be simulated within the DisCoCat framework by setting $S$ to a boolean space. This can be done in one of two ways: namely either by setting $S$ to be a one dimensional space $B_1$ with basis vector $\overrightarrow{\top}$, where the vector $\overrightarrow{\top} \equiv \top$ and $\overrightarrow{0} \equiv \bot$; or by setting $S$ to be a two dimensional space $B_2$ with basis $\{\overrightarrow{\top},\overrightarrow{\bot}\}$, where:
\[
\overrightarrow{\top}
=
\left[
\begin{tabular}{c}
1\\
0
\end{tabular}
\right]
\equiv \top
\qquad
\overrightarrow{\bot}
=
\left[
\begin{tabular}{c}
0\\
1
\end{tabular}
\right]
\equiv \bot
\]
To represent a logical model in vector spaces using either of the above options for boolean space, I consider the following translation mechanism:
\begin{itemize}
    \item I assign to the domain $\mathcal{D}$---the set of objects in our logic---a vector space $N$ over $\mathbb{N}$ of $|\mathcal{D}|$ dimensions where each basis vector of $N$ is in one-to-one correspondence with elements of $\mathcal{D}$.
    \item An element of $\mathcal{D}$ is therefore represented as a one-hot vector in $N$ (a sparse unit-length vector with a single non-zero value), the single-non null value of which is the weight for the basis vector mapped to that element of $\mathcal{D}$. 
    \item Similarly, a subset of $\mathcal{D}$ is a vector of $N$ where those elements of $\mathcal{D}$ in the subset have $1$ as their corresponding basis weights in the vector, and those not in the subset have $0$. Therefore there is a one-to-one correspondence between the vectors in $N$ with basis weights of either 0 or 1 and the elements of the power set $\mathcal{P}(\mathcal{D})$.
    \item Each unary predicate $P$ in the logic is represented in the logical model as a set $M_P \subseteq \mathcal{D}$ containing the elements of the domain for which the predicate is true. Traditionally, we could view the predicate as a function $f_P : \mathcal{D} \to \mathbb{B}$ where:
    \[
    f_P(x) =
    \left\{
    \begin{tabular}{l}
     $\top\quad$ if $x \in M_P$\\
     $\bot\quad$ otherwise
    \end{tabular}
    \right.
    \]
    Given that our sentence space is set to a boolean space, using one of the forms suggested above (i.e.~$B_1$ or $B_2$), we could directly model this function as a tensor in $S \otimes N$ as discussed in $\S$\ref{sub:formal_semantics_with_tensors}. However, I suggest a different way of distributionally modelling predicates which is more in line with how predicates work in a DisCoCat. Let us instead view predicates $P$ as functions $f_P: \mathcal{P}(\mathcal{D}) \to \mathcal{P}(\mathcal{D})$, defined as:
    \[
    f_P(X) = X \cap M_P
    \]
    Therefore the distributional form of these functions will be tensors in $N \otimes N$. Through tensor contractions, they map subsets of $\mathcal{D}$ (elements of $N$) to subsets of $\mathcal{D}$ containing only those objects of the original subset for which $P$ holds (i.e.~yielding another vector in $N$).
    \item Finally, $n$-ary relations $R$ such as verbs are represented in a logical model as the set $M_R$ of $n$-tuples of elements from $\mathcal{D}$ for which $R$ holds. Therefore such relations can be represented as functions $f_R: \mathcal{D}^n \to \mathbb{B}$ where:
    \[
    f_R(x_1,\ldots,x_n) =
    \left\{
    \begin{tabular}{l}
    $\top\quad$ if $(x_1,\ldots,x_n) \in M_R$\\
    $\bot\quad$ otherwise
    \end{tabular}
    \right.
    \]
    Considering our choice of $S$ as a boolean space, and using the process described in $\S$\ref{sub:formal_semantics_with_tensors}, we can represent these relations distributionally as tensors in $S \otimes \underbrace{N \otimes \ldots \otimes N}_n$.
\end{itemize}

To give a concrete example about how predicates and relations (namely verbs) would work in our setting, let our domain be the individuals John ($j$), Mary ($m$) and Peter ($p$). John and Mary love each other and love themselves, but are indifferent about Peter. Peter hates himself and John, but loves Mary. The logical model for this budding romantic tragedy is as follows:
\begin{align*}
  & \mathcal{D} = \{j,m,p\}\\
  & M_\text{loves} = \{ (j,j),\, (m,m),\, (j,m),\, (m,j),\, (p,m) \}\\
  & M_\text{hates} = \{ (p,p),\, (p,j) \}
\end{align*}
Distributionally speaking, the elements of the domain will be mapped to the following one-hot vectors in some three-dimensional space $N$ as follows:
\[
j \mapsto \overrightarrow{j} =
\left[
\begin{tabular}{c}
1\\
0\\
0
\end{tabular}
\right]
\qquad
m \mapsto \overrightarrow{m} =
\left[
\begin{tabular}{c}
0\\
1\\
0
\end{tabular}
\right]
\qquad
p \mapsto \overrightarrow{p} =
\left[
\begin{tabular}{c}
0\\
0\\
1
\end{tabular}
\right]
\]
The representation of the verbs will depend on our choice of sentence space $S$. If we set $S = B_1$, we define the verbs as follows, using the Kronecker product:
\begin{align*} 
    & f_\text{loves} \mapsto (\overrightarrow{\top} \otimes \overrightarrow{j} \otimes \overrightarrow{j}) + (\overrightarrow{\top} \otimes \overrightarrow{m} \otimes \overrightarrow{m}) + (\overrightarrow{\top} \otimes \overrightarrow{j} \otimes \overrightarrow{m}) + (\overrightarrow{\top} \otimes \overrightarrow{m} \otimes \overrightarrow{j}) + (\overrightarrow{\top} \otimes \overrightarrow{p} \otimes \overrightarrow{m})\\
    & f_\text{hates} \mapsto (\overrightarrow{\top} \otimes \overrightarrow{p} \otimes \overrightarrow{p}) + (\overrightarrow{\top} \otimes \overrightarrow{p} \otimes \overrightarrow{j})
\end{align*} 
Using the distributivity of $\otimes$ over $+$ we can express this more compactly as:
\begin{align*} 
    & f_\text{loves} \mapsto T^{\text{loves}} = \overrightarrow{\top} \otimes ((\overrightarrow{j} \otimes \overrightarrow{j}) + (\overrightarrow{m} \otimes \overrightarrow{m}) + (\overrightarrow{j} \otimes \overrightarrow{m}) + (\overrightarrow{m} \otimes \overrightarrow{j}) + (\overrightarrow{p} \otimes \overrightarrow{m}))\\
    & f_\text{hates} \mapsto T^{\text{hates}} = \overrightarrow{\top} \otimes ((\overrightarrow{p} \otimes \overrightarrow{p}) + ( \overrightarrow{p} \otimes \overrightarrow{j}))
\end{align*} 
These definitions implicitly set the basis weight for sets of elements not in $M_\text{loves}$ and $M_\text{hates}$ to zero.

If we set $S = B_2$ we simply have to explicitly state all the element pairs for which the relation is false, in addition to those pairs for which it is true. Using the more compact notation used above, this gives:
\begin{align*} 
    & f_\text{loves} \mapsto T^{\text{loves}} = \overrightarrow{\top} \otimes ((\overrightarrow{j} \otimes \overrightarrow{j}) + (\overrightarrow{m} \otimes \overrightarrow{m}) + (\overrightarrow{j} \otimes \overrightarrow{m}) + (\overrightarrow{m} \otimes \overrightarrow{j}) + (\overrightarrow{p} \otimes \overrightarrow{m}))
    \\ & \qquad + \overrightarrow{\bot} \otimes ( (\overrightarrow{p} \otimes \overrightarrow{p}) + (\overrightarrow{m} \otimes \overrightarrow{p}) + (\overrightarrow{j} \otimes \overrightarrow{p}) + (\overrightarrow{p} \otimes \overrightarrow{j}))
    \\
    & f_\text{hates} \mapsto T^{\text{hates}} = \overrightarrow{\top} \otimes ((\overrightarrow{p} \otimes \overrightarrow{p}) + (\otimes \overrightarrow{p} \otimes \overrightarrow{j}))\\
    & \qquad + \overrightarrow{\bot} \otimes ( (\overrightarrow{j} \otimes \overrightarrow{j}) + (\overrightarrow{m} \otimes \overrightarrow{m}) + (\overrightarrow{j} \otimes \overrightarrow{m}) + (\overrightarrow{m} \otimes \overrightarrow{j}) + (\overrightarrow{j} \otimes \overrightarrow{p}) + (\overrightarrow{m} \otimes \overrightarrow{p}) + (\overrightarrow{p} \otimes \overrightarrow{m}))
\end{align*} 

Computing the value of a sentence such as ``John loves Mary'' would then involve producing the distributional representation of $f_\textrm{loves}(j,m)$. In the distributional setting, this would be equivalent to the following computation with $S=B_1$, and with $\times$ as the tensor contraction operation:
\begin{align*}
  & (T^{\text{loves}} \times \overrightarrow{m}) \times \overrightarrow{j} \\
  & \quad =  ((\overrightarrow{\top} \otimes ((\overrightarrow{m} \otimes \overrightarrow{m}) +  (\overrightarrow{j} \otimes \overrightarrow{m}) + (\overrightarrow{p} \otimes \overrightarrow{m}))) \times \overrightarrow{m}) \times \overrightarrow{j}\\
  & \quad = (\overrightarrow{\top} \otimes (\overrightarrow{m} +  \overrightarrow{j} + \overrightarrow{p} )) \times \overrightarrow{j}\\
  & \quad = (\overrightarrow{\top} \otimes \overrightarrow{j}) \times \overrightarrow{j}\\
  & \quad = \overrightarrow{\top}
\end{align*}
In contrast, ``Peter hates Mary'' would yield the computation:
\begin{align*}
  & (T^{\text{hates}} \times \overrightarrow{m}) \times \overrightarrow{p} \\
  & \quad = ((\overrightarrow{\top} \otimes ((\overrightarrow{p} \otimes \overrightarrow{p}) + ( \overrightarrow{p} \otimes \overrightarrow{j}))) \times \overrightarrow{m}) \times \overrightarrow{p}\\
  & \quad = \overrightarrow{\top}\cdot 0 \times \overrightarrow{p}\\
  & \quad = \overrightarrow{\top}\cdot 0\\
  & \quad = \overrightarrow{0}\\
\end{align*}
which is the origin in $S = B_1$, which I interpret as false. In the case of false statements, the nature of the computation is a lot less clear mathematically. In contrast, it is explicit in the case of $S = B_2$, for which we reproduce both computations initially shown above here. For ``John loves Mary'' we can notice very little change, since everybody loves Mary:
\begin{align*}
  & (T^{\text{loves}} \times \overrightarrow{m}) \times \overrightarrow{j} \\
  & \quad =  ((\overrightarrow{\top} \otimes ((\overrightarrow{m} \otimes \overrightarrow{m}) +  (\overrightarrow{j} \otimes \overrightarrow{m}) + (\overrightarrow{p} \otimes \overrightarrow{m})) + \overrightarrow{\bot} \cdot 0 \times \overrightarrow{m}) \times \overrightarrow{j}\\
  & \quad = (\overrightarrow{\top} \otimes (\overrightarrow{m} +  \overrightarrow{j} + \overrightarrow{p} ) + \overrightarrow{\bot} \cdot 0) \times \overrightarrow{j}\\
  & \quad = (\overrightarrow{\top} \otimes \overrightarrow{j} + \overrightarrow{\bot} \cdot 0) \times \overrightarrow{j}\\
  & \quad = \overrightarrow{\top}\cdot 1 + \overrightarrow{\bot} \cdot 0\\
  & \quad =
  \left[
  \begin{tabular}{c}
  1\\
  0
  \end{tabular}
  \right]\\
  & \quad = \overrightarrow{\top}
\end{align*}
But for ``Peter hates Mary'' the falsehood of the statement is more clearly computed:
\begin{align*}
  & (T^{\text{hates}} \times \overrightarrow{m}) \times \overrightarrow{p} \\
  & \quad = ((\overrightarrow{\top} \otimes ((\overrightarrow{p} \otimes \overrightarrow{p}) + ( \overrightarrow{p} \otimes \overrightarrow{j})) + \overrightarrow{\bot} \otimes ( (\overrightarrow{m} \otimes \overrightarrow{m}) + (\overrightarrow{j} \otimes \overrightarrow{m}) + (\overrightarrow{p} \otimes \overrightarrow{m}) )) \times \overrightarrow{m}) \times \overrightarrow{p}\\
  & \quad = (\overrightarrow{\top}\cdot 0 + \overrightarrow{\bot} \otimes ( \overrightarrow{m} + \overrightarrow{j} + \overrightarrow{p} )) \times \overrightarrow{p}\\
  & \quad = \overrightarrow{\top}\cdot 0 + \overrightarrow{\bot} \cdot 1\\
  & \quad =
  \left[
  \begin{tabular}{c}
  0\\
  1
  \end{tabular}
  \right]\\
  & \quad = \overrightarrow{\bot}
\end{align*}

Let us quickly illustrate how predicates work distributionally: let us consider a domain with two dogs ($a$ and $b$) and a cat ($c$). One of the dogs is brown, as is the cat. Let $D$ be the set of dogs, and $B$ the predicate ``brown''. I represent these statements in the model as follows:
\begin{align*}
  & \mathcal{D} = \{a,b,c\}\\
  & D = \{a,b\}\\
  & M_B = \{b,c\}
\end{align*}
Distributionally, I model the domain as a three dimensional vector space, the set of dogs as a vector
\[
D \mapsto
\left[
\begin{tabular}{c}
1\\
1\\
0
\end{tabular}
\right]
\]
and the predicate `brown' as a tensor in $N \otimes N$
\[
f_B \mapsto T^{\text{brown}} =
\left[
\begin{tabular}{ccc}
  0 & 0 & 0\\
  0 & 1 & 0\\
  0 & 0 & 1
\end{tabular}
\right]
\]
The set of brown dogs is obtained by computing $f_B(D)$, which distributionally corresponds to applying the tensor $T^{\text{brown}}$ to the vector representation of $D$ via tensor contraction, as follows:
\[
T^{\text{brown}} \times D =
\left[
\begin{tabular}{ccc}
  0 & 0 & 0\\
  0 & 1 & 0\\
  0 & 0 & 1
\end{tabular}
\right]
\left[
\begin{tabular}{c}
1\\
1\\
0
\end{tabular}
\right]
=
\left[
\begin{tabular}{c}
0\\
1\\
0
\end{tabular}
\right]
=
\overrightarrow{b}
\]

Therefore using tensors and vectors, we can effectively simulate a simple predicate calculus within either the DisCoCat framework or the formal distributional semantics presented earlier in this chapter.

% subsection simulating_simple_predicate_calculi (end)

\subsection{Logical Operations, and Integrating Non-Linearity} % (fold)
\label{sub:integrating_non_linearity}

In $\S$\ref{sub:formal_semantics_with_tensors}, I presented a way of combining tensor-based compositional distributional models with simple formal semantic models. In $\S$\ref{sub:simulating_simple_predicate_calculi}, I showed how a simple predicate logic could be simulated within such a distributional formal semantics or the DisCoCat framework discussed throughout this thesis. However, both in the case of formal semantics and in the natural language statements we wish to model the semantics of with DisCoCat models, there are not only predicates and relations, but also logical words such `and', `or' and expressions such as `every', `some', `for all', which we model logically as logical connectives ($\land$, $\lor$, $\lnot$, and so on) and as quantifiers ($\forall$, $\exists$). How are these to be represented within tensor-based models? In this subsection, I show that the linear maps encoded by tensors do not \emph{prima-facie} suffice to simulate all these logical elements, but that the distributional formal semantics described above can be enhanced with non-linear operations that fit the bill.

\paragraph{Logical Operations}
Consider the following additional rules for the formal semantics presented in Figure~\ref{fig:simpleformalsemantics}:
\begin{center}
  \begin{tabular}{l|l}
\textbf{Syntax} & \textbf{Semantics}\\
\hline
S $\Rightarrow$ S$_1$ and S$_2$ & $\semantics{S} \Rightarrow \land\left(\semantics{S_1}, \semantics{S_2}\right)$\\
S $\Rightarrow$ S$_1$ or S$_2$ & $\semantics{S} \Rightarrow \lor\left(\semantics{S_1}, \semantics{S_2}\right)$\\
S $\Rightarrow$ not S & $\semantics{S} \Rightarrow \lnot\left(\semantics{S}\right)$\\
\end{tabular}
\end{center}
Can these additional semantic functions, corresponding to conjunction, disjunction and negation be modelled by linear maps (and thereby represented as tensors)?

I considered two options for a boolean sentence space, $B_1$ and $B_2$, as suggested by \cite{Coecke2010}. For $B_1$ it is trivial to show that negation cannot be modelled by a linear map. Assume there is some linear map $not: B_1 \to B_1$ such that:
\begin{align*}
  & not(\overrightarrow{\top}) = \overrightarrow{0}\\
  & not(\overrightarrow{0}) = \overrightarrow{\top}
\end{align*}
All linear maps $f$ must satisfy
\[
f(\overrightarrow{0}) = \overrightarrow{0}
\]
so $not$ is not a linear map, and hence no tensor models negation for $S = B_1$. Similarly, the obvious representations for conjunction and disjunction, namely
\begin{align*}
  & and(\overrightarrow{v},\overrightarrow{w}) = min(\overrightarrow{v},\overrightarrow{w})
  & or(\overrightarrow{v},\overrightarrow{w}) = max(\overrightarrow{v},\overrightarrow{w})
\end{align*}
where $min$ and $max$ are component-wise minimum and maximum operators, are not multilinear maps.

For $S= B_2$, \cite{Coecke2010} show that the swap matrix
\[
T^{\text{not}} =
\left[
\begin{tabular}{cc}
0 & 1\\
1 & 0
\end{tabular}
\right]
\]
models negation, which can easily be verified:
\begin{align*}
  & T^{\text{not}} \times \overrightarrow{\top} =
\left[
\begin{tabular}{cc}
0 & 1\\
1 & 0
\end{tabular}
\right]
\left[
\begin{tabular}{c}
1\\
0
\end{tabular}
\right]
= 
\left[
\begin{tabular}{c}
0\\
1
\end{tabular}
\right]
= \overrightarrow{\bot}\\
& T^{\text{not}} \times \overrightarrow{\bot} =
\left[
\begin{tabular}{cc}
0 & 1\\
1 & 0
\end{tabular}
\right]
\left[
\begin{tabular}{c}
0\\
1
\end{tabular}
\right]
= 
\left[
\begin{tabular}{c}
1\\
0
\end{tabular}
\right]
= \overrightarrow{\top}
\end{align*}

I go further here and show that various other logical operators can be modelled with $S=B_2$. To make talking about order-3 tensors used to model binary operations easier, I will use the following block matrix notation for $2 \times 2 \times 2$ order-3 tensors $T$:
\[
T =
\left[
\begin{tabular}{cc|cc}
$a_1$ & $b_1$ & $a_2$  & $b_2$\\
$c_1$ & $d_1$ & $c_2$  & $d_2$ 
\end{tabular}
\right]
\]
which allows us to express tensor contractions as follows, for some $\overrightarrow{v}=[\alpha\ \beta]^{\top}$:
\[
T \times \overrightarrow{v} =
\left[
\begin{tabular}{cc|cc}
$a_1$ & $b_1$ & $a_2$  & $b_2$\\
$c_1$ & $d_1$ & $c_2$  & $d_2$ 
\end{tabular}
\right]
\left[
\begin{tabular}{c}
$\alpha$\\
$\beta$
\end{tabular}
\right]
=
\left[
\begin{tabular}{cc}
$\alpha \cdot a_1 + \beta \cdot a_2$ & $\alpha \cdot b_1 + \beta \cdot b_2$\\
$\alpha \cdot c_1 + \beta \cdot c_2$ & $\alpha \cdot d_1 + \beta \cdot d_2$
\end{tabular}
\right]
\]
or more concretely:
\begin{align*}
& T \times \overrightarrow{\top} =
\left[
\begin{tabular}{cc|cc}
$a_1$ & $b_1$ & $a_2$  & $b_2$\\
$c_1$ & $d_1$ & $c_2$  & $d_2$ 
\end{tabular}
\right]
\left[
\begin{tabular}{c}
$1$\\
$0$
\end{tabular}
\right]
=
\left[
\begin{tabular}{cc}
$a_1$ & $b_1$\\
$c_1$ & $d_1$
\end{tabular}
\right]\\
& T \times \overrightarrow{\bot} =
\left[
\begin{tabular}{cc|cc}
$a_1$ & $b_1$ & $a_2$  & $b_2$\\
$c_1$ & $d_1$ & $c_2$  & $d_2$ 
\end{tabular}
\right]
\left[
\begin{tabular}{c}
$0$\\
$1$
\end{tabular}
\right]
=
\left[
\begin{tabular}{cc}
$a_2$ & $b_2$\\
$c_2$ & $d_2$
\end{tabular}
\right]
\end{align*}

Using this notation, we can define tensors for the following operations:
\begin{align*}
& (\lnot) \mapsto T^{\lnot} =
\left[
\begin{tabular}{cc}
$0$ & $1$\\
$1$ & $0$
\end{tabular}
\right]\\
& (\lor) \mapsto T^{\lor} =
\left[
\begin{tabular}{cc|cc}
1 & 1 & 1 & 0\\
0 & 0 & 0 & 1
\end{tabular}
\right]\\
& (\land) \mapsto T^{\land} =
\left[
\begin{tabular}{cc|cc}
1 & 0 & 0 & 0\\
0 & 1 & 1 & 1
\end{tabular}
\right]\\
& (\to) \mapsto T^{\to} =
\left[
\begin{tabular}{cc|cc}
1 & 0 & 1 & 1\\
0 & 1 & 0 & 0
\end{tabular}
\right]
\end{align*}
The design behind these tensors is not particularly complicated. They are applied to two arguments: if the first argument is true ($\overrightarrow{\top}$) then the left matrix of the block matrix is applied to the second argument; if the first argument is false ($\overrightarrow{\bot}$) then the right matrix of the block matrix is applied to the second argument. It is fairly trivial, using the truth tables for any logical connective, to design such partitioned matrices for the tensor representation of any logical operator in a propositional calculus. For example, for the case of $\land$, the truth table states that $a \land b$ is true if and only if $a$ is true and $b$ is true. Therefore if $a$ is true, $a \land b$ holds the truth value of $b$, and if $a$ is false, $a \land b$ is false regardless of the value of $b$. From this, we design the tensor for $\land$ as follows
\[
T^{\land} =
\left[
\begin{tabular}{cc|cc}
1 & 0 & 0 & 0\\
0 & 1 & 1 & 1
\end{tabular}
\right]
\]
When $a$ is true, the partial application of the tensor to the first argument should yield an identity matrix, as its application to the second argument should yield the truth value of $b$:
\[
T^{\land} \times \overrightarrow{\top} =
\left[
\begin{tabular}{cc|cc}
1 & 0 & 0 & 0\\
0 & 1 & 1 & 1
\end{tabular}
\right]
\left[
\begin{tabular}{c}
1\\
0
\end{tabular}
\right]
=
\left[
\begin{tabular}{cc}
1 & 0\\
0 & 1
\end{tabular}
\right]
\]
When $a$ is false, the partial application of the tensor to the first argument should yield a matrix which ignores the truth value of the second argument, mapping both true and false to false:
\[
T^{\land} \times \overrightarrow{\bot} =
\left[
\begin{tabular}{cc|cc}
1 & 0 & 0 & 0\\
0 & 1 & 1 & 1
\end{tabular}
\right]
\left[
\begin{tabular}{c}
0\\
1
\end{tabular}
\right]
=
\left[
\begin{tabular}{cc}
0 & 0\\
1 & 1
\end{tabular}
\right]
\]
Having established these partial applications, we can verify that the truth table for $\land$ is replicated:
\begin{align*}
    & \left(T^{\land} \times \overrightarrow{\top}\right) \times \overrightarrow{\top} =
\left[
\begin{tabular}{cc}
1 & 0\\
0 & 1
\end{tabular}
\right]
\left[
\begin{tabular}{c}
1\\
0
\end{tabular}
\right]
=
\left[
\begin{tabular}{c}
1\\
0
\end{tabular}
\right]
= \overrightarrow{\top}\\
& \left(T^{\land} \times \overrightarrow{\top}\right) \times \overrightarrow{\bot} =
\left[
\begin{tabular}{cc}
1 & 0\\
0 & 1
\end{tabular}
\right]
\left[
\begin{tabular}{c}
0\\
1
\end{tabular}
\right]
=
\left[
\begin{tabular}{c}
0\\
1
\end{tabular}
\right]
= \overrightarrow{\bot}\\
& \left(T^{\land} \times \overrightarrow{\bot}\right) \times \overrightarrow{\top} =
\left[
\begin{tabular}{cc}
0 & 0\\
1 & 1
\end{tabular}
\right]
\left[
\begin{tabular}{c}
1\\
0
\end{tabular}
\right]
=
\left[
\begin{tabular}{c}
0\\
1
\end{tabular}
\right]
= \overrightarrow{\bot}\\
& \left(T^{\land} \times \overrightarrow{\bot}\right) \times \overrightarrow{\bot} =
\left[
\begin{tabular}{cc}
0 & 0\\
1 & 1
\end{tabular}
\right]
\left[
\begin{tabular}{c}
0\\
1
\end{tabular}
\right]
=
\left[
\begin{tabular}{c}
0\\
1
\end{tabular}
\right]
= \overrightarrow{\bot}\\
\end{align*}
A similar set of steps can be used to verify the other tensor encodings of logical connectives specified above.

\paragraph{Non-linearity and Quantification}
It may therefore seem that for the case where $S=B_2$, logical operators may be dealt with using only multi-linear maps represented as tensors. However, when it comes to dealing with quantification, it is not obvious that a solution using multilinear maps exists.

An intuitive way of modelling universal quantification is as follows: expressions of the form ``All $X$s are $Y$s'' are true if and only if $M_X = M_X \cap M_Y$, where $M_X$ and $M_Y$ are the sets of $X$s and the set of $Y$s, respectively. We saw earlier that sets of objects of the domain could be represented as vectors $\overrightarrow{X}$ and $\overrightarrow{Y}$ in a vector space $N$. The intersection of two such sets can be represented distributionally using the component-wise minimum function $min$:
\[
M_X \cap M_Y \mapsto min(\overrightarrow{X},\overrightarrow{Y})
\]
Using this, we can define the map $forall$ for distributional universal quantification modelling expressions of the form ``All $X$s are $Y$s'' as follows:
\[
forall(\overrightarrow{X},\overrightarrow{Y}) =
\left\{
\begin{tabular}{ll}
$\overrightarrow{\top}$ & $\quad$ if $\overrightarrow{X} = min(\overrightarrow{X},\overrightarrow{Y})$\\
$\overrightarrow{\bot}$ & $\quad$ otherwise
\end{tabular}
\right.
\]
Existential statements of the form ``There exists X'' can be modelled using the function $exists$, which tests whether or not $M_X$ is empty, and is defined as follows:
\[
exists(\overrightarrow{X}) =
\left\{
\begin{tabular}{ll}
$\overrightarrow{\top}$ & $\quad$ if $|\overrightarrow{X}| > 0$\\
$\overrightarrow{\bot}$ & $\quad$ otherwise
\end{tabular}
\right.
\]
In both cases, the definitions are given for $S = B_2$, but are adaptable to $S = B_1$ by setting $\overrightarrow{\bot} = \overrightarrow{0}$.

To give a simple example, let us take the domain $\mathcal{D} = \{dog_1, dog_2, cat_1\}$, where $dog_1$ is a black dog, $dog_2$ is a brown dog, and $cat_2$ is a brown cat. Let the set $\overrightarrow{B}$ of brown things, the set $\overrightarrow{C}$ of cats and $\overrightarrow{P}$ of dogs be represented in vector form as follows:
\[
\overrightarrow{B} = 
\left[
\begin{tabular}{c}
1\\
0\\
1
\end{tabular}
\right]
\qquad
\overrightarrow{C} = 
\left[
\begin{tabular}{c}
0\\
0\\
1
\end{tabular}
\right]
\qquad
\overrightarrow{P} = 
\left[
\begin{tabular}{c}
1\\
1\\
0
\end{tabular}
\right]
\]
To evaluate a sentence such as ``all brown things are dogs'' we would compute $forall(\overrightarrow{B}, \overrightarrow{P})$ by first computing
\[
min(\overrightarrow{B}, \overrightarrow{P}) =
\left[
\begin{tabular}{c}
1\\
0\\
0
\end{tabular}
\right]
\]
then checking if the result is equal to $\overrightarrow{B}$, which it is not, and hence $forall(\overrightarrow{B}, \overrightarrow{P}) = \overrightarrow{\bot}$. We can check if there exist any brown cats by first computing the intersection of the set of brown things and the set of cats:
\[\overrightarrow{B}\cap\overrightarrow{C} = min(\overrightarrow{B},\overrightarrow{C}) = \left[
\begin{tabular}{c}
0\\
0\\
1
\end{tabular}
\right]
\]
We then check the size of this vector, which is $1$. The definition of $exists$ tells us that $exists(\overrightarrow{B} \cap \overrightarrow{C}) = \overrightarrow{\top}$.

Neither of the quantification functions defined above are multi-linear, since a multilinear function must be linear in all arguments. A counter example for $forall$ is to consider the case where $M_X$ and $M_Y$ are empty, and multiply their vector representations by non-zero scalar weights $\alpha$ and $\beta$.
\begin{align*}
  & \alpha\overrightarrow{X} = \overrightarrow{X}\\
  & \beta\overrightarrow{Y} = \overrightarrow{Y}\\
  & forall(\alpha\overrightarrow{X},\beta\overrightarrow{Y}) = forall(\overrightarrow{X},\overrightarrow{Y}) = \overrightarrow{\top}\\
  & forall(\alpha\overrightarrow{X}, \beta\overrightarrow{Y}) \neq \alpha\beta\overrightarrow{\top}
\end{align*}
I observe that the equations above demonstrate that $forall$ is not a multilinear map. This proof holds for $S=B_1$ and $S=B_2$.

The proof that $exists$ is not a multilinear map is equally trivial. Assume $M_X$ is an empty set and $\alpha$ is a non-zero scalar weight:
\begin{align*}
  & \alpha \overrightarrow{X} = \overrightarrow{X}\\
  & exists(\alpha \overrightarrow{X}) = exists(\overrightarrow{X}) = \overrightarrow{\bot}\\
  & exists(\alpha \overrightarrow{X}) \neq \alpha \overrightarrow{\bot}
\end{align*}
This proof holds for $S = B_2$.

To conclude this section, we have seen how the two choices for boolean sentence space differ when we consider how to `implement' logical operations as tensors. For $S = B_1$, most operations need be non-linear maps, which can be defined in our distributional formal semantic models but not in those constructed within the DisCoCat formalism. For $S = B_2$, logical connectives can be modelled as multilinear maps (and therefore as tensors), and can thus be used in the DisCoCat setting as well as in the distributional formal semantics model presented in this section, however the (perhaps na\"ive) functions used to model basic quantification were shown to be non-linear maps for both settings. The question of how to approach quantification and logic in the DisCoCat framework, either using only multilinear maps or by using some semantic representation other than $\mathbf{FVect}$, as well as the question of how to go from simulating predicate logic with tensors to defining (or learning) logical connectives for the non-truth theoretic compositional distributional models described in this thesis (i.e.~going from vector encodings of classical logic to logic for continuous vector space models), are both difficult and fundamental questions which would merit attention in future research on this topic. 
% subsection integrating_non_linearity (end)

% section distributional_logic (end)

\section{Learning Tensors by Multi-Step Linear Regression} % (fold)
\label{sec:learning_tensors_by_multi_step_regression}

In Chapter~\ref{cha:learning_procedures_for_a_discocat}, I presented learning procedures that could be used to produce distributional representations of words and relations within the DisCoCat framework based on information provided by a corpus. A parallel effort to describe distributional compositionality has been presented in the work of Marco Baroni, Roberto Zamparelli and colleagues, which I briefly described in $\S$\ref{sub:adjective_matrices}.

To quickly repeat how their approach, first presented in \cite{Baroni2010}, works: nouns are vectors in some vector space $N$, and adjectives are square matrices in $N \otimes N$. Adjective-noun composition is simply matrix-vector multiplication. In sum, this is exactly how things work in the DisCoCat models I discussed in Chapter~\ref{cha:learning_procedures_for_a_discocat}. The novel aspect of their approach is the learning procedure for adjective matrices. To learn the matrix for an adjective $adj$, the following steps are taken:
\begin{enumerate}
  \item For each instance of the adjective applied to some noun $n$, the vector $\overrightarrow{adj\ n}$ is learned using the same procedure used to learn noun vectors.
  \item Let $I$ be the set of noun vectors $adj$ takes as `input' and $O$ be the set of adjective-noun vectors learned in the previous steps. Therefore for each input vector in $I$ there is an output vector in $O$.
  \item The matrix for $adj$ is learned through linear regression, such that the output set $O'$ produced by multiplying the matrix for $adj$ with the elements of $I$ is minimally different from $O$.
\end{enumerate}
This is an interesting machine-learning approach, with a high degree of parametric freedom: we are free to select the procedure for building the noun and adjective-noun vectors, the linear-regression algorithm used, the distance metric used to compare $O$ and $O'$, what sort of dimensionality reduction techniques are applied, and so on.

It should be noted that this approach can directly be applied to how we dealt with intransitive verbs in Chapter~\ref{cha:learning_procedures_for_a_discocat}, since they are also matrices in $N \otimes N$. The question is now: can this learning procedure be scaled to deal with the higher order tensors used for other constructs, such as transitive verbs? In \cite{GrefSadrBarIWCS13}\footnote{I am indebted to my collaborators Georgiana Dinu, Yao-Zhong Zhang, Mehrnoosh Sadrzadeh and Marco Baroni for their contribution to this work, and for allowing me to reproduce the contents. The figure in this section was produced by Marco Baroni. The parameters for the learning procedure were described by Georgiana Dinu.}, in collaboration with Baroni and his colleagues, Mehrnoosh Sadrzadeh and I have produced a generalised multi-step linear regression learning procedure which permits the construction of higher order semantic tensors for use in DisCoCat models. In this section, I will first present this procedure before discussing how well tensors produced using this new learning mechanism performed in the experiments presented in Chapter~\ref{cha:evaluating_a_discocat}.

\subsection{Multi-Step Linear Regression} % (fold)
\label{sub:multi_step_regression}

Multi-step regression learning is a generalisation of linear
regression learning for tensors of order 3 or higher, as procedures
already exist for tensors of order 1 (lexical semantic vectors) and
order 2 (cf.~\cite{Baroni2010}). For order 1 tensors, we suggest
learning vectors using any standard lexical semantic vector learning
model, and present sample parameters in $\S$\ref{sub:experiments} below. Learning order 2 tensors
(matrices) can be treated as a multivariate multiple regression
problem, where the matrix components are chosen to optimise (in a
least squares error sense) the mapping from training instances of
input (argument) and output (composed expression) vectors. Consider
for example the task of estimating the components of the matrix
representing an intransitive verb, that maps subject vectors to
(subject-verb) sentence vectors (Baroni and Zamparelli discuss the
analogous adjective-noun composition case):
\[ \overrightarrow{s} = V \times \overrightarrow{subj} \]
The weights of the matrix are estimated by least-squares regression
from example pairs of input subject and output sentence vectors
directly extracted from the corpus. For example, the matrix for
\emph{sing} is estimated from corpus-extracted vectors representing
pairs such as $($\emph{mom}, \emph{mom sings}$)$, $($\emph{child},
\emph{child sings}$)$, etc. Note that if the input and output vectors
are $n$ dimensional, we must estimate an $n \times{} n$ matrix, each
row corresponding to a separate regression problem (the $i$-th row
vector of the estimated matrix will provide the weights to linearly
combine the input vector components to predict the $i$-th output
vector component).  Regression is a supervised technique requiring
training data. However, as we are extracting the training data automatically
from the corpus, this approach does not incur an extra knowledge
cost with respect to unsupervised methods.

\begin{figure*}[htb]
    \centering
    \includegraphics[scale=0.5]{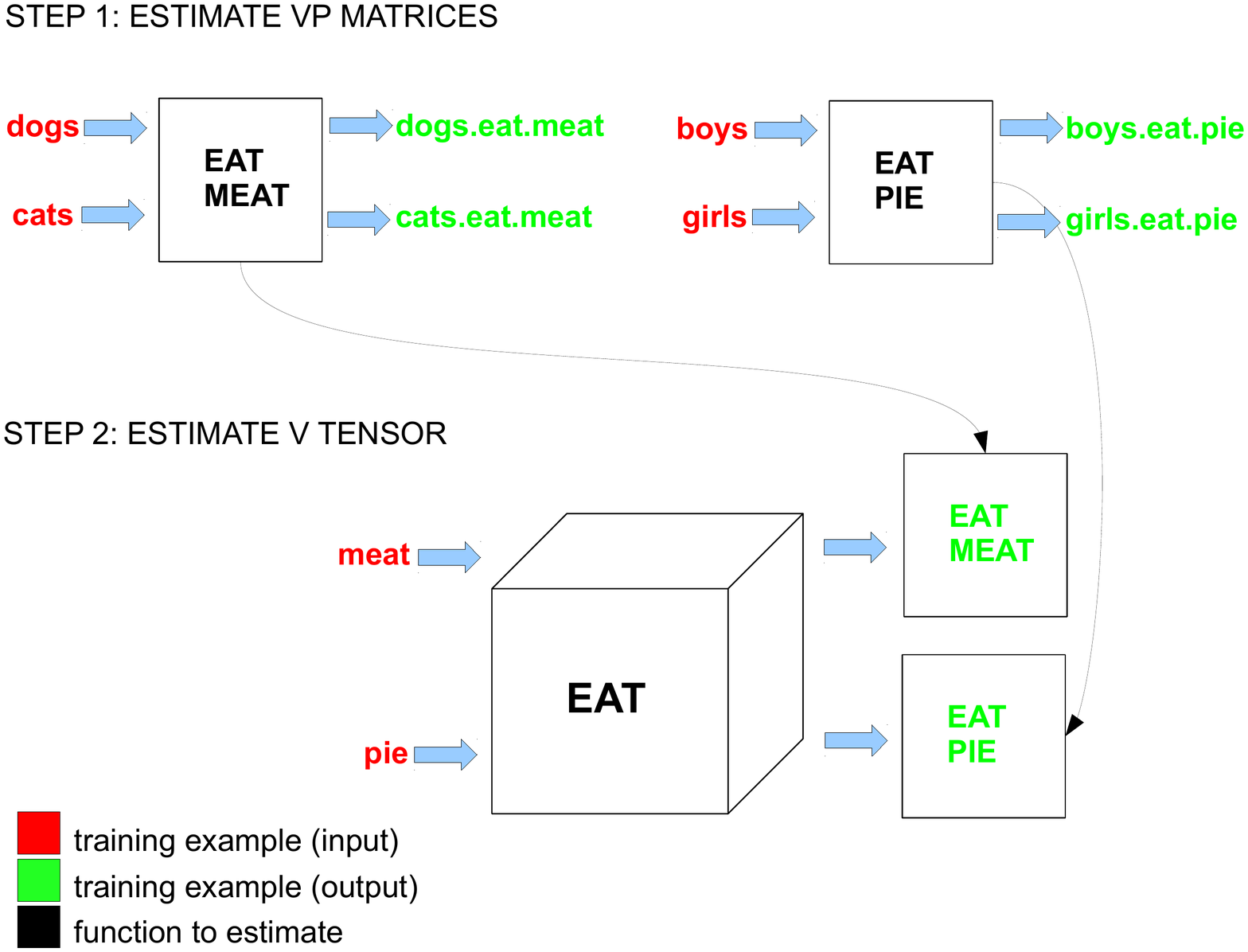}
    \caption{Estimating a tensor for \emph{eat} in two steps.}
    \label{fig:two-step-estimation}
\end{figure*}

Learning tensors of higher order by linear regression involves
iterative application of the linear regression learning method
described above. The idea is to progressively learn the functions of
arity two or higher encoded by such tensors by recursively learning
the partial application of these functions, thereby reducing the
problem to the same matrix-learning problem as addressed by Baroni and
Zamparelli. To start with an example: the matrix-by-vector operation
of \cite{Baroni2010} is a special case of the general
tensor-based function application model we are proposing, where a
`mono-argumental' function (intransitive verbs) corresponds to a order
2 tensor (a matrix). The approach is naturally extended to
bi-argumental functions, such as transitive verbs, where the verb will
be a order 3 tensor to be multiplied first by the object vector and
then by the subject, to return a sentence-representing vector:
\[ 
\overrightarrow{s} = V \times \overrightarrow{obj} \times \overrightarrow{subj} 
\]
The first multiplication of a $n\times{}n\times{}n$ tensor by a
$n$-dimensional vector will return a $n$-by-$n$ matrix (equivalent to
an intransitive verb, as it should be: both \emph{sings} and
\emph{eats meat} are VPs requiring a subject to be saturated). Note
that given $n$-dimensional input vectors, the $ij$-th $n$-dimensional
vector in the estimated tensor provides the weights to linearly
combine the input object vector components to predict the $ij$-th
output component of the unsaturated verb-object matrix. The matrix is
then multiplied by the subject vector to obtain a $n$-dimensional
vector representing the sentence. Again, we estimate the tensor
components by linear regression on input-output examples. In the first
stage, we apply linear regression to obtain examples of semi-saturated
matrices representing \emph{verb-object} constructions with a specific
verb. These matrices are estimated, like in the intransitive case,
from corpus-extracted examples of $($subject, subject-verb-object$)$
pairs. After estimating a suitable number of such matrices for a
variety of objects of the same verb, we use pairs of corpus-derived
object vectors and the corresponding estimated verb-object matrices as
input-output pairs for another regression, where we estimate the verb
tensor components. The estimation procedure is schematically
illustrated for \emph{eat} in Fig.~\ref{fig:two-step-estimation}. We first
estimate matrices for the VPs \emph{eat-meat}, \emph{eat-pie}
etc.~by linear regression on input subject and output sentence
vector pairs. We then estimate the tensor for \emph{eat} by
linear regression with the matrices estimated in the previous
step as output examples, and the vectors for the corresponding
objects as input examples.

We can generalise this learning procedure to functions of arbitrary arity. Consider an $n$-ary function $f: X_1 \to \ldots \to X_n \to Y$. Let $L_i$ be the set of $i$-tuples $\{w^j_1,\ldots,w^j_i\}_{i\in[1,k]}$, where $k = |L_i|$, corresponding to the words which saturate the first $i$ arguments of $f$ in a corpus. For each tuple in some set $L_i$, let $f\ w^j_1\ \ldots\ w^j_i = f^j_i: X_{i+1} \to \ldots \to X_n \to Y$. Trivially, there is only one such $f^j_0$---namely $f$ itself---since $L_0 = \emptyset$ (as there are no arguments of $f$ to saturate for $i = 0$). The idea behind multi-step regression is to learn, at each step, the tensors for functions $f^j_i$ by linear regression over the set of pairs $(w^{j'}_{i+1},f^{j'}_{i+1})$, where the tensors $f^{j'}_{i+1}$ are the expected outcomes of applying $f^j_i$ to $w^{j'}_{i+1}$ and are learned during the previous step. We bootstrap this algorithm by learning the vectors in $Y$ of the set $\{f_n^j\}_j$ by treating the word which each $f^j_n$ models combined with the words of its associated tuple in $L_n$ as a single token. We then learn the vector for this token from the corpus using our preferred distributional semantics method. By recursively learning the sets of functions from $i=n$ down to $0$, we obtain smaller and smaller sets of increasingly de-saturated versions of $f$, which converge towards $f_0 =f$.

To specify how the set of pairs used for recursion is determined, let
there exist a function $super$ which takes the index of a tuple from
$L_i$ and returns the set of indices from $L_{i+1}$ which denote
tuples identical to the first tuple, excluding the last element:
{\small
\[
super: \mathbb{N} \times \mathbb{N} \to \mathcal{P}(\mathbb{N}) :: (i,j) \mapsto \{j' | \forall j' \in [1,k'].[w_1^j = w_1^{j'} \land \ldots \land w_i^j = w_i^{j'}]\}\ \text{where}\ k' = |L_{i+1}|
\]
}
Using this function, the regression set for some $f_i^j$ can be defined as $\{(w^{j'}_{i+1},f^{j'}_{i+1})|j' = super(i,j)\}$.

% subsection multi_step_regression (end)

\subsection{Experiments} % (fold)
\label{sub:experiments}

We evaluated tensors produced using this multi-step regression method against the datasets presented in Chapter~\ref{cha:evaluating_a_discocat}, producing three new sets of experimental results. In this section, I will first report the model parameters initially described by Georgiana Dinu, before reporting the experimental results that show this new learning procedure to produce models that perform competitively against other leading models.

\paragraph{Source corpus} We extract co-occurrence data from the
concatenation of the Web-derived ukWaC corpus\footnote{\url{http://wacky.sslmit.unibo.it/}}, a mid-2009 dump of the English
Wikipedia \footnote{\url{http://en.wikipedia.org}} and the British National
Corpus \footnote{\url{http://www.natcorp.ox.ac.uk/}}. The corpus has been
tokenised, POS-tagged and lemmatised with the TreeTagger
\cite{Schmid1995} and dependency-parsed with the MaltParser
\cite{Hall2006}. It contains about 2.8 billion tokens.

\paragraph{Co-occurrence extraction} We collect vector representations
for the top 8K most frequent nouns and 4K verbs in the corpus, as well
as for the subject-verb (320K) and subject-verb-object (1.36M) phrases
containing one of the verbs to be used in one of the experiments below
and subjects and objects from the list of top 8K nouns. For all target
items, we collect within-sentence co-occurrences with the top 10K most
frequent content words (nouns, verbs, adjectives and adverbs), save
for a stop list of the 300 most frequent words. We extract
co-occurrence statistics at the lemma level, ignoring inflectional
information.

\paragraph{Weighting} Following standard practice, raw co-occurrence
counts are transformed into statistically weighted scores.  We tested
various weighting schemes of the semantic space on a word similarity
task. We used a subset of the MEN data-set
\cite{Bruni2012} containing 2000 pairs of words (present
in our vocabulary) together with human-assigned similarity judgments
obtained through crowdsourcing. We observed that non-negative
pointwise mutual information (PMI) and local mutual information (raw
frequency count multiplied by PMI score) generally outperform other weighting schemes by a large margin, and that PMI in particular works
best when combined with dimensionality reduction by non-negative
matrix factorization (described below). Consequently, we pick PMI
weighting for our experiments.

\paragraph{Dimensionality reduction} Reducing co-occurrence vectors to
lower dimensionality is a common step in the construction of
distributional semantic models. Extensive evidence suggests that
dimensionality reduction does not affect, and might even improve the
quality of lexical semantic vectors
\cite{Landauer1997,Sahlgren2006,Schuetze1997}. In our
setting, dimensionality reduction is virtually a necessary step, since
working with 10K-dimensional vectors is problematic for the Regression
approach, which requires learning matrices and tensors with dimensionalities which are quadratic
and cubic in the dimensionality of the input vectors, respectively.

We consider two dimensionality reduction methods, the Singular Value
Decomposition (SVD) and Non-negative Matrix Factorization (NMF). SVD
is the most common technique in distributional semantics, and it was
used by \cite{Baroni2010}. NMF is a less commonly
adopted method, but it has also been shown to be an effective
dimensionality reduction technique for distributional semantics
\cite{Dinu2010}. It has a fundamental advantage from our
point of view: the Multiply and Kronecker composition approaches, because of their
multiplicative nature, cannot be meaningfully applied to vectors
containing negative values. NMF, unlike SVD, produces non-negative
vectors, and thus allows a fair comparison of all composition methods
in the same reduced space. For this reason, we
 follow \cite{Baroni2010} in performing dimensionality
reduction to obtain 300-dimensional representations.

We perform the Singular Value Decomposition of the input matrix $X$:
$X = U\Sigma V^{\top}$ and, like Baroni and Zamparelli and many others,
pick the first $k=300$ columns of $U\Sigma$ to obtain reduced
representations. Non-negative Matrix Factorization factorizes a $(m
\times n)$ non-negative matrix $X$ into two $(m \times k)$ and $(k
\times n)$ non-negative matrices: $X\approx WH$ (we normalize the
input matrix to $\sum_{i,j}X_{ij} = 1$ before applying NMF). We use
the Matlab implementation of the projected gradient algorithm\footnote{Available at
\url{http://www.csie.ntu.edu.tw/~cjlin/nmf/}.} proposed
in \cite{Lin2007}, which minimizes the
squared error of Frobenius norm $F(W,H) = \|X - WH\|^2_F$.  We set
$k=300$ and we use $W$ as reduced representation of input
matrix $X$. 
For both SVD and NMF, the latent dimensions are computed using a
``core'' matrix containing nouns and verbs only, subsequently
projecting phrase vectors onto the same space. In this way, the
dimensions of the reduced space do not depend on the ad-hoc choice of
phrases required by our experiments.

By experimenting with the MEN data-set of word similarity judgments
(see above), we found that the performance using distributional
semantic vectors from the original 10K-dimensional space or from the
two reduced spaces is very similar, confirming the hypothesis that
dimensionality reduction does not have a negative impact on the
quality of distributional word representations.

We evaluated this new learning procedure against other models of composition, most of which were discussed in Chapter~\ref{cha:evaluating_a_discocat}. Here are the ones we used in the new set of experiments.

\textbf{Verb} is a baseline measuring the cosine between the verbs in
two sentences as a proxy for sentence similarity (e.g., similarity of
\emph{mom sings} and \emph{boy dances} is approximated by the cosine
of \emph{sing} and \emph{dance}).

We adopt the widely used and generally successful multiplicative and
additive models of \cite{Mitchell2010} and
others. Composition with the \textbf{Multiply} and \textbf{Add
}methods is achieved by, respectively, component-wise multiplying and
adding the vectors of the constituents of the sentence we want to
represent. Vectors are normalised before addition, as this has
consistently shown to improve the performance of additive models in earlier experiments run by Baroni and his colleagues.

In $\S$\ref{sub:an_efficient_alternative}, I proposed an alternative
implementation of the general DisCoCat approach to compositional
distributional semantics \cite{Coecke2010}, namely
\textbf{Kronecker}. Under this approach, a transitive sentence is
a matrix $S$ derived from:
\[ S = (\overrightarrow{v} \otimes \overrightarrow{v}) \odot (\overrightarrow{subj}
\otimes \overrightarrow{obj}) \]
That is, if nouns and verbs live in a $n$-dimensional space, a
transitive sentence is a $n\times{}n$ matrix given by the
component-wise multiplication of two Kronecker products: that of the
verb vector with itself and that of the subject and object
vectors. I showed, in $\S$\ref{sec:second_experiment}, that this method outperforms
other implementations of the same formalism and is the current state
of the art on the transitive sentence task, which we also tackle below. 
For intransitive sentences, the same approach reduces to the
component-wise multiplication of verb and subject vectors, that is, to
the Multiply method.

Training of nouns and verbs under the proposed (multi-step)
\textbf{Regression} model is implemented using Ridge Regression (RR)
\cite{Hastie2009}.  RR, also known as $L_{2}$ regularized
regression, is a different approach from the Partial Least Square
Regression (PLSR) method that was used in previous related work
\cite{Baroni2010,Guevara2010} to deal with the
multicollinearity problem.  When multicollinearity exists, the matrix
$X^{\top}X$ ($X$ here is the input matrix after dimensionality reduction)
becomes nearly singular and the diagonal elements of $(X^{\top}X)^{-1}$
become quite large, which makes the variance of weights too large.  In
RR, a positive constant $\lambda$ is added to the diagonal elements of
$X^{\top}X$ to strengthen its non-singularity.  Compared with PLSR, RR
has a simpler solution for the learned weight matrix $B = (X^{\top}X +
\lambda I)^{-1}X^{\top}Y$ and produces competitive results at a faster
speed.  For each verb matrix or tensor to be learned, we tuned the
parameter $\lambda$ by generalized cross-validation
\cite{Golub1979}. The objective function used for tuning
minimizes least square error when predicting corpus-observed sentence
vectors or intermediate VP matrices; the data sets we evaluate the
models on are \emph{not} touched during tuning.

The training examples for Regression are constructed by combining the
8K nouns we have vectors for with any verb in the
evaluation sets into subject-verb-(object)
constructions, and extracting the corresponding vectors from the
corpus, where attested (vectors are normalised before feeding them to
the regression routine). We use only example vectors with at least 10 dimensions with non-zero basis weights before dimensionality reduction, and we require at
least 3 training examples per regression. For the first experiment
(intransitives), these (untuned) constraints result in an average of
281 training examples per verb. In the second experiment, in the
verb-object matrix estimation phase, we estimate on average 324
distinct matrices per verb, with an average of 15 training examples
per matrix. In the verb tensor estimation phase we use all relevant
verb-object matrices as training examples.

\paragraph{Experimental Results}

We present the results for two experiments testing our set of models in Table~\ref{tab:results}. `Humans' is inter-annotator correlation.  The multiplication-based Multiply and Kronecker methods are not well-suited for the SVD space and their performance is reported in NMF space only. `Kronecker' is only defined for the transitive case, with `Multiply' functioning as its intransitive-case equivalent.

We first tested our set of models against the intransitive verb dataset presented in $\S$\ref{sec:second_experiment}, to verify that the regression approach developed for adjectives worked suitably for intransitive verbs as well. 
The results in Table \ref{tab:results}\subref{tab:intransitive-results} show that the
Regression-based model achieves the best correlation when applied to
SVD space, confirming that the approach proposed by Baroni and
Zamparelli for adjective-noun constructions can be straightforwardly
extended to composition of a verb with its subject with good empirical
results. The Regression model also achieves good performance in NMF
space, where it is comparable to Multiply.  Multiply was found to be
the best model by Mitchell and Lapata, and we confirm their results
here (recall that Multiply can also be seen as the natural extension
of Kronecker to the intransitive setting). The correlations attained
by Add and Verb are considerably lower than those of the other
methods.

\begin{table}[ht]
\centering
{\small
\subtable[Intransitive Sentences]{
  \centering
  \begin{tabular}{l|r}
    \emph{method}&$\rho$\\
    \hline
    Humans&0.40\\
    \hline
    Multiply.nmf&0.19\\
    Regression.nmf&0.18 \\
    Add.nmf&0.13\\
    Verb.nmf&0.08\\
    \hline
    Regression.svd&0.23\\
    Add.svd&0.11\\
    Verb.svd&0.06\\
  \end{tabular}
  \label{tab:intransitive-results}
  $\quad\quad\quad$
}
\subtable[Transitive Sentences]{
\centering
  \begin{tabular}{l|rl}
    \emph{method}&$\rho$\\
    \hline
    Humans&0.62\\
    \hline
    Regression.nmf&0.29 \\
    Kronecker.nmf&0.25\\
    Multiply.nmf&0.23\\
    Add.nmf&0.07\\
    Verb.nmf&0.04\\
    \hline
    Regression.svd&0.32\\
    Add.svd&0.12\\
    Verb.svd&0.08\\
  \end{tabular}
\label{tab:transitive-results}
  $\quad\quad\quad$
  }
}

\caption{Spearman correlation of composition methods with human similarity intuitions on two sentence similarity data sets (all correlations significantly above chance).}
\label{tab:results}
\end{table}

We then applied our regression method to the task of learning transitive verbs from a corpus, and tested these representations against the transitive verb dataset presented in $\S$\ref{sec:second_experiment}. As the results in Table
\ref{tab:results}\subref{tab:transitive-results} show, the Regression
model performs very well again, better than any other methods in NMF
space, and with a further improvement when SVD is used, similarly to
the first experiment.  The Kronecker model is also competitive,
confirming the results from $\S$\ref{sec:second_experiment}. Neither Add nor Verb achieve very good results, although
even for them the correlation with human ratings is significant.

% subsection experiments (end)

\subsection{Discussion} % (fold)
\label{sub:discussion_and_next_steps}

The results presented in the previous section show that our iterative linear regression
algorithm outperforms the leading multiplicative method on
intransitive sentence similarity when using SVD (and it is on par with
it when using NMF), and outperforms both the multiplicative method and
the leading Kronecker model in predicting transitive sentence
similarity.

Here again, we saw that Kronecker also performs very well in our modified experimental setup (although
not as well as Regression). The main advantage of Kronecker over
Regression lies in its simplicity: there is no training involved, all
it takes is two vector outer products and a component-wise
multiplication.

While our new regression-based model's estimation procedure is
considerably more involved than for Kronecker, the model has much to
recommend it, both from a statistical and from a linguistic point of
view. On the statistical side, there are many aspects of the
estimation routine that could be tuned on automatically collected
training data, thus bringing up the Regression model performance. We
could, for example, harvest a larger number of training phrases (not
limiting them to those that contain nouns from the 8K most frequent in
the corpus, as we did), or \emph{vice versa} limit training to more
frequent phrases, whose vectors are presumably of better
quality. Moreover, Ridge Regression is only one of many estimation
techniques that could be tried to come up with better matrix and
tensor weights.

On the linguistic side, the model is clearly motivated as an
instantiation of the vector-space ``dual'' of classic composition by
function application via the tensor contraction operation, as
discussed earlier in this chapter.  Moreover, Regression
produces vectors of the same dimensionality for sentences formed with
intransitive and transitive verbs, whereas for Kronecker, if the
former are $n$-dimensional vectors, the second are $n\times{}n$
matrices. Thus, under Kronecker composition, sentences with
intransitive and transitive verbs are not directly comparable, which
is counter-intuitive (being able to measure the similarity of, say,
\emph{kids sing} and \emph{kids sing songs} is both natural and
practically useful).
% subsection discussion_and_next_steps (end)

% section learning_tensors_by_multi_step_regression (end)

\section{Further Syntactic Extensions: Combinatory Categorial Grammar} % (fold)
\label{sec:supporting_combinatory_categorial_grammar}

In Chapter~\ref{cha:syntactic_extensions}, I discussed procedures for extending the DisCoCat framework to work with other grammatical formalisms, and showed how Context Free Grammars and Lambek Grammars could be included into the formalism in such a way. In this section, I sketch the foundations for extending DisCoCat to work with a more powerful grammatical formalism called Combinatory Categorial Grammar, a weakly context sensitive grammar which is widely used and comes with efficient parsing tools such as the C\&C parser of \cite{Curran2007}. I show that at least some aspects of the grammar can be given categorical semantics, and briefly discuss aspects which should be the subject of future work, should we wish to fully integrate variants of CCG into the DisCoCat formalism.

\subsection{Combinatory Categorial Grammar} % (fold)
\label{sub:combinatory_categorial_grammar}

Combinatory Categorial Grammars (CCGs) are very similar in spirit---and notation---to Lambek Grammar, discussed in $\S$\ref{sec:supporting_lambek_grammar}. They are categorial, in that every word in a natural language belongs to one or more syntactic categories; and they are combinatorial in that syntactic categories act as functions which combine to produce syntactical analyses of phrases. Because I also talk about category theory, I will talk of CCG types instead of CCG categories to avoid confusion.

CCG types are generally defined recursively as follows:
\begin{itemize}
  \item Let there be a set of atomic types $N$, $S$, etc.
  \item $A\backslash_C B$ is a type if $A$ and $B$ are types.
  \item $A / B$ is a type if $A$ and $B$ are types.
\end{itemize}
This completes the type definition. So far, it is very similar to the type definition for Lambek Grammar. However, the backslash $\backslash_C$ has a slightly different meaning than the backslash $\backslash_L$ used for Lambek grammars, as will become clear below. Generally speaking the two notations can be used interchangeably if the combination operations are rewritten appropriately: $A \backslash_L B$ is equivalent to writing $B \backslash_C A$, and vice-versa. These are merely two different notational conventions used in categorial grammars.

CCGs come equipped with a set of reduction rules. The principal ones surveyed by \cite{fowler2010accurate}, are as follows:
\begin{itemize}
  \item Application:
  \begin{align*}
    & X/Y,\, Y \to X\\
    & Y,\, X \backslash_C Y \to X
  \end{align*}
  \item Composition:
  \begin{align*}
    & X/Y,\, Y/Z \to X/Z\\
    & Y \backslash_C Z,\, X \backslash_C Y \to X \backslash_C Z
  \end{align*}
  \item Type raising:
  \begin{align*}
    & X \to T/(T\backslash_C X)\\
    & X \to T \backslash_C (T/X)
  \end{align*}
\end{itemize}
So far, these rules are almost exactly similar to the combination rules for Lambek Grammar presented in $\S$\ref{sec:supporting_lambek_grammar}, with the difference that $\cdot$ operations become commas  and the expressions of the form $X \backslash_L Y$ are turned into $Y\backslash_C X$ in the rules.

Two sets of rules which are not found in Lambek Grammar are the following:
\begin{itemize}
  \item Crossed composition:
  \begin{align*}
    & X /Y,\, Y \backslash_C Z \to X \backslash_C Z\\
    & Y/Z,\, X \backslash_C Y \to X/Z
  \end{align*}
  \item Substitution:
  \begin{align*}
    & (X/Y)/Z,\, Y/Z \to X/Z\\
    & (X/Y)\backslash_C Z,\, Y\backslash_C Z \to X\backslash_C Z\\
    & Y / Z,\, (X \backslash_C Y) / Z \to X/Z\\
    & Y \backslash_C Z,\, (X \backslash_C Y) \backslash_C Z \to X \backslash_C Z
  \end{align*}
\end{itemize}
Additionally, both composition and crossed composition have a generalised form which allows the rule to be applied when $Z$ is not an atomic type.

These rules are used to perform grammatical analysis of sentences as follows: if there is some type-assignment for each word in the sentence such that the sequence of types assigned to the sentence reduces to some sentence type $S$ through the combinatorial rules described above, then the sentence is grammatical.

There are many additional rules, restrictions and modifications of the above that are available to generate CCGs with certain properties. For example, CCGs are generally mildly-context sensitive \cite{Vijay-Shanker1994}, meaning that they can identify or generate sentence structures which context free grammars such as CFGs, LGs or pregroup grammars could not. However, with certain rule restrictions such as removing the unrestricted type raising rule described above, it can be shown \cite{fowler2010accurate} that some CCGs are in fact context free.

% subsection combinatory_categorial_grammar (end)

\subsection{Categorical Semantics for CCGs} % (fold)
\label{sub:categorical_semantics_for_ccgs}

I now turn to the task of trying to represent CCGs as categories. This section will provide foundations for this task and discuss some of the issues faced. It does not aim to provide a complete solution, leaving this for future work.

In $\S$\ref{sub:closed_monoidal_categories}, we saw that Lambek Grammars could be viewed as bi-closed monoidal categories. Since CCGs and LGs have some very similar combination rules, let us first consider how far bi-closed monoidal categories get us.

We begin by assuming that the categorical representation $\mathbf{CCG}$ of a CCG is a bi-closed monoidal category, with the following objects:
\begin{itemize}
  \item For each atomic type $A$ in the CCG there is some object $A$ in $ob(\mathbf{CCG})$.
  \item For each type of the form $A/B$ in the CCG there is some object $A \multimapinv B$ in $ob(\mathbf{CCG})$, where $A$ and $B$ are in $ob(\mathbf{CCG})$.
  \item For each type of the form $A \backslash_C B$ in the CCG there is some object $B \multimap A$ in $ob(\mathbf{CCG})$, where $A$ and $B$ are in $ob(\mathbf{CCG})$.
  \item For each sequence of types $A,\, B$ permitted by the CCG, there is an object $A \otimes B$ in $ob(\mathbf{CCG})$, where $A$ and $B$ are in $ob(\mathbf{CCG})$.
\end{itemize}
So far, this is effectively the same set of objects as one would expect to find in the categorical representation $\mathbf{LG}$ of a Lambek Grammar. The same goes for the first set of combination operations. As a reminder, the biclosed category has the following general morphisms for any $A$, $B$ and $C$ in $ob(\mathbf{CCG})$:
\begin{itemize}
    \item $ev^l_{A,B} : A \otimes (A \multimap B) \to B$
    \item $ev^r_{A,B} : (A \multimapinv B) \otimes B \to A$
    \item $\Lambda^l(f) : C \to A \multimap B$ for any $f: A \otimes C \to B$
    \item $\Lambda^r(g) : C \to A \multimapinv B$ for any $g: C \otimes B \to A$
\end{itemize}
Using these we can define the first set of composition operations categorically, as was done in $\S$\ref{sub:closed_monoidal_categories}:
\begin{itemize}
    \item \textbf{Application operations} are simply the evaluation morphisms.
    \item \textbf{Composition:} for any pair of objects $A \multimap B$ and $B \multimap C$ in the category, there is a morphism 
    \[comp^r_{A \multimap B, B \multimap C}: (A \multimap B) \otimes (B \multimap C) \to A \multimap C\]
     Likewise, for any pair of objects $A \multimapinv B$ and $B \multimapinv C$ in the category, there is a morphism 
     \[comp^l_{A \multimapinv B, B \multimapinv C}: (A \multimapinv B) \otimes (B \multimapinv C) \to A \multimapinv C\]
     \item \textbf{Type raising} morphisms $raise^l_{A,B}$ and $raise^r_{A,B}$ are just the right and left currying of the left and right evaluation morphisms:
     \begin{itemize}
         \item $raise^l_{A,B} = \Lambda^r(ev^l_{A,B}) : A \to B \multimapinv (A \multimap B)$
         \item $raise^r_{A,B}  = \Lambda^l(ev^r_{B,A}) : A \to (B \multimapinv A) \multimap B$
     \end{itemize}
\end{itemize}
Note that the composition morphisms make no assumptions about the structure of $C$ in the first case and $A$ in the second, thereby representing both composition and generalised composition at the same time. The diagrams for these operations are shown in Figures~\ref{fig:dlangspecific}--\ref{fig:typeraising} of Chapter~\ref{cha:syntactic_extensions}.

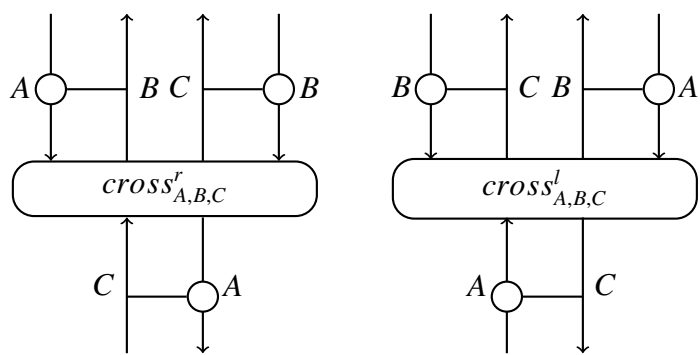
\begin{figure}[ht]
\centering
\tikzset{func/.style={shape=rectangle,rounded corners=8,minimum width=2cm,minimum height=.5cm,draw}}
\tikzset{claspnode/.style={shape=circle,minimum width=0.25cm,fill=white,draw}}
\tikzset{wideclasp/.style={shape=rectangle, rounded corners=3,minimum width=1.3cm,fill=white,draw}}
\begin{tikzpicture}[thick]
    \begin{scope}
      \draw[->] (0,4) -- node[left] {$A\ $} (0,2.05);
      \draw[->] (1,2.05) -- node[right] {$B$} (1,4);
      \node[claspnode] (c1) at (0,3) {};
      \draw (1,3) -- (c1);

      \draw[<-] (2,4) -- node[left] {$C$} (2,2.05);
      \draw[<-] (3,2.05) -- node[right] {$\ B$} (3,4);
      \node[claspnode] (c2) at (3,3) {};
      \draw (2,3) -- (c2);

      \node[func, minimum width=4cm] at (1.5,1.68) {$cross^r_{A,B,C}$};

      \draw [->] (1,-.5) -- node[left] {$C$} (1,1.3);
      \draw [<-] (2,-.5) -- node[right] {$\ A$} (2,1.3);
      \node[claspnode] (c3) at (2,0.25) {};
      \draw (1,0.25) -- (c3);
    \end{scope}
    \begin{scope}[xshift=5cm]
      \draw[->] (0,4) -- node[left] {$B\ $} (0,2.07);
      \draw[->] (1,2.07) -- node[right] {$C$} (1,4);
      \node[claspnode] (c1) at (0,3) {};
      \draw (1,3) -- (c1);

      \draw[<-] (2,4) -- node[left] {$B$} (2,2.07);
      \draw[<-] (3,2.07) -- node[right] {$\ A$} (3,4);
      \node[claspnode] (c2) at (3,3) {};
      \draw (2,3) -- (c2);

      \node[func, minimum width=4cm] at (1.5,1.68) {$cross^l_{A,B,C}$};

      \draw [->] (1,-.5) -- node[left] {$A\ $} (1,1.3);
      \draw [<-] (2,-.5) -- node[right] {$C$} (2,1.3);
      \node[claspnode] (c3) at (1,0.25) {};
      \draw (2,0.25) -- (c3);
    \end{scope}
  \end{tikzpicture}
\caption{Simple diagrammatic form of $cross$ morphisms.}
\label{fig:crossdiagsimple}
\end{figure}

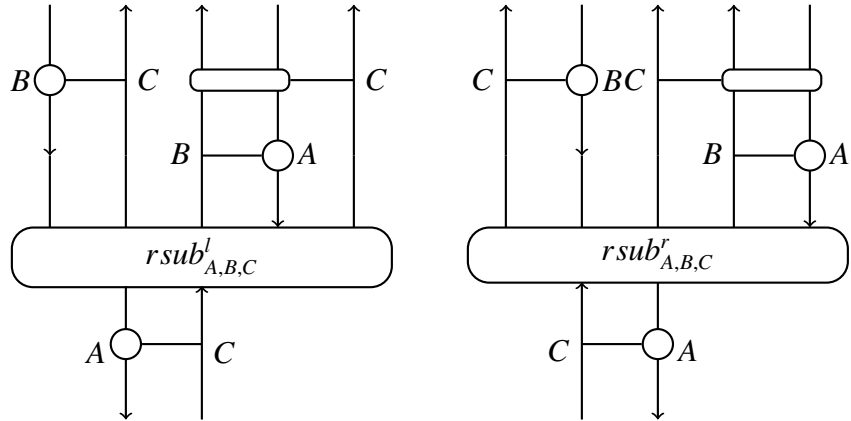
\begin{figure}[ht]
\centering
\tikzset{func/.style={shape=rectangle,rounded corners=8,minimum width=2cm,minimum height=.5cm,draw}}
\tikzset{claspnode/.style={shape=circle,minimum width=0.25cm,fill=white,draw}}
\tikzset{wideclasp/.style={shape=rectangle, rounded corners=3,minimum width=1.3cm,fill=white,draw}}
\begin{tikzpicture}[thick]
    \begin{scope}
    \draw[-] (3,6) -- (3,5);
    \draw[->] (2,5) -- (2,6);
    \draw[->] (3,5) -- node[right] {$\ A$} (3,3.05);
    \draw[-] (2,5) -- node[left] {$B$} (2,3.05);
    \draw[->] (4,4) -- node[right] {$C$} (4,6);
    \draw (4,4) -- (4,3.05);
    \node[claspnode] (c1) at (3,4) {};
    \draw (c1) -- (2,4);
    \node[wideclasp] (w1) at (2.5,5) {};
    \draw (w1) -- (4,5);

    \draw[->] (0,6) -- node[left] {$B\ $} (0,4);
    \draw (0,4) -- (0,3.05);
    \draw[->] (1,4) -- node[right] {$C$} (1,6);
    \draw (1,4) -- (1,3.05);
    \node[claspnode] (c2) at (0,5) {};
    \draw (c2) -- (1,5);

    \node[func, minimum width=5cm] at (2,2.65) {$rsub^l_{A,B,C}$};

    \draw[->] (1,2.27) -- node[left] {$A\ $} (1,0.5);
    \draw[<-] (2,2.27) -- node[right] {$C$} (2,0.5);
    \node[claspnode] (c3) at (1,1.5) {};
    \draw (c3) -- (2,1.5);
    \end{scope}

    \begin{scope}[xshift=6cm]
    \draw[-] (4,6) -- (4,5);
    \draw[->] (3,5) -- (3,6);
    \draw[->] (4,5) -- node[right] {$\ A$} (4,3.05);
    \draw[-] (3,5) -- node[left] {$B$} (3,3.05);
    \draw[->] (2,4) -- node[left] {$C$} (2,6);
    \draw (2,4) -- (2,3.05);
    \node[claspnode] (c1) at (4,4) {};
    \draw (c1) -- (3,4);
    \node[wideclasp] (w1) at (3.5,5) {};
    \draw (w1) -- (2,5);

    \draw[->] (1,6) -- node[right] {$\ B$} (1,4);
    \draw (0,4) -- (0,3.05);
    \draw[->] (0,4) -- node[left] {$C$} (0,6);
    \draw (1,4) -- (1,3.05);
    \node[claspnode] (c2) at (1,5) {};
    \draw (c2) -- (0,5);

    \node[func, minimum width=5cm] at (2,2.68) {$rsub^r_{A,B,C}$};

    \draw[->] (2,2.32) -- node[right] {$\ A$} (2,0.5);
    \draw[<-] (1,2.32) -- node[left] {$C$} (1,0.5);
    \node[claspnode] (c3) at (2,1.5) {};
    \draw (c3) -- (1,1.5);
    \end{scope}
  \end{tikzpicture}
\caption{Simple diagrammatic form of $rsub$ morphisms.}
\label{fig:rsubdiagsimple}
\end{figure}

So far, it seems like bi-closed monoidal categories are a suitable categorical representation for CCGs. However, a problem comes up when we consider crossed composition and substitution, which have the following forms:
\begin{align*}
  & cross^r_{A,B,C} : (A \multimapinv B) \otimes (C \multimap B) \to C \multimap A\\
  & cross^l_{A,B,C} : (B \multimapinv C) \otimes (B \multimap A) \to A \multimapinv C\\
  & lsub^l_{A,B,C} : ((A \multimapinv B) \multimapinv C) \otimes (B \multimapinv C) \to A \multimapinv C\\
  & lsub^r_{A,B,C} : (C \multimap (A \multimapinv B)) \otimes (C \multimap B) \to C \multimap B \\
  & rsub^l_{A,B,C} : (B \multimapinv C) \otimes ((A \multimap B) \multimapinv C) \to A \multimapinv C\\
  & rsub^r_{A,B,C} : (C \multimap B) \otimes (C \multimap (A \multimap B)) \to C \multimap A\\
\end{align*}
It should be fairly clear from the domain and codomains of these morphisms that bi-closed monoidal categories will most likely not support these operations using just the evaluation and currying morphisms provided above. This becomes especially clear when we consider the simple diagrammatic form of these morphisms, shown in Figure~\ref{fig:crossdiagsimple} for the $cross$ morphisms, and in Figures~\ref{fig:lsubdiagsimple} and~\ref{fig:rsubdiagsimple} for the $lsub$ and $rsub$ morphisms, respectively.

\begin{figure}[ht]
\centering
\tikzset{func/.style={shape=rectangle,rounded corners=8,minimum width=2cm,minimum height=.5cm,draw}}
\tikzset{claspnode/.style={shape=circle,minimum width=0.25cm,fill=white,draw}}
\tikzset{wideclasp/.style={shape=rectangle, rounded corners=3,minimum width=1.3cm,fill=white,draw}}
\begin{tikzpicture}[thick]
    \begin{scope}
    \draw[-] (0,6) -- (0,5);
    \draw[->] (1,5) -- (1,6);
    \draw[->] (0,5) -- node[left] {$A\ $} (0,3.05);
    \draw[-] (1,5) -- node[right] {$B$} (1,3.05);
    \draw[->] (2,4) -- node[right] {$C$} (2,6);
    \draw (2,4) -- (2,3.05);
    \node[claspnode] (c1) at (0,4) {};
    \draw (c1) -- (1,4);
    \node[wideclasp] (w1) at (.5,5) {};
    \draw (w1) -- (2,5);

    \draw[->] (3,6) -- node[left] {$B\ $} (3,4);
    \draw (3,4) -- (3,3.05);
    \draw[->] (4,4) -- node[right] {$C$} (4,6);
    \draw (4,4) -- (4,3.05);
    \node[claspnode] (c2) at (3,5) {};
    \draw (c2) -- (4,5);

    \node[func, minimum width=5cm] at (2,2.65) {$lsub^l_{A,B,C}$};

    \draw[->] (2,2.27) -- node[left] {$A\ $} (2,0.5);
    \draw[<-] (3,2.27) -- node[right] {$C$} (3,0.5);
    \node[claspnode] (c3) at (2,1.5) {};
    \draw (c3) -- (3,1.5);
    \end{scope}

    \begin{scope}[xshift=6cm]
    \draw[-] (1,6) -- (1,5);
    \draw[->] (2,5) -- (2,6);
    \draw[->] (1,5) -- node[left] {$A\ $} (1,3.05);
    \draw[-] (2,5) -- node[right] {$B$} (2,3.05);
    \draw[->] (0,4) -- node[left] {$C$} (0,6);
    \draw (0,4) -- (0,3.05);
    \node[claspnode] (c1) at (1,4) {};
    \draw (c1) -- (2,4);
    \node[wideclasp] (w1) at (1.5,5) {};
    \draw (w1) -- (0,5);

    \draw[->] (4,6) -- node[right] {$\ B$} (4,4);
    \draw (3,4) -- (3,3.05);
    \draw[->] (3,4) -- node[left] {$C$} (3,6);
    \draw (4,4) -- (4,3.05);
    \node[claspnode] (c2) at (4,5) {};
    \draw (c2) -- (3,5);

    \node[func, minimum width=5cm] at (2,2.67) {$lsub^r_{A,B,C}$};

    \draw[->] (3,2.32) -- node[right] {$\ A$} (3,0.5);
    \draw[<-] (2,2.32) -- node[left] {$C$} (2,0.5);
    \node[claspnode] (c3) at (3,1.5) {};
    \draw (c3) -- (2,1.5);
    \end{scope}
  \end{tikzpicture}
\caption{Simple diagrammatic form of $lsub$ morphisms.}
\label{fig:lsubdiagsimple}
\end{figure}

For the $cross$ morphisms in Figure~\ref{fig:crossdiagsimple} we can see that within the internal structure of the morphisms, there must be some point where the wires for $B$ and the wire for $C$ cross over each other in order for the two $B$ wires to form a cup, which is not possible in a bi-closed monoidal category. For the substitutions morphisms in Figures~\ref{fig:lsubdiagsimple} and~\ref{fig:rsubdiagsimple}, the problem is a little more subtle: not only must wires cross for the $B$ wires to form a cup, but there are two $C$ wires as input and only one as output. This means that there must be some mechanism within the morphism that takes two wires bearing the same letter and direction, and produces one wire bearing the same letter and direction as the other two.

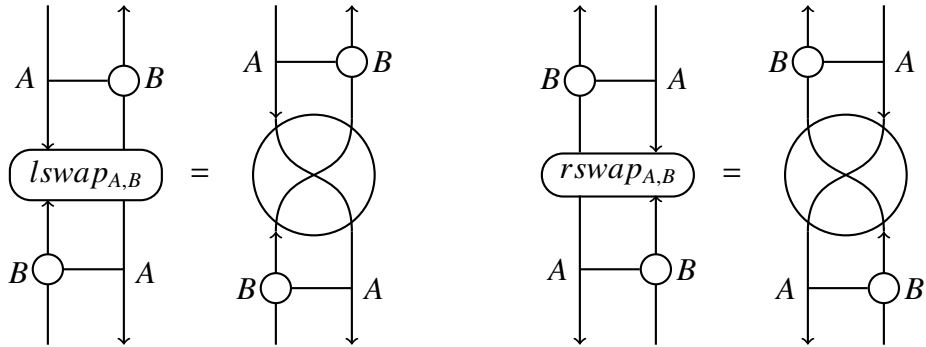
\begin{figure}[ht]
  \centering
    \begin{tikzpicture}[thick]
    \begin{scope}
      \draw[->] (0,3) -- node[left] {$A$} (0,1.08);
      \draw[<-] (1,3) -- node[right] {$\ B$} (1,1.08);
      \node[claspnode] (c1) at (1,2) {};
      \draw (c1) -- (0,2);
      
      \node[func] at (0.5,0.75) {$lswap_{A,B}$};

      \draw[->] (1,0.42) -- node[right] {$A$} (1,-1.5);
      \draw[<-] (0,0.42) -- node[left] {$B\ $} (0,-1.5);
      \node[claspnode] (c2) at (0,-.5) {};
      \draw (c2) -- (1,-.5);

      \node at (2,0.75) {$=$};

      \draw[->] (3,3) -- node[left] {$A$} (3,1.5);
      \draw[<-] (4,3) -- node[right] {$\ B$} (4,1.5);
      \node[claspnode] (c3) at (4,2.25) {};
      \draw (c3) -- (3,2.25);

      \draw (3.5,.75) circle (.8cm);

      \draw (3,1.5) .. controls +(0,-1) and +(0,1) .. (4,0);
      \draw (4,1.5) .. controls +(0,-1) and +(0,1) .. (3,0);
      
      \draw[->] (4,0) -- node[right] {$A$} (4,-1.5);
      \draw[<-] (3,0) -- node[left] {$B\ $} (3,-1.5);
      \node[claspnode] (c4) at (3,-.75) {};
      \draw (c4) -- (4,-.75);

    \end{scope}

    \begin{scope}[xshift=7cm]
      \draw[->] (1,3) -- node[right] {$A$} (1,1.05);
      \draw[<-] (0,3) -- node[left] {$B\ $} (0,1.05);
      \node[claspnode] (c1) at (0,2) {};
      \draw (c1) -- (1,2);
      
      \node[func] at (0.5,0.75) {$rswap_{A,B}$};

      \draw[->] (0,0.48) -- node[left] {$A$} (0,-1.5);
      \draw[<-] (1,0.48) -- node[right] {$\ B$} (1,-1.5);
      \node[claspnode] (c2) at (1,-.5) {};
      \draw (c2) -- (0,-.5);

      \node at (2,0.75) {$=$};

      \draw[->] (4,3) -- node[right] {$A$} (4,1.5);
      \draw[<-] (3,3) -- node[left] {$B\ $} (3,1.5);
      \node[claspnode] (c3) at (3,2.25) {};
      \draw (c3) -- (4,2.25);

      \draw (3.5,.75) circle (.8cm);

      \draw (3,1.5) .. controls +(0,-1) and +(0,1) .. (4,0);
      \draw (4,1.5) .. controls +(0,-1) and +(0,1) .. (3,0);
      
      \draw[->] (3,0) -- node[left] {$A$} (3,-1.5);
      \draw[<-] (4,0) -- node[right] {$\ B$} (4,-1.5);
      \node[claspnode] (c4) at (4,-.75) {};
      \draw (c4) -- (3,-.75);
    \end{scope}
    \end{tikzpicture}
  \caption{Diagrams for $lswap$ and $rswap$ morphisms.}
  \label{fig:swapdiags}
\end{figure}

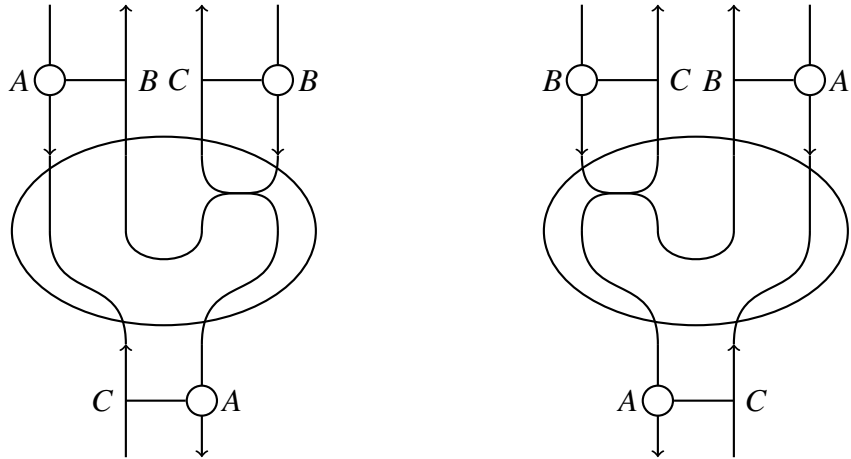
\begin{figure}[ht]
\centering
\begin{tikzpicture}[thick]
    \begin{scope}
      \draw[->] (0,4) -- node[left] {$A\ $} (0,2);
      \draw[->] (1,2) -- node[right] {$B$} (1,4);
      \node[claspnode] (c1) at (0,3) {};
      \draw (1,3) -- (c1);

      \draw[<-] (2,4) -- node[left] {$C$} (2,2);
      \draw[<-] (3,2) -- node[right] {$\ B$} (3,4);
      \node[claspnode] (c2) at (3,3) {};
      \draw (2,3) -- (c2);

      \draw (2,2) .. controls +(0,-1) and +(0,1) .. (3,1);
      \draw (3,2) .. controls +(0,-1) and +(0,1) .. (2,1);

      \draw (1,2) -- (1,1);
      \draw (1,1) .. controls +(0,-.5) and +(0,-.5) .. (2,1);

      \draw (0,2) -- (0,1);
      \draw (0,1) .. controls +(0,-1) and +(0,1) .. (1,-.5);
      \draw (3,1) .. controls +(0,-1) and +(0,1) .. (2,-.5);

      \draw (1.5,1) ellipse (2 and 1.25);

      \draw [->] (1,-2) -- node[left] {$C$} (1,-.5);
      \draw [<-] (2,-2) -- node[right] {$\ A$} (2,-.5);
      \node[claspnode] (c3) at (2,-1.25) {};
      \draw (1,-1.25) -- (c3);
    \end{scope}

    \begin{scope}[xshift=7cm]
       \draw[->] (0,4) -- node[left] {$B\ $} (0,2);
      \draw[->] (1,2) -- node[right] {$C$} (1,4);
      \node[claspnode] (c1) at (0,3) {};
      \draw (1,3) -- (c1);

      \draw[<-] (2,4) -- node[left] {$B$} (2,2);
      \draw[<-] (3,2) -- node[right] {$\ A$} (3,4);
      \node[claspnode] (c2) at (3,3) {};
      \draw (2,3) -- (c2);

      \draw (0,2) .. controls +(0,-1) and +(0,1) .. (1,1);
      \draw (1,2) .. controls +(0,-1) and +(0,1) .. (0,1);

      \draw (3,2) -- (3,1);
      \draw (1,1) .. controls +(0,-.5) and +(0,-.5) .. (2,1);

      \draw (2,2) -- (2,1);
      \draw (0,1) .. controls +(0,-1) and +(0,1) .. (1,-.5);
      \draw (3,1) .. controls +(0,-1) and +(0,1) .. (2,-.5);

      \draw (1.5,1) ellipse (2 and 1.25);

      \draw [->] (2,-2) -- node[right] {$C$} (2,-.5);
      \draw [<-] (1,-2) -- node[left] {$A\ $} (1,-.5);
      \node[claspnode] (c3) at (1,-1.25) {};
      \draw (2,-1.25) -- (c3);
    \end{scope}
  \end{tikzpicture}
\caption{Expanded diagrammatic form of $cross$ morphisms.}
\label{fig:crossmorphisms}
\end{figure}

Let us begin by addressing the issue of crossing wires: we modify our hypothesis about the categorical structure of $\mathbf{CCG}$ to state that it is a closed monoidal category. Typically a closed monoidal category is defined identically to a bi-closed monoidal category, except for the fact that instead of the pair of objects $A \multimap B$ and $B \multimapinv A$ for each $A$ and $B$ in the category, there is a single object $A \multimap B$ which satisfies $A \otimes (A \multimap B) \cong (A \multimap B) \otimes A \cong B$. In short, $A \multimap B$ in the closed monoidal category acts like both $A \multimap B$ and $B \multimapinv A$ in a bi-closed monoidal category. Here, we will modify our categorical definition explicitly to turn the bi-closed category into a closed category by adding the following morphisms for each $A$ and $B$:
\[
lswap_{A,B}: A \multimap B \to B \multimapinv A \qquad rswap_{A,B}: B \multimapinv A \to A \multimap B
\]
such that $rswap_{A,B} \circ lswap_{A,B} = id_{A \multimap B}$ and $lswap_{A,B} \circ rswap_{A,B} = id_{B \multimapinv A}$ (i.e.~$lswap$ and $rswap$ morphisms are isomorphisms). We give these morphisms the  diagrammatic representation shown in Figure~\ref{fig:swapdiags}. Using these morphisms, we can diagrammatically show the inner structure for the $cross$ morphisms initially shown in Figure~\ref{fig:crossdiagsimple}, as demonstrated in Figure~\ref{fig:crossmorphisms}.

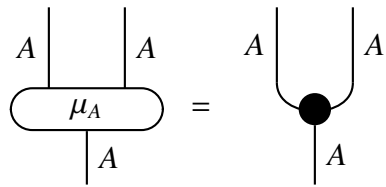
\begin{figure}[ht]
  \centering
  \tikzset{darkdot/.style={shape=circle,minimum width=0.15cm,fill=black,draw}}
  \begin{tikzpicture}[thick]
  \begin{scope}
  \draw (0,3) -- node[left] {$A$} (0,1.93);
  \draw (1,3) -- node[right] {$A$} (1,1.93);
  \node[func] (d1) at (0.5,1.65) {$\mu_A$};
  \draw (d1) -- node [right] {$A$} (0.5,0.65);

  \node at (2,1.65) {$=$};

  \draw (3,3) -- node[left] {$A$} (3,2);
  \draw (4,3) -- node[right] {$A$} (4,2);
  \draw (3,2) .. controls +(0,-.5) and +(0,-.5) .. (4,2);
  \node[darkdot] (d2) at (3.5,1.65) {};
  \draw (d2) -- node [right] {$A$} (3.5,0.65);
  \end{scope}
\end{tikzpicture}
\caption{Diagrammatic representation of $\mu$ morphisms.}
\label{fig:mumap}
\end{figure}

To deal with the substitution morphisms, I suggest borrowing a particular map from Frobenius Algebras represented in categories as done in \cite{Kartsaklis2012}, defined as follows for any object $A$ in our category:
\[
\mu_A: A \otimes A \to A
\]
This map effectively works like a cup, cancelling out some of the information provided as input, but instead of completely eliminating the input information, it preserves some of it and outputs it. Diagrammatically, it is represented as shown in Figure~\ref{fig:mumap}. Using this morphism, we can diagrammatically show the inner structure for the substitution morphisms initially shown in Figures~\ref{fig:lsubdiagsimple} and~\ref{fig:rsubdiagsimple}, as demonstrated in Figures~\ref{fig:lsubdiags} and~\ref{fig:rsubdiags}.

\begin{figure}[ht]
  \centering
  \tikzset{func/.style={shape=rectangle,rounded corners=8,minimum width=2cm,minimum height=.5cm,draw}}
  \tikzset{claspnode/.style={shape=circle,minimum width=0.25cm,fill=white,draw}}
  \tikzset{wideclasp/.style={shape=rectangle, rounded corners=3,minimum width=1.3cm,fill=white,draw}}
  \tikzset{darkdot/.style={shape=circle,minimum width=0.15cm,fill=black,draw}}
  \begin{tikzpicture}[thick]
    \begin{scope}
    \draw[-] (0,6) -- (0,5);
    \draw[->] (1,5) -- (1,6);
    \draw[->] (0,5) -- node[left] {$A\ $} (0,3);
    \draw[-] (1,5) -- node[right] {$B$} (1,3);
    \draw[->] (2,4) -- node[right] {$C$} (2,6);
    \draw (2,4) -- (2,3);
    \node[claspnode] (c1) at (0,4) {};
    \draw (c1) -- (1,4);
    \node[wideclasp] (w1) at (.5,5) {};
    \draw (w1) -- (2,5);

    \draw[->] (3,6) -- node[left] {$B\ $} (3,4);
    \draw (3,4) -- (3,3);
    \draw[->] (4,4) -- node[right] {$C$} (4,6);
    \draw (4,4) -- (4,3);
    \node[claspnode] (c2) at (3,5) {};
    \draw (c2) -- (4,5);

    \draw (3,3) .. controls +(0,-.5) and +(0,.5) .. (4, 2);
    \draw (4,3) .. controls +(0,-.5) and +(0,.5) .. (3, 2);

    \draw (2,3) -- (2,2);
    \draw (2,2) .. controls +(0,-.5) and +(0,-.5) .. (3,2);
    \node[darkdot] (d1) at (2.5,1.6) {};
    \draw (d1) -- (2.5,1);

    \draw (4,2) -- (4,1);
    \draw (4,1) .. controls +(0,-.75) and +(0,1) .. (2.5, 0);
    \draw (2.5,1) .. controls +(0,-.75) and +(0,1) .. (4, 0);

    \draw (0,3) -- (0,0);
    \draw (1,3) -- (1,0);
    \draw (0,0) .. controls +(0,-1) and +(0,1) .. (2,-2);
    \draw (4,0) .. controls +(0,-1) and +(0,1) .. (3,-2);
    \draw (1,0) .. controls +(0,-.75) and +(0,-.75) .. (2.5,0);

    \draw[->] (2,-2) -- node[left] {$A\ $} (2,-4);
    \draw[<-] (3,-2) -- node[right] {$C$} (3,-4);
    \node[claspnode] (c3) at (2,-3) {};
    \draw (c3) -- (3,-3);
    \draw (2,1) circle (2.7);
    \end{scope}

    \begin{scope}[xshift=7cm]
    \draw[-] (1,6) -- (1,5);
    \draw[->] (2,5) -- (2,6);
    \draw[->] (1,5) -- node[left] {$A\ $} (1,3);
    \draw[-] (2,5) -- node[right] {$B$} (2,3);
    \draw[->] (0,4) -- node[left] {$C$} (0,6);
    \draw (0,4) -- (0,3);
    \node[claspnode] (c1) at (1,4) {};
    \draw (c1) -- (2,4);
    \node[wideclasp] (w1) at (1.5,5) {};
    \draw (w1) -- (0,5);

    \draw[->] (4,6) -- node[right] {$\ B$} (4,4);
    \draw (3,4) -- (3,3);
    \draw[->] (3,4) -- node[left] {$C$} (3,6);
    \draw (4,4) -- (4,3);
    \node[claspnode] (c2) at (4,5) {};
    \draw (c2) -- (3,5);

    \draw (0,3) .. controls +(0,-0.5) and +(0,.5) .. (2,2);
    \draw (1,3) .. controls +(0,-0.5) and +(0,.5) .. (0,2);
    \draw (2,3) .. controls +(0,-0.5) and +(0,.5) .. (1,2);
    
    \draw (3,3) -- (3,2);
    \draw (2,2) .. controls +(0,-.5) and +(0,-.5) .. (3,2);
    \node[darkdot] (d1) at (2.5,1.6) {};
    \draw (d1) -- (2.5,1);

    \draw (4,3) -- (4,1);
    \draw (2.5,1) .. controls +(0,-.75) and +(0,.75) .. (4,0);
    \draw (4,1) .. controls +(0,-.75) and +(0,.75) .. (2.5,0);

    \draw (0,2) -- (0,0);
    \draw (1,2) -- (1,0);

    \draw (1,0) .. controls +(0,-.75) and +(0,-.75) .. (2.5,0);

    \draw (0,0) .. controls +(0,-1) and +(0,1) .. (3,-2);
    \draw (4,0) .. controls +(0,-1) and +(0,1) .. (2,-2);

    \draw[->] (3,-2) -- node[right] {$\ A$} (3,-4);
    \draw[<-] (2,-2) -- node[left] {$C$} (2,-4);
    \node[claspnode] (c3) at (3,-3) {};
    \draw (c3) -- (2,-3);

    \draw (2,1) circle (2.7);

    \end{scope}
  \end{tikzpicture}
  \caption{Expanded diagrammatic form of $lsub$ morphisms.}
  \label{fig:lsubdiags}
\end{figure}

\begin{figure}[ht]
  \centering
  \tikzset{func/.style={shape=rectangle,rounded corners=8,minimum width=2cm,minimum height=.5cm,draw}}
  \tikzset{claspnode/.style={shape=circle,minimum width=0.25cm,fill=white,draw}}
  \tikzset{wideclasp/.style={shape=rectangle, rounded corners=3,minimum width=1.3cm,fill=white,draw}}
  \tikzset{darkdot/.style={shape=circle,minimum width=0.15cm,fill=black,draw}}
  \begin{tikzpicture}[thick]
  \begin{scope}
    \draw[-] (3,6) -- (3,5);
    \draw[->] (2,5) -- (2,6);
    \draw[->] (3,5) -- node[right] {$\ A$} (3,3);
    \draw[-] (2,5) -- node[left] {$B$} (2,3);
    \draw[->] (4,4) -- node[right] {$C$} (4,6);
    \draw (4,4) -- (4,3);
    \node[claspnode] (c1) at (3,4) {};
    \draw (c1) -- (2,4);
    \node[wideclasp] (w1) at (2.5,5) {};
    \draw (w1) -- (4,5);

    \draw[->] (0,6) -- node[left] {$B\ $} (0,4);
    \draw (0,4) -- (0,3);
    \draw[->] (1,4) -- node[right] {$C$} (1,6);
    \draw (1,4) -- (1,3);
    \node[claspnode] (c2) at (0,5) {};
    \draw (c2) -- (1,5);

    \draw (4,3) .. controls +(0,-0.5) and +(0,.5) .. (2,2);
    \draw (2,3) .. controls +(0,-0.5) and +(0,.5) .. (3,2);
    \draw (3,3) .. controls +(0,-0.5) and +(0,.5) .. (4,2);

    \draw (1,3) -- (1,2);
    \draw (1,2) .. controls +(0,-.5) and +(0,-.5) .. (2,2);
    \node[darkdot] (d1) at (1.5,1.6) {};
    \draw (d1) -- (1.5,1);

    \draw (0,3) -- (0,1);
    \draw (1.5,1) .. controls +(0,-.75) and +(0,.75) .. (0,0);
    \draw (0,1) .. controls +(0,-.75) and +(0,.75) .. (1.5,0);

    \draw (3,2) -- (3,0);
    \draw (4,2) -- (4,0);

    \draw (3,0) .. controls +(0,-.75) and +(0,-.75) .. (1.5,0);

     \draw (0,0) .. controls +(0,-1) and +(0,1) .. (2,-2);
     \draw (4,0) .. controls +(0,-1) and +(0,1) .. (1,-2);

    \draw[->] (1,-2) -- node[left] {$A\ $} (1,-4);
    \draw[<-] (2,-2) -- node[right] {$C$} (2,-4);
    \node[claspnode] (c3) at (1,-3) {};
    \draw (c3) -- (2,-3);
    \draw (2,1) circle (2.7);
    \end{scope}

    \begin{scope}[xshift=7cm]
    \draw[-] (4,6) -- (4,5);
    \draw[->] (3,5) -- (3,6);
    \draw[->] (4,5) -- node[right] {$\ A$} (4,3);
    \draw[-] (3,5) -- node[left] {$B$} (3,3);
    \draw[->] (2,4) -- node[left] {$C$} (2,6);
    \draw (2,4) -- (2,3);
    \node[claspnode] (c1) at (4,4) {};
    \draw (c1) -- (3,4);
    \node[wideclasp] (w1) at (3.5,5) {};
    \draw (w1) -- (2,5);

    \draw[->] (1,6) -- node[right] {$\ B$} (1,4);
    \draw (0,4) -- (0,3);
    \draw[->] (0,4) -- node[left] {$C$} (0,6);
    \draw (1,4) -- (1,3);
    \node[claspnode] (c2) at (1,5) {};
    \draw (c2) -- (0,5);

    \draw (0,3) .. controls +(0,-.5) and +(0,.5) .. (1, 2);
    \draw (1,3) .. controls +(0,-.5) and +(0,.5) .. (0, 2);

    \draw (2,3) -- (2,2);
    \draw (2,2) .. controls +(0,-.5) and +(0,-.5) .. (1,2);
    \node[darkdot] (d1) at (1.5,1.6) {};
    \draw (d1) -- (1.5,1);

    \draw (0,2) -- (0,1);
    \draw (0,1) .. controls +(0,-.75) and +(0,1) .. (1.5, 0);
    \draw (1.5,1) .. controls +(0,-.75) and +(0,1) .. (0, 0);

    \draw (3,3) -- (3,0);
    \draw (4,3) -- (4,0);
    \draw (4,0) .. controls +(0,-1) and +(0,1) .. (2,-2);
    \draw (0,0) .. controls +(0,-1) and +(0,1) .. (1,-2);
    \draw (3,0) .. controls +(0,-.75) and +(0,-.75) .. (1.5,0);

    \draw[->] (2,-2) -- node[right] {$\ A$} (2,-4);
    \draw[<-] (1,-2) -- node[left] {$C$} (1,-4);
    \node[claspnode] (c3) at (2,-3) {};
    \draw (c3) -- (1,-3);
    \draw (2,1) circle (2.7);
    \end{scope}
    
  \end{tikzpicture}
  \caption{Expanded diagrammatic form of $rsub$ morphisms.}
  \label{fig:rsubdiags}
\end{figure}

So to summarise what we have seen here, I hypothesise that closed monoidal categories with an additional set of $\mu$ morphisms model those composition operations in CCGs which we discussed here. I say `hypothesise' for two reasons. Firstly, because the inner structure of the cross composition and substitution morphisms presented here is obtained through `diagrammatic reverse-engineering', rather than through any real appeal to the algebraic structure of CCGs (as there is not a tidy definition of this structure, especially compared to Pregroup Grammars or Lambek Grammars). As such, future work should seek to determine whether or not the inner structure of these morphisms presented here makes sense from an algebraic---or indeed a linguistic---standpoint to confirm or reject this hypothesis. Secondly, this is a hypothesis because it does not fully capture all the variants of CCGs present in the literature. Earlier in this section, I talked about optional restrictions placed on rules such as composition and cross composition. Other work in the CCG literature (e.g.~\cite{Baldridge2003}) goes further in suggesting that the slashes should be augmented with compositional modalities, and that restrictions should apply to most rules based on the modalities of the slashes involved. For both cases, there is no obvious way of modelling restrictions in closed monoidal categories. I leave it to further work to determine whether or not this matters, and if so, how these aspects of CCG can be incorporated into our categorical discourse. 

An additional topic for further work on this subject would be to check that no issues arise from the introduction of swap morphisms into this category. In \cite{van1988semantics,van1995language}, van Benthem shows that associative and commutative extensions of the Lambek calculus underlying Lambek grammar recognises/generates all permutation closures of context-free languages, which makes it unsuitable for syntactic analysis. In practical terms, this means that the grammar will overgenerate, recognising sentences as grammatical which are not meant to be recognised as such. This `overgenerative' property also holds for the categorical representation of CCG presented above, for the case of unrestricted cross-composition morphisms. Insofar as this is an issue for CCG as a whole, rather than the categorical representation, I will consider this outside of the scope of this thesis. However, this strongly supports the case for further investigation of how to include rule restrictions into the categorical representation of CCGs.
% subsection categorical_semantics_for_ccgs (end)

\subsection{Defining a Functor} % (fold)
\label{sub:defining_a_functor_ccg}

To complete this section, I will briefly present how a functorial passage from the closed monoidal category $\mathbf{CCG}$ defined here to the category of vector spaces $\mathbf{FVect}$ can be defined. Because of the similarity between $\mathbf{LG}$ and $\mathbf{CCG}$, we can re-use substantial parts of the work presented in $\S$\ref{sub:defining_a_functor_lg} here. 

Let $F$ be a functor between $\mathbf{CCG}$ and $\mathbf{FVect}$. We begin by assigning each atomic type $A$ in the CCG to a vector space $V = F(A)$. We then define the compound objects as was done in $\mathbf{LG}$:
\begin{align*}
    & F(A \otimes B) = F(A) \otimes F(B)\\
    & F(A \multimap B) = F(A) \otimes F(B)\\
    & F(A \multimapinv B) = F(A) \otimes F(B)
\end{align*}
The functors mapping those composition operations in CCGs that are also in LGs are exactly as defined for the case of a functor between $\mathbf{LG}$ and $\mathbf{FVect}$, and are as follows:
\begin{align*}
    & F(ev^l_{A,B}) = \epsilon_{F(A)} \otimes 1_{F(B)} : F(A) \otimes F(A) \otimes F(B) \to F(B)\\
    & F(ev^r_{A,B}) = 1_{F(B)} \otimes \epsilon_{F(A)} : F(B) \otimes F(A) \otimes F(A) \to F(B)\\
    & F(comp^r_{A,B,C}) =  1_{F(A)} \otimes \epsilon_{F(B)} \otimes 1_{F(C)} : F(A) \otimes F(B) \otimes F(B) \otimes F(C) \to F(A) \otimes F(C)\\
    & F(comp^l_{A,B,C}) = 1_{F(A)} \otimes \epsilon_{F(B)} \otimes 1_{F(C)} : F(A) \otimes F(B) \otimes F(B) \otimes F(C) \to F(A) \otimes F(C)\\
    & F(raise^l_{A,B}) = (\eta_{F(B)} \otimes 1_{F(A)}): F(A) \to F(B) \otimes F(B) \otimes F(A)\\
    & F(raise^r_{A,B}) = (1_{F(A)} \otimes \eta_{F(B)}): F(A) \to F(A) \otimes F(B) \otimes F(B)\\
\end{align*}
This leaves us only with the task of defining a functorial passage for the cross composition and substitution morphisms. We begin with the internal structure of the cross composition morphisms, which we can read from the diagrams:
\begin{align*}
     & F(cross^r_{A,B,C}) = F(comp^l_{A,B,C} \circ (1_{A \multimapinv B} \otimes lswap_{A,B})) \\
    & \qquad = (1_{F(A)} \otimes \epsilon_{F(B)} \otimes 1_{F(C)}) \circ (1_{F(A)} \otimes 1_{F(B)} \otimes F(lswap_{A,B}))\\
    & F(cross^l_{A,B,C}) = F(comp^r_{A,B,C} \circ (rswap_{A,B}) \otimes 1_{B \multimap A} )\\
    & \qquad = (1_{F(A)} \otimes \epsilon_{F(B)} \otimes 1_{F(C)}) \circ (F(rswap_{A,B}) \otimes 1_{F(A)} \otimes 1_{F(B)})\\
\end{align*}
We know how the functorial passage for every element of the above expressions above is defined, except for the swap morphisms. To complete the functorial passage for the cross composition morphisms, we must therefore describe what the swap morphisms correspond to in $\mathbf{FVect}$. As $\mathbf{FVect}$ is a symmetric monoidal category, there is an isomorphism $swap_{A,B}$ for all objects $A$ and $B$ in $ob(\mathbf{FVect})$ defined as follows:
\[
swap_{A,B}: A \otimes B \to B \otimes A
\]
This is the obvious target for the swap morphisms in $\mathbf{CCG}$, and hence:
\[
  F(lswap_{A,B}) = F(rswap_{A,B}) = swap_{F(A),F(B)}
\]
Linear algebraically, this morphism of $\mathbf{FVect}$ corresponds to the permutation of tensor indices. The simplest and most intuitive case to illustrate this is that of applying $swap(A,B)$ to a matrix in $A \otimes B$, which yields a matrix in $B \otimes A$ which is the transpose of the original matrix.

So we can rewrite the functorial definition for the cross composition morphisms as follows:
\begin{align*}
     & F(cross^r_{A,B,C}) = (1_{F(A)} \otimes \epsilon_{F(B)} \otimes 1_{F(C)}) \circ (1_{F(A)} \otimes 1_{F(B)} \otimes swap_{F(A),F(B)})\\
    & F(cross^l_{A,B,C}) = (1_{F(A)} \otimes \epsilon_{F(B)} \otimes 1_{F(C)}) \circ (swap_{F(A),F(B)} \otimes 1_{F(A)} \otimes 1_{F(B)})\\
\end{align*}

With this in mind, we turn to the functorial passage for substitution morphisms. First for the $lsub$ morphisms:
\begin{align*}
    & F(lsub^l_{A,B,C}) = F(comp^l_{A,B,C} \circ (1_{A \multimapinv B} \otimes lswap_{B,C}) \circ (1_{A \multimapinv B} \otimes \mu_{C} \otimes 1_B) \\
    & \qquad\qquad \circ (1_{(A \multimapinv B) \multimapinv C} \otimes rswap_{B,C}))\\
    &\qquad = (1_{F(A)} \otimes \epsilon_{F(B)} \otimes 1_{F(C)}) \circ (1_{F(A)} \otimes 1_{F(B)} \otimes swap_{F(A),F(B)})\\ 
    & \qquad\qquad \circ (1_{F(A)} \otimes 1_{F(B)} \otimes F(\mu_A) \otimes 1_{F(B)}) \circ (1_{F(A)} \otimes 1_{F(B)} \otimes 1_{F(C)} \otimes swap_{F(B),F(C)})\\
    & F(lsub^r_{A,B,C}) = F(rswap_{A,C} \circ comp^l_{A,B,C} \circ (1_{A \multimapinv B} \otimes lswap_{B,C}) \circ (1_{A \multimapinv B} \otimes \mu_C \otimes 1_B)\\
    & \qquad\qquad \circ (lswap_{A \multimapinv B,C} \otimes 1_{C \multimap B}))\\
    & \qquad = swap_{F(A),F(C)} \circ (1_{F(A)} \otimes \epsilon_{F(B)} \otimes 1_{F(C)}) \circ (1_{F(A)} \otimes 1_{F(B)} \otimes F(\mu_C) \otimes 1_{F(B)})\\
    & \qquad\qquad \circ (swap_{F(A) \otimes F(B), F(C)} \otimes 1_{F(C)} \otimes 1_{F(B)}) \\
\end{align*}
Next, for the $rsub$ morphisms:
\begin{align*}
    & F(rsub^l_{A,B,C}) = F(lswap_{A,C} \circ comp^r_{C,B,A} \circ (rswap_{C,B} \otimes 1_{B \multimap A}) \circ (1_{B} \otimes \mu_C \otimes 1_{B \multimap A})\\
    & \qquad\qquad \circ (1_{B \multimapinv C} \otimes rswap_{C,B \multimap A})) \\
    & \qquad = swap_{F(A),F(C)} \circ (1_{F(C)} \otimes \epsilon_{F(B)} \otimes 1_{F(A)}) \circ (swap_{F(C),F(B)} \otimes 1_{F(B)} \otimes 1_{F(A)}) \\
    & \qquad\qquad \circ (1_{F(B)} \otimes F(\mu_C) \otimes 1_{F(B)} \otimes 1_{F(A)}) \circ (1_{F(B)} \otimes 1_{F{C}} \otimes swap_{F(C),F(B) \otimes F(A)})\\
    & F(rsub^r_{A,B,C}) = F(comp^r_{C,B,A} \circ (rswap_{C,B} \otimes 1_{B \multimap A}) \circ (1_B \otimes \mu_C \otimes 1_{B \multimap A})\\
    & \qquad\qquad \circ (lswap_{B,C} \otimes 1_{C \multimap (B\multimap A)})) \\
    & \quad = (1_{F(C)} \otimes \epsilon_{F(B)} \otimes 1_{F(A)}) \circ (swap_{F(C),F(B)} \otimes 1_{F(B)} \otimes 1_{F(A)})\\
    & \qquad\qquad \circ (1_{F(B)} \otimes F(\mu_C) \otimes 1_{F(B)} \otimes 1_{F(A)}) \circ (swap_{F(B),F(C)} \otimes 1_{F(C)} \otimes 1_{F(B)} \otimes 1_{F(A)} )  \\  
\end{align*}
The remaining unknown here is now the functorial passage for $\mu$ maps. An interpretation of such a Frobenius operation is provided in \cite{Kartsaklis2012}, and should suit our needs. In $\mathbf{FVect}$ there are morphisms $\mu_A$ for any object $A$, defined as follows:
\[
\mu_A : A \otimes A \to A :: \overrightarrow{v} \otimes \overrightarrow{w} \mapsto \overrightarrow{v} \odot \overrightarrow{w}
\]
So we can define the functorial passage for $\mu$ maps in $hom(\mathbf{CCG})$ as follows:
\[
F(\mu_A) = \mu_{F(A)}
\]
Using this, we can rewrite the functorial passage for $lswap$ morphisms as follows:
\begin{align*}
    & F(lsub^l_{A,B,C}) = (1_{F(A)} \otimes \epsilon_{F(B)} \otimes 1_{F(C)}) \circ (1_{F(A)} \otimes 1_{F(B)} \otimes swap_{F(A),F(B)})\\ 
    & \qquad\qquad \circ (1_{F(A)} \otimes 1_{F(B)} \otimes \mu_{F(A)} \otimes 1_{F(B)}) \circ (1_{F(A)} \otimes 1_{F(B)} \otimes 1_{F(C)} \otimes swap_{F(B),F(C)})\\
    & F(lsub^r_{A,B,C}) = swap_{F(A),F(C)} \circ (1_{F(A)} \otimes \epsilon_{F(B)} \otimes 1_{F(C)}) \circ (1_{F(A)} \otimes 1_{F(B)} \otimes \mu_{F(C)} \otimes 1_{F(B)})\\
    & \qquad\qquad \circ (swap_{F(A) \otimes F(B), F(C)} \otimes 1_{F(C)} \otimes 1_{F(B)}) \\
\end{align*}
And the functorial passage for $rswap$ morphisms as follows:
\begin{align*}
    & F(rsub^l_{A,B,C}) = swap_{F(A),F(C)} \circ (1_{F(C)} \otimes \epsilon_{F(B)} \otimes 1_{F(A)}) \circ (swap_{F(C),F(B)} \otimes 1_{F(B)} \otimes 1_{F(A)}) \\
    & \qquad\qquad \circ (1_{F(B)} \otimes \mu_{F(C)} \otimes 1_{F(B)} \otimes 1_{F(A)}) \circ (1_{F(B)} \otimes 1_{F{C}} \otimes swap_{F(C),F(B) \otimes F(A)})\\
    & F(rsub^r_{A,B,C}) = (1_{F(C)} \otimes \epsilon_{F(B)} \otimes 1_{F(A)}) \circ (swap_{F(C),F(B)} \otimes 1_{F(B)} \otimes 1_{F(A)})\\
    & \qquad\qquad \circ (1_{F(B)} \otimes \mu_{F(C)} \otimes 1_{F(B)} \otimes 1_{F(A)}) \circ (swap_{F(B),F(C)} \otimes 1_{F(C)} \otimes 1_{F(B)} \otimes 1_{F(A)} )  \\  
\end{align*}
This completes the functor definition, and this section.

% subsection defining_a_functor_ccg (end)

% section supporting_combinatory_categorial_grammar (end)

\section{Next Steps} % (fold)
\label{sec:further_work}

Throughout this chapter, we have seen three areas in which further progress has been established based on the foundations provided by the earlier parts of this thesis. First, I discussed the difficulties behind integrating logical aspects of natural language into the framework, and the problems that arise from our reliance on multilinear maps within the DisCoCat compositional formalism. Second, I discussed ways in which machine learning methods could be integrated into the framework to replace or supplement the learning methods I had initially developed with Mehrnoosh Sardzadeh. Third, I discussed how grammatical formalisms with a more complex structure than those previously examined, and with different linguistic and grammatical properties, could possibly be integrated into the DisCoCat formalism. To conclude this chapter, I offer some thoughts on how these three areas can form the foundations for future work, and what issues other researchers in the field may wish to examine.

In $\S$\ref{sec:distributional_logic}, I showed how aspects of formal semantics and predicate logic could be simulated within a tensor-based compositional framework. In some cases, such as logical connectives, the possibility of modelling operations using multilinear maps depended on the shape of the sentence space. While I showed that some sentence spaces (e.g.~$S=B_2$) did allow multilinear maps to model connectives, it was not obvious how quantification could be modelled without appealing to non-linear maps. Researchers wishing to develop this aspect of the DisCoCat framework would therefore do well to consider how this obstacle may be surpassed using categorical logic, or by changing the semantic model from $\mathbf{FVect}$ to some other form of semantic representation which would naturally allow the integration of non-linearity into the formalism.

A related problem is that of how to model logical operations in non-truth-theoretic semantic representations, such as those developed in Chapter~\ref{cha:learning_procedures_for_a_discocat}. While I have no specific suggestions concerning where to begin addressing this problem, it should be fairly clear from my attempt to simulate predicate logic that the nature and structure of the sentence space $S$ will crucially affect the nature of logical operations. It is possible to simulate predicate logic precisely because we can interpret basis elements of the sentence space as truth values; therefore when considering how logical connectives and quantification would operate over other kinds of values, we must first consider how to interpret the elements of the sentence space in linguistically or logically sensible ways. Arguably, what is lacking in our current non-truth-theoretic models is such an interpretation, and therefore researchers interested in further developing this topic may wish to begin by re-thinking what, precisely, it is that we are representing in our sentence vectors.

In $\S$\ref{sec:learning_tensors_by_multi_step_regression}, I discussed joint work performed with Georgiana Dinu, Yao-Zhong Zhang, Mehrnoosh Sadrzadeh and Marco Baroni, in which we showed that sophisticated machine learning methods could be adapted to learn the semantic representations used in the DisCoCat framework from data. As a by-product, this approach also addressed the problem present in the approach presented in Chapter~\ref{cha:learning_procedures_for_a_discocat} whereby sentences with verbs of different valency as their head would reside in different sentence spaces. Continuing to develop a data-driven machine-learning approach to learning semantic representations should constitute an important future research direction for this formalism, as we move away from the task of simply evaluating it against simple experiments, and towards the task of applying it to concrete problems such as machine translation evaluation and machine translation itself, paraphrase detection, information retrieval, sentiment analysis, or any other text-based problems which would benefit from including deeper semantic knowledge into solutions, as is being done in a variety of work produced since the writing of this thesis (e.g.~\cite{hermann2013role,socherEMNLP12}).

Finally, in $\S$\ref{sec:supporting_combinatory_categorial_grammar}, I laid down foundations for integrating CCGs into the DisCoCat formalism, with the hope that fully integrating variants of CCGs into the formalism will be taken up as a subject for further work. This is not only useful because of the powerful parsing tools available for this grammatical formalism, but also for the rich linguistic aspects (cf.~\cite{Steedman2009}) underlying this class of grammars. One of the main benefits of the categorical approach behind DisCoCat is that it allows information to pass between syntactic and semantic structures. Investigating how to exploit the expressive nature of CCGs to represent more sophisticated linguistic phenomena in our semantic models should be both interesting and help give us a better idea about exactly what semantic properties we are modelling, and how our semantic representations may need to evolve to cater to the linguistic properties CCGs address.
% section further_work (end)

% chapter applications_and_future_work (end)

\mbox{}
\newpage
%!TEX root = ../grefenstettethesis.tex

\chapter{Conclusions} % (fold)
\label{cha:conclusions}

To conclude this thesis, let me begin by summarising the work presented throughout this document. In Chapter~\ref{cha:literature_review}, I provided a critical overview of the some attempts, new and old, to address the problem of compositionality in distributional models of semantics. We observed a tension between the need to produce tractable representations which could easily be learned and compared, and the need to integrate syntactic, relational, or knowledge-based information into both our semantic representations and our compositional process.

In Chapter~\ref{cha:foundations_of_discocat}, I reported a recent effort to produce a general framework, DisCoCat, within which syntax and semantics are put into relation in order to produce compositional distributional models of semantics that would take grammatical structure into account both in the semantic representation of words, and into how they are composed to produce representations of sentences. This established the foundation on which the rest of the thesis was built, with the goal of showing that this framework could indeed produce high quality, learnable models of compositional distributional semantics.

In Chapter~\ref{cha:syntactic_extensions}, I presented various syntactic extensions to the DisCoCat framework, allowing us to use a wider variety of grammatical formalisms as part of the creation of compositional distributional models of semantics. I also described a procedure by which grammatical formalisms not described in this document could be integrated into the framework, by giving them a categorical representation and defining a functorial passage between such a representation and the categorical representation of whichever semantic model we choose to use.

In Chapter~\ref{cha:learning_procedures_for_a_discocat}, I presented a learning procedure for the production of semantic representations to be used within the DisCoCat formalism. This procedure allows us to develop concrete models from the abstract framework. I also discussed a reduced representation for semantic objects used in such models, which allows us to efficiently compute sentence representations. Finally, I introduced a variant on this learning procedure for reduced representations, based on the Kronecker product of lexical vectors, and showed that reduced representations can generally be viewed as embedded within the full representation of semantic relations, leading to a reduction in computational complexity without modifying the mathematical nature of composition operations.

In Chapter~\ref{cha:evaluating_a_discocat}, I evaluated the concrete models from the previous chapter in the context of three phrase similarity detection experiments, against other leading models discussed earlier in the thesis. I showed that the models produced by the DisCoCat framework match or outperform competing models, and that the difference between DisCoCat models and rivals grows with the complexity of the sentences used in the experiments.

In Chapter~\ref{cha:applications_and_future_work}, I presented three further extensions to the work done in the rest of the thesis. I discussed options for integrating logical operations into distributional compositional semantic models, and outlined the difficulties and obstacles faced by the DisCoCat framework in trying to accommodate such operations. I presented a new machine learning algorithm developed with colleagues, which improved upon certain aspects of the learning procedures presented earlier in this thesis. I also discussed foundations for the inclusion of Combinatory Categorial Grammars into the DisCoCat framework, and surveyed some of the difficulties faced by trying to fully integrate all variants of such grammars into our categorical formalism. I concluded the chapter by providing suggestions as to future directions research on the topic of categorical compositional distributional models of semantics might take.

Throughout this thesis, I have shown that the DisCoCat framework is not a theoretical toy example of how category theory can be applied to problems in linguistics. Indeed, not only can learning procedures be developed to generate concrete models from it, but the category theoretic aspects themselves provide powerful tools when it comes to the expansion of the framework itself. I discussed how such expansions could take place in the form of integrating new syntactic formalisms into the framework, but also suggested that the categorical properties of the framework may provide ways to surpass the limitations of our semantic representations when it comes to modelling logical aspects of language. In discussing both its potential for expansion and showing its ability to admit machine-learning-inspired learning algorithms, I hope to have demonstrated that the categorical approach to linguistics walks the path between theory and practice. On these grounds, I am convinced that it will constitute an important and fascinating area for future research, as others have already noted \cite{HeunenSadrGref2013}.
% chapter conclusions (end)

\singlespacing
\addcontentsline{toc}{chapter}{Bibliography}
\bibliography{grefenstettethesis}        %use a bibtex bibliography file refs.bib

\begin{thebibliography}{10}

\bibitem{abramsky2004categorical}
S.~Abramsky and B.~Coecke.
\newblock A categorical semantics of quantum protocols.
\newblock In {\em Logic in Computer Science, 2004. Proceedings of the 19th
  Annual IEEE Symposium on}, pages 415--425. IEEE, 2004.

\bibitem{Alshawi1992}
H.~Alshawi, editor.
\newblock {\em The Core Language Engine}.
\newblock MIT Press, 1992.

\bibitem{Baez2011}
J.~Baez and M.~Stay.
\newblock Physics, topology, logic, and computation: A rosetta stone.
\newblock In B.~Coecke, editor, {\em New Structures in Physics}, volume 813 of
  {\em Lecture Notes in Physics}. Springer, 2011.

\bibitem{Baldridge2003}
J.~Baldridge and G.~J.~M. Kruijff.
\newblock Multi-modal combinatory categorial grammar.
\newblock In {\em Proceedings of the tenth conference of the European chapter
  of the Association for Computational Linguistics-Volume 1}, pages 211--218.
  Association for Computational Linguistics, 2003.

\bibitem{Baroni2010}
M.~Baroni and R.~Zamparelli.
\newblock Nouns are vectors, adjectives are matrices: Representing
  adjective-noun constructions in semantic space.
\newblock In {\em Proceedings of the 2010 Conference on Empirical Methods in
  Natural Language Processing}, pages 1183--1193. Association for Computational
  Linguistics, 2010.

\bibitem{Bourbaki:1989}
N.~Bourbaki.
\newblock {\em Commutative Algebra: Chapters 1-7}.
\newblock Springer-Verlag (Berlin and New York), 1989.

\bibitem{Bruni2012}
E.~Bruni, G.~Boleda, M~Baroni, and N.~K. Tran.
\newblock Distributional semantics in technicolor.
\newblock In {\em Proceedings of the 50th Annual Meeting of the Association for
  Computational Linguistics (Volume 1: Long Papers)}, pages 136--145, Jeju
  Island, Korea, July 2012. Association for Computational Linguistics.

\bibitem{buszkowski2001lambek}
W.~Buszkowski.
\newblock Lambek grammars based on pregroups.
\newblock {\em Logical aspects of computational linguistics}, pages 95--109,
  2001.

\bibitem{Buszkowski2008}
W.~Buszkowski and K.~Moroz.
\newblock Pregroup grammars and context-free grammars.
\newblock {\em Computational Algebraic Approaches to Natural Language,
  Polimetrica}, pages 1--21, 2008.

\bibitem{choi1975completely}
M.D. Choi.
\newblock Completely positive linear maps on complex matrices.
\newblock {\em Linear algebra and its applications}, 10(3):285--290, 1975.

\bibitem{chomsky1956three}
N.~Chomsky.
\newblock Three models for the description of language.
\newblock {\em Information Theory, IRE Transactions on}, 2(3):113--124, 1956.

\bibitem{Clarkbookchap2013}
S.~Clark.
\newblock {\em Type-Driven Syntax and Semantics for Composing Meaning Vectors}.
\newblock 2013.

\bibitem{Clark2008}
S.~Clark, B.~Coecke, and M.~Sadrzadeh.
\newblock A compositional distributional model of meaning.
\newblock In {\em Proceedings of the Second Quantum Interaction Symposium
  (QI-2008). College Publications}, 2008.

\bibitem{Clark2006}
S.~Clark and S.~Pulman.
\newblock {Combining symbolic and distributional models of meaning}.
\newblock In {\em AAAI Spring Symposium on Quantum Interaction}, 2006.

\bibitem{cocke1969programming}
J.~Cocke.
\newblock Programming languages and their compilers: Preliminary notes.
\newblock 1969.

\bibitem{coecke2006kindergarten}
B.~Coecke.
\newblock Kindergarten quantum mechanics: Lecture notes.
\newblock In {\em AIP Conference Proceedings}, volume 810, 2006.

\bibitem{Sadrzadeh2013}
B.~Coecke, E.~Grefenstette, and M.~Sadrzadeh.
\newblock Lambek vs. lambek: Functorial vector space semantics and string
  diagrams for lambek calculus.
\newblock {\em Annals of Pure and Applied Logic}, 2013.

\bibitem{coecke2009categories}
B.~Coecke and \'{E}.O. Paquette.
\newblock {Categories for the practising physicist}.
\newblock {\em Arxiv preprint arXiv:0905.3010}, 2009.

\bibitem{Coecke2010}
B.~Coecke, M.~Sadrzadeh, and S.~Clark.
\newblock {Mathematical Foundations for a Compositional Distributional Model of
  Meaning}.
\newblock March 2010.

\bibitem{curran-clark-bos:2007:PosterDemo}
J.~Curran, S.~Clark, and J.~Bos.
\newblock Linguistically motivated large-scale nlp with c\&c and boxer.
\newblock In {\em Proceedings of the 45th Annual Meeting of the Association for
  Computational Linguistics Companion Volume Proceedings of the Demo and Poster
  Sessions}, pages 33--36, Prague, Czech Republic, June 2007. Association for
  Computational Linguistics.

\bibitem{Curran2004}
J.~R. Curran.
\newblock {\em {From distributional to semantic similarity}}.
\newblock PhD thesis, 2004.

\bibitem{Curran2007}
J.R. Curran, S.~Clark, and J.~Bos.
\newblock Linguistically motivated large-scale nlp with c\&c and boxer.
\newblock In {\em Proceedings of the 45th Annual Meeting of the ACL on
  Interactive Poster and Demonstration Sessions}, pages 33--36. Association for
  Computational Linguistics, 2007.

\bibitem{Dean2008a}
J.~Dean and S.~Ghemawat.
\newblock {MapReduce: simplified data processing on large clusters}.
\newblock {\em Communications of the ACM}, 51(1):107--113, 2008.

\bibitem{Dinu2010}
G.~Dinu and M.~Lapata.
\newblock Measuring distributional similarity in context.
\newblock In {\em Proceedings of EMNLP}, pages 1162--1172, Cambridge, MA, 2010.

\bibitem{earley1970efficient}
J.~Earley.
\newblock An efficient context-free parsing algorithm.
\newblock {\em Communications of the ACM}, 13(2):94--102, 1970.

\bibitem{Erk2008}
K.~Erk and S.~Pad\'{o}.
\newblock {A structured vector space model for word meaning in context}.
\newblock {\em Proceedings of the Conference on Empirical Methods in Natural
  Language Processing - EMNLP '08}, (October):897, 2008.

\bibitem{finkelstein2001placing}
L.~Finkelstein, E.~Gabrilovich, Y.~Matias, E.~Rivlin, Z.~Solan, G.~Wolfman, and
  E.~Ruppin.
\newblock Placing search in context: The concept revisited.
\newblock In {\em Proceedings of the 10th international conference on World
  Wide Web}, pages 406--414. ACM, 2001.

\bibitem{Firth1957}
J.~R. Firth.
\newblock {A synopsis of linguistic theory 1930-1955}.
\newblock {\em Studies in linguistic analysis}, 1957.

\bibitem{fowler2009parsing}
T.~A.~D. Fowler.
\newblock Parsing ccgbank with the lambek calculus.
\newblock In {\em Parsing with Categorial Grammars Workshop ESSLLI 2009
  Bordeaux, France Book of Abstracts}, 2009.

\bibitem{fowler2010accurate}
T.~A.~D. Fowler and G.~Penn.
\newblock {Accurate context-free parsing with combinatory categorial grammar}.
\newblock In {\em Proceedings of the 48th Annual Meeting of the Association for
  Computational Linguistics}, pages 335--344. Association for Computational
  Linguistics, 2010.

\bibitem{Frege1892}
G.~Frege.
\newblock Uber sinn und bedeutung.
\newblock {\em Zeitschrift f{\"u}r Philosophie und philosophische Kritik},
  100(1):25--50, 1892.

\bibitem{Golub1979}
G.~H. Golub, M.~Heath, and G.~Wahba.
\newblock Generalized cross-validation as a method for choosing a good {R}idge
  parameter.
\newblock {\em Technometrics}, pages 215--223, 1979.

\bibitem{MScGrefenstette2009}
E.~Grefenstette.
\newblock {Analysing Document Similarity Measures}.
\newblock Master's thesis, University of Oxford, September 2009.

\bibitem{GrefenstetteTFDS}
E.~Grefenstette.
\newblock Towards a formal distributional semantics: Simulating logical calculi
  with tensors.
\newblock {\em Proceedings of the Second Joint Conference on Lexical and
  Computational Semantics}, 2013.

\bibitem{GrefSadrBarIWCS13}
E.~Grefenstette, G.~Dinu, Y.~Zhang, M.~Sadrzadeh, and M.~Baroni.
\newblock Multi-step regression learning for compositional distributional
  semantics.
\newblock {\em Proceedings of the 10th International Conference on
  Computational Semantics (IWCS 2013)}, 2013.

\bibitem{Grefenstette2011a}
E.~Grefenstette and M.~Sadrzadeh.
\newblock Experimental support for a categorical compositional distributional
  model of meaning.
\newblock In {\em Proceedings of the 2011 Conference on Empirical Methods in
  Natural Language Processing}, 2011.

\bibitem{Grefenstette2011b}
E.~Grefenstette and M.~Sadrzadeh.
\newblock Experimenting with transitive verbs in a discocat.
\newblock In {\em Proceedings of the 2011 EMNLP Workshop on Geometric Models of
  Natural Language Semantics}, 2011.

\bibitem{grefenstette2011concrete}
E.~Grefenstette, M.~Sadrzadeh, S.~Clark, B.~Coecke, and S.~Pulman.
\newblock Concrete sentence spaces for compositional distributional models of
  meaning.
\newblock In {\em Proceedings of the Ninth International Conference on
  Computational Semantics}, pages 125--134. Association for Computational
  Linguistics, 2011.

\bibitem{Grefenstette2010}
E.~Grefenstette, M.~Sadrzadeh, B.~Coecke, S.~Pulman, and S.~Clark.
\newblock {Concrete Compositional Sentence Spaces}.
\newblock {\em ESSLLI'10 Workshop on Compositionality and Distributional
  Semantic Models}, 2010.

\bibitem{Grefenstette1992}
G.~Grefenstette.
\newblock {Use of syntactic context to produce term association lists for text
  retrieval}.
\newblock In {\em Proceedings of the 15th annual international ACM SIGIR
  conference on Research and development in information retrieval}, pages
  89--97. ACM, 1992.

\bibitem{Grefenstette1994}
G.~Grefenstette.
\newblock {\em {Explorations in automatic thesaurus discovery}}.
\newblock 1994.

\bibitem{Guevara2010}
E.~R. Guevara.
\newblock {Modelling Adjective-Noun Compositionality by Regression}.
\newblock {\em ESSLLI'10 Workshop on Compositionality and Distributional
  Semantic Models}, 2010.

\bibitem{Hall2006}
J.~Hall.
\newblock {\em MaltParser: An Architecture for Labeled Inductive Dependency
  Parsing}.
\newblock Licentiate thesis, V\"axj\"o University, V\"axj\"o, Sweden, 2006.

\bibitem{Harris1968}
Z.~S. Harris.
\newblock {\em {Mathematical structures of language}}.
\newblock Wiley, 1968.

\bibitem{Hastie2009}
T.~Hastie, T.~Tibshirani, and J.~Friedman.
\newblock {\em The Elements of Statistical Learning, 2nd ed.}
\newblock Springer, New York, 2009.

\bibitem{hermann2013role}
K.M. Hermann and P.~Blunsom.
\newblock The role of syntax in vector space models of compositional semantics.
\newblock {\em Proceedings of ACL, Sofia, Bulgaria, August. Association for
  Computational Linguistics}, 2013.

\bibitem{HeunenSadrGref2013}
C.~Heunen, M.~Sadrzadeh, and E.~Grefenstette, editors.
\newblock {\em Quantum Physics and Linguistics: A Compositional, Diagrammatic
  Discourse}.
\newblock Oxford University Press, 2013.

\bibitem{hindley1969principal}
R.~Hindley.
\newblock The principal type-scheme of an object in combinatory logic.
\newblock {\em Transactions of the american mathematical society}, 146:29--60,
  1969.

\bibitem{Kartsaklis2012}
D.~Kartsaklis, M.~Sadrzadeh, and S.~Pulman.
\newblock A unified sentence space for categorical distributional-compositional
  semantics: Theory and experiments.
\newblock In {\em Proceedings of 24th International Conference on Computational
  Linguistics (COLING 2012): Posters}, pages 549--558, Mumbai, India, December
  2012.

\bibitem{kasami1965efficient}
T.~Kasami.
\newblock An efficient recognition and syntax analysis algorithm for
  context-free languages.
\newblock Technical report, Air Force Cambridge Research Laboratory, 1965.

\bibitem{Lambek1958}
J.~Lambek.
\newblock {The Mathematics of Sentence Structure}.
\newblock {\em The American Mathematical Monthly}, 65(3):154 -- 170, 1958.

\bibitem{Lambek1999}
J.~Lambek.
\newblock {Type grammar revisited}.
\newblock {\em Logical aspects of computational linguistics}, 1999.

\bibitem{Lambek2008}
J.~Lambek.
\newblock {\em {From word to sentence. A computational algebraic approach to
  grammar}}.
\newblock Polimetrica, 2008.

\bibitem{Lambek2010}
J.~Lambek.
\newblock {\em {Compact Monoidal Categories from Linguistics to Physics}},
  pages 451--469.
\newblock 2010.

\bibitem{Landauer1997}
T.~K. Landauer and S.~T. Dumais.
\newblock {A solution to Plato's problem: The latent semantic analysis theory
  of acquisition, induction, and representation of knowledge.}
\newblock {\em Psychological review}, 1997.

\bibitem{Lee:1997}
J.~Lee.
\newblock {\em Riemannian manifolds: An introduction to curvature}, volume 176.
\newblock Springer Verlag, 1997.

\bibitem{Lin2007}
C.~Lin.
\newblock Projected gradient methods for nonnegative matrix factorization.
\newblock {\em Neural Computation}, 19(10):2756--2779, 2007.

\bibitem{mac1998categories}
S.~{Mac Lane}.
\newblock {\em {Categories for the working mathematician}}.
\newblock Springer verlag, 1998.

\bibitem{Manning}
C.~D. Manning, P.~Raghavan, and H.dum Sch\"{u}tze.
\newblock {An introduction to information retrieval}.
\newblock {\em dspace.cusat.ac.in}, 2009.

\bibitem{marcus1993building}
M.~P. Marcus, M.~A. Marcinkiewicz, and B.~Santorini.
\newblock Building a large annotated corpus of english: The penn treebank.
\newblock {\em Computational linguistics}, 19(2):313--330, 1993.

\bibitem{milner1978theory}
R.~Milner.
\newblock A theory of type polymorphism in programming.
\newblock {\em Journal of computer and system sciences}, 17(3):348--375, 1978.

\bibitem{minnen2001applied}
G.~Minnen, J.~Carroll, and D.~Pearce.
\newblock Applied morphological processing of english.
\newblock {\em Natural Language Engineering}, 7(03):207--223, 2001.

\bibitem{mitchell2008vector}
J.~Mitchell and M.~Lapata.
\newblock {Vector-based models of semantic composition}.
\newblock In {\em Proceedings of ACL}, volume~8, 2008.

\bibitem{Mitchell2010}
J.~Mitchell and M.~Lapata.
\newblock {Composition in Distributional Models of Semantics}.
\newblock {\em Cognitive Science}, 2010.

\bibitem{Montague1974}
R.~Montague.
\newblock {English as a Formal Language}.
\newblock {\em Formal Semantics: The Essential Readings}, 1974.

\bibitem{moortgat2010}
M.~Moortgat.
\newblock 2 categorial type logics.
\newblock {\em Handbook of logic and language}, page~95, 2010.

\bibitem{plate1991holographic}
T.~A. Plate.
\newblock {Holographic reduced representations: Convolution algebra for
  compositional distributed representations}.
\newblock In {\em Proceedings of the 12th International Joint Conference on
  Artificial Intelligence}, pages 30--35. Citeseer, 1991.

\bibitem{Preller2010}
A.~Preller and M.~Sadrzadeh.
\newblock Bell states and negative sentences in the distributed model of
  meaning.
\newblock In P.~Selinger B.~Coecke, P.~Panangaden, editor, {\em Electronic
  Notes in Theoretical Computer Science, Proceedings of the 6th QPL Workshop on
  Quantum Physics and Logic}. University of Oxford, 2010.

\bibitem{sadrzadeh2007high}
M.~Sadrzadeh.
\newblock High-level quantum structures in linguistics and multi-agent systems.
\newblock In {\em AAAI Spring symposium on quantum interactions}, 2007.

\bibitem{Sahlgren2006}
M.~Sahlgren.
\newblock {\em The {W}ord-{S}pace {M}odel}.
\newblock Dissertation, Stockholm University, 2006.

\bibitem{Schmid1995}
H.~Schmid.
\newblock Improvements in part-of-speech tagging with an application to
  {G}erman.
\newblock In {\em Proceedings of the EACL-SIGDAT Workshop}, Dublin, Ireland,
  1995.

\bibitem{Schuetze1997}
H.~Sch\"utze.
\newblock {\em Ambiguity Resolution in Natural Language Learning}.
\newblock CSLI, Stanford, CA, 1997.

\bibitem{Selinger2010}
P.~Selinger.
\newblock {A survey of graphical languages for monoidal categories}.
\newblock {\em New Structures for Physics}, pages 275--337, 2010.

\bibitem{smolensky1990tensor}
P.~Smolensky.
\newblock {Tensor product variable binding and the representation of symbolic
  structures in connectionist systems}.
\newblock {\em Artificial intelligence}, 46(1-2):159--216, 1990.

\bibitem{Smolensky2006}
P.~Smolensky and G.~Legendre.
\newblock {\em {The Harmonic Mind: From Neural Computation to
  Optimality-Theoretic Grammar Volume I: Cognitive Architecture}}.
\newblock 2006.

\bibitem{socherEMNLP12}
R.~Socher, B.~Huval, C.~D. Manning, and A.Y Ng.
\newblock Semantic compositionality through recursive matrix-vector spaces.
\newblock {\em Proceedings of the 2012 Conference on Empirical Methods in
  Natural Language Processing}, pages 1201--1211, 2012.

\bibitem{steedman2000syntactic}
M.~Steedman.
\newblock {\em The syntactic process}.
\newblock The MIT press, 2001.

\bibitem{Steedman2009}
M.~Steedman and J.~Baldridge.
\newblock {Combinatory categorial grammar}.
\newblock {\em Non-Transformational Syntax}, 2009.

\bibitem{stolcke2002srilm}
A.~Stolcke.
\newblock Srilm-an extensible language modeling toolkit.
\newblock In {\em Seventh International Conference on Spoken Language
  Processing}, 2002.

\bibitem{turney2010frequency}
P.~D. Turney and P.~Pantel.
\newblock From frequency to meaning: Vector space models of semantics.
\newblock {\em Journal of Artificial Intelligence Research}, 37(1):141--188,
  2010.

\bibitem{van1988semantics}
J.~{van Benthem}.
\newblock The semantics of variety in categorial grammar.
\newblock {\em Categorial grammar}, 25:37--55, 1988.

\bibitem{van1995language}
J.~{van Benthem}.
\newblock {\em Language in action: Categories, lambdas, and dynamic logic}.
\newblock MIT Press, 1995.

\bibitem{van2004geometry}
C.~J. {Van Rijsbergen}.
\newblock {\em {The geometry of information retrieval}}.
\newblock Cambridge Univ Pr, 2004.

\bibitem{Vijay-Shanker1994}
K.~Vijay-Shanker and D.~J. Weir.
\newblock {The equivalence of four extensions of context-free grammars}.
\newblock {\em Theory of Computing Systems}, 27(6):511--546, 1994.

\bibitem{walters1991categories}
R.~F. Walters.
\newblock {\em {Categories and computer science}}.
\newblock Cambridge Univ Press, 1991.

\bibitem{widdowsgeometry}
D.~Widdows.
\newblock {\em {Geometry and Meaning}}.
\newblock University of Chicago Press, 2005.

\bibitem{widdows2008semantic}
D.~Widdows.
\newblock {Semantic vector products: Some initial investigations}.
\newblock In {\em Proceedings of the Second Quantum Interaction Symposium
  (QI-2008). College Publications}. Citeseer, 2008.

\bibitem{Wittgenstein1953}
L.~Wittgenstein.
\newblock {\em Philosophical investigations}.
\newblock Blackwell, 1953.

\bibitem{younger1967recognition}
D.~H. Younger.
\newblock Recognition and parsing of context-free languages in time $n^3$.
\newblock {\em Information and control}, 10(2):189--208, 1967.

\bibitem{zanzotto2010estimating}
F.~M. Zanzotto, I.~Korkontzelos, F.~Fallucchi, and S.~Manandhar.
\newblock Estimating linear models for compositional distributional semantics.
\newblock In {\em Proceedings of the 23rd International Conference on
  Computational Linguistics}, pages 1263--1271. Association for Computational
  Linguistics, 2010.

\bibitem{zyczkowski2004duality}
K.~{\.Z}yczkowski and I.~Bengtsson.
\newblock On duality between quantum maps and quantum states.
\newblock {\em Open systems \& information dynamics}, 11(01):3--42, 2004.

\end{thebibliography}
\bibliographystyle{plain}  %use the plain bibliography style

\end{document}